%% file: paper.tex
\documentclass[]{fairmeta}
\usepackage{arydshln}
\usepackage{makecell}
\usepackage{amssymb}
\usepackage{pifont}
\usepackage{tabularx}
\usepackage{wrapfig}
\usepackage{bbding}
\newcolumntype{C}{>{\centering\arraybackslash}p{0.19\textwidth}}
\newcolumntype{D}{>{\centering\arraybackslash}p{0.24\textwidth}}
\newcolumntype{Y}{>{\centering\arraybackslash}X}

\title{V-JEPA 2.1: Unlocking Dense Features in Video Self-Supervised Learning}

\author[1,2,*]{Lorenzo Mur-Labadia}
\author[1]{Matthew Muckley}
\author[1]{Amir Bar}
\author[1]{Mido Assran}
\author[1]{Koustuv Sinha}
\author[1]{Mike Rabbat}
\author[1]{Yann LeCun}
\author[1,\dagger]{Nicolas Ballas}
\author[1,\dagger]{Adrien Bardes}

\affiliation[1]{FAIR at Meta}
\affiliation[2]{Universidad de Zaragoza}

\contribution[*]{Work done at Meta}
\contribution[\dagger]{Joint last author}

\abstract{
We present \textbf{V-JEPA 2.1}, a family of self-supervised models that learns \emph{dense}, high-quality representations for visual scenes in both images and videos, while retaining strong \emph{global} scene understanding. V-JEPA 2.1 combines four key ingredients: (i) a \textbf{Dense Predictive Loss}, a masking-based objective in which \emph{all} tokens---visible context and masked tokens alike---contribute to the training loss, encouraging explicit spatial and temporal grounding; (ii) \textbf{Deep Self-Supervision}, which applies the self-supervised objective hierarchically at multiple intermediate encoder layers to improve representation quality ; (iii) \textbf{Multi-Modal Tokenizers} that support unified training over images and videos; and (iv) effective \textbf{model and data scaling}. These design choices substantially improve dense feature quality, yielding representations that are spatially structured, semantically coherent, and temporally consistent. 
. Empirically, V-JEPA 2.1 achieves state-of-the-art results on a range of benchmarks: 7.71 mAP on Ego4D for short-term object-interaction anticipation, 40.8 Recall@5 on EPIC-KITCHENS for high-level action anticipation, and a 20\% improvement in real-robot grasping success rate over VJEPA-2 AC. The model also demonstrates state-of-art performances in robotic navigation (5.687 ATE on Tartan Drive), depth estimation (0.307 RMSE on NYUv2 with a linear probe), and global recognition (77.7\% on Something-Something-V2).
Our results demonstrate that V-JEPA 2.1 advances the state of the art in dense visual understanding and world modeling.

}

\date{\today}
\correspondence{\email{lmur@unizar.es}, \email{abardes@meta.com}}

\metadata[Code]{\url{https://github.com/facebookresearch/vjepa2}}
\begin{document}

\maketitle

\begin{figure}[h]
    \centering
    \begin{minipage}{0.015\linewidth}
        \centering
        \rotatebox{90}{\textbf{V-JEPA 2.1 \quad \quad V-JEPA 2 \quad \quad Image/Video}}
    \end{minipage}\hfill
    \begin{minipage}{0.98\linewidth}
    \centering
        \includegraphics[width=\linewidth]{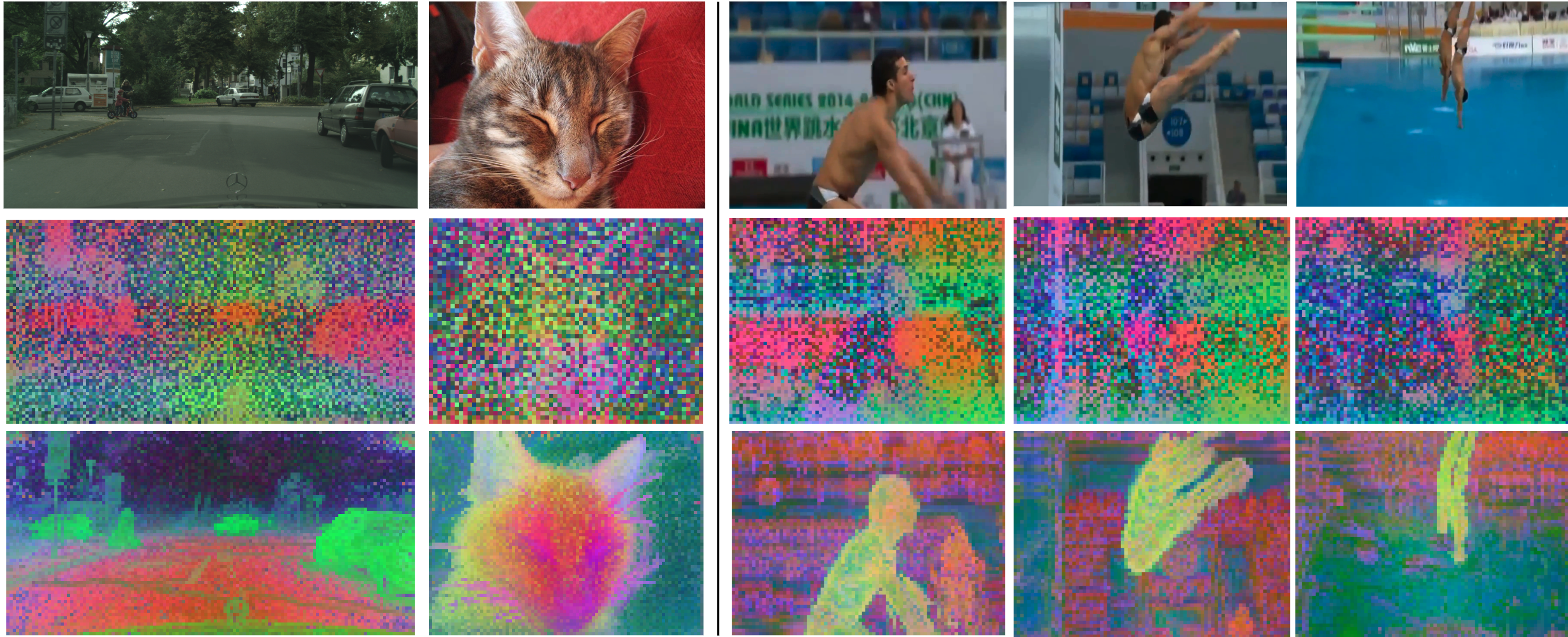}
    \end{minipage}
    \caption{\textbf{V-JEPA 2.1 unlocks high-quality dense features.} \normalfont We compute PCA on patch features extracted from the same image or video and map the top three components to RGB channels for both V-JEPA 2 (ViT-g) and V-JEPA 2.1 (ViT-G).
    Our novel V-JEPA 2.1 produces dense representations with strong spatial and temporal consistency, learning semantically coherent features where similar objects map to the same PCA components.}
    \label{fig:teaser1}
\end{figure}

\section{Introduction} \label{section:intro}

World models hold the promise of enabling agents to perceive, predict, and plan effectively in the physical world~\citep{sutton1981adaptive,ha2018world,assran2023self}. At the core of these models lies the state-estimation problem: learning representations that reliably summarize the current world state from low-level, noisy perceptual inputs. Self-Supervised Learning (SSL) from video has recently emerged as a powerful route to this goal~\citep{caron2021dino,assran2025v}, because it can exploit large-scale, label-free data to learn representations that capture scene geometry, dynamics, and intrinsic physical properties~\citep{simeoni2025dinov3,garrido2025intuitive}.

\begin{figure}[t]
    \centering
    \includegraphics[width=0.99\linewidth]{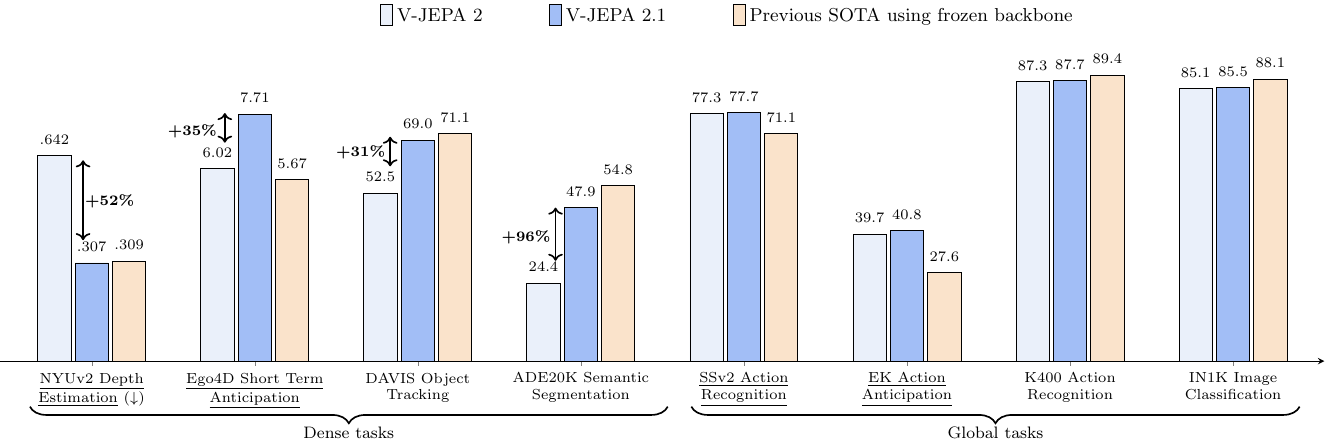}
    \caption{\textbf{V-JEPA 2.1 ViT-G performance across dense and global prediction tasks.} \normalfont We show the relative improvements of V-JEPA 2.1 compared to the previous V-JEPA 2 ViT-g model~\citep{assran2025v}. We also report the performance of previous SOTA models using a frozen backbone evaluation: DINOv3~\citep{simeoni2025dinov3} is the reference model for depth estimation, object tracking, semantic segmentation, SSv2 action recognition, and image classification; InternVideo2s~\citep{wang2024internvideo2} for K400 action recognition, STAformer~\citep{mur2024aff} for short-term object interaction anticipation, and PlausiVL~\citep{mittal2024can} for action anticipation.
    Tasks where V-JEPA 2.1 ViT-G obtains SOTA in frozen-backbone evaluation are \underline{underlined}.}
    % \mike{Are the underlined ones absolute SOTA, or SOTA in frozen backbone eval? It would be good to clarify here and throughout.}}
    \label{fig:teaser2}
\end{figure}

Despite rapid progress, learning representations that simultaneously preserve \emph{dense} spatio-temporal structure (needed for localization, geometry, and tracking) while also capturing \emph{dynamics} and supporting \emph{global} understanding (needed for high-level recognition) remains an open challenge.
Among recent advances, Joint Embedding Predictive Architectures (JEPA)~\citep{lecun2022path}—and in particular the V-JEPA family~\citep{bardes2024revisiting, assran2025v}—have demonstrated strong global video understanding, especially in settings that require modeling motion and dynamics, and have shown promise for enabling prediction and planning in embodied agents~\citep{assran2025v}. However, as illustrated in Figure~\ref{fig:teaser1}, their learned representations can be less amenable to extracting fine-grained local spatial structure. In contrast, other SSL approaches such as DINO~\citep{caron2021dino, oquab2023dinov2, simeoni2025dinov3} yield high-quality dense features for detection and segmentation, but are primarily image-based and therefore do not directly learn temporal dynamics from video.

In this work, we study self-supervised learning with a latent mask-denoising objective, where the model predicts masked segments of an image or video directly in a learned representation space. Our central finding is that high-quality \emph{dense} spatio-temporal features—preserving fine-grained spatial layout and motion dynamics—do not emerge reliably when the prediction loss is applied only to masked regions. Instead, extending the predictive loss to the \emph{entire} input, both masked and unmasked segments, substantially improves the low-level (dense) representations.

Building on this insight, we introduce \textbf{V-JEPA 2.1}, a self-supervised approach for learning unified image and video representations. V-JEPA 2.1 uses a \emph{dense predictive loss} applied to \emph{all tokens} (both visible context and masked tokens), grounding each token in its spatio-temporal location and preventing visible tokens from acting as global aggregators—an effect that is key to \emph{improving} dense feature quality (Figure~\ref{fig:methods_1}). Additionally, we find that \emph{deep self-supervision}—applying the loss hierarchically at multiple intermediate encoder layers to provide training signals throughout the network—yields consistent gains on both dense and global downstream tasks. To enable native joint training across modalities, we use modality-specific learned tokenizers for images and videos within a single shared encoder. Finally, we show these improvements scale with data and model capacity: expanding the image component from 1M to 142M images using VisionMix-163M and scaling the model from 300M to 2B parameters leads to systematic downstream gains.

We train and release a suite of V-JEPA 2.1 models (ViT-g/G, 1B/2B), along with two distilled, smaller variants (ViT-B/L, 80M/300M). Empirically, V-JEPA 2.1 achieves state-of-the-art performance on \emph{predictive} video benchmarks spanning both fine-grained and semantic forecasting: it reaches 7.71 mAP on Ego4D short-term object-interaction anticipation, which requires predicting \emph{where} and \emph{when} interactions will occur (localized interaction regions and time-to-interaction), and 40.8 Recall@5 on EPIC-KITCHENS-100 action anticipation, which evaluates the ability to forecast upcoming actions from partial temporal context. 

We further show that better dense-features also improves performances on world modelling tasks. V-JEPA 2.1 dense-features leads to +20\% success rate compared to VJEPA-2 AC~\citep{assran2025v} on grasping when when deploy our model on real Franka arms in new environment zero-shot. V-JEPA 2.1 is also suitable for robot navigation where it achieves state-of-art performances  (5.687 ATE on Tartan Drive) while having 10x faster planning speed compared to previous work~\citep{Bar_2025_CVPR}.

Beyond prediction and planning, V-JEPA 2.1 also delivers strong performance in both \emph{dense} and \emph{global} understanding tasks. For dense tasks, V-JEPA 2.1 ViT-G sets a new state of the art in linear-probe monocular depth estimation (0.307 RMSE on NYUv2), achieves competitive linear-probe semantic segmentation (85.0 mIoU on Pascal VOC), and produces temporally consistent features for video object segmentation (72.7 $\mathcal{J}\&\mathcal{F}$-Mean on YouTube-VOS). At the global level, it also attains state-of-the-art accuracy in action recognition (77.7\% on Something-Something-v2) and achieves competitive performances in Video Question Answering (VQA) tasks (83.1 accuracy on PerceptionTest).

We hope that these contributions will foster research in learning strong representations for physical world modelling, while empowering many applications in video understanding. We make our code and pretrained models publicly available to facilitate further research and applications.

\input{sections/methods}
\input{sections/results}

\section{Related work} \label{section:related_work}

\paragraph{\textbf{Self-Supervised Learning.}}

Early SSL approaches in vision focus on simple pretext tasks such as predicting the relative position of image patches~\citep{doersch2015context}, reordering shuffled patches~\citep{noroozi2016jigsaw, misra2020pirl}, inpainting missing regions~\citep{pathak2016context}, re-colorizing grayscale images~\citep{zhang2016colorful}, or predicting applied image transformations~\citep{gidaris2018rotnet}. Beyond hand-crafted tasks, view-invariant approaches proposed to use a joint-embedding architecture to learn invariances to visual transformations, which led to significant advances in SSL, with contrastive approaches~\citep{henaff2019data, he2020moco, chen2020simclr}, non-contrastive approaches~\citep{chen2020simsiam, grill2020byol, bardes2021vicreg}, or clustering-based techniques~\citep{caron2018clustering, caron2020swav}. Later, the adoption of vision transformers~\citep{dosovitskiy2020image} became the standard in self-supervised learning, first by revisiting existing view-invariant approaches ~\citep{chen2021mocov3, caron2021dino}, and then with masked image modeling approaches, which can operate in pixel space~\citep{xie2021simmim, wei2021masked}, or in the latent space of the transformer~\citep{Bao2021beit}. Most powerful approaches employ a combination of view-invariant and masked image-modeling techniques~\citep{zhou2021ibot, oquab2023dinov2}. The concept of learning representations by predicting missing information in latent space is formalized by the Joint-Embedding Predictive Architecture (JEPA)~\citep{lecun2022path}, which have been successfully applied across multiple modalities, including audio~\citep{baevski2022data2vec}, images~\citep{assran2023self,barstochastic}, and video~\citep{bardes2024revisiting}.

\paragraph{\textbf{Video Models.}}

Motion cues from videos have inspired many early video pretext tasks. Early methods predicted camera transformations between image pairs~\citep{agrawal2015moving} or future frames from ego-motion~\citep{jayaraman2015ego}. Others brought consecutive frame patches closer in feature space~\citep{wang2015videos} or used unsupervised object retrieval for segmentation~\citep{pathak2017objects}. The rise a large labeled video datasets such as Kinetics~\citep{carreira2017quovadis} changed the focus of the community towards supervised video models. Early approaches used image sequences with late fusion~\citep{donahue2015long} or 3D convolutions~\citep{tran2015learning}, but 3D ConvNets are computationally expensive. Recent work improves efficiency with 2D spatial and 1D temporal convolutions~\citep{tran2018look, xie2018rethinking}, dual-pathways~\citep{feichtenhofer2019slowfast}, and vision transformers adapted for video~\citep{arnab2021vivit, bertasius2021spacetime}. Despite rapid progress, supervised learning requires extensive labeled data and may not fully exploit temporal information, and the focus went back to self-supervised learning.

Early self-supervised learning methods focused on temporal order verification~\citep{misra2016shuffle, xu2019self}, contrastive learning~\citep{han2020self, dave2022tclr}, and future prediction~\citep{han2019video}. Recent masked modeling approaches, like VideoMAE~\citep{tong2022videomae, feichtenhofer2022masl}, extend image-based masking to video. Other advances include masked modeling in CLIP space~\citep{li2023unmasked}, joint image-video training (OmniMAE~\citep{girdhar2023omnimae}), and architectural innovations such as hierarchical transformers~\citep{ryali2023hiera} and decoupled encoders/decoders~\citep{gupta2023decoupled}. Despite progress, defining optimal pretext tasks and efficient video processing remain open challenges.

Scaling existing approaches, both in terms of data and model size, has demonstrated that generalist video encoders can be learned from extensive observation datasets of videos using self-supervised learning~\citep{carreira2024scaling, wang2023videomae, rajasegaran2025empirical}. In particular, Joint-Embedding Predictive architectures show promising efficiency gains at large scale compared to generative approaches~\citep{bardes2024revisiting}, and open the door to applications in world modeling~\citep{assran2025v}. Self-supervised models can also incorporated weak language supervision~\citep{zhao2024videoprism, wang2024internvideo2, bolya2025perception}, unlocking language-based tasks.

\paragraph{\textbf{Learning dense features.}}

Learning dense features is often achieved through designing \textit{local} loss functions and objectives, such as leveraging spatio-temporal consistency in videos~\citep{jabri2020space}, spatial alignment between image crops~\citep{pinheiro2020unsupervised, bardes2022vicregl}, and patch consistency~\citep{yun2022patch}. Contrastive learning approaches such as DetCon~\citep{henaff2021efficient} and ORL~\citep{xie2021unsupervised} use region proposals, while newer methods relax this requirement~\citep{henaff2022object, wen2022slotcon}. Recent distillation-based methods combine strengths from multiple encoders, such as AM-RADIO~\citep{ranzinger2024radio}, Perception Encoder~\citep{bolya2025perception}, and DINOv3~\citep{simeoni2025dinov3}, using objectives like cosine similarity and Gram matrix regularization to improve dense features. Some works focus on post-hoc improvements, including fine-tuning with clustering objectives~\citep{ziegler2022self}, patch alignment~\citep{salehi2023time}, and patch-sorting~\citep{pariza2025near}. Others enhance features without fine-tuning, such as STEGO~\citep{hamilton2022unsupervised} and feature augmentation~\citep{simoncini2024no}.

Beyond playing with the loss function, the standard Vision Transformer (ViT)~\citep{dosovitskiy2021vit} can be adapted for dense prediction tasks, \citet{ranftl2021dpt} introduced the Dense Prediction Transformer (DPT), evolving from their prior CNN-based work on MiDaS~\citep{ranftl2020midas}. DPT solved the token-to-pixel problem using a "reassemble" operation that converts the 1D ViT sequence into multi-scale 2D feature maps, which are then fused by a convolutional decoder adapted from RefineNet~\citep{lin2017refinenet}. This design, which leverages the global receptive field of the transformer backbone, proved effective for obtaining structural coherence, moving beyond the limitations of purely hierarchical CNNs. The work also catalyzed parallel research, including the development of purely transformer-based decoders like Segmenter~\citep{strudel2021segmenter} and hierarchical transformer backbones like the Swin Transformer~\citep{liu2021swin} and the Multiscale Vision Transformer~\citep{fan2021multiscale}. DPT architectural framework has proven to be the most influential design for scaling up, serving as the basis for current state-of-the-art foundation models like Depth Anything~\citep{yang2024depthanything}.

\section{Conclusion and Future work} \label{section:conclusion}

This work demonstrates how Joint-Embedding Predictive Architectures and Self-Supervised Learning from video unlock high-quality dense local features with strong global semantic understanding.
Our key ingredient is a novel deep spatio-temporal self-supervision, composed by a weighted context loss and a multi-level predictor that enables supervision across the intermediate encoder layers.
This design unlocks high-quality dense features that are spatially structured, semantically coherent and temporally consistent, as evidenced by PCA visualizations.
Beyond the training objective, we demonstrate the importance of modality-specific tokenizers, large-scale and balanced image–video data curation, and scaling to ViT-G, which together enable an effective distillation to compact models.
Our V-JEPA 2.1 ViT-G achieves absolute state-of-the-art performance in short-term object interaction anticipation, action anticipation and action recognition, showcasing its excellent predictive and motion understanding capabilities. 
Moreover, the dense feature maps achieve linear-probe state-of-the-art performance in monocular depth estimation and strong results in semantic segmentation and video object tracking. We hope these findings encourage further research on predictive physical world modeling.

%, with a careful design of the loss function, unlock dense video features, which yield excellent downstream performance on dense tasks. The key ingredient is to apply the loss at the spatio-temporal locations corresponding to the context of the prediction \textit{in addition} to the locations to predict, which is highlighted by high-quality and temporally consistent PCA visualizations. V-JEPA 2.1 significantly improves the performance of V-JEPA 2 on image, video segmentation and depth estimation tasks, reaching comparable performance to DINOv3, while preserving the global video understanding capabilities. V-JEPA 2.1 also sets new state-of-the-art performances on prediction tasks: action anticipation, and short-term action anticipation that requires predicting bounding boxes of objects that will interact in the future in ego-centric videos.

\paragraph{\bf Future work.} 

Future work will focus on several promising directions. First, scaling along two axes: a) model size, we observe a very positive trend scaling from 1B to 2B and other work such as DINOv3 showed the benefit of scaling to 7B in vision; b) data size, our work on data curation highlighted that video self-supervised learning really benefit from large-scale datasets. Second, this work focus on learning better representation, whereas V-JEPA 2 showed the potential of building world models on top of these representations. Future work in this direction will explore world modeling aspects with the prism of learning dense prediction capabilities~\citep{karypidis2025dinof, baldassarre2025dinow}. Finally, world models with dense understanding and prediction abilities will unlock many applications in robotics and autonomous agents, where a precise estimation of the state down to the pixel level is required, for example navigation in challenging real world environment, or fine-grained manipulation.

\clearpage
\newpage
\bibliographystyle{assets/plainnat}
\bibliography{paper}

\clearpage
\newpage
\beginappendix

\section{Pretraining details} \label{app:pretraining}

\begin{table}
\centering
\caption{\textbf{Pretraining hyperparameters for the Primary and Cooldown phases.} Hyperparameters are identical across all our ViT architectures (ViT-L, -g and -G).}
\begin{tabular}{lcc}
\toprule
\textbf{Parameter} & \textbf{Primary Phase} & \textbf{Cooldown Phase} \\
\midrule
Number of frames               & 16                      & 64                      \\
Frames per Second             & 4.0                     & 4.0                     \\
Video Crop Size& 256                     & 384\\
 Image Crop Size& 256&512\\
Random Resize Aspect Ratio    & [0.75, 1.35]            & [0.75, 1.35]            \\
Random Resize Scale           & [0.3, 1.0]              & [0.3, 1.0]              \\
Steps                         & 135,000& 12000                   \\
Warmup Steps                  & 12,000& N/A                     \\
Image batch Size (global)& 2,304& 2,304\\
Video batch Size (global)& 128&128\\
Starting Learning Rate        & 1e-4                    & 6e-4\\
Final Learning Rate           & 5.25e-4                 & 1e-6                    \\
Weight Decay                  & 0.04                    & 0.04                    \\
EMA                           & 0.99925                 & 0.99925                 \\
Spatial Mask Scale            & [0.15, 0.7]             & [0.15, 0.7]             \\
Temporal Mask Scale           & [1.0, 1.0]              & [1.0, 1.0]              \\
Mask Aspect Ratio             & [0.75, 1.5]             & [0.75, 1.5]             \\
Tubelet Size                  & 2                       & 2                       \\
Patch Size                    & 16                      & 16                      \\
Predictor Blocks              & 24                      & 24                      \\
Predictor Embedding Size      & 384                     & 384                      \\
Encoder Intermediate Blocks (ViT-G)   & [12, 24, 36, 48]        & [12, 24, 36, 48]         \\
Context Loss $\lambda$   & 0.5 for video, 0.7 for image        & 0.5 for videos, 0.7 for images        \\
\bottomrule
\end{tabular}
\vspace{5mm}
\label{tab:training_phases}
\end{table}

In this section, we detail our pretraining protocol and provide hyper-parameters. We use identical hyper-parameters for all our ViT architectures (ViT-L, -g and -G). Following~\cite{assran2025v}, pretraining consists of two phases: 1) the primary phase, lasting for 135k iterations, where the learning rate follows a warmup (from 1e-4 to 5.25e-4) for 12k iterations, then a constant schedule (5.25e-4); and 2) a cooldown phase, lasting for 12k iterations, where the learning rate is decayed to a small value (from 6e-4 to 1e-6).

During the primary phase, video samples consist of 16-frames clips sampled at 4 frames per second, both video and images have a resolution of $256 \times 256$. The cooldown phase is shorter, allowing for higher input resolution with manageable compute resources, video samples consist of 64-frames clips at resolution $384 \times 384$ and images are sampled at resolution $512 \times 512$. During both phases, the masking strategy follows the one introduced in~\cite{bardes2024revisiting}, the image and video batch sizes are set to, respectively, 2304 and 128, the weight-decay is set to 0.04 and the EMA coefficient is set to 0.99925.

We use the standard ViT architecture with a patch size of 16 for both images and videos, and a tubelet size of 2 for videos. Our predictor has 24 blocks (versus 12 in V-JEPA 2) and an embedding size of 384. For our deep self-supervision loss, we use 4 intermediate blocks of the encoder, at equally spaced indices. The parameter $\lambda$ we use for Eq.~\ref{eq:lambda_i} is different for video and images, with a value of 0.5 for video and 0.7 for images.

\section{Distillation details} \label{app:distillation}

Our distillation protocol is adapted from the pretraining recipe, maintaining the same JEPA loss, masking, architecture, and hyperparameters, except for a few key modifications: the Exponential Moving Average (EMA) encoder is replaced by a frozen Teacher network. The predictor's final layer embedding dimension is adapted to match the Teacher's dimension to enable the computation of the loss. We do not use deep-self-supervision for distillation: the predictor operates only on the last layer of the encoder, and the loss is only calculated on this last layer. We do not initialize the predictor with the pretraining weights and instead retrain a new predictor, which has 12 blocks (instead of 24 for pretraining), greatly improving distillation stability. Throughout training, a separate EMA of the context encoder is maintained, corresponding to Polyak averaging, which is never used in the loss calculation but serves as the final, high-performing model released for downstream tasks. 

Distillation follows a multi-stage training approach similar to pretraining. Stage 1 (Primary Phase) with a warmup-then-constant schedule and low-resolution data (16 frames, $256 \times 256$ pixels) using a pre-cooldown low-resolution V-JEPA 2.1 ViT-G as the teacher, and Stage 2 (Cooldown Phase) with a short annealing schedule and high-resolution data (64 frames, $384 \times 384$) using the corresponding post-cooldown high-resolution V-JEPA 2.1 ViT-G as the teacher. The Stage 2 student network is initialized with the EMA student network from Stage 1, a two-stage strategy found to produce the best-performing models compared to alternatives such as single-stage approaches or always distilling from the final high-resolution V-JEPA 2.1 teacher.

\section{Evaluation protocols} \label{app:evaluation}

\subsection{Depth Estimation}

\paragraph{\bf Datasets and Metrics.} We evaluated the geometric quality of the dense features of V-JEPA 2.1 in depth estimation on NYUv2~\citep{silberman2018nyuv2} and KITTI~\citep{geiger2013kitti}, reporting the root mean square error (RMSE).

\paragraph{\bf Evaluation Protocol.} For fair comparison with the state-of-the-art, we adopted the same protocol reported by \cite{simeoni2025dinov3}, training a corresponding linear classifier on the training set of each data set. 
The linear layer is applied only on top of the V-JEPA 2.1 visual encoder frozen patch features, and it is further normalized using a learned BatchNorm layer.
Note that we both train and evaluate single-image classifiers, as we process individual images using the 2D Convolution image tokenizer described in Sec.\ref{sec:tokenizer}. 
We sweep learning rates $\{1.5\!\times\!10^{-3}, 1\!\times\!10^{-3}, 8\!\times\!10^{-4}, 3\!\times\!10^{-4}, 1\!\times\!10^{-4}, 8\!\times\!10^{-5}, 3\!\times\!10^{-5}\}$ and weight decay $\{10^{-3}, 10^{-4}\}$.

\subsection{Semantic Segmentation}

\paragraph{\bf Datasets and Metrics.} We evaluate semantic segmentation with linear probing on the image semantic segmentation datasets ADE20K~\citep{zhou2017ade20k}, Pascal VOC12~\citep{everingham2015pascal}, and Cityscapes~\citep{cordts2016cityscapes}, reporting mean intersection-over-union (mIoU).

\paragraph{\bf Evaluation Protocol.} Similarly, we also follow the same protocol reported in DINOv3 \cite{simeoni2025dinov3}.
We train a linear classifier on top of last-block patch features from the frozen encoder (already normalized via LayerNorm).
Similar to the depth estimation evaluation protocol, we both train and evaluate single-image classifiers.
Linear probes are optimized with AdamW using learning rates 
$\{1.5\!\times\!10^{-3}, 8\!\times\!10^{-4}, 5\!\times\!10^{-4}, 8\!\times\!10^{-5}\}$ and weight decay  $\{0.1, 0.01, 10^{-4}\}$.
We use a $512$px resolution for ADE20K and VOC12, and $1024$px height for Cityscapes.

\subsection{Video Object Segmentation Tracking}

\paragraph{\bf Dataset and Metrics.} We evaluate on DAVIS 2017~\citep{pont20172017} and YouTube-VOS~\citep{xu2018youtube}. DAVIS provides dense annotations for train/val (60/30 videos). In YouTube-VOS, as only the training set is publicly available, we divided this set both training (80$\%$) and validation (20$\%$) videos.

\paragraph{\bf Evaluation Protocol.}
Following \citep{rajasegaran2025empirical} and \cite{simeoni2025dinov3}, we implement a non-parametric label-propagation approach based on cosine similarity between patch features from a frozen backbone. 
Segmentation labels from the first frame are encoded as one-hot patch assignments. For each subsequent frame, we compute similarities with patches from the first frame and a small set of past frames, and propagate labels via a weighted $k$-NN scheme within a spatial neighborhood. 

We sweep over several hyperparameters, including context length $\{4,7,10,15\}$, neighborhood size $\{12,24,36\}$, neighborhood shape $\{\text{square},\text{circle}\}$, top-$k$ neighbors $\{3,5,10\}$, and similarity temperature $\{0.01, 0.1, 0.2, 0.7\}$. 
Hyperparameters are selected on the DAVIS training set and applied to all test splits. We use an input resolution of 480px.

\subsection{Short Term Object Interaction Anticipation}

\paragraph{\bf Dataset and Metrics.}
We evaluate V-JEPA 2.1 on the STA v2 split of Ego4D \citep{grauman2022ego4d}, which comprises 243 hours of annotated video clips spanning 128 noun and 81 verb categories, with a total of 98,276 training and 47,395 validation samples.
Following the established evaluation protocol \citep{grauman2022ego4d}, we report Top-5 Average Precision (AP) and Top-5 mean Average Precision (mAP) metrics. As in standard mAP, predictions are matched to ground-truth boxes using IoU > 0.5 and additional class-specific criteria. For instance, the Top-5 mAP All variant requires a correct noun, correct verb, IoU above 0.5, and a time-to-contact $\delta$ prediction within a 0.25-second tolerance. To address the inherently multi-modal nature of future anticipation—where multiple plausible next-active objects may exist—the metric discounts up to four highest-scoring false positives per example, thereby avoiding penalties for plausible but unannotated predictions.
Additionally, we report individual metrics to evaluate the temporal ($\delta$), spatial (bounding boxes), and semantic (noun and verb) dimensions of the task.

\paragraph{\bf Evaluation Protocol.}

For a fair comparison with the previous state-of-the-art, we follow the official StillFast baseline implementation~\citep{ragusa2023stillfast}, which extends Faster R-CNN~\citep{girshick2015fast}. 
Our attentive probe consists of four attention blocks followed by the frame-guided temporal pooling mechanism of~\citep{mur2024aff}, since STA predictions must remain spatially aligned with the last video frame.
This pooling mechanism adopts a residual cross-attention module between the 3D tokens of the full video clip and the 2D tokens of the last frame. Queries are computed from the last-frame tokens via a linear projection, while keys and values are obtained from the video tokens using corresponding projections.
Once the video tokens are pooled in 2D, we apply a bilinear interpolation to match the resolution of the high-resolution image tokens. Both representations are summed and fused through a 2D convolution layer. Finally, we upsample this representation to produce a multi-scale pyramid at $1/4$, $1/8$, $1/16$, $1/32$, and $1/64$ of the image input resolution. This pyramid forms the input to the prediction head, which follows the design used in prior work~\citep{ragusa2023stillfast, mur2024aff, thakur2023guided}.

\paragraph{\bf Probe Architecture.} In this prediction head, first a Region Proposal Network (RPN) generates proposals from a feature pyramid, and RoIAlign extracts corresponding local features. 
To enrich these features with scene-level context, we concatenate to each local RoI feature the resulting pooled token from the 4-layers attentive probe.
The fused representation is processed by two fully connected layers and added back to the local features through a residual connection, enabling global cues to modulate––rather than replace––local information.
Noun, verb and time to contact $\delta$ are predicted using respective linear layers from the fused RoI features. A Softplus activation layers is applied to the $\delta$ output in order to obtain positive predictions.
Training is end-to-end, combining standard Faster R-CNN losses with an additional verb loss $\mathcal{L}_v$ and a Smooth-$L_1$ TTC loss $\mathcal{L}_{\text{ttc}}$, weighted by 0.1 and 0.5 respectively. 

\paragraph{\bf Optimization.} Models are optimized with Adam using learning rates in  $\{1\!\times\!10^{-5},\, 8\!\times\!10^{-5},\, 1.2\!\times\!10^{-4},\, 2\!\times\!10^{-4},\, 5\!\times\!10^{-4}\}$.  We additionally explore different input configurations, including sampling rates  $\{16, 8, 4, 2\}\,\text{fps}$, clip lengths $\{32, 16, 8, 4\}$ frames, and spatial resolutions $\{256\text{px},\, 384\text{px}\}$, covering videos with varying temporal spans. We follow the multi-scale training strategy of Faster R-CNN, sampling image short sides uniformly from $\{640,672,704,736,768,800\}$ with a maximum long side of 1333~px. Our final best configuration consist of 16 frames videos at 384px with a sampling rate of 2 fps, and using a learning rate of $1.2\!\times\!10^{-4}$. We apply the same protocol to DINOv2 \cite{oquab2023dinov2}, DINOv3 \cite{simeoni2025dinov3} and V-JEPA 2 \cite{assran2025v} models, reporting the best respective performance.

\subsection{Video and Image Classification} \label{app:classification}

\paragraph{\bf Dataset and Metrics.} We use the evaluation benchmarks of V-JEPA 2 and evaluate our models on image classification on ImageNet~\cite{deng2009imagenet}, and video classification on Kinetics-400~\cite{kay2017kinetics}, Something-Something-v2~\cite{goyal2017something} and Diving-48~\cite{li2018resound}.

\paragraph{\bf Probe Architecture.} We train an attentive probe on top of the frozen encoder output using the training data from each downstream task. Our attentive probe is composed of four transformer blocks, each using 16 heads in the attention layer. The first three blocks use standard self-attention; the final block uses a cross-attention layer with a learnable query token. The output of the cross-attention layer in the final block is added back to the query token as a residual connection before applying the rest of the block (LayerNorm, followed by MLP with a single GeLU activation). The transformer blocks are followed by a final linear classifier layer.

\paragraph{\bf Evaluation Protocol.} We follow the evaluation protocol of V-JEPA 2. For video evaluations, we sampled multiple clip segments from each input video. During validation, we also extracted three spatial views from each segment (instead of one view during training). The number of clip segments, frame step parameter, and global batch size vary for each evaluation. We use $64 \times 2 \times 3$ input for SSv2 (16 frames clip, 2 temporal crops, 3 spatial crops), $16 \times 8 \times 3$ for K400, and $32 \times 4 \times 3$ for Diving-48. For all benchmarks, we use a spatial resolution of $384 \times 384$. We use the 2D or 3D tokenizer, respectively, for image and video benchmarks. We use a batch size of 256 except for ImageNet, we use 1024. For ImageNet, we do not use multiple clips or views per sample. For Diving-48, we employ a multilayer strategy. Instead of attending to the tokens from only the last layer of the encoder, we extract tokens from four encoder layers (the last layer and three intermediate layers) and attend to all of them.

\paragraph{\bf Optimization.} For each evaluation, we simultaneously train multiple classifier heads with different hyper-parameters (learning rate and weight decay), reporting the accuracy of the best-performing classifier. For most of our evaluations (Kinetics, SSv2, and ImageNet), we train for 20 epochs and use 20 heads, each
using one of five learning rate values and four weight decay values, and the learning rate decays according to a cosine schedule. For Diving-48, which benefit from longer training, we train for 50 epochs.

\subsection{Action Anticipation}

\paragraph{\bf Dataset and Metrics.} We evaluated V-JEPA 2.1 on the Action Anticipation task of the Epic-Kitchen-100 benchmark~\cite{Damen2022rescaling}. The EK100 dataset is comprised of 100 hours of cooking activities recorded from an egocentric perspective across 45 kitchen environments. Each video in EK100 is annotated with action segments, which include a start timestamp, an end timestamp, and an action label. There are 3,568 unique action labels, each consisting of a verb and a noun category, with a total of 97 verb categories and 300 noun categories. The EK100 action anticipation task involves predicting noun, verb, and action (i.e., predicting verb and noun jointly) from a video clip, referred to as context, that occurs before the start timestamp of an action segment.
The interval between the end of the context and the beginning of the action segment is the anticipation time, which is set to 1 second by default. Given that different future actions may be possible from a given context, mean-class recall-at-5 for noun, verb and action, is used as the metric to measure performance~\citep{Damen2022rescaling}.

\paragraph{\bf Evaluation Protocol.} We follow the same evaluation protocol introduced in V-JEPA 2 and use a focal loss~\cite{lin2017focal} with $\alpha$ = 0.25 and $\gamma$ = 2.0 when training the probe. We used a context of 32 frames with a frame-rate of 8 frames per second at resolution $384 \times 384$. During probe training, we randomly sample an anticipation time between 0.25 and 1.75 seconds and an anticipation point between 0.0 and 0.25. Our probe architecture for action anticipation follows the architecture introduced in V-JEPA 2 and described in Appendix~\ref{app:classification}, consisting of four transformer blocks, including a last cross-attention layer with a set of learnable query tokens, followed by a final linear classifier layer for each query token.

\section{Additional Ablations}

\subsection{Multi-scale evaluation} \label{app:multiscale_evaluation}

We evaluate the impact of Deep Self-Supervision on the need for multi-scale evaluation, i.e. using multiple layers of the encoder as input to the evaluation probe. Table~\ref{tab:multilayers} compares a model trained with Deep Self-Supervision with a model trained without, and we measure performance on image segmentation on ADE20k, depth estimation on NYU, and action recognition on Diving-48, using either; a) only the last layer of the encoder as input to the probe (Last-Layer) or b) 4 equally spaced intermediate layers (4-Layers). We observe that the model trained \textit{without} Deep Self-Supervision has a wide gap in performance between using Last-Layer and 4-Layers, and benefits a lot from multi-scale evaluation. On the other hand, the model trained \textit{with} Deep Self-Supervision is already stronger using Last-Layer, and only benefits a little from using 4-Layers.

\begin{table}
\centering
\caption{\textbf{Impact of deep self-supervision on the need for multi-scale evaluation.} We conduct experiments with  our V-JEPA 2.1 ViT-L trained from scratch. Last-Layer indicates that evaluation is conducted with only the last layer of the encoder as input to the probe. 4-Layers indicates that evaluation is conducted by concatenating 4 intermediate layers of the encoder and feeding it to the probe.}
\resizebox{\columnwidth}{!}{%
\begin{tabular}{ccccccc}
\toprule

\textbf{Deep Self-Supervision} &
\multicolumn{2}{c}{\textbf{ADE20k}}&
\multicolumn{2}{c}{\textbf{NYU}}& \multicolumn{2}{c}{\textbf{Diving-48}}\\
% \cmidrule(lr){3-4} \cmidrule(lr){5-6}
& Last-Layer & 4-Layers & Last-Layer & 4-Layers & Last-Layer & 4-Layers \\

% \textbf{Method} & \textbf{Multi-Layers Pred.} & \textbf{ADE20k Last-Layer} & \textbf{ADE20k 4-Layers} & \textbf{NYU Last-Layer} & \textbf{NYU 4-Layers} & \textbf{Diving-48 Last-Layer} & \textbf{Diving 4-Layers}\\
\midrule
% V-JEPA 2 ViT-L & & 24.4 &	33.4&	0.649&	0.591&	84.3&	89.0 \\
% \hdashline
\XSolidBrush	& 34.9&	39.1	&0.513	&0.463	&85.8	& 86.9 \\
\CheckmarkBold & 42 &	43.9&	0.381	&0.370	&87.2&	88.1 \\
\bottomrule
\end{tabular}
}
\label{tab:multilayers}
\end{table}

\subsection{Pretraining Spatial Resolution}

Table~\ref{tab:lowres_pretraining} presents the impact of the resolution used during the cooldown phase of pretraining on performance on most of our downstream tasks. We compare models trained with a video spatial resolution of $256 \times 256$ to the models we present in the main paper that are trained with a video spatial resolution of $384 \times 384$. V-JEPA 2.1 benefits from higher resolution pretraining across all of our downstream tasks.

\begin{table}
\caption{\textbf{Impact of pretraining spatial resolution.} We compare models trained during the cooldown phase at $256 \times 256$ spatial resolution against $384 \times 384$.}
\resizebox{\columnwidth}{!}{%
\begin{tabular}{lccccccccccccc}
\toprule

\textbf{Model} & \textbf{Res} &
\bf SSv2 &	\bf D-48 &	\bf K400 & \bf IN1K & \bf EK100 & \bf ADE20k & \bf Citys. & \bf VOC &	\bf NYU	& \bf KITTI & \bf DAVIS & \bf YT-VOS \\
\midrule
ViT-g  & 256 &	76.7&	88.6&	86.2&	84.1&	35.7	&47.8&	71.7&	84.6&	0.359&	2.611&	68.0&	72.0 \\

ViT-g  & 384&	76.9&	89	&87	&84.8&	38.4	&47.8	&71.8&	84.7&	0.350&	2.601&	67.4	&71.3\\

\hdashline

ViT-G & 256&	77.1&	89.5&	87.3&	84.8	&38.8	&47.8	&73.2	&84.7&	0.313&	2.472&	68.6	&72.7\\

ViT-G & 384 &	77.7	&89.2	&87.7	&85.5	&40.8	&47.9	&73.5&	85.0&	0.307	&2.461	&69.0	&72.6 \\
\bottomrule
\end{tabular}
}
\label{tab:lowres_pretraining}
\end{table}

\end{document}

%% file: sections/methods.tex
\section{Methodology} \label{section:method}

%\nicolas{I would split Methodology into three subsection 1) Preliminary: (i.e. stuff prior to the contribution) 2) Identifying the issue: analysis showing that applying the loss on masked only leads to bad PCA MAPS, norm distribution unrelated to the input patches... 3) VJEPA 2.1 which present the fix to the problem. For 3), I would distinguish fix that are novel and specific for fixing low-level recipe (predictall, hierarchy, dual tokenizer) vs fix that are more generic (pretraining data/scale) so novelty stands more. Also we need to discuss empirical setup and results in the section, see the VJEPA 2  method section for instance. }

%\nicolas{maybe put the first part in a preliminary section as this is not new/contribution}

%\textcolor{red}{GENERAL TWO PARAGRAPHS EXPLAINING OVERVIEW VJEPA, MAKING REFERENCE TO FIG 2 LEFT.}

\subsection{Preliminaries: Joint-Embedding Predictive Architectures}

Joint-Embedding Predictive Architecture (JEPA)~\citep{lecun2022path} is a self-supervised learning framework designed to learn representations of data by making predictions in a learned latent space, rather than directly in the observation (input) space.
JEPA models operate by encoding both a noise-corrupted version and an uncorrupted (clean) version of the same input. A predictor network is then trained to predict the representation of the clean input from the representation of the corrupted input.

Corrupted and clean inputs are processed by the encoder $E_{\theta}(\cdot)$, which produces latent representations.
The predictor, $P_{\phi}(\cdot)$, is then used to map the representation of the corrupted input to the representation of the clean input. Given that the encoder and predictor are learned simultaneously, the system admits a trivial and uninformative solution in which the encoder $E_{\theta}(\cdot)$ outputs a constant vector regardless of its input.
To avoid this representation collapse, explicit~\citep{bardes2021vicreg,balestriero2025lejepa,mo2024connecting} or implicit~\citep{grill2020bootstrap,assran2023self} regularization is used to promote representations that preserve input information.

%With an optimal predictor, the encoder is implicitly driven to retain all information necessary to infer unseen content.

In this work, we build upon the V-JEPA family of models, where video representations are learned through a mask-denoising objective in the representation space~\citep{bardes2024revisiting,assran2025v}.
The V-JEPA objective aims to predict the representation of a video $y$ from another view $x$ of that video that has been corrupted through masking, i.e., from which patches have been randomly dropped.
%\nicolas{Maybe add back figure to illustrate}.
The encoder $E_\theta(\cdot)$ processes the masked video $x$, and outputs an embedding vector, or context tokens, for each visible patch. The
outputs of the encoder are concatenated with a set of learnable mask tokens $\Delta_y$ that specify the spatio-temporal position of the masked patches. The predictor network processes the combined tokens sequence, and outputs an embedding vector for each input token.
The encoder and predictor are trained by minimizing the following objective:
\begin{equation}
\mathcal{L_{\text{predict}}} =  \frac{1}{|M|} \sum_{i \in M}  \| P_{\phi}(E_{\theta}(x), \Delta_y)_i - \mathrm{sg}(E_{\overline{\theta}}(y)_{i}) \|_{1}  ,
\label{eq: l1_main_loss}
\end{equation}
where $M$ is a set containing the masked patch indexes from the $x$ view.
The loss use a stop-gradient operator, $\mathrm{sg}$, to prevent representation collapse~\citep{grill2020byol} and an exponential moving average $\overline{\theta}$  of $\theta$ is used to update the weight of the encoder $E_{\overline{\theta}}(y)$  processing the video $y$. $\mathcal{L_{\text{predict}}}$ is only applied on masked tokens, and not on the context tokens.

Both encoder $E_{\theta}(\cdot)$ and predictor $P_{\phi}(\cdot)$ are parametrized with Vision Transformer~\citep{dosovitskiy2020image} and we rely on the same masking strategy as~\citet{assran2025v}. Refer to Appendix~\ref{app:pretraining} for more details.

\subsection{Analysis of V-JEPA  Features for Dense Vision Tasks}

While V-JEPA has proven to be an effective approach for understanding global semantic information from video, predicting future actions, and planning to reach specific goals~\citep{bardes2024revisiting,assran2025v}, previous works have not investigated the suitability of V-JEPA 2 features for dense vision tasks. To address this gap, we analyze the V-JEPA 2 feature maps through qualitative visualizations and dense downstream tasks evaluation.

For the qualitative visualizations, we compute the Principal Component Analysis (PCA) of patch features extracted from the V-JEPA 2 encoder, and we map the first three components to the RGB color channels.
We assess the encoder performance on dense tasks using a linear probing protocol, in which we train a single linear layer on top of the frozen encoder features. We evaluate V-JEPA 2 on semantic segmentation using the ADE20K dataset~\citep{zhou2017ade20k} and depth estimation on NYUv2~\citep{silberman2018nyuv2}. Refer to Appendix~\ref{app:evaluation} for more details on the evaluation setup.

\paragraph{\bf Observations and Hypothesis.}

\begin{wrapfigure}{r}{0.49\textwidth}
    \centering
    \vspace{-4mm}
    \begin{tabularx}{\linewidth}{%
        >{\centering\arraybackslash}m{0.1cm}   % columna 1 estrecha
        >{\centering\arraybackslash}m{0.36\linewidth}  % columna 2 MÁS grande
        >{\centering\arraybackslash}m{0.47\linewidth}  % columna 3 MÁS pequeña
    }

        \rotatebox{90}{Image} &
        \includegraphics[width=\linewidth]{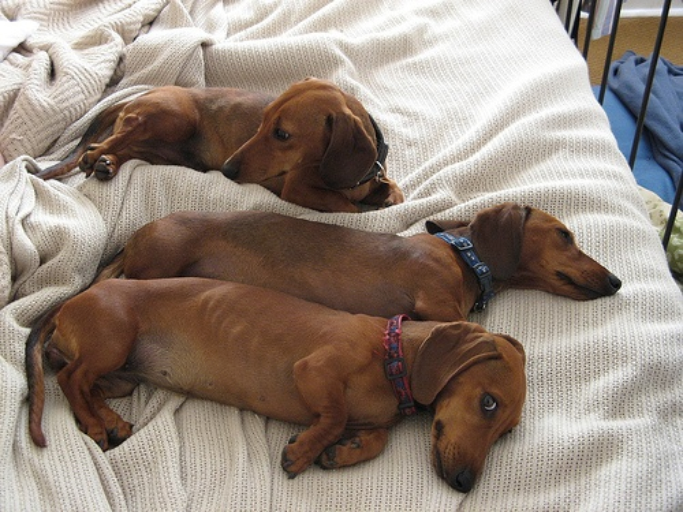} &
        \includegraphics[width=\linewidth]{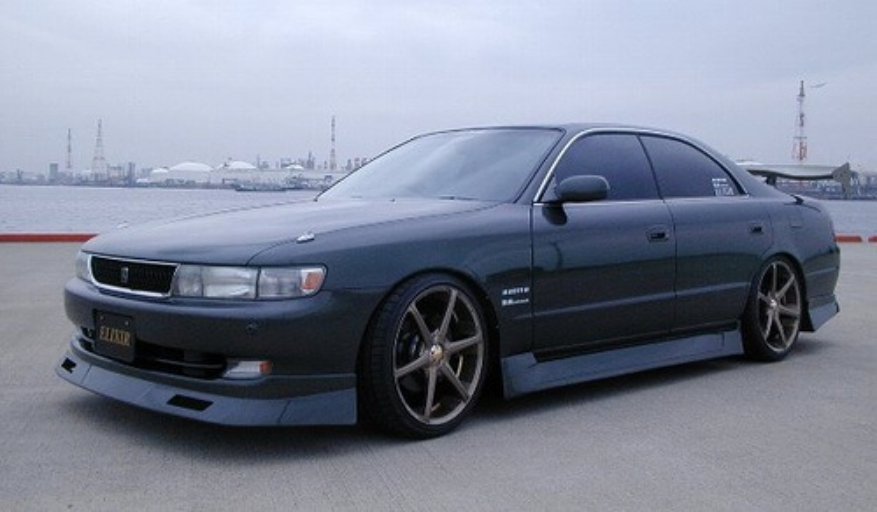} \\

        \rotatebox{90}{V-JEPA 2} &
        \includegraphics[width=\linewidth]{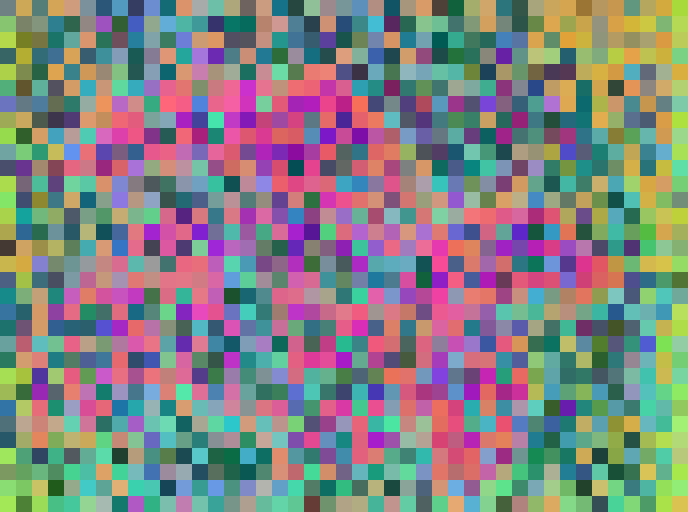} &
        \includegraphics[width=\linewidth]{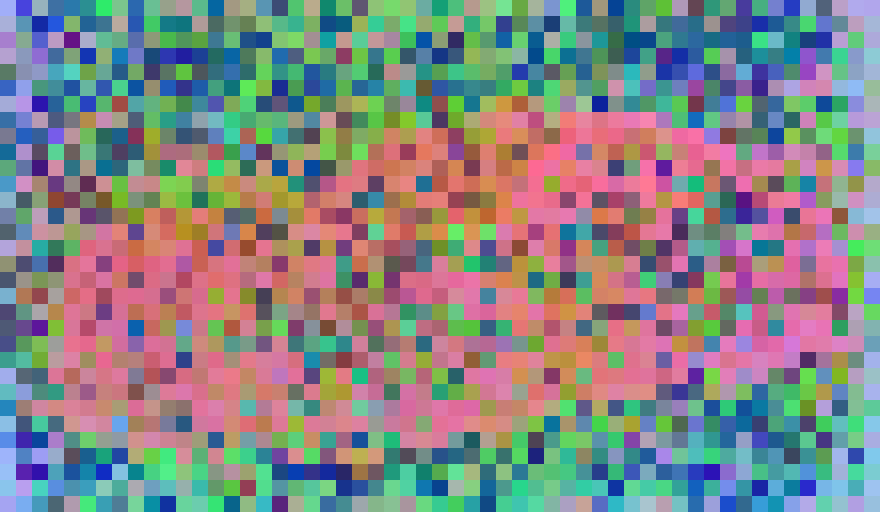} \\

        \rotatebox{90}{$+\mathcal{L}_{ctx}$} &
        \includegraphics[width=\linewidth]{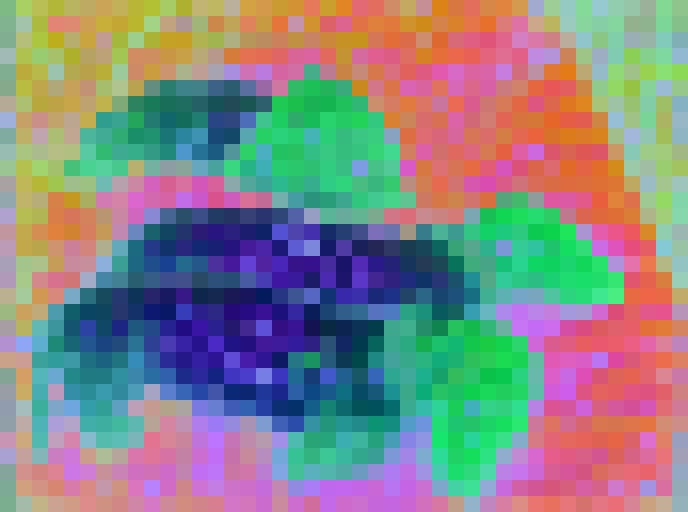} &
        \includegraphics[width=\linewidth]{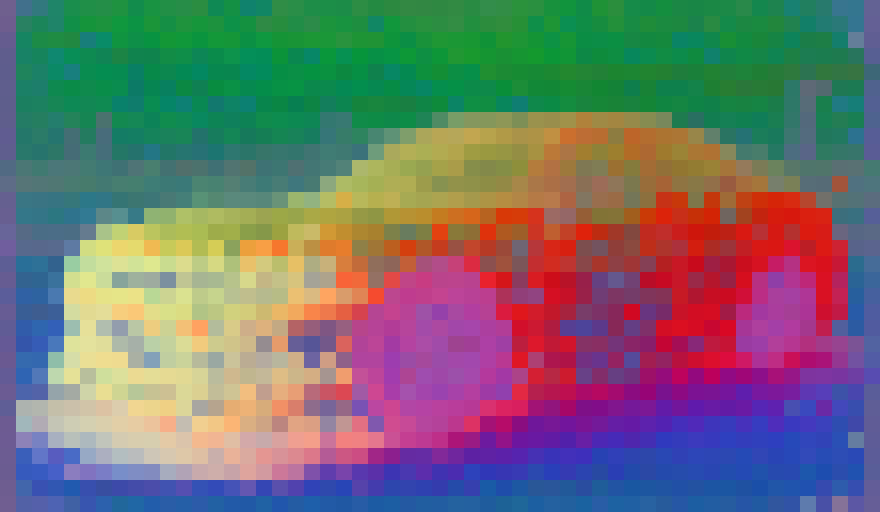} \\

    \end{tabularx}

    \caption{\textbf{Influence of the context loss $\mathcal{L}_{ctx}$.} \normalfont We show PCA visualizations of the feature map representations learned with V-JEPA 2 and with a model trained using V-JEPA 2 plus our $\mathcal{L}_{ctx}$ loss.
    While V-JEPA features only show fragmented local spatial structure, explicitly supervising unmasked regions with $\mathcal{L}_{ctx}$ leads to feature maps that exhibit coherent spatial structure. Similar semantic parts (e.g., heads of dogs, wheels of cars) are mapped to the same PCA components.}
    \label{fig:methods_1}
\end{wrapfigure}

Feature map visualizations of V-JEPA 2 are shown in Figure~\ref{fig:teaser1} and Figure~\ref{fig:methods_1}. We observe that feature maps are noisy and show only fragmented local spatial structure. Additionally, V-JEPA 2 features obtain limited performance on dense tasks when using a simple linear probing protocol, such as semantic segmentation (22.2 mIoU on ADE20K) or depth estimation (0.682 RMSE on NYUv2), as reported in Table~\ref{tab:ablation_table}. Overall, these results support the conclusion that local information about the visual scene is not easily extractable from the V-JEPA 2 representation.

We hypothesize that the absence of local structure in the feature maps is due to the lack of self-supervision on patches that are not masked, i.e. the context patches. 
The predictor $P_{\phi}(\cdot)$ takes as input the concatenation of context tokens computed by $E_{\theta}(x)$ and a set of mask tokens $\Delta_y$ that specify the masked positions to predict. The predictor outputs one token for each input, i.e., for both context and masked tokens. However, the original loss from V-JEPA 2~\citep{assran2025v} is applied only to the masked tokens, as Equation~\ref{eq: l1_main_loss} shows.
Therefore, the model has no incentive to encode local information within the context tokens and can instead devote this computation to aggregating global information to minimize $\mathcal{L_{\text{prediction}}}$, similarly to register tokens~\citep{darcet2023vision}.

\paragraph{\bf Context Self-Supervision.}
To verify this hypothesis, we propose to self-supervise both the mask and context patches and introduce a context loss $\mathcal{L}_{\text{ctx}}$, which is a weighted version of $\mathcal{L}_{\text{predict}}$ applied on the context tokens:
\begin{equation}
\mathcal{L_{\text{context}}} =  \frac{1}{|C|}  \sum_{i \in C} \lambda_i  \| P_{\phi}(E_{\theta}(x), \Delta_y)_i - \mathrm{sg}(E_{\overline{\theta}}(y)_{i}) \|_{1}  ,
\label{eq:l1_ctx_loss}
\end{equation}
where $C$ is the set of indexed context tokens, and $\lambda_i$ is a patch-specific weighting parameter described in the next section. The model is trained to minimize $\mathcal{L}_{\text{predict}} + \mathcal{L}_{\text{ctx}}$. Figure~\ref{fig:methods_1} shows that adding $\mathcal{L}_{\text{ctx}}$ has a significant effect on the learned feature maps. With the context loss, local structure now clearly appears in the feature maps, and similar semantic parts (e.g., head of the dogs, wheel of the car) are mapped to the same PCA components. Additionally, adding $\mathcal{L}_{\text{ctx}}$ significantly improves performance on dense-prediction tasks, achieving $33.9$ mIoU on ADE20K (up from $22.2$), and $0.473$ RMSE on NYUv2 (down from $0.682$). Hence, those results validate that by explicitly supervising context tokens, the model learns features that encode coherent local structure.

\subsection{V-JEPA 2.1: Improving Dense Video SSL Features}

Building on the previous observation, we introduce V-JEPA 2.1, a self-supervised training recipe for learning representations that combine high-quality dense local features with global semantic understanding.
Our key algorithmic innovations are  (1) Dense Prediction loss that applies self-supervision on both masked and unmasked tokens (Section \ref{sec:loss}) and, (2) Deep Self-Supervision of the encoder intermediate layers via a multi-level predictor (Section \ref{sec:predictor}).
Additionally, we explore (3) a Multi-Modal Tokenizer with modality-specific patch embeddings for images and videos (Section \ref{sec:tokenizer}); (4) Data Scaling through a more diverse and balanced image–video training distribution (Section \ref{sec:dataset}); and Model Scaling to ViT-G (Section \ref{sec:scaling}), enabling state-of-the-art downstream performance and effective distillation to smaller models (ViT-L, ViT-B, Section \ref{sec:results_distillation}).

\begin{figure}[t]
    \centering
    \includegraphics[width=\linewidth]{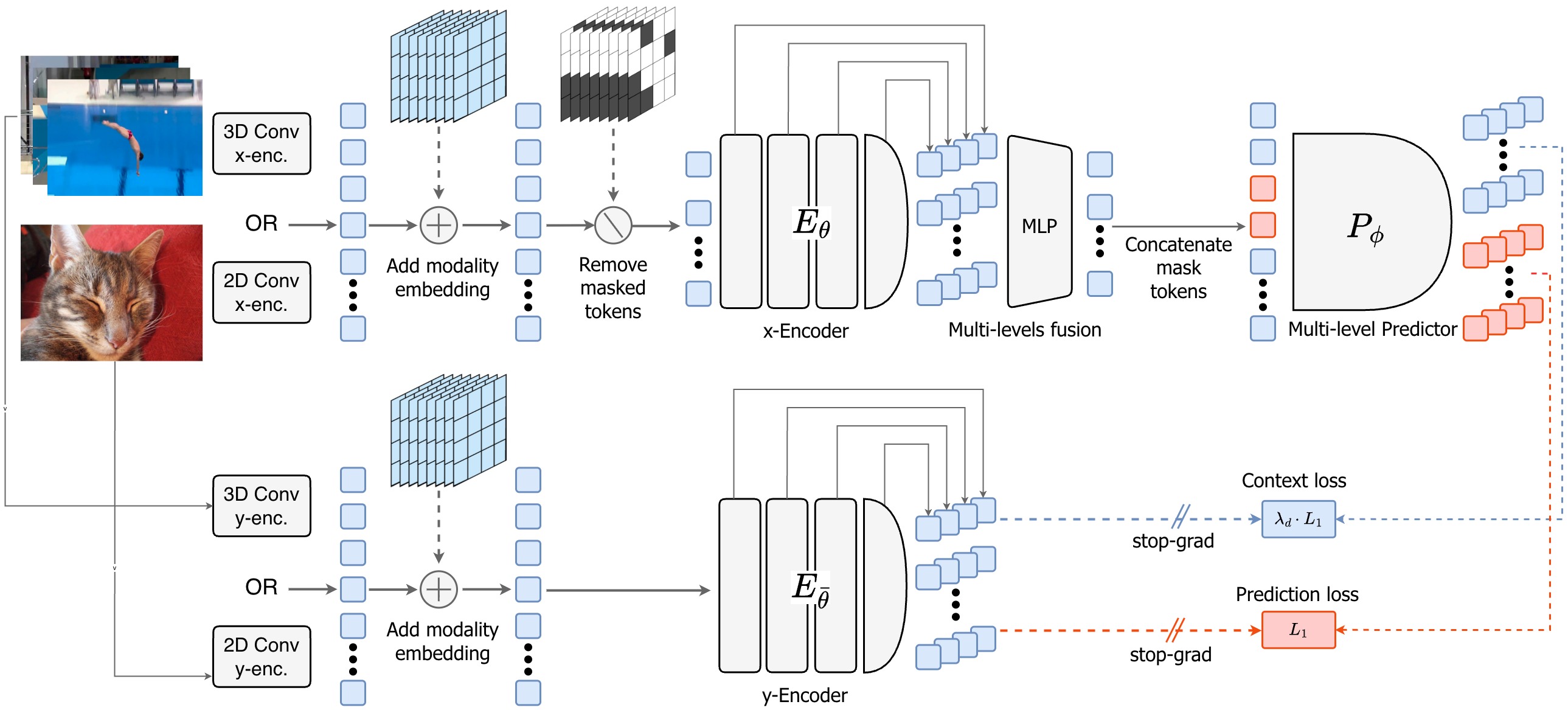}
    \caption{\textbf{V-JEPA 2.1 Detailed Architecture.} \normalfont
    %We first flatten the input—either a video or an image—into a sequence of into a sequence of $L$ tokens. For a video of $T$ frames and spatial resolution of $H \times W$, the number of tokens is $L = (T / \text{tubelet size}) \times (H / \text{patch size}) \times (W / \text{patch size})$, with a tubelet size of 2 and a patch size of 16. For images, $L = (H / \text{patch size}) \times (W / \text{patch size})$.
    Images and videos are processed by respectively either a 2D or 3D Convolutional patch embedding. Then, 3D Rotational Positional Encoding (RoPE) and learnable modality embedding are added. The $x-$encoder processes the visible tokens and outputs multi-level embeddings formed by concatenating the normalized output from intermediate encoder blocks. Then, a MLP fuses this multi-level representation and reduces its dimensionality. These context tokens are concatenated with learnable mask tokens carrying spatio-temporal positional information. The predictor processes the combined sequence and produces multi-level predictions for the masked tokens. 
    Training uses two different losses: (i) an L1 loss on masked-token predictions (the original V-JEPA objective), and (ii) a distance-weighted L1 loss on nearby context tokens, both supervised using the $y$-encoder multi-level outputs.}
    \label{fig:architecture}
\end{figure}

\paragraph{\bf VJEPA 2.1 Architecture.} 
We illustrate the V-JEPA 2.1 architecture in Figure~\ref{fig:architecture}.
An input, either an image or a video, is projected into a sequence of embedding vectors, or tokens, using a modality-specific patch embedding. Mask corruption is then applied to the sequence by randomly dropping patch tokens.
The $x$-encoder processes the remaining visible context tokens and outputs representations from multiple encoder levels in addition to the final output. The multi-level representations are then concatenated along the channel axis and fed to an MLP to reduce their dimensionality. Context tokens are concatenated, along the sequence axis, with learnable mask tokens that carry spatio-temporal positional information of the masked patches. The predictor processes the combined sequence and produces multi-level predictions for each token.
Training uses two different losses: (i) an L1 loss on masked-token predictions (the original V-JEPA objective), and (ii) a distance-weighted L1 loss for context tokens. Both use the $y$-encoder outputs as targets, which process the unmasked sequence of patches from the input images or videos.  Losses are applied to several intermediate representation levels in addition to the encoder output.

\paragraph{\bf Implementation Details.} We follow the warmup-constant learning rate schedule of V-JEPA 2 and we train models for 135{,}000 iterations. We maintain the teacher EMA coefficient and weight decay at fixed values. Each video sample is a clip of 16 frames at a resolution of $256 \times 256$, and each image sample has a resolution of $256 \times 256$. 
Additionally, in a second stage we explore the effect of applying a cool-down phase, i.e., decaying the learning rate and increasing the input images and videos resolution. We further train our models for 12{,}000 iterations during this cool-down phase, increasing the input resolution: video clips now have 64 frames at a resolution of $384 \times 384$, and images have a resolution of $512 \times 512$. 
Ablation results are reported after the first training phase, whereas final downstream tasks results use the full warmup–constant–cooldown schedule.
%Unless stated otherwise, we report performance after completing the warmup-constant learning rate schedule.
More details and all hyper-parameters are provided in Appendix~\ref{app:pretraining}.
%We use the VideoMix22M dataset~\citep{assran2025v} as our initial pretraining dataset, and we study the effect of increasing visual diversity by incorporating a large amount of image data with VisionMix163M (see Section~\ref{sec:dataset}). 

\begin{figure}[t]
\centering
    \includegraphics[width=0.80\linewidth]{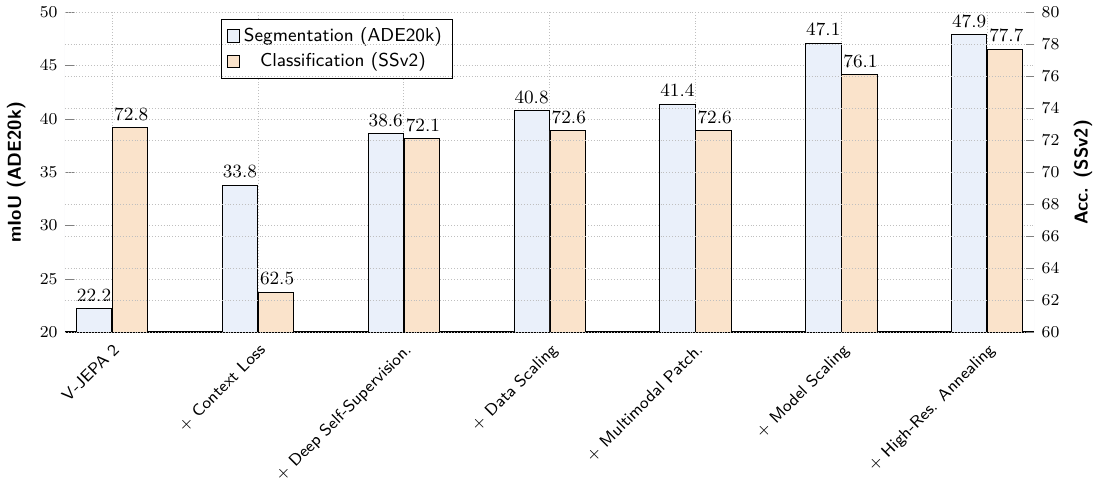}
    \captionof{figure}{\textbf{Impact of individual components of our novel V-JEPA 2.1 training recipe.} \normalfont The ablation is conducted starting from a ViT-L architecture, on single-image semantic segmentation on ADE20k, and action classification on SSv2. Introducing \textit{Weighted-Context Self-Supervision} with the context loss significantly improves segmentation, at the cost of classification. \textit{Deep Self-Supervision} allows to retrieve the performance, and further improving segmentation. Scaling the image data with VisionMix 163M dataset, combined with a \textit{Multi-Modal Tokenizer} further improve the results. Finally, the recipe \textit{scales} with model size and high resolution \textit{cool-down}.}
    %\nicolas{Update anneal to cooldown in the plot}
    \label{fig:ablation_figure}
\end{figure}

\paragraph{\bf Empirical Evaluation.}
To evaluate our design choice, we rely on a set of dense-vision tasks (ADE20K and NYUv2) using a linear probing evaluation protocol following \citet{simeoni2025dinov3} and global recognition tasks (Something-Somethingv2 for action recognition and ImageNet for object recognition) with an attentive probing protocol following~\citet{assran2025v}.
We ablate the effect of each architecture component in Figure~\ref{fig:ablation_figure} and Table~\ref{tab:ablation_table}. 
%Additionally, we show that our V-JEPA 2.1 can benefit from data scaling by leveraging a new dataset, VisionMix163M, and from model scaling, increasing model size from a ViT-L model size with 300 millions of parameters to ViT-G leveraging 2 billion parameters.
In the following, we describe each component in more detail, as well as their impact on downstream performance.

%We introduce three key architectural components that substantially improve dense feature quality while preserving the strong global understanding of V-JEPA 2: (1) a \textit{Context Loss} (Sec.~\ref{sec:loss}) that explicitly supervises context tokens and encourages coherent local structure, (2) a \textit{Multi-Level Predictor} (Sec.~\ref{sec:predictor}) that leverages intermediate encoder representations for multi-scale supervision, and (3) a \textit{Multi-Modal Tokenizer} (Sec.~\ref{sec:tokenizer}) that uses modality-specific patch embeddings for images and videos. Additionally, we increase data diversity with our novel (4) \textit{VisionMix163M dataset} (Sec.~\ref{sec:dataset}) , providing a curated and balance image video dataset. Finally, we (5) \textit{scale model size} by training a ViT-G (2B) model, which we distill into compact ViT-L and ViT-B variants.

%For evaluation, we rely on simple prediction heads to discriminate the model's behaviour. Dense evaluation includes depth estimation (NYUv2, KITTI), semantic segmentation (ADE20K, VOC12, Cityscapes), non-parametric label propagation for video object segmentation (DAVIS, YT-VOS), and short-term object-interaction anticipation (Ego4D). Global understanding is assessed through attentive-probe for action recognition (SSv2, Diving48, K400), image classification (ImageNet-1K), and action anticipation (EPIC-Kitchens). We also visualize PCA maps on images and videos representations.

\begin{figure}[t]
\centering
\begin{minipage}[t]{0.485\textwidth}
    \captionof{table}{\textbf{Impact of individual components of our novel V-JEPA 2.1 training recipe.} \normalfont Results on image classification (IN1K), video classification (SSv2), depth estimation (NYU), and semantic segmentation (ADE20K). Introducing the Context Loss improves dense tasks but reduces classification performance on SSv2. Incorporating our Deep Self-Supervision restores the classification performance. The VisionMix 163M dataset, the Multi-Modal Tokenizer, and scaling model size further improve the results.}
    \resizebox{\textwidth}{!}{
        \begin{tabular}{c c c c l}
        \makecell{IN1K \\ Acc.} & \makecell{SSv2 \\ Acc.} & \makecell{NYU \\RMSE $\downarrow$} & \makecell{ADE20K \\ mIoU} & \\ \hline
        82.2 & 72.8 & 0.682 & 22.2 & V-JEPA 2 \\
        \hdashline
        72.6 & 62.5 & 0.474 & 33.8 & + Context Loss \\
        80.8 & 72.1 & 0.463 & 38.6 & + Multi-level Pred. \\ 
        81.6 & 72.6 & 0.418 & 40.8 & + Vision Mix \\ 
        81.6 & 72.6 & 0.415 & 41.4 & + Multi-modal Tok. \\ 
        84.8 & 76.1 & 0.365 & 47.1 & + Model Scaling \\
        85.5 & 77.7 & 0.307 & 47.9 & + Cool-down. \\
        \end{tabular}
    }
    \label{tab:ablation_table}
\end{minipage}
\hfill
\begin{minipage}[t]{0.485\textwidth}
    \captionof{table}{\textbf{Impact of the context loss weighting scheme.} \normalfont Results on semantic segmentation (ADE20K) and video classification (SSv2). Introducing the context loss with a fixed coefficient has a significant positive impact on the segmentation performance, however at the cost of classification performance. Warmup on the coefficient allows to mitigate part of the lost performance. Our weighted scheme further improves the performance on both segmentation and classification.}
    \resizebox{\textwidth}{!}{
    \begin{tabular}{c c c c c}
        \makecell{ADE20K \\ mIoU} & \makecell{SSv2 \\ Acc.} & $\lambda$ & Scheme & Warmup \\
        \hline
         22.2& 72.8& V-JEPA 2, $\lambda = 0$ & cte & \\
         \hdashline
         26.4& 71.0& 0.05 & cte & \\
         29.6& 62.5& 0.2 & cte & \\ 
         27.5& 53.8& 0.5 & cte & \\ 
         24.6& 51.1 & 1.0 & cte  \\ 
         30.5& 60.5 & 0.2 & cte & \checkmark \\ 
         32.2& 61.5& 0.5 & cte & \checkmark \\ 
         33.8& 62.5& 0.5 & weighted & \checkmark \\
    \end{tabular}
    }
    \label{tab:weighting_table}
\end{minipage}
\end{figure}

\subsubsection{Dense Prediction Loss}
\label{sec:loss}

% \nicolas{Table~\ref{tab:weighting_table} does not crjjbirnfuishow really training unstability, more a performance trade-off, should we motivate $\lambda$ differently ?}
We propose applying our self-supervised loss to both masked and visible patches by minimizing $\mathcal{L_{\text{dense}}} = \mathcal{L}_{\text{predict}} + \mathcal{L}_{\text{ctx}}$, where $\mathcal{L}_{\text{predict}}$ is defined in Eq.~\ref{eq: l1_main_loss} and $\mathcal{L}_{\text{ctx}}$ is defined in Eq.~\ref{eq:l1_ctx_loss}. Naive application of $\mathcal{L}_{\text{ctx}}$ loss leads to poor performance on global semantic tasks, as the system can potentially find trivial solutions, such as copying the context features. 
We therefore explore various weighting coefficients $\lambda_i$ in Eq.~\ref{eq:l1_ctx_loss}. Table~\ref{tab:weighting_table} presents an ablation on various weighting schemes. First, we experiment with fixed values and set all $\lambda_i$ to a constant $\lambda$ and try values in the range $\lambda = [0.0, 0.05, 0.2, 0.5, 1.0]$. We observe that as we increase $\lambda$, the performance in semantic segmentation in the ADE20k dataset increases significantly up to certain point, but at the cost of the performance in action recognition in the SSv2 dataset decreases.  
We then introduce a progressive warm-up of $\lambda$ to restore action-recognition performance, with a schedule from epochs 50–100. We found empirically that it greatly stabilizes training.
Following, we introduce a dynamic weighting scheme where $\mathcal{L}_{\text{ctx}}$ for a given patch $i$ is weighted by the inverse square root of its minimum spatio-temporal distance to any masked token in the video sequence: i.e. setting 
\begin{equation}\lambda_i = \frac{\lambda}{\sqrt{\text{d}_{\text{min}} (i, M)}}\label{eq:lambda_i}\end{equation}
in $\mathcal{L}_{\text{ctx}}$ in Eq.~\ref{eq:l1_ctx_loss}, where $\text{d}_{\text{min}}$ is the distance, in number of blocks, between a context token and its closest mask token. This weighting emphasizes patches near masked regions by enforcing local continuity between masked and context areas, yielding a good trade-off between segmentation and action recognition performance.

Introducing our novel context loss $\mathcal{L}_{ctx}$ with our weighted scheme improves the performance on dense vision tasks (22.2 $\rightarrow$ 33.9 mIoU on ADE20k, 0.682 $\rightarrow$ 0.473 RMSE on NYUv2). Qualitatively, this loss smooths the feature maps by removing noisy artifacts, as shown in Figure~\ref{fig:methods_1}.
However, Table~\ref{tab:ablation_table} shows that there is still a degradation in video understanding (72.8 $\rightarrow$ 62.5 on SSv2) and image classification (82.2 $\rightarrow$ 72.6 on IN1K).

\subsubsection{Deep Self-Supervision}
\label{sec:predictor}

%In prior JEPA models \citep{assran2023self, bardes2024revisiting, assran2025v}, the predictor only infer the last-layer representation of the ViT encoder. 
%While effective for capturing semantic information, this design limits the model’s ability to leverage lower-level cues. CNNs naturally build multiscale representations through progressive down-sampling of the feature maps, where early layers encode fine-grained spatial detail and deeper layers encode abstract semantic content. Vision Transformers, however, lack this inductive hierarchy and rely on global self-attention, which does not inherently encourage multiscale feature organization. \textcolor{orange}{I do not know if this motivation of the multiscale predictor is correct. This idea was suggested with Yann, but I think that the motivation is not yet correctly expressed.}\nicolas{I don't know if we show lot results of hierarchical representation in the sense of representation that gets more abstract through hierchical layer. I would ground the description in 1) method 2) empirical benefit (stable training, better perf... ?)}

We self-supervise the encoder representation not only at the output but also at multiple intermediate levels. We first concatenate, along the channel dimension, the outputs of three intermediate $x$-encoder blocks, in addition to the output layer. Then, a lightweight MLP fuses these multi-level representations and reduces their dimensionality before feeding them into the predictor. The predictor processes the fused multi-level sequence of context and mask tokens and produces four outputs corresponding to the four encoder layers. Both the prediction loss and the context loss are then applied at each one of these four levels.

%Supervising predictions across multiple encoder levels encourages the model to preserve information at different spatial and semantic scales. Low-level cues originating in early layers of the encoder are propagated through the network, while high-level structure remains reinforced through the deeper layers. As a result, the last layers of encoder capture both fine local details and high level temporal semantics.

%Thus, in the case of an optimal predictor, the encoder must learn to capture as much information about the video as possible to minimize the deviation of the target. Now, this information is multi-scale, as both the first levels and the last levels are supervised.

Deep Self-Supervision leads to significant improvement on downstream performance for both global and dense tasks, as shows Figure~\ref{fig:ablation_figure}.
Furthermore, it allows local information to flow towards the final layers, effectively removing the need for intermediate layers in dense downstream tasks as we show in Appendix \ref{app:multiscale_evaluation}.
Deep Self-Supervision allows to recover the global understanding capabilities of V-JEPA 2 (72.0 on SSv2, 80.8 on IN1K) while improving on dense tasks with the context loss (38.6 mIoU on ADE20K, 0.463 RMSE on NYU).

\subsubsection{Scaling Image Data.}
\label{sec:dataset}

DINOv2 \citep{oquab2023dinov2} introduced a cluster-based retrieval strategy to select images from a large pool of raw internet data, resulting in a curated set of 142 million images, referred to as the LVD-142M dataset.
Using a similar approach, V-JEPA 2 \citep{assran2025v} collected and curated video scenes from YT1B videos \citep{zellers2022merlot}, combined with other publicly available video datasets, yielding a large-scale collection of 19 million video samples from the internet, corresponding to $1.6$ million hours. Both works demonstrated the positive effect of data scaling for SSL pretraining.

Building on these insights, we construct our VisionMix163M Dataset, combining large-scale curated sources from these two prior works. As we show in Table~\ref{tab: datasets}, we replace the 1M-image ImageNet subset from VJEPA-2 pretraining data with LVD-142M, providing a broader and more diverse appearance distribution.
Since this extensive image collection already covers many static visual concepts, we shift the video sampling strategy towards more dynamic, motion-rich content, increasing the SSv2 sampling weight from 0.056 to 0.170.
We also found beneficial to increase the contribution of YT-1B from 0.188 to 0.720, which contains much more heterogeneous video samples.

Rather than mixing images and videos within the same training batch, we leverage distributed training by assigning separate workers to each modality. After each iteration, gradients from video-only and image-only nodes are aggregated before updating the model.
The image/video ratio is controlled through per-modality batch sizes.
Empirically, we found optimal performance with 128 video clips (16 frames each) and 2,304 images per global batch. As we report in Table~\ref{tab:ablation_table}, the improvement of training on a more diverse and extend database is beneficial in all tasks (72.1 $\rightarrow$ 72.6 on SSv2, 80.8 $\rightarrow$ 81.6 on ImageNet, 38.6 $\rightarrow$ 40.8 on ADE20K, 0.463 $\rightarrow$ 0.418 RMSE on NYUv2).

\begin{table}[t]
\centering
\caption{\textbf{Comparison between VideoMix22M and VisionMix163M.} \normalfont We increase the visual diversity of the pretraining dataset by increasing the number of images, replacing ImageNet with LVD-142M, and update the YT-1B sampling weight to promote data-source with more visual diversity.}

\resizebox{\columnwidth}{!}{
\begin{tabular}{lccccc}

\toprule
\textbf{Source} & \textbf{Samples} & \textbf{Type} & \textbf{Total Hours} & \makecell{\textbf{V-JEPA2} \\ \textbf{weights}} &\makecell{\textbf{V-JEPA2.1} \\ \textbf{weights}} \\
\midrule
SSv2 \citep{goyal2017something}         & 168K  & EgoVideo & 168     & 0.056 & 0.170 \\
Kinetics \citep{carreira2019short}   & 733K  & ExoVideo & 614     & 0.188  & 0.010 \\
Howto100M \citep{miech2019howto100m}      & 1.1M  & ExoVideo & 134K    & 0.318& 0.100 \\
ImageNet \citep{deng2009imagenet}        & 1M    & Images   & n/a   & 0.250 & 0.\\
YT-1B \citep{zellers2022merlot} & 19M & ExoVideo & 1.6M    & 0.188  & 0.720 \\
\hdashline
LVD-142M \citep{oquab2023dinov2} & 142 M & Curated images & n/a &  & \\
\bottomrule
\label{tab: datasets}
\end{tabular}
}
\end{table}

\subsubsection{Multi-Modal Tokenizer}
\label{sec:tokenizer}

Previous work exploring joint training on image and video, such as V-JEPA 2, was suboptimal: it employed a single 3D convolution as the patch-embedding layer. Images were duplicated temporally and treated as a 16-frame static video, which significantly increased their computational cost and introduced an incorrect representational bias (i.e., images were interpreted as static videos).
Instead, we introduce a multi-modal tokenizer, which applies a 3D convolution of $16\times16\times2$ for processing videos and a 2D convolution of $16\times16$ for images.
We also add a modality-learnable token to both the encoder and the predictor inputs, which explicitly encodes whether the input comes from the image or video pathway. These learnable tokens condition the processing on the input modality, helping the model disentangle stronger static appearance cues in images and temporal motion information in videos.

This design allows each modality to be processed in its native form, eliminating the need for temporal duplication of images and improving computational efficiency. Additionally, we observe that using the Multi-Modal Tokenizer has a positive effect on dense-task performance: it improves mIoU on ADE20K from $40.8$ to $41.4$, while performance on action or object recognition remains stable.

\subsubsection{Scaling Model Size to ViT-G (2B) and High-Resolution Cool-Down}
\label{sec:scaling}

Next, we explore the effect of model scaling and high-resolution cooldown. Scaling the V-JEPA encoder from a ViT-L with 300 million parameters to a ViT-G with 2 billion parameters leads to significant improvements on all downstream tasks (72.6 $\rightarrow$ 76.1 on SSv2, 81.6 $\rightarrow$ 84.8 accuracy on ImageNet, 41.4 $\rightarrow$ 47.1 mIoU on ADE20K, 0.415 $\rightarrow$ 0.365 RMSE on NYUv2).

Additionally, introducing a cool-down phase for the learning rate, during which we increase both the spatial resolution of images (from $256\times256$ to $512\times512$) and the spatio-temporal resolution of videos (from 16 frames at $256\times256$ to 64 frames at $384\times384$), further improves performance on all tasks.
This achieves an accuracy of $77.7$ on SSv2, $85.5$ on ImageNet, an mIoU of $47.9$ on ADE20K, and an RMSE of $0.307$ on NYUv2, resulting in our best performing model. 
The benefits of a cool-down training phase are particularly strong in depth estimation (0.365 $\rightarrow0.307$ RMSE on NYUv2).

\subsubsection{Model Distillation}
\label{sec: distillation}

We scaled the size of the model not only to achieve top-tier performance but also to enable effective compression of the model through distillation \cite{hinton2015distilling}. We distill our ViT-G (2B) model into smaller variants (ViT-B, 80M; ViT-L, 300M). The distillation protocol is adapted from our pretraining recipe, with key differences: i) we replace the Exponential-Moving-Average (EMA) of the target encoder by a frozen teacher model, ii) we keep an EMA copy of the student encoder, that is not used in the loss, but serves as the final model, iii) the distillation loss is identical to our pretraining loss, except it is only computed on the last layer of the teacher encoder, and it does not use deep self-supervision, iv) we use a predictor with only 12 blocks and a final linear layer matching the teacher embedding dimension. All other hyper-parameters—including masking ratios, cool-down schedules, learning rates, and data augmentations—remain identical to the original pretraining recipe. We provide more details on the distillation protocol in Appendix~\ref{app:distillation}.

%% file: sections/results.tex
\section{Results} \label{section:results}

In this section, we evaluate the performance of V-JEPA 2.1 on a large variety of downstream tasks. Throughout the experiments, we employ V-JEPA 2.1 as a \textit{frozen encoder}, demonstrating the versatility of its features.
We first demonstrate the predictive capabilities of V-JEPA 2.1 in two forecasting tasks: short-term object interaction anticipation (Section~\ref{sec:short_term_anticipation}) and action anticipation (Section~\ref{sec:action_anticipation}).
We then demonstrate that can leverage V-JEPA 2.1 to learn an action-condition world model and performs robot manipulation (Section~\ref{sec:robotic_arm_planning}) and navigation tasks (Section~\ref{sec:navigation_planning}) in zero-shot setup.
Then, we evaluate the quality of the learned dense features by assessing the V-JEPA 2.1 performance on single-image depth estimation and semantic segmentation (Section~\ref{sec:depth_segmentation}).
Following, we analyze the temporal consistency of V-JEPA 2.1 representations through the video object segmentation task (Section~\ref{sec:depth_segmentation}).
We also analyze V-JEPA 2.1 global understanding on two high-level understanding tasks: probe-based video classification and image classification (Section~\ref{sec:probe_classification}).
Finally, we present results for our smaller distilled models (Section~\ref{sec:results_distillation}). 
We provide more details on the experimental setup and all pretraining hyper-parameters in Appendices~\ref{app:pretraining} and~\ref{app:evaluation}.

%We then evaluate the dense features learned by V-JEPA 2.1 by assessing their performance on image depth estimation and semantic segmentation (Section~\ref{sec:depth_segmentation}), video object segmentation (Section~\ref{sec:video_segmentation}), and . 
%We then show the quality of the global features on higher-level understanding tasks, including probe-based classification (Section~\ref{sec:probe_classification}) and action anticipation . 

\subsection{Short-Term Object Interaction Anticipation} \label{sec:short_term_anticipation}

\begin{figure}[t]
    \centering
    \includegraphics[width=\linewidth]{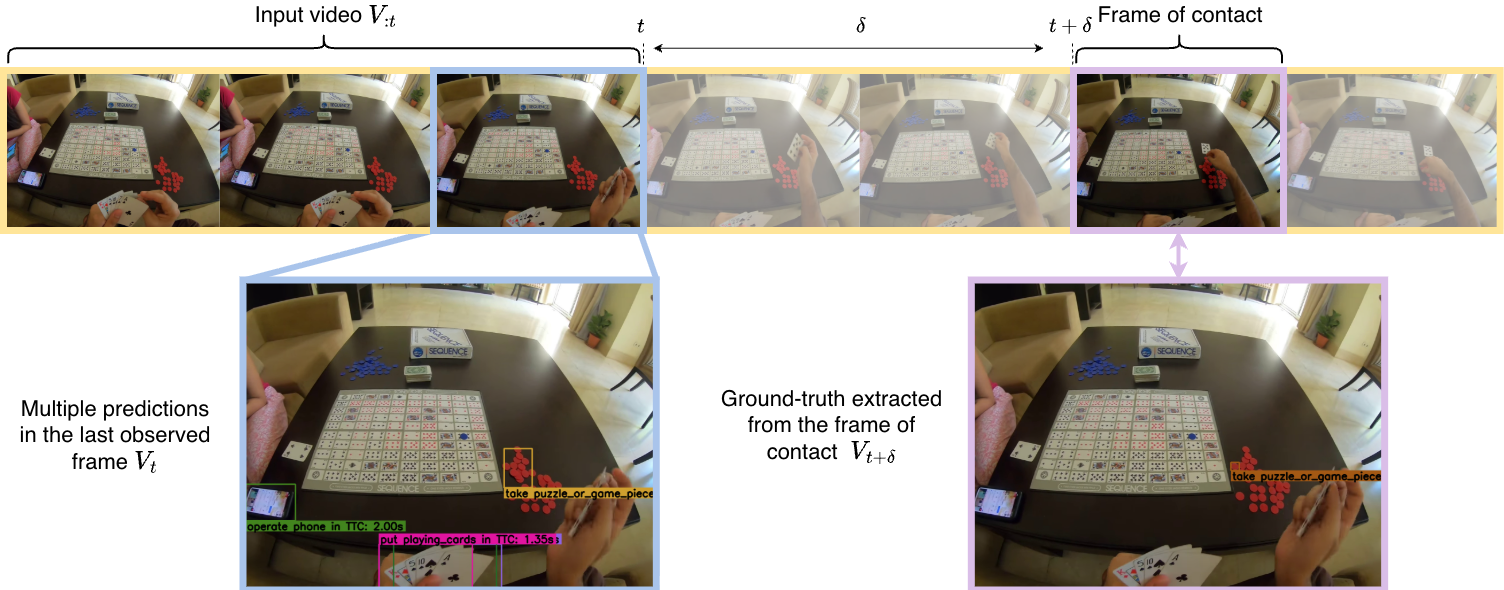}
    \caption{\textbf{Short-Term Anticipation task.} \normalfont Given a video observed up to time $t$ (denoted $V_{:t}$), the model predicts a set of future object--interaction events. Each prediction includes: (i) a bounding box localizing the object that will be interacted with, (ii) the noun class of that object, (iii) a verb class describing the upcoming interaction, and (iv) the anticipation time $\delta$, i.e., the number of seconds between the current frame and the start of the interaction. Ground-truth labels are derived from the contact frame $V_{t+\delta}$, but are expressed in the coordinate frame of the last observed frame $V_t$.}

    \label{fig:enter-label}
\end{figure}

\begin{table}[t]
\centering
\caption{\textbf{Short-Term Object Interaction Anticipation on Ego4D.} \normalfont Following \cite{grauman2022ego4d}, we report different Top-5 Average Precision (AP) and mean Average Precision (mAP) metrics, where the winning score is the mAP All. We compare results reported in the literature with different frozen SSL encoders (DINOv2, DINOv2, V-JEPA 2 and our V-JEPA 2.1) extended with an attentive-based predicted head. V-JEPA 2.1 obtains the state-of-the-art due to its high-quality dense features and predictive capabilities.}
\footnotesize
\resizebox{\columnwidth}{!}{%
\begin{tabular}{lcccccccccccc}
%\toprule
\bf{Model} &
\makecell{AP\\ b} &
\makecell{AP\\ b+N} &
\makecell{AP \\ b+V} &
\makecell{AP \\ b+$\delta$} &
\makecell{AP\\ b+N+V} &
\makecell{AP\\ b+$\delta$+N} &
\makecell{AP\\ b+$\delta$+V} &
\makecell{AP\\ All} &
\makecell{mAP\\ N} &
\makecell{mAP\\ N+V} &
\makecell{mAP\\ N+$\delta$} &
\makecell{mAP\\ All} \\
\midrule
\multicolumn{13}{l}{\textit{Results Reported in the Literature}}\\
StillFast~\citep{ragusa2023stillfast} &  - & -& -& -& -& -& -&-& 20.2& 10.3& 7.17 & 3.96 \\
STAformer~\citep{mur2024aff} &  38.3& 28.3& 16.6& 12.27& 12.8& 8.89& 5.47&4.06& 29.4& 15.3& 9.94 & 5.67 \\
GANO~\citep{thakur2023guided} & 45.3& 27.0& 12.2& 16.6& 6.54& 9.0& 4.18& 2.47& -& -& -&-\\
\midrule
\multicolumn{13}{l}{\textit{Encoders using our same protocol}}\\
%DINOv2 ViT-L & Img. & 46.8 & 29.4 & 15.73 & 9.10 & 5.06 \\
DINOv2 ViT-L~\citep{oquab2023dinov2} & 45.1& 37.8& 21.9& 14.4& 17.6& 10.6& 7.27&5.25& 28.3& 15.2& 9.22& 5.25\\
%DINOv2 ViT-g & Img. & \bf{49.4} & \bf{31.6} & \bf{17.23} & 9.88 & 5.50 \\
DINOv2 ViT-g~\citep{oquab2023dinov2} &  47.8 & 38.1& 23.6& 14.1& 19.1& 10.9& 7.29&5.73&  30.6&  16.8&  9.37&  5.44\\
%DINOv3 ViT-H+ & Img. & 48.15& 32.44& 16.70& 9.55 & 5.20 \\
DINOv3 ViT-H+~\citep{simeoni2025dinov3} &  48.9& 40.2& 23.4& 13.6& 19.4& 10.8& 6.71&5.45&  32.4&  17.2& 9.53&  5.20\\
%DINOv3 ViT-7B & Img. &  &  &  &  &  \\
DINOv3 ViT-7B~\citep{simeoni2025dinov3} & 50.2& \bf{40.9}& 23.8& 14.2& 19.8& 11.5& 7.33&5.82& \bf{33.8}& \bf{17.4}& 10.3 & 5.68 \\
\midrule
%vJEPA 2 ViT-g\textsubscript{384}& Img & TR. & TR & TR & TR &\\
V-JEPA 2 ViT-g~\citep{assran2025v} & 45.7& 34.4& 22.2& 16.7& 16.9& 12.1& 8.56&6.26& 27.2& 14.9& 10.4& 6.02 \\
\hdashline
%vJEPA 2.1 ViT-L (dist.) &   & & & & & & &&  &  &  &  \\
%vJEPA 2.1 ViT-L &   & & & & & & &&  &  &  &  \\
%vJEPA 2.1 ViT-g\textsubscript{384} & 45.8 & 25.6 & 14.01 & 7.82 & 4.60 \\
V-JEPA 2.1 ViT-g & 47.3  & 36.5& 23.6& 18.1& 18.2& 13.5& 9.55&7.21& 28.7& 15.5& 11.4& 6.75 \\
%\bf{Model} & \bf{AP b}  & AP b + N& AP b + V& AP b + ttc & AP b + N + V & AP b + ttc + N & AP b + ttc + V & AP b + ttc + V + N& \bf{mAP N} & \bf{mAP N + V} & \bf{mAP N + ttc} &\bf{Overall} \\
V-JEPA 2.1 ViT-G & \bf{50.7}  & 39.3& \bf{25.8}& \bf{20.2}& \bf{20.3}& \bf{15.1}& \bf{10.8}&\bf{8.20}& 30.9 & 17.2 & \bf{12.8}& \bf{7.71} \\
\bottomrule
\end{tabular}
}

\label{tab:staformer_results_extended}
\end{table}

Short-Term object-interaction Anticipation (STA) consists in predicting future object interaction in a ego-centric scenario, predicting the next active object bounding box $b$, the object noun category $N$ the action verb $V$ and the time until contact ($\delta$).
This formulation evaluates fine-grained 2D localization, semantic understanding of objects, temporal reasoning over video dynamics, and the ability to forecast user intentions.
Using the Ego-4D STA benchmark~\citep{grauman2022ego4d}, we show that V-JEPA 2.1 outperforms previous state-of-the-art performance by a significant margin due to its high-quality dense features and predictive capabilities.

%The ability to anticipate near-future states is a key requirement for world models, as it supports robust understanding and planning in dynamic environments.
%While the previous section focused on evaluating V-JEPA 2.1 global predictive skills with the Epic Kitchens Action Anticipation benchmark, in this section we asses its dense forecasting capabilities on the the Short-Term object–interaction Anticipation (STA) \citep{grauman2022ego4d} task, which consists in the simultaneous prediction of the next active object bounding box $b$, the object noun category $N$ the action verb $V$ and the remaining time until contact ($\delta$).

\paragraph{\bf{Task.}} 
We evaluate V-JEPA 2.1 on the STA v2 split of Ego4D \citep{grauman2022ego4d}, which comprises 243 hours of annotated video clips spanning 128 noun and 81 verb categories, with a total of 98,276 training and 47,395 validation samples.
We compare against state-of-the-art methods \citep{ragusa2023stillfast, mur2024aff, thakur2023guided}, and we further asses the performance of DINOv2 \citep{darcet2023vision} and DINOv3 \citep{simeoni2025dinov3} image encoders using our proposed attentive probe.
Following the evaluation protocol of \citet{grauman2022ego4d}, we report Top-5 Average Precision (AP) and Top-5 mean Average Precision (mAP) metrics. As in standard mAP, predictions are matched to ground-truth boxes using IoU > 0.5 and additional class-specific criteria. For instance, the Top-5 mAP All variant requires a correct noun, correct verb, IoU above 0.5, and a time-to-contact $\delta$ prediction within a 0.25-second tolerance. To address the inherently multi-modal nature of future anticipation—where multiple plausible next-active objects may exist—the metric discounts up to four highest-scoring false positives per example, thereby avoiding penalties for plausible but unannotated predictions.
Additionally, we report individual metrics to evaluate the temporal ($\delta$), spatial (bounding boxes), and semantic (noun and verb) dimensions of the task.

\begin{figure}[h]
    \centering
    \setlength{\tabcolsep}{0.5pt}
    \renewcommand{\arraystretch}{1.2}
    % \resizebox{\columnwidth}{!}{%
    \begin{tabularx}{\textwidth}{c c c c}
        \textbf{V-JEPA 2.1 Prediction} & \textbf{Ground-truth} & \textbf{V-JEPA 2.1 Prediction} & \textbf{Ground-truth} \\
        \includegraphics[width=0.247\linewidth]{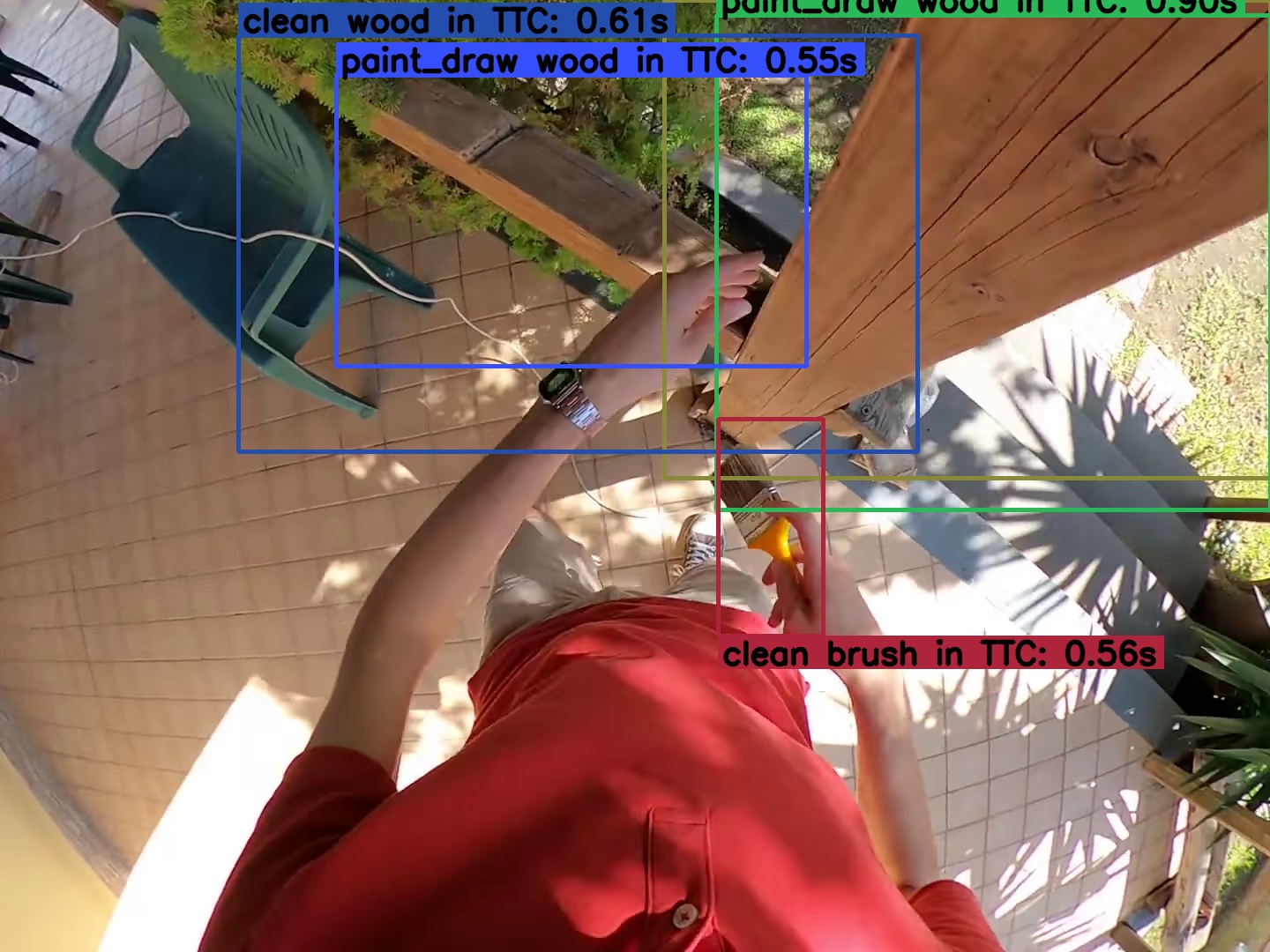} &
         \includegraphics[width=0.247\linewidth]{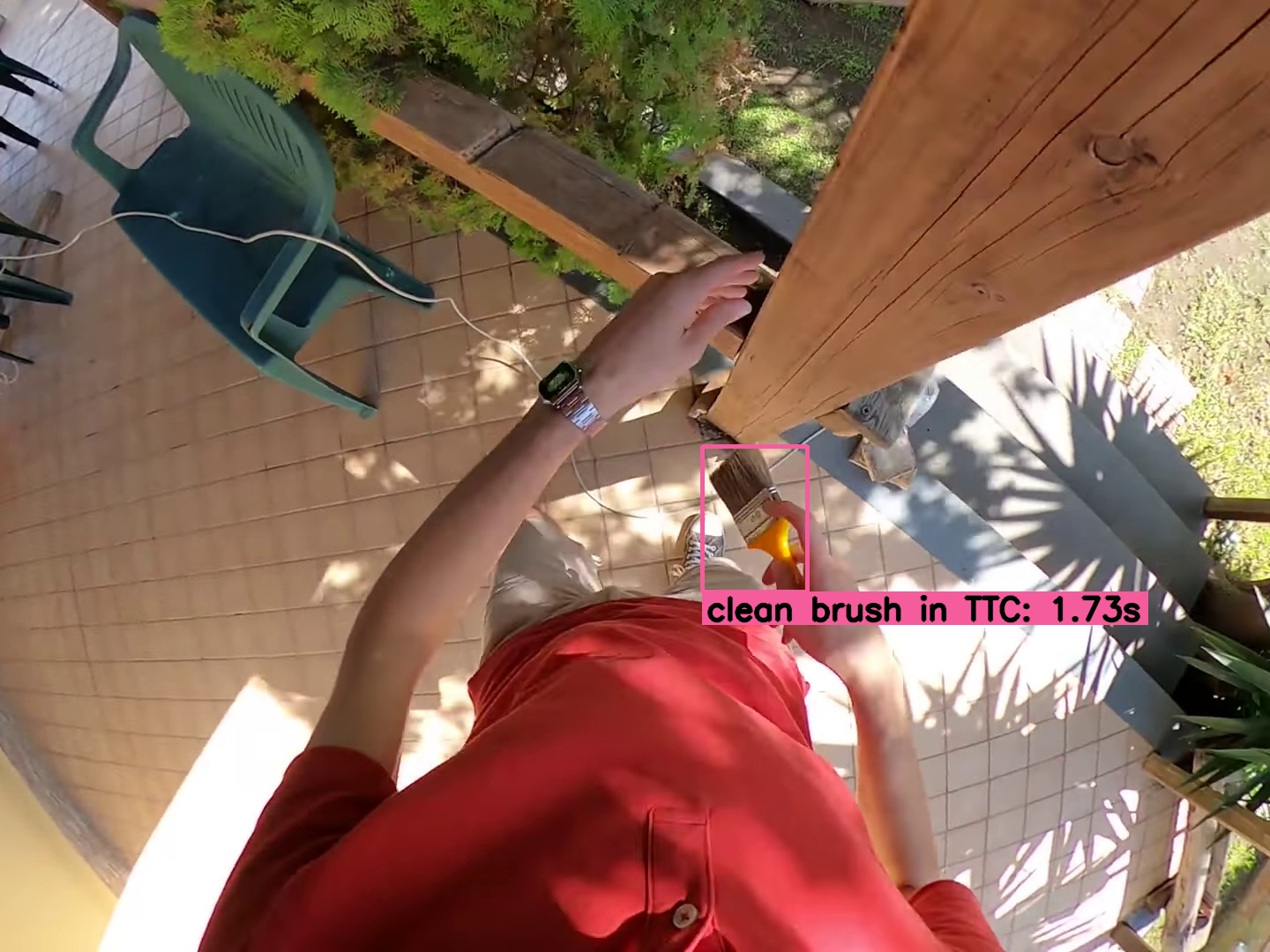} &
        \includegraphics[width=0.247\linewidth]{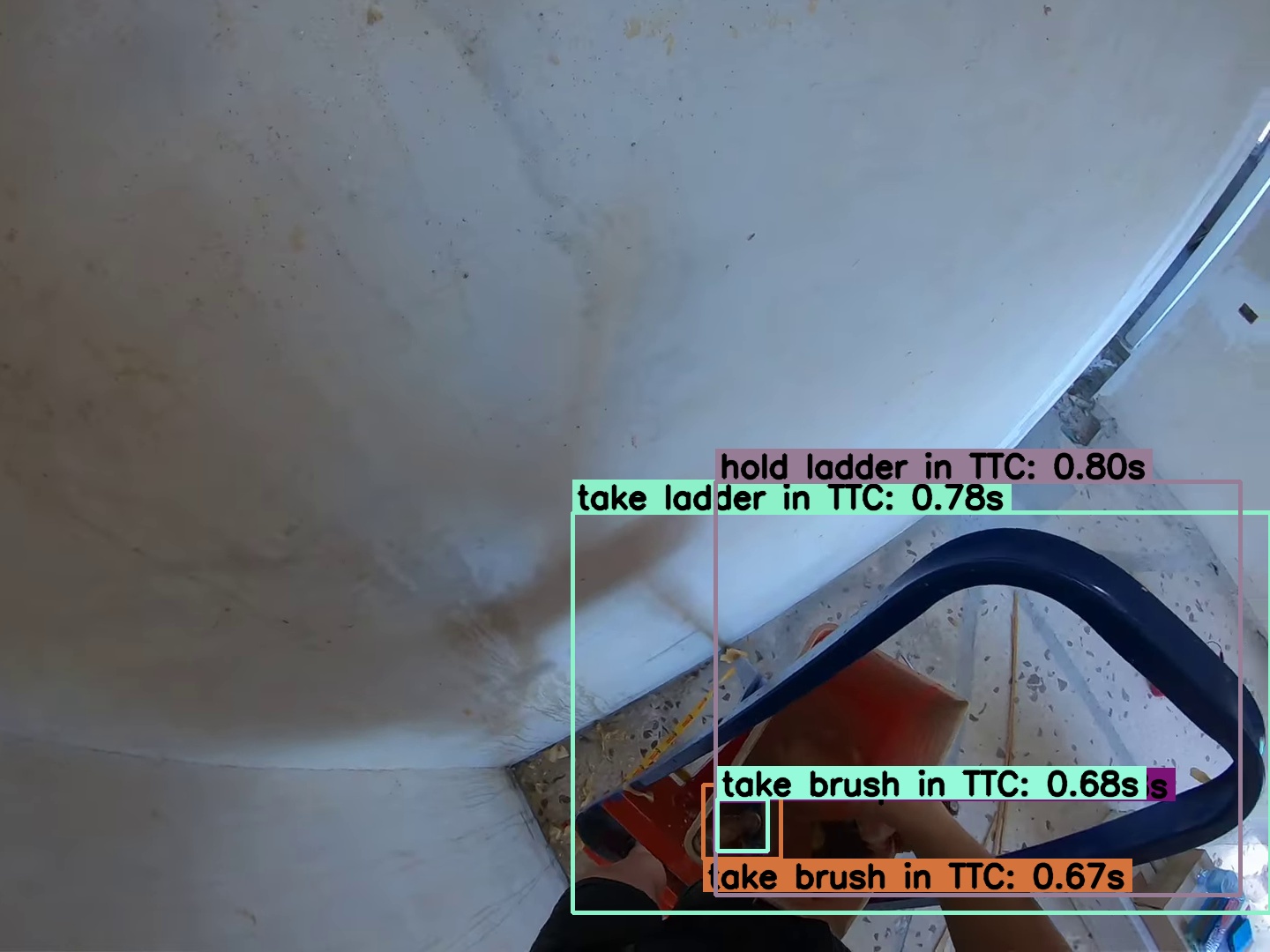} &
        \includegraphics[width=0.247\linewidth]{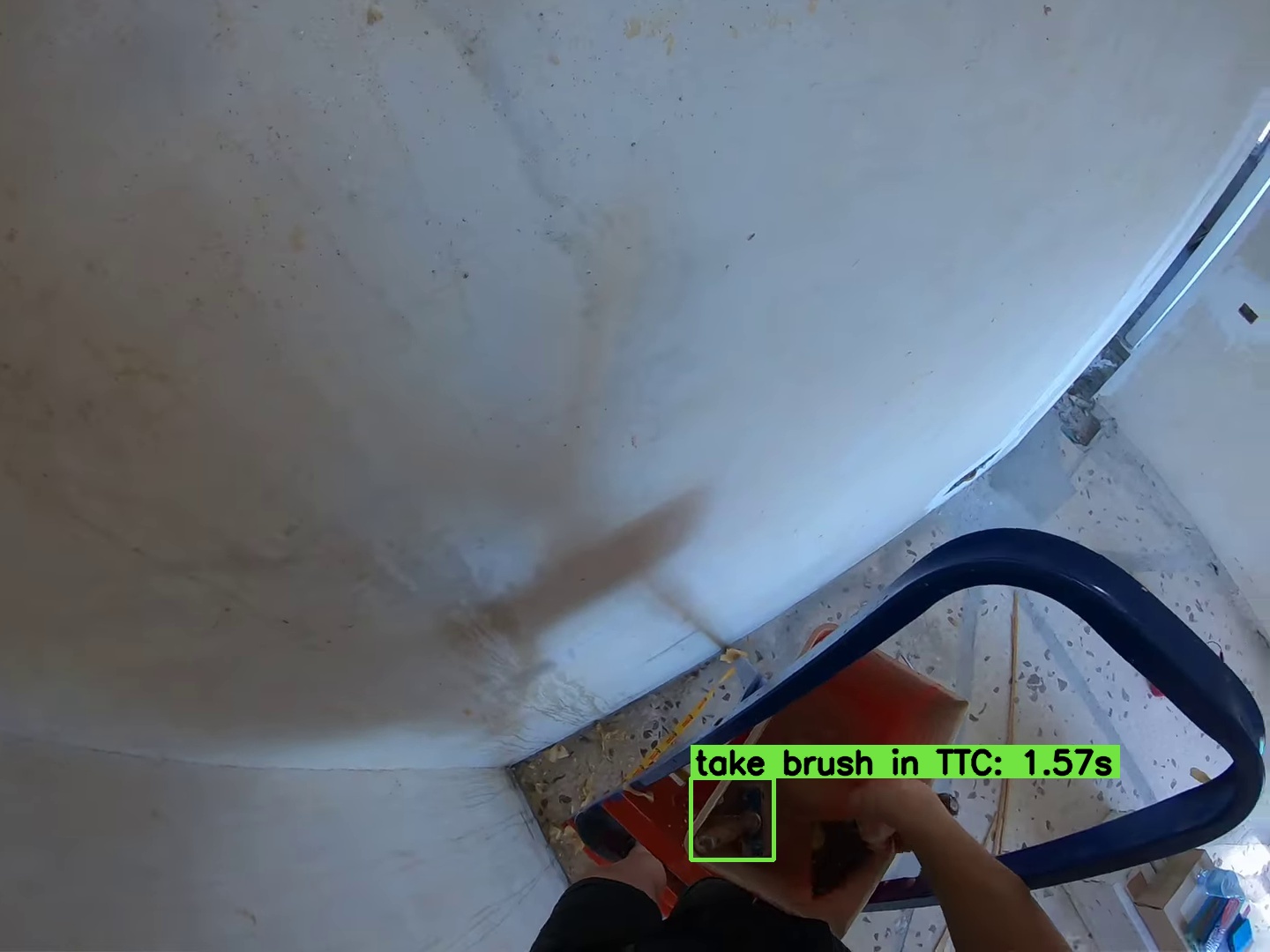} \\

        \includegraphics[width=0.247\linewidth]{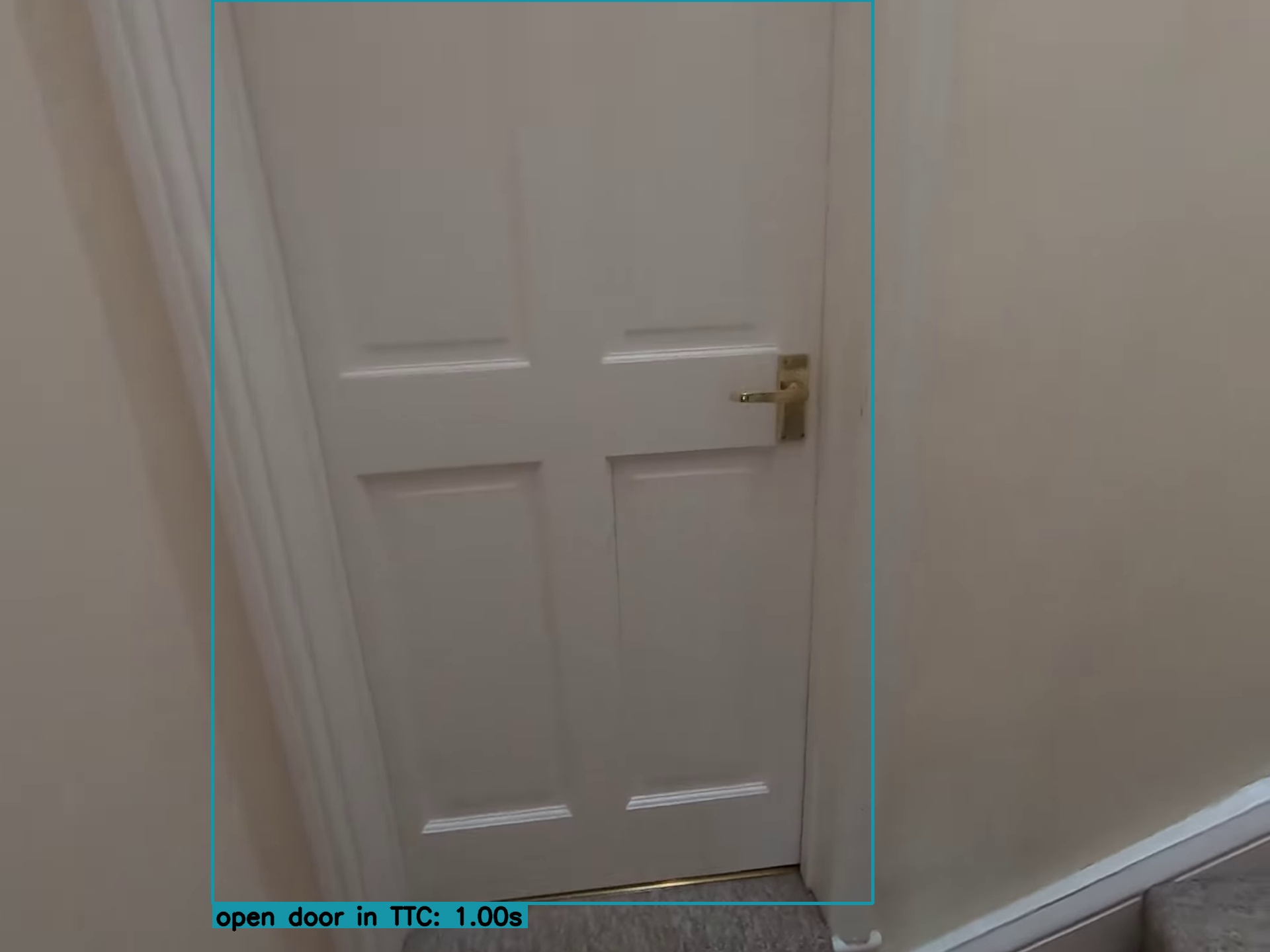} &
        \includegraphics[width=0.247\linewidth]{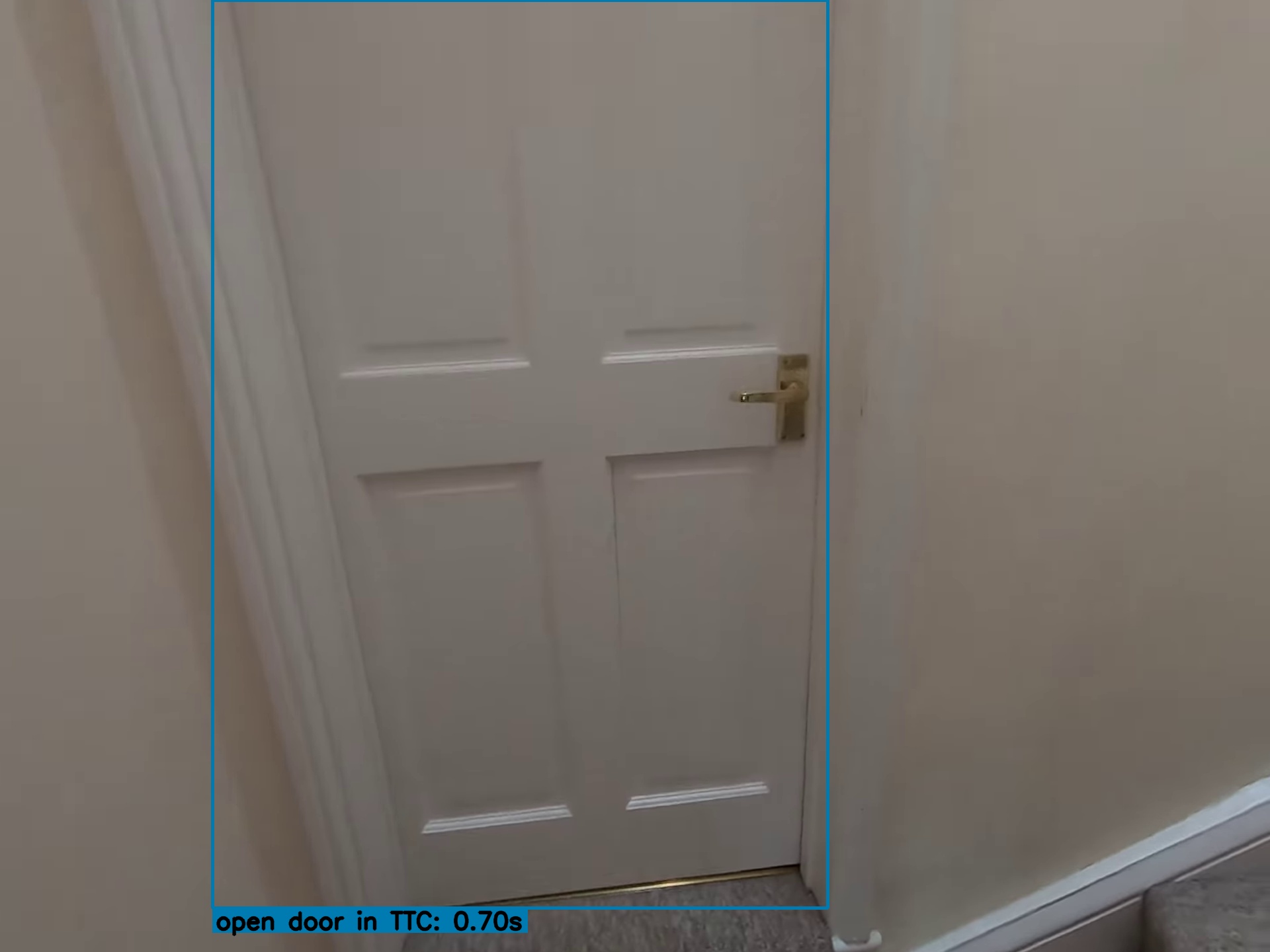} &
        \includegraphics[width=0.247\linewidth]{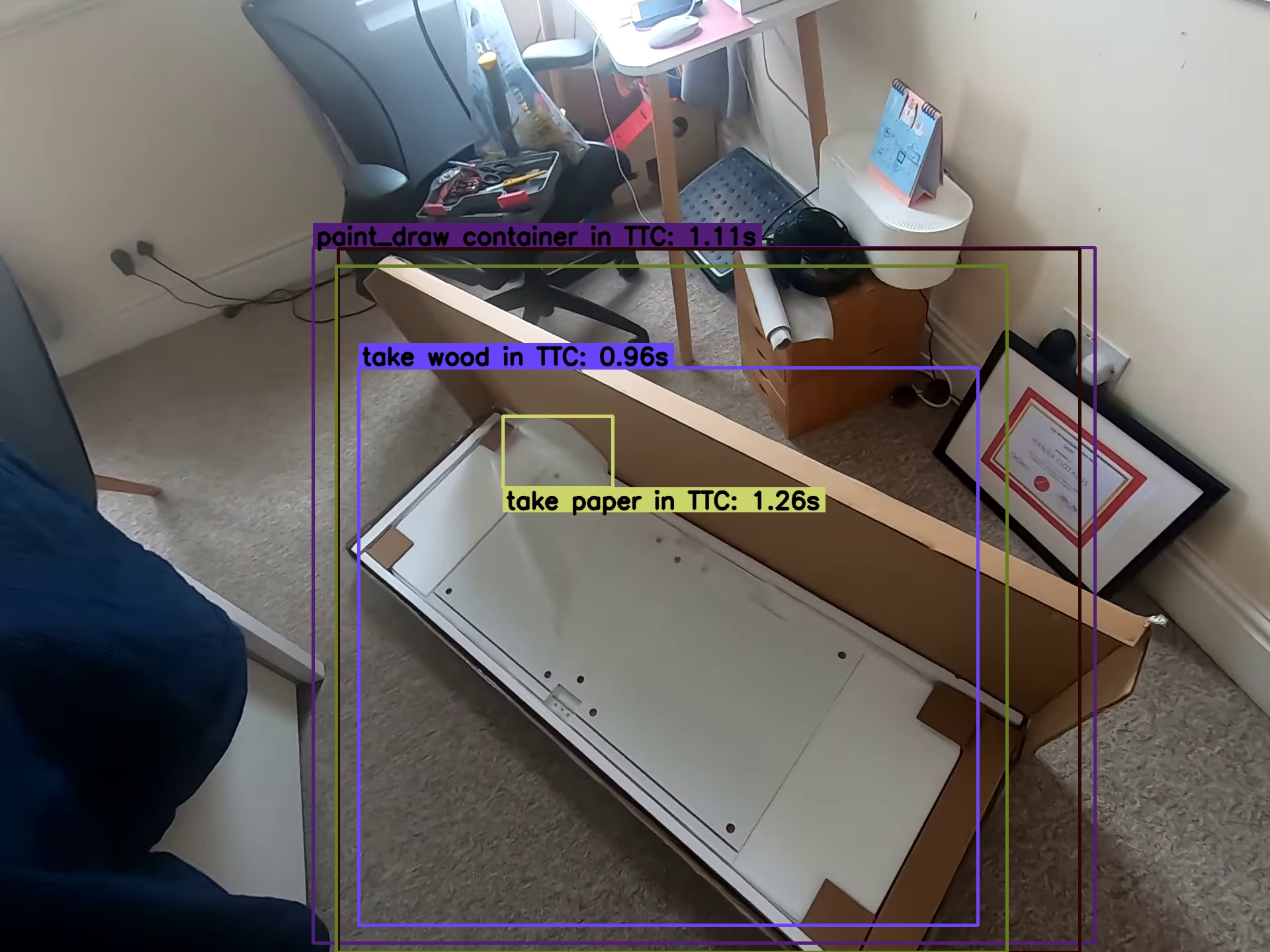} &
        \includegraphics[width=0.247\linewidth]{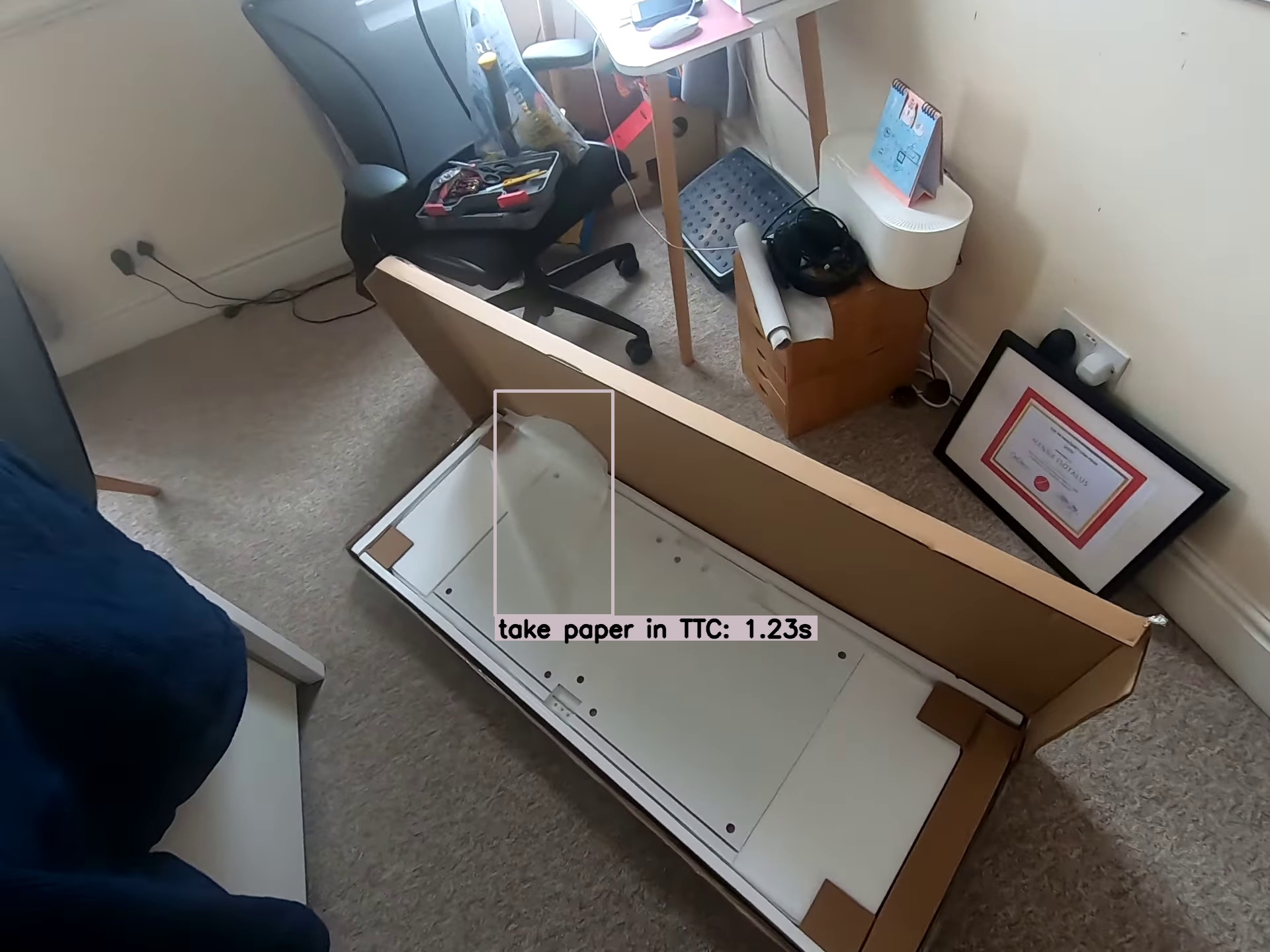} \\
    \end{tabularx}
    \caption{\textbf{Short-Term Anticipation (STA) qualitative results.} \normalfont We compare our model predictions with the ground-truth data. V-JEPA 2.1 predicts multiple scenarios, where the predicted object and verb classes correspond with plausible human-object interactions.}
    \label{fig:sta_qualitative}
\end{figure}

\paragraph{\bf{Evaluation protocol.}} 
The model receives as input a low-resolution video clip (384 px, 8 frames spanning 0.5 seconds) and a high-resolution image (1080 px) corresponding to the last frame of the clip. Using the frozen encoder of V-JEPA 2.1, we extract last-level features independently from both inputs using the respective patchifier (a 2D convolution for the high-resolution image and a 3D convolution for the video) and adding the corresponding modality embedding.
For the video features, we train a four-layer attentive probe, followed by the frame-guided temporal pooling module from \cite{mur2024aff}. This module aggregates the 3D video tokens into a 2D representation that is spatially aligned with the spatial reference of the last frame.
The pooled video features are then summed to the image features, and the resulting fused representation is rescaled into four multiscale feature maps, which serve as input to the detection head. 
To ensure a fair comparison with state-of-the-art methods \citep{ragusa2023stillfast, mur2024aff, thakur2023guided}, we adopt the same detection head \citep{ragusa2023stillfast} utilized in these approaches. This head extends Faster R-CNN by adding linear layers to additionally predict the verb category and the time to contact.

\paragraph{\bf{Results.}} 
Table~\ref{tab:staformer_results_extended} presents a comparison with state-of-the-art methods on the Short Term Anticipation task. V-JEPA 2.1 demonstrates achieves the absolute state-of-the-art performance with an overall mAP of 7.71. This represents a relative improvement of approximately 35$\%$ over the previous best method \citep{mur2024aff}, which introduces task-specific training components.
As it is shown by the individual metrics, this gain is primarily driven by improved understanding of the next action 25.8 AP\textsubscript{b+V} and the time to contact 20.20 AP\textsubscript{b+$\delta$}, 12.9 mAP \textsubscript{b+$\delta$}. Furthermore, the high-quality V-JEPA 2.1 dense features enable the precise 2D detection of the next-active object bounding boxes, achieving 50.7 AP and surpassing DINOv3 ViT-7B in this category.
Qualitative results are shown in Figure~\ref{fig:sta_qualitative}. V-JEPA 2.1 predicts plausible short-term interactions, with accurate bounding box localization, robust understanding of object categories, and plausible inference of the user next action intentions.

\subsection{Action Anticipation} \label{sec:action_anticipation}

Action anticipation consists in predicting the future action given a contextual video clip leading up to some time before the action. Using the Epic-Kitchens-100 (EK100) benchmark~\citep{Damen2022rescaling}, we show that V-JEPA 2.1 outperforms the action anticipation performance of V-JEPA 2, and sets a new state-of-the-art for the task.

\paragraph{\bf{Task.}} The EK100 dataset is comprised of 100 hours of cooking activities recorded from an egocentric perspective across 45 kitchen environments. Each video in EK100 is annotated with action segments, which include a start timestamp, an end timestamp, and an action label. There are 3,568 unique action labels, each consisting of a verb and a noun category, with a total of 97 verb categories and 300 noun categories. The EK100 action anticipation task involves predicting noun, verb, and action (i.e., predicting verb and noun jointly) from a video clip, referred to as context, that occurs before the start timestamp of an action segment.
The interval between the end of the context and the beginning of the action segment is the anticipation time, which is set to 1 second by default. Given that different future actions may be possible from a given context, mean-class recall-at-5 is used as the metric to measure performance~\citep{Damen2022rescaling}.

\paragraph{\bf{Evaluation protocol.}} We employ the same protocol as V-JEPA 2 and use an attentive probe trained on top of the frozen V-JEPA 2.1 encoder and predictor to anticipate future actions. Specifically, we sample a video clip that ends 1 second before an action starts. This video context is fed to the encoder. The predictor takes the encoder representation, along with the mask tokens corresponding to the frame 1 second into the future, and predicts the representation of the future video frame. The outputs of the predictor and encoder are concatenated along the token dimension and fed to an attentive probe with a similar architecture to those used in our probe-based video classification protocol, with the difference being that the anticipation probe’s final cross-attention layer learns three query tokens (as opposed to one), and each query output is fed to a different linear classifier to predict the action category, the verb category, and the noun category respectively. A focal loss~\citep{lin2017focal} is applied to each classifier independently and then summed before back-propagating through the shared attention blocks of the probe.

\begin{table}[t]
  \centering
    % \small
    \caption{\textbf{Action Anticipation on EK100.} \normalfont 
    Comparison with the state-of-the-art on the EK100 Action Anticipation benchmark. We report mean-class recall-at-5 for verb, noun and action on the validation set of EK100. V-JEPA 2.1 offers comparable performance to V-JEPA 2 for the ViT-g 1B model, but shows improved performance for the ViT-G 2B model, setting a new state-of-the-art for the task.}
    \label{tb:prediction_main}
    \begin{tabular}{l c c c c}
      \bf Method & \bf Param. & \multicolumn{3}{c}{\bf Action Anticipation}  \\
      & & \small Verb & \small Noun & \small Action \\
        \toprule
        InAViT~{\small\citep{roy2024interaction}} & 160M & 51.9 & 52.0 & 25.8 \\
        Video-LLaMA~{\small\citep{zhang2023video}}  & 7B & 52.9 & 52.0 & 26.0 \\
        PlausiVL~{\small\citep{mittal2024can}}  & 8B & 55.6 & 54.2 & 27.6 \\
        \midrule
        % \multicolumn{5}{l}{\bf\it Frozen Backbone}\\[1ex]
        V-JEPA 2 ViT-g {\small\citep{assran2025v}} & 1B & 63.6 & 57.1 & 39.7 \\
        \hdashline
        V-JEPA 2.1 ViT-g & 1B & 63.6 & 56.2 & 38.4 \\
        V-JEPA 2.1 ViT-G & 2B & \bf 64.3 & \bf 59.9 & \bf 40.8 \\
\end{tabular}
\end{table}

\paragraph{\bf{Results.}} Table~\ref{tb:prediction_main} summarizes the results on the validation set of the EK100 action anticipation benchmark.  We compare V-JEPA 2.1 ViT-g 1B and ViT-G 2B encoders with V-JEPA 2 ViT-g 1B encoder. Both leverage 32 frames with 8 frames per second at resolution 384 × 384 as video context. V-JEPA 2.1 is comparable to V-JEPA 2 on the same 1B model size, but show scaling of the performance to the 2B models size, setting a new absolute state-of-the-art performance at 40.8 Action Recall@5, corresponding to a $+2.8\%$ improvement.

\subsection{Robotic Arm Planning}
\label{sec:robotic_arm_planning}
Next, we evaluate V-JEPA 2.1 for robot manipulation.
To that end, we train a frame-causal action-conditioned predictor, following the protocol of~\citet{assran2025v}.
In particular, we use the public codebase from~\citet{assran2025v} and modify it to compute features using our V-JEPA 2.1 video encoder; the predictor is an identical 24-layer transformer network containing approximately 300M parameters, and is trained on the Droid raw dataset~\citep{khazatsky2024droid} using both a teacher-forcing and two-step rollout loss.
The model is then deployed in our lab, zero-shot, on a table-top Franka Panda robot arm with a parallel-jaw gripper, and evaluated on reach, grasp, and pick-and-place tasks with visual goal specification via model-predictive control.
We use the same exact task configuration files provided in~\citet{assran2025v}, and similar planning hyper-parameters.
\begin{table}[b]
    \caption{
    {\bf Planning Performance.} 
    \normalfont
    Comparing closed-loop robot manipulation of a cup using MPC with VJEPA 2 versus VJEPA 2.1.
    In both cases we leverage the cross-entropy method~\citep{rubinstein1997optimization} for optimizing the sequence of actions using a single A100 GPU.
    For each robot skill, we evaluate each model across 10 tasks and average the results.
    The improved depth understanding and dense features of VJEPA 2.1 lead to a 20\% improvement in the success rate of Grasp. Moreover, qualitatively, we find that any failures on Pick-and-Place and Grasp are a result of poor planning over gripper actions, as opposed to failures in spatial understanding (e.g., closing the gripper too soon, such that you cannot grasp the object, or slightly opening the gripper while in transit leading to the object being dropped.
    }
    \label{tb:robot}
    \centering
    \begin{tabular}{lcccc|ccc}
        & \multicolumn{4}{c}{\bf Planning Details} & & \\
        \bf Method & \#Samples & Iter. & Horizon & Time & \bf Reach & \bf Grasp & \bf Pick-\&-Place \\\midrule
        VJEPA 2~\citep{assran2025v} & 800 & 10 & 1 & 3 sec. & 100\% & 60\% & 80\%\\ \midrule
        \multirow{2}{*}{VJEPA 2.1 (ours)} & 800 & 10 & 1 & 3 sec. & 100\% & 70\% & 80\%\\ 
        & 300 & 15 & 8 & 14 sec. & 100\% & 80\% & 80\%\\ 
        \bottomrule
    \end{tabular}
\end{table} 

\begin{figure}[t]
    \centering
    \includegraphics[width=\linewidth]{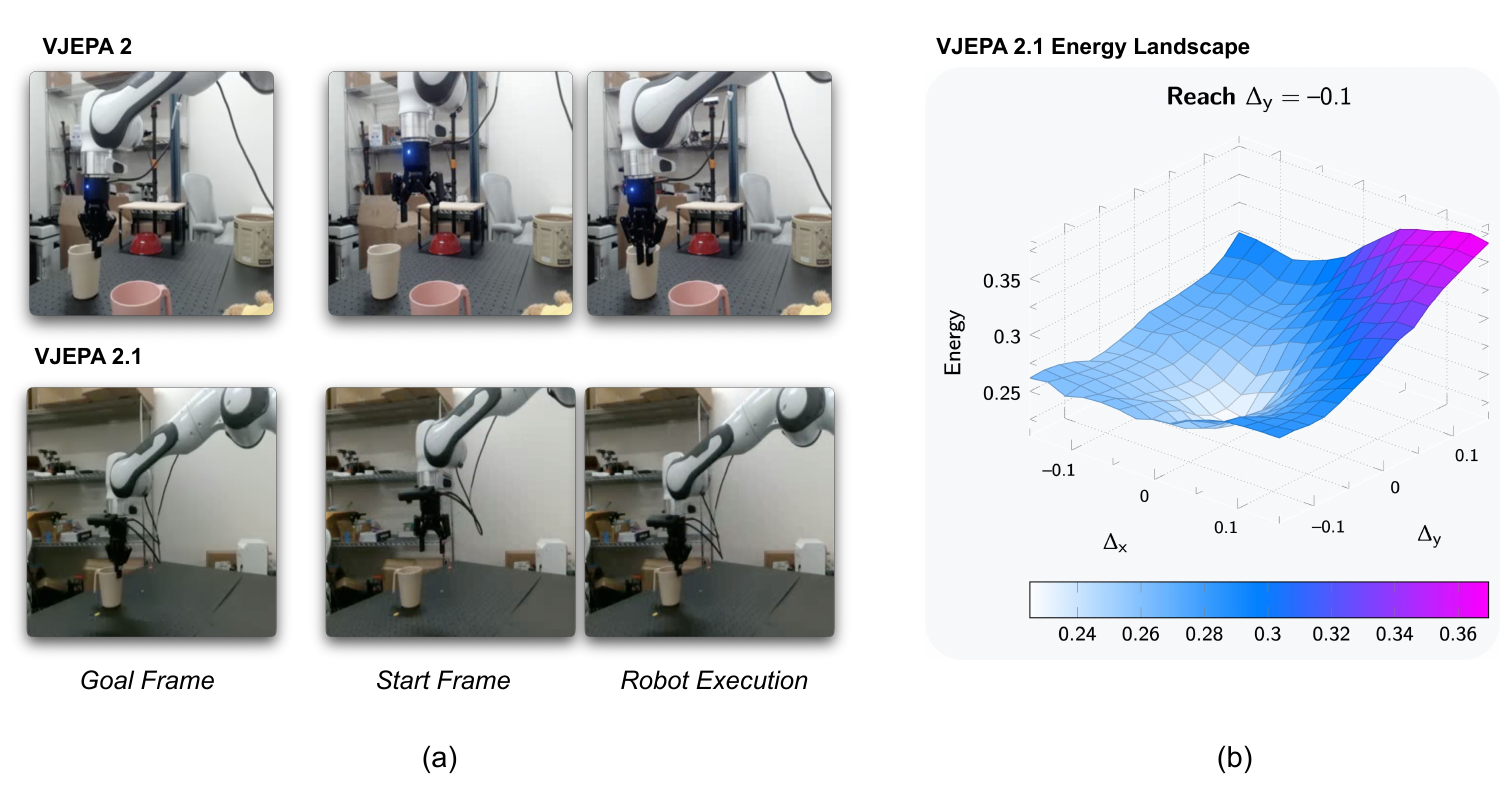}
    \caption{
    Zero-shot evaluation of VJEPA 2 and VJEPA 2.1 models on a table-top Franka Panda robot arm with a parallel-jaw gripper, demonstrating {\it Grasp} of a cup using visual goal specification via model-predictive control.
    VJEPA 2.1 features better encode depth information, leading to an improvement in robot manipulation when grasping the cup involves reasoning over actions along the depth axis of the camera.}
    \label{fig:robot_depth}
\end{figure}

Success rates are reported in Table~\ref{tb:robot}.
VJEPA 2.1 exhibits an improved depth understanding, leading to a 10\% improvement in the success rate of Grasp compared to VJEPA 2 (cf.~Figure~\ref{fig:robot_depth}).
Moreover, we find that the VJEPA 2.1 model unlocks the benefit of planning with slightly longer rollouts, perhaps due to the improvement in dense features afforded by our model.
In particular, by reducing the number of CEM samples and planning over 8 steps, we observe an additional improvement in the success rate on Grasp.
By contrast, we actually observe a degradation in the success rate of VJEPA 2 when planning over longer horizons.
Qualitatively, we further observe that task failures using the VJEPA 2.1 model on Pick-and-Place and Grasp are a result of poor planning over gripper actions, as opposed to failures in spatial understanding; e.g., closing the gripper too soon such that you cannot grasp the object, or slightly opening the gripper while in transit leading to the object being dropped.

\subsection{Navigation Planning}
\label{sec:navigation_planning}
\begin{figure}[t]
    \centering
    \includegraphics[width=\linewidth]{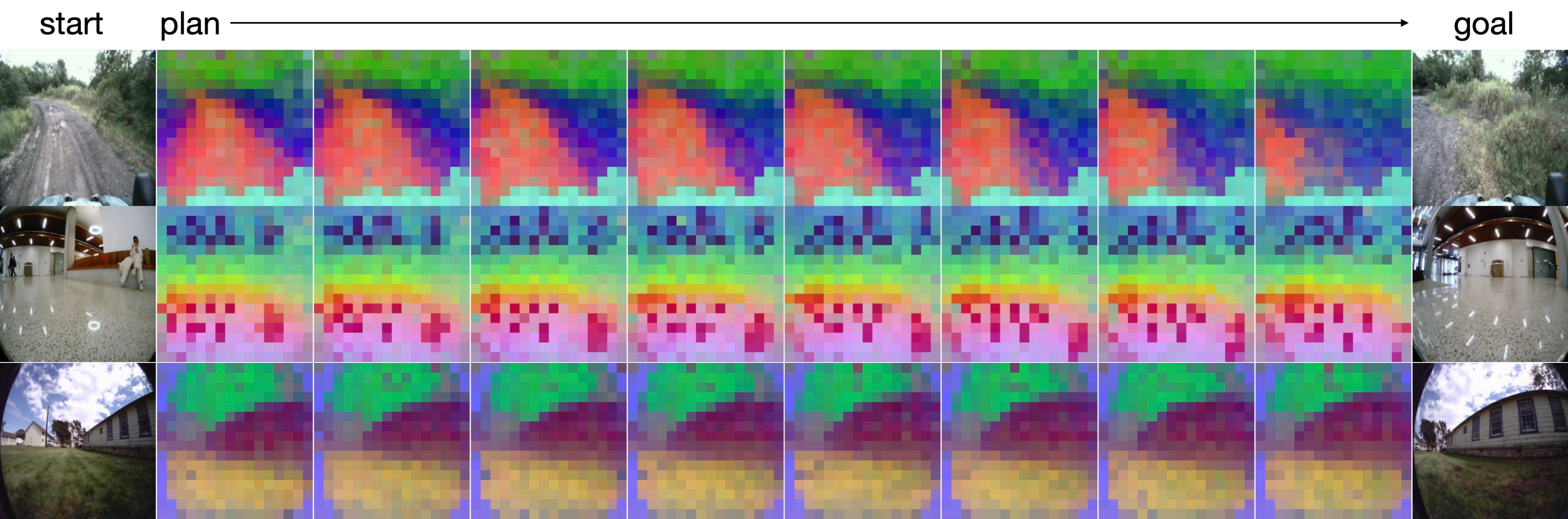}
    \caption{\textbf{Planning Navigation Trajectories in Latent Space.} V-JEPA 2.1 enables $10\times$ faster and more accurate navigation planning compared to~\citep{Bar_2025_CVPR}. We show PCA visualizations of 8 denoising steps of the planned latent trajectory between a start frame and a goal frame.}
    \label{fig:nav-planning}
\end{figure}

% \nicolas{Could you add a small paragraph describing the tasks at a high-level (i.e. action space, trajectory length..}

Next, we evaluate the utility of V-JEPA 2.1 representations for short-term robotic navigation. Specifically, given an agent’s most recent observation and a goal location specified by an image, we predict a 2-second navigation trajectory toward the goal. We adopt the navigation planning setup of NWM~\citep{Bar_2025_CVPR} and train a latent world model on top of the V-JEPA 2.1 representations. We find that V-JEPA 2.1 enables \textbf{more accurate planning} while achieving \textbf{$10\times$ faster planning speed} than the previously used SD-VAE~\citep{rombach2022high}. Results are reported in Table~\ref{tab:planning_comparison}, with qualitative examples shown in Figure~\ref{fig:nav-planning}.

We train a Conditional Diffusion Transformer (CDiT) on top of the V-JEPA 2.1 representations, similarly to NWM. Our initial experiments indicate that directly training a diffusion model in the high-dimensional embedding space of the V-JEPA 2.1 is challenging: because the representations are high-dimensional (at least $80\times$ larger than those of an SD-VAE), injecting noise in arbitrary directions can easily push samples off the data manifold. To improve robustness, we introduce two modifications. First, we train the model to predict a clean representation rather than noise~\citep{li2025back}. Second, we use DDIM sampling~\citep{ddim}, which is less stochastic than DDPM~\citep{ddpm} and empirically improves stability.

For planning, given an initial context embedding and a goal, we use the Cross-Entropy Method to sample $N=480$ candidate trajectories of length $2$ seconds at $4$ FPS. We select the best candidate by simulating it in the world model and choosing the trajectory that minimizes the distance to the goal embedding. To simplify the search, we plan in a reduced 3-DoF action space (translation and yaw rotation). With V-JEPA 2.1, we require only $8$ denoising steps, compared to the optimal SD-VAE setting which requires at least $128$ steps. This yields lower Absolute Trajectory Error (ATE) and Relative Trajectory Error (RTE) on the validation set while reducing planning time by $10\times$.\begin{table}[t]
\centering
\caption{\textbf{Open Loop Navigation Planning.}\normalfont~We finetune V-JEPA 2.1 on robot navigation datasets and compare the open-loop planning performance to NWM. Reporting the average planning time (seconds), normalized Average Trajectory Error (ATE) and Relative Trajectory Error (RTE). V-JEPA 2.1 representation enables $10\times$ speed-up without sacrificing performance.}
\label{tab:planning_comparison}
\resizebox{\textwidth}{!}{
\begin{tabular}{lccccccccccc}
\toprule

Model & Time &
% Model & Avg. Time (seconds) &
\multicolumn{2}{c}{Recon} &
\multicolumn{2}{c}{Tartan Drive} &
\multicolumn{2}{c}{Scand} &
\multicolumn{2}{c}{Sacson} &
\multicolumn{2}{c}{Average} \\
\cmidrule(lr){3-4}
\cmidrule(lr){5-6}
\cmidrule(lr){7-8}
\cmidrule(lr){9-10}
\cmidrule(lr){11-12}
 & & ATE & RTE & ATE & RTE & ATE & RTE & ATE & RTE & ATE & RTE \\
\midrule
NWM~\citep{Bar_2025_CVPR} & $103.2$\,{\footnotesize(s)} &
1.146 & 0.339 & 5.831 & 1.219 & 1.113 & {0.297} & 4.037 & 0.928 & 3.032 & 0.696 \\
\hdashline
V-JEPA 2.1 ViT-g & {10.6}\,{\footnotesize(s)} &
\textbf{1.146} & \textbf{0.338} & {5.758} & {1.200} & {1.094} & {0.299} & \textbf{3.904} & \textbf{0.921} & \textbf{2.975} & {0.690} \\
V-JEPA 2.1 ViT-G & \textbf{10.6}\,{\footnotesize(s)} & 1.179 & 0.340 & \textbf{5.687} & \textbf{1.187} & \textbf{1.038} & \textbf{0.285} & 4.054 & 0.938 & 2.990 & \textbf{0.688} \\
\bottomrule
\end{tabular}
}
\end{table}

% \begin{table}[t]
% \centering
% \caption{Planning performance comparison (Jan 7 sweep with 5 seeds).}
% \label{tab:planning_jan7}
% \resizebox{\textwidth}{!}{%
% \begin{tabular}{lcccccccccc}
% \toprule
% Model & \multicolumn{2}{c}{Recon} & \multicolumn{2}{c}{Tartan Drive} & \multicolumn{2}{c}{Scand} & \multicolumn{2}{c}{Sacson} & \multicolumn{2}{c}{Average} \\
% \cmidrule(lr){2-3} \cmidrule(lr){4-5} \cmidrule(lr){6-7} \cmidrule(lr){8-9} \cmidrule(lr){10-11} 
%  & ATE & RTE & ATE & RTE & ATE & RTE & ATE & RTE & ATE & RTE \\
% \midrule
% VAE & 1.146 & 0.339 & 5.831 & 1.219 & 1.113 & 0.297 & 4.037 & 0.928 & 3.032 & 0.696 \\
% V-JEPA 2.1 & \textbf{1.146} & \textbf{0.338} & 5.758 & 1.200 & 1.094 & 0.299 & \textbf{3.904} & \textbf{0.921} & \textbf{2.975} & 0.690 \\
% V-JEPA 2.1 Gigantic & 1.179 & 0.340 & \textbf{5.687} & \textbf{1.187} & \textbf{1.038} & \textbf{0.285} & 4.054 & 0.938 & 2.990 & \textbf{0.688} \\
% \bottomrule
% \end{tabular}
% }%
% \end{table}

\begin{table}[h]
\caption{\textbf{Performance on dense visual tasks.} \normalfont
We report the Root Mean Squared Error (RMSE) for depth estimation (NYUv2, KITTI), mean Intersection-over-Union (mIoU) for semantic segmentation (ADE20K, Cityscapes, VOC12), and the $\mathcal{J}\&\mathcal{F}$-Mean for video object segmentation (DAVIS, YouTube-VOS). Following~\cite{simeoni2025dinov3}, ADE20K and VOC12 are evaluated at an input resolution of $448\times448$ for models with patch size 14 and $512\times512$ for those with patch size 16. Cityscapes, NYUv2, and KITTI use their default evaluation resolutions. For DAVIS and YouTube-VOS, we evaluate with the shorter image side set to 480 px (patch size 16) or 420 px (patch size 14). V-JEPA 2.1 VIT G demonstrates strong performance on dense vision tasks. Notably, it achieves state-of-art performances on NYUv2 depth estimation, outperforming a DINOv3 backbone with 7 billion parameters.}
\centering
\resizebox{\columnwidth}{!}{%
\begin{tabular}{lccccccccc}
%\toprule
\multirow{2}{*}{\bf{Method}} & \multirow{2}{*}{\bf{Param.}} &
\multicolumn{2}{c}{\bf{Depth}} &
\multicolumn{3}{c}{\bf{Segmentation}} &
\multicolumn{2}{c}{\bf{VOS}} \\ % ← VOS se agrega en la misma fila
\cmidrule(lr){3-4} \cmidrule(lr){5-7} \cmidrule(lr){8-9}
& &  NYUv2$\downarrow$ & KITTI$\downarrow$ & ADE20k & Citysc. & VOC & DAVIS-S & YT VOS-S \\
\bottomrule

\multicolumn{9}{l}{\textit{7B models}} \\
Web-DINO~\citep{fan2025webdino} & 7B/14 & 0.466 & 3.158 & 42.7 & 68.3 & 76.1 & 57.2 & 43.9 \\
DINOv3~\citep{simeoni2025dinov3} & 7B/16 & \bf{0.309} & \bf{2.346} & \bf{55.9} & \bf{81.1} & \bf{86.6} & \bf{71.1} & \bf{74.1} \\
%\midrule
\bottomrule

\multicolumn{9}{l}{\textit{Models less than 2B parameters}} \\
PEcore~\citep{bolya2025perception} & 2B/14 & 0.590 & 4.119 & 38.9 & 61.1 & 69.2 & 48.2 & 34.7 \\
PEspatial~\citep{bolya2025perception} & 2B/14 & 0.362 & 3.082 &49.3 & 73.2 & 82.7 &  68.4 & 68.5 \\
AM-RADIOv2.5~\citep{ranzinger2024radio} & 1B/14 & 0.340 & 2.918 & 53.0 & 78.4 & 85.4 & 66.5 & 70.1 \\
%\bf{Franca}~\citep{venkataramanan2025franca} & 1B/14  & 0.445 & 3.140 & 46.3 & 68.7 & 82.9 &  61.8 & 67.3 \\
InternVideo2-1B~\citep{wang2024internvideo2} & 1B/14  & 0.471 & 4.739 & 28.4 & 35.6 & 65.2 &  50.6 & 51.2 \\
SigLIP 2~\citep{tschannen2025siglip} & 1B/16 &  0.494 & 3.273 & 42.7 & 64.8 & 72.7 & 56.1 & 52.0 \\
DINOv2~\citep{oquab2023dinov2} & 1B/14  & 0.372 & 2.624 & 49.5 & 75.6 & 83.1 & 63.9 & 65.6 \\
DINOv3 ViT-H+~\citep{simeoni2025dinov3} & 0.8B/16 & 0.352 & 2.635 & \bf{54.8} & \bf{79.5} & \bf{85.8} & \bf{71.1} & \bf{74.0} \\
\hdashline

V-JEPA2 ViT-g~\citep{assran2025v} & 1B/16 & 0.642 & 4.650 & 24.4 & 45.9 & 63.9 &  52.5 & 53.7 \\
\hdashline

%\bf{V-JEPA 2.1 ViT-g}  & 1B/16 &  0.359 & 2.611 & 47.8 & 71.7 & 84.6 & 68.0 & 72.0 \\
V-JEPA 2.1 ViT-g  & 1B/16 & 0.350 & 2.601 & 47.8 & 71.8 & 84.7 &  68.1 & 72.3 \\
%\bf{V-JEPA 2.1 ViT-G} & 2B/16 &  0.313 & 2.472 & 47.8 & 73.2 & 84.7 & 68.6 & 72.7 \\
V-JEPA 2.1 ViT-G & 2B/16 & \bf{0.307} & \bf{2.461} & 47.9 & 73.5 & 85.0 & 69.0 & 72.7 \\
%\midrule
\bottomrule

%\multicolumn{9}{l}{\textit{Models less than 500M parameters}} \\
%\bf{DINOv2 ViT-L}  & 300M/14 & 48.8 & 70.3 & 82.1 & 0.394 & 2.780 & 64.4 & 68.4 \\
%\bf{DINOv3 ViT-L}  & 300M/16 & 54.9 & 78.4 & 86.2 & 0.352 & 2.645 & 72.2 & 76.7 \\
%\hdashline

%\bf{V-JEPA 2 ViT-L}  & 300M/16 & 24.2 & 43.7 & 60.9 & 0.649 & 4.708 & 52.3 & 52.8 \\
%\hdashline
%\bf{V-JEPA 2.1 ViT-L}  & 300M/16 & 42.0 & 66.7 & 79.6 & 0.381 & 2.914 & todo. & todo \\
%\bf{V-JEPA 2.1 ViT-L dist.}  & 300M/16 & 42.0 & EV. & TR. & TR. & TR. & TR. & TR. \\

%\cmidrule
\end{tabular}
}

\label{tab:segmentation_depth_extended}
\end{table}

%\subsubsection{Linear probing on image segmentation and depth estimation} 

\begin{figure}[t]
    \centering
    \renewcommand{\arraystretch}{1.2} % Espaciado entre filas
    \setlength{\tabcolsep}{0.5pt}       % Espaciado entre columnas
    \begin{tabularx}{\textwidth}{c c c c c}
         & \multicolumn{3}{c}{\textbf{NYUv2}} & \textbf{KITTI} \\
         \raisebox{0.4\height}{\rotatebox{90}{\textbf{Image}}} & 
         \includegraphics[width=0.17\linewidth]{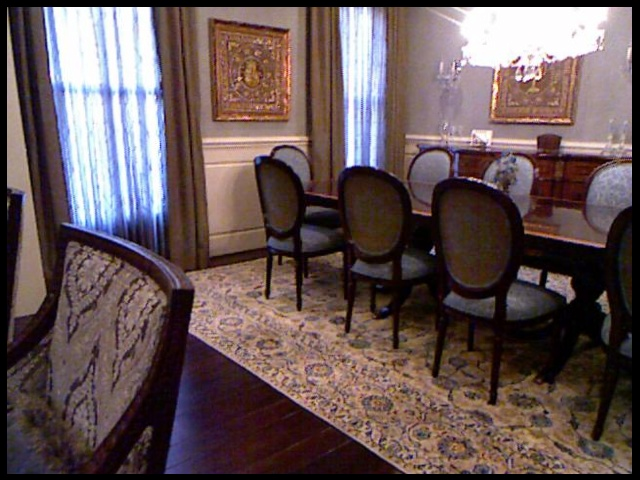} & 
         \includegraphics[width=0.17\linewidth]{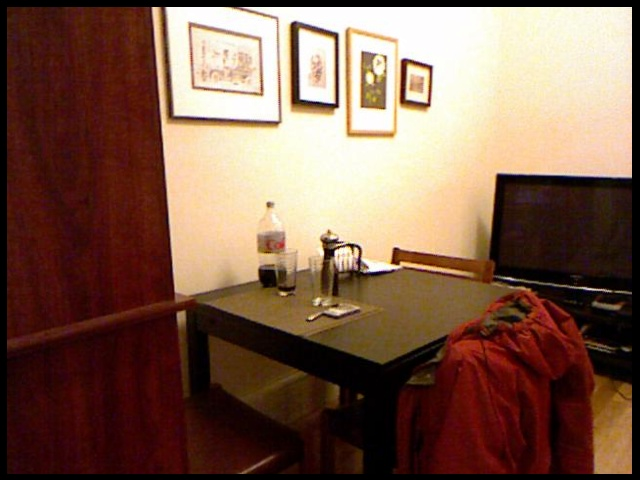} &
         \includegraphics[width=0.17\linewidth]{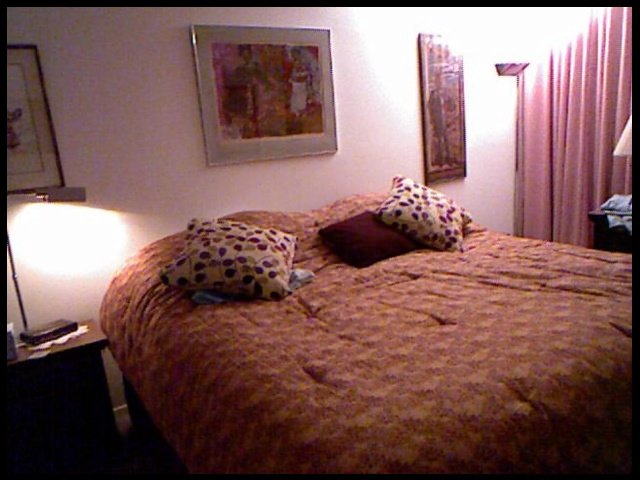} &
         \includegraphics[width=0.44\linewidth]{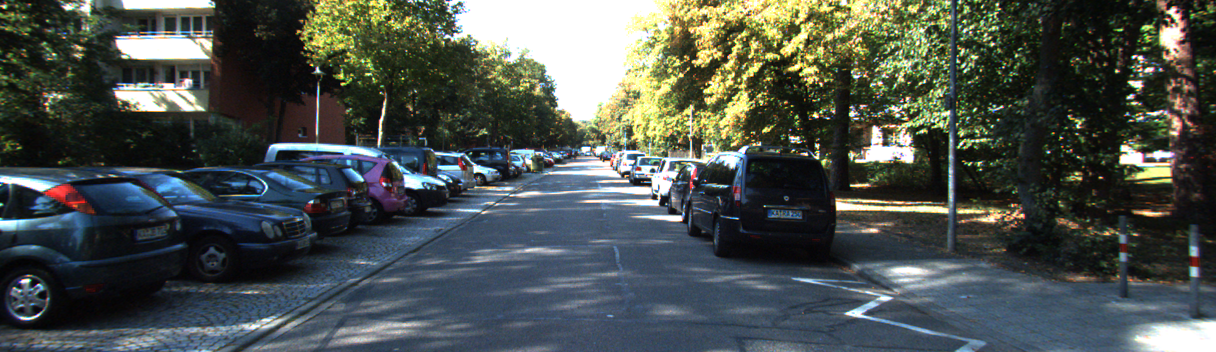} \\

         \raisebox{0.2\height}{\rotatebox{90}{\textbf{V-JEPA 2}}}& 
         \includegraphics[width=0.17\linewidth]{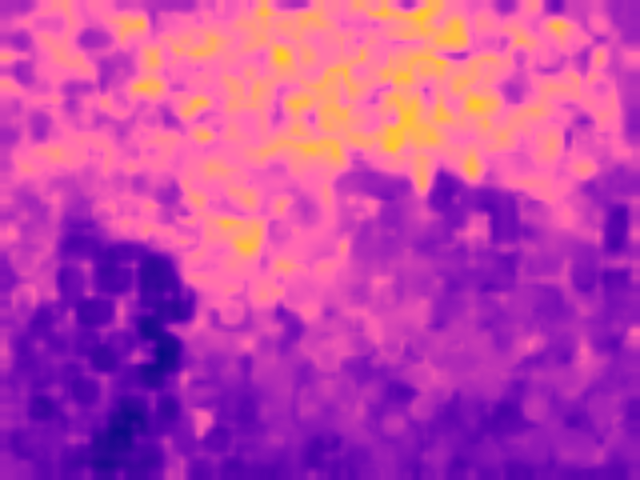} & 
         \includegraphics[width=0.17\linewidth]{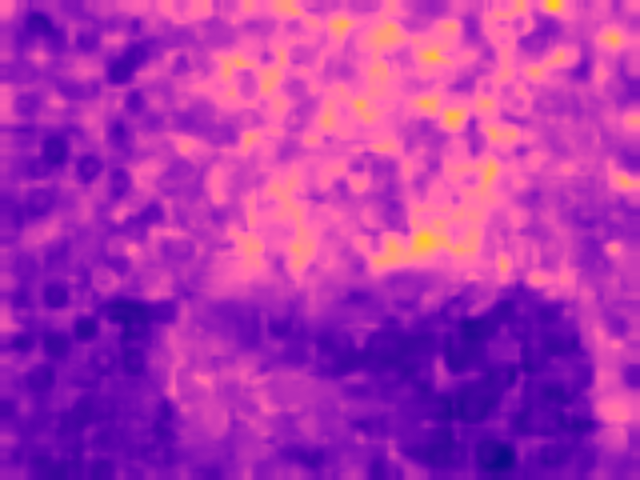} &
         \includegraphics[width=0.17\linewidth]{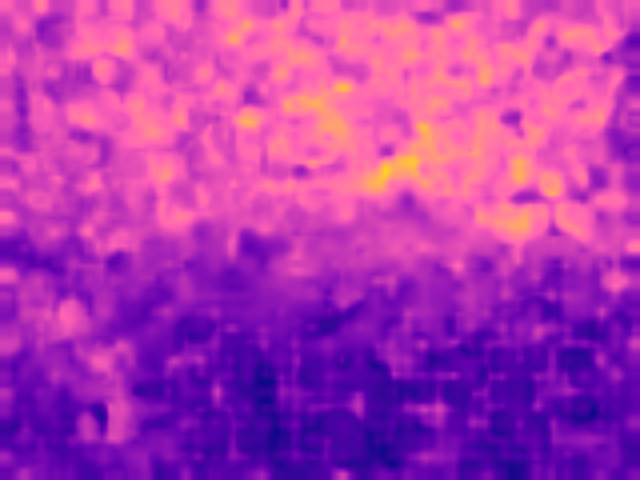} &
         \includegraphics[width=0.44\linewidth]{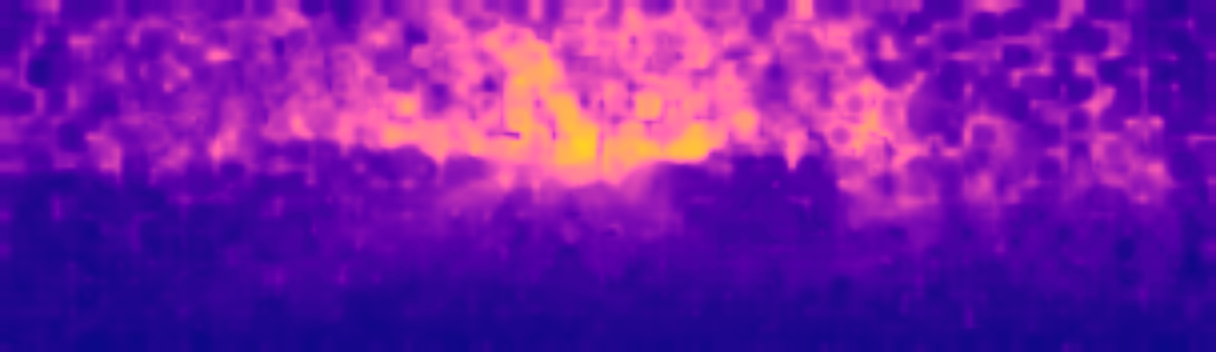} \\

         {\rotatebox{90}{\textbf{V-JEPA 2.1}}} & 
         \includegraphics[width=0.17\linewidth]{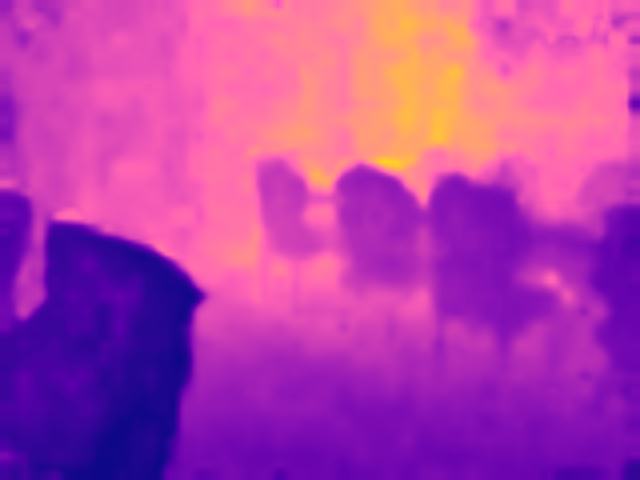} & 
         \includegraphics[width=0.17\linewidth]{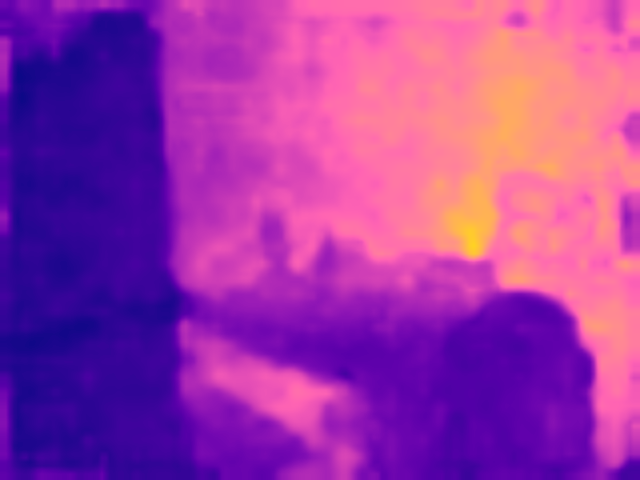} &
         \includegraphics[width=0.17\linewidth]{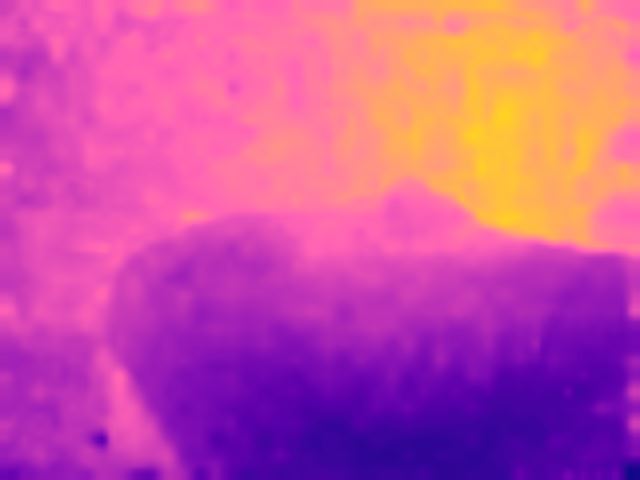} &
         \includegraphics[width=0.44\linewidth]{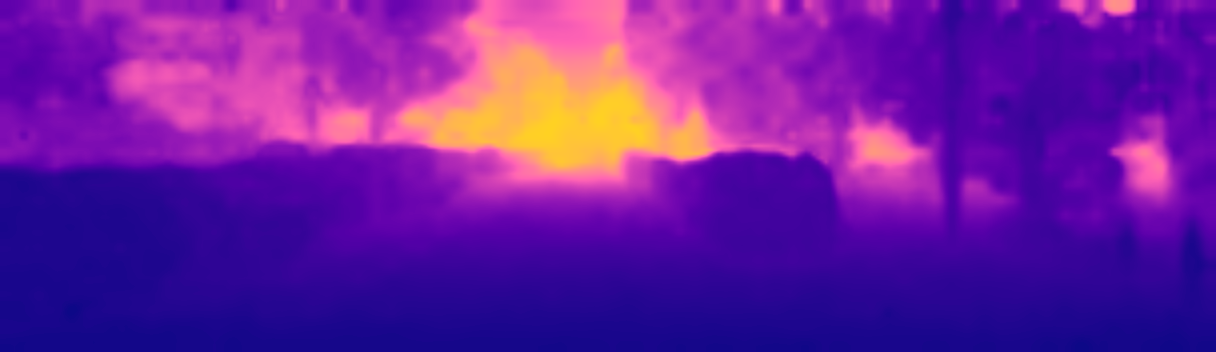} \\

         %\rotatebox{90}{\parbox{1.8cm}{\centering \bf{Image}}} & 
         %\includegraphics[width=0.20\linewidth]{depth_paper/imgs/image_385.png} & %\includegraphics[width=0.50\linewidth]{depth_paper/imgs/image_319.png} \\

         %\rotatebox{90}{\parbox{1.8cm}{\centering \bf{V-JEPA 2}}} & 
         %\includegraphics[width=0.20\linewidth]{depth_paper/vjepa2/pred_385.png} & %\includegraphics[width=0.50\linewidth]{depth_paper/vjepa2/pred_319.png} \\

         %\rotatebox{90}{\parbox{1.8cm}{\centering{\bf{V-JEPA 2.1}}} & 
         %\includegraphics[width=0.20\linewidth]{depth_paper/ours/pred_385.png} & \includegraphics[width=0.5\linewidth]{depth_paper/ours/pred_319.png} \\
    \end{tabularx}
    \caption{\textbf{Depth estimation comparison on NYU and KITTI datasets.} \normalfont While V-JEPA 2 captures the overall scene geometry, its predictions lack local consistency and precise boundary structure. In contrast, our V-JEPA 2.1 produces sharper, more coherent, and fine-grained depth maps.}
    \label{fig:depth_qual}
\end{figure}

\begin{figure}[h]
    \centering
    \renewcommand{\arraystretch}{1.2} % Espaciado entre filas
    \setlength{\tabcolsep}{0.5pt}       % Espaciado entre columnas
    \begin{tabularx}{\textwidth}{Y Y Y Y Y}
        \textbf{Image} & \textbf{Ground-truth} & \textbf{DINOv3} & \textbf{V-JEPA 2} & \textbf{V-JEPA 2.1} \\
        %\multirow{5}{*}{\rotatebox{90}{\bf{VOC 12}}} &
        \includegraphics[width=\linewidth]{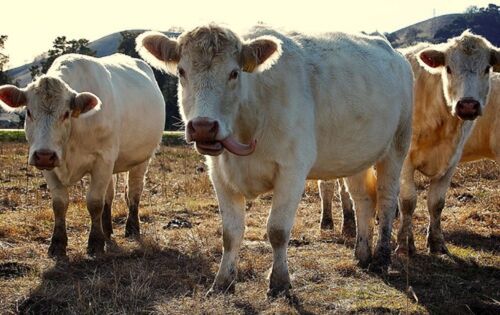} &
        \includegraphics[width=\linewidth]{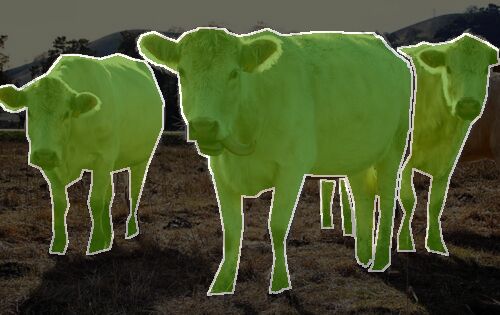} &
        \includegraphics[width=\linewidth]{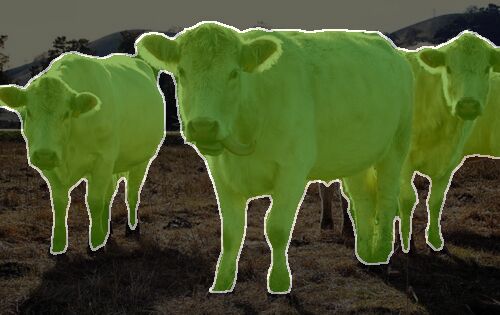} &
        \includegraphics[width=\linewidth]{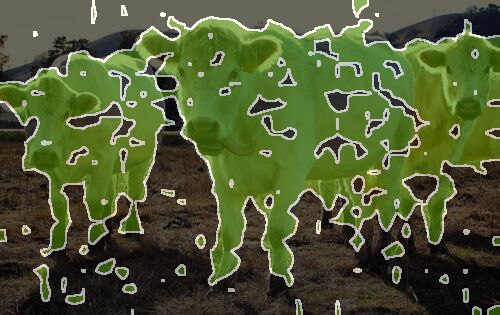} &
        \includegraphics[width=\linewidth]{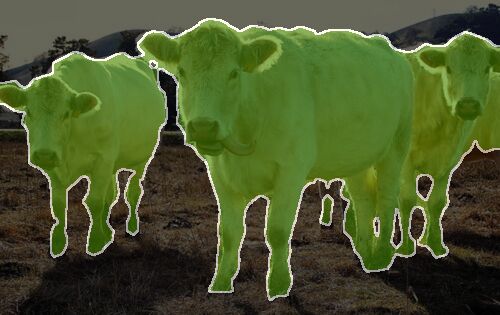} \\
        
        %&  
        \includegraphics[width=\linewidth]{qualitative_voc12/image_decoder_linear_lr_0_00150000_wd_0_01000000_23.jpg} &
        \includegraphics[width=\linewidth]{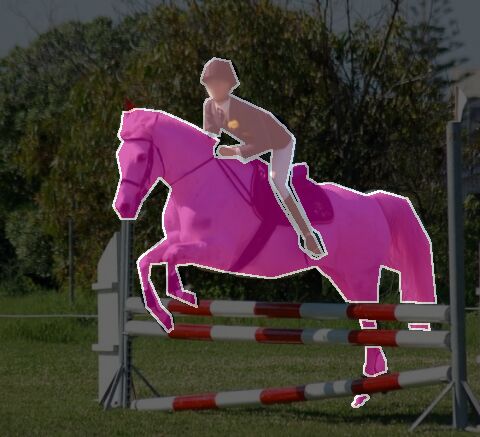} &
        \includegraphics[width=\linewidth]{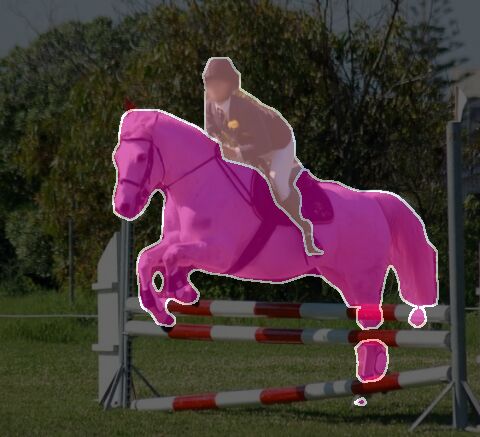} &
        \includegraphics[width=\linewidth]{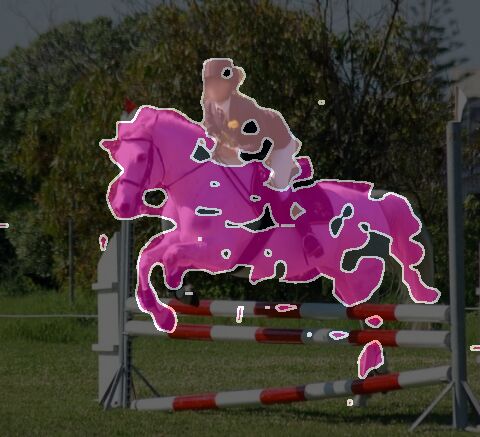} &
        \includegraphics[width=\linewidth]{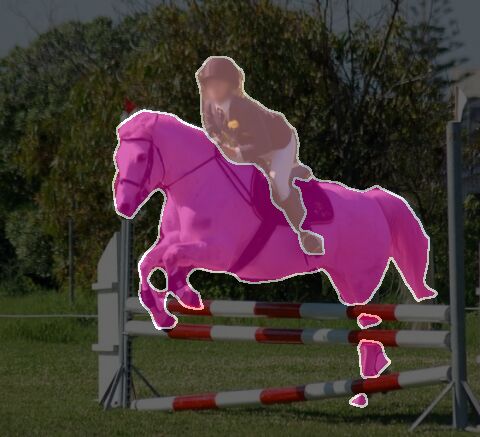} \\

        %& % 
        %\includegraphics[width=\linewidth]{qualitative_voc12/image_decoder_linear_lr_0_00150000_wd_0_01000000_252.jpg} &
        %\includegraphics[width=\linewidth]{qualitative_voc12/gt_decoder_linear_lr_0_00008000_wd_0_00010000_252.jpg} &
        %\includegraphics[width=\linewidth]{qualitative_voc12/dino_252.jpg} &
        %\includegraphics[width=\linewidth]{qualitative_voc12/pred_decoder_linear_lr_0_00008000_wd_0_00010000_252.jpg} &
        %\includegraphics[width=\linewidth]{qualitative_voc12/ours_252.jpg} \\

        %& % 
        %\includegraphics[width=\linewidth]{qualitative_voc12/image_decoder_linear_lr_0_00150000_wd_0_01000000_254.jpg} &
        %\includegraphics[width=\linewidth]{qualitative_voc12/gt_decoder_linear_lr_0_00008000_wd_0_00010000_254.jpg} &
        %\includegraphics[width=\linewidth]{qualitative_voc12/dino_254.jpg} &
        %\includegraphics[width=\linewidth]{qualitative_voc12/pred_decoder_linear_lr_0_00008000_wd_0_00010000_254.jpg} &
        %\includegraphics[width=\linewidth]{qualitative_voc12/ours_254.jpg} \\

        %& % 
        \includegraphics[width=\linewidth]{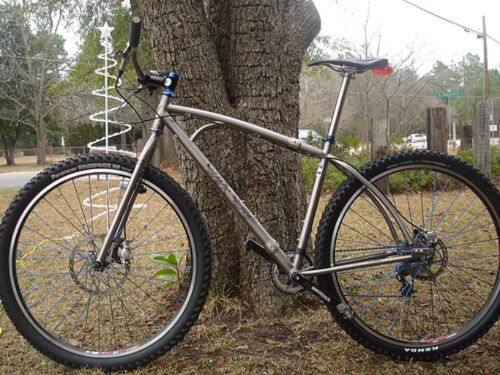} &
        \includegraphics[width=\linewidth]{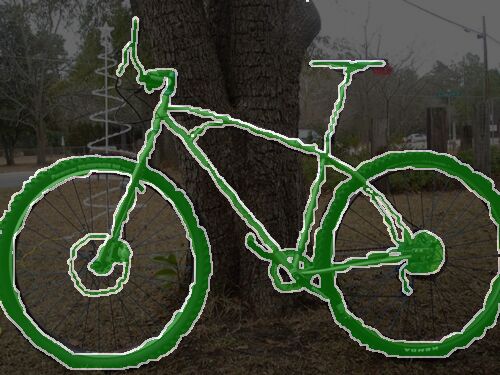} &
        \includegraphics[width=\linewidth]{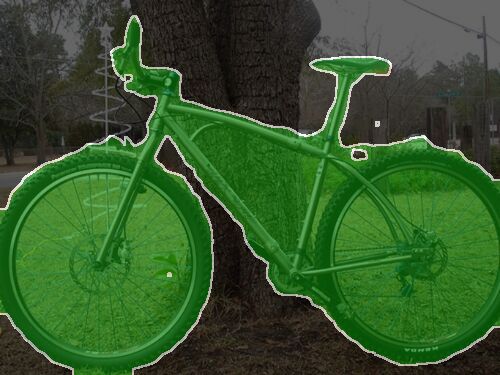} &
        \includegraphics[width=\linewidth]{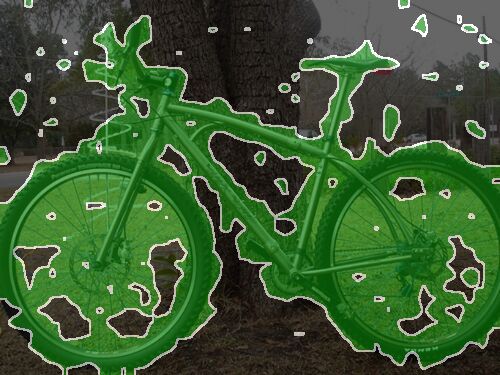} &
        \includegraphics[width=\linewidth]{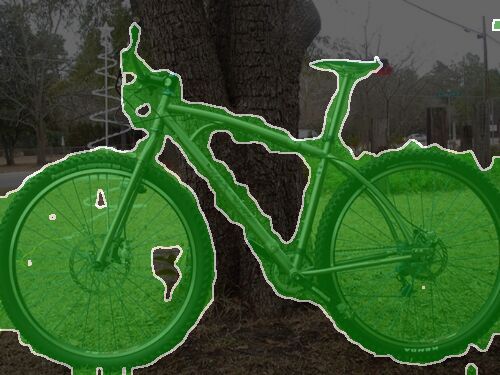} \\
        
        \includegraphics[width=\linewidth]{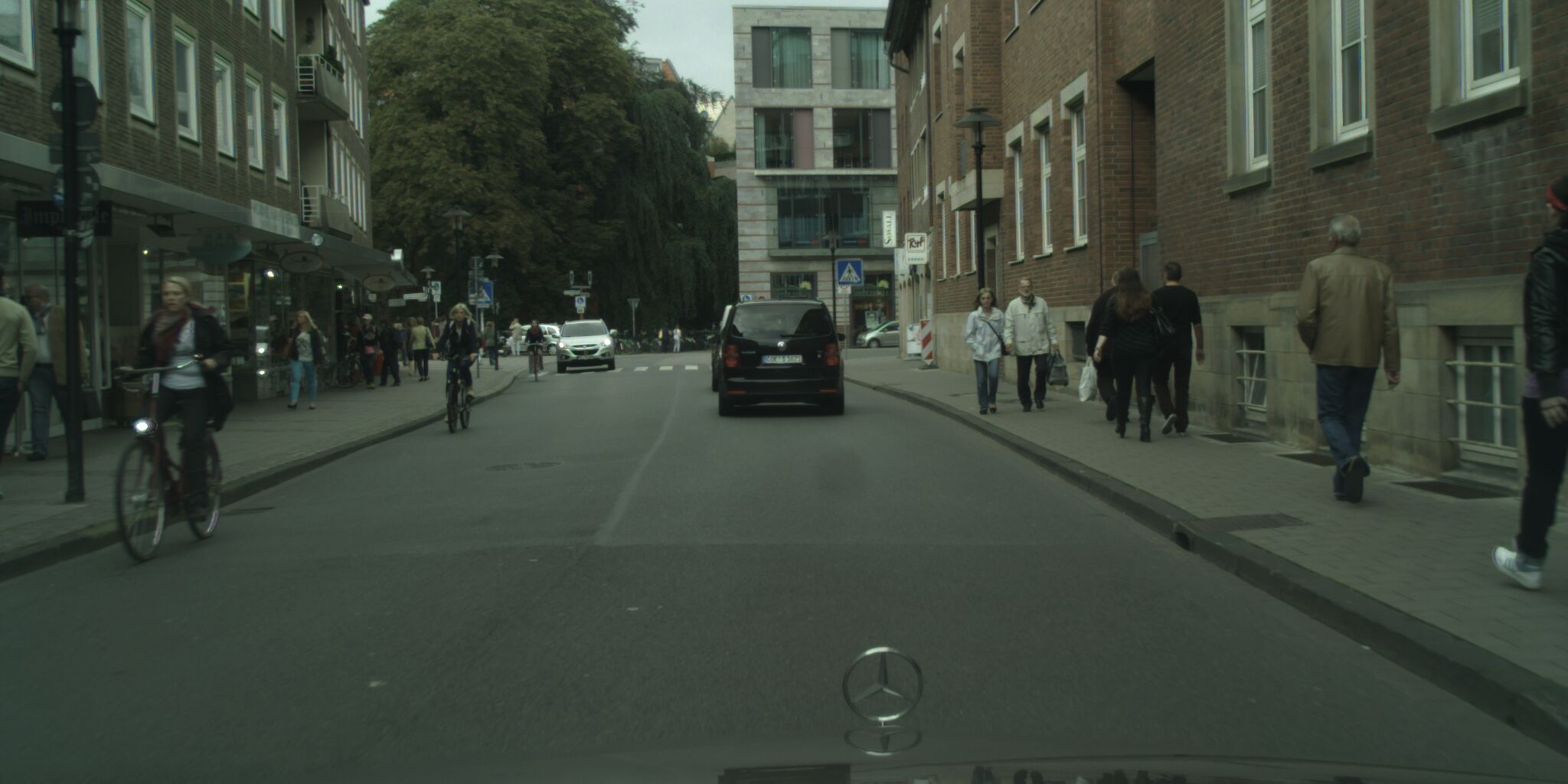} &
        \includegraphics[width=\linewidth]{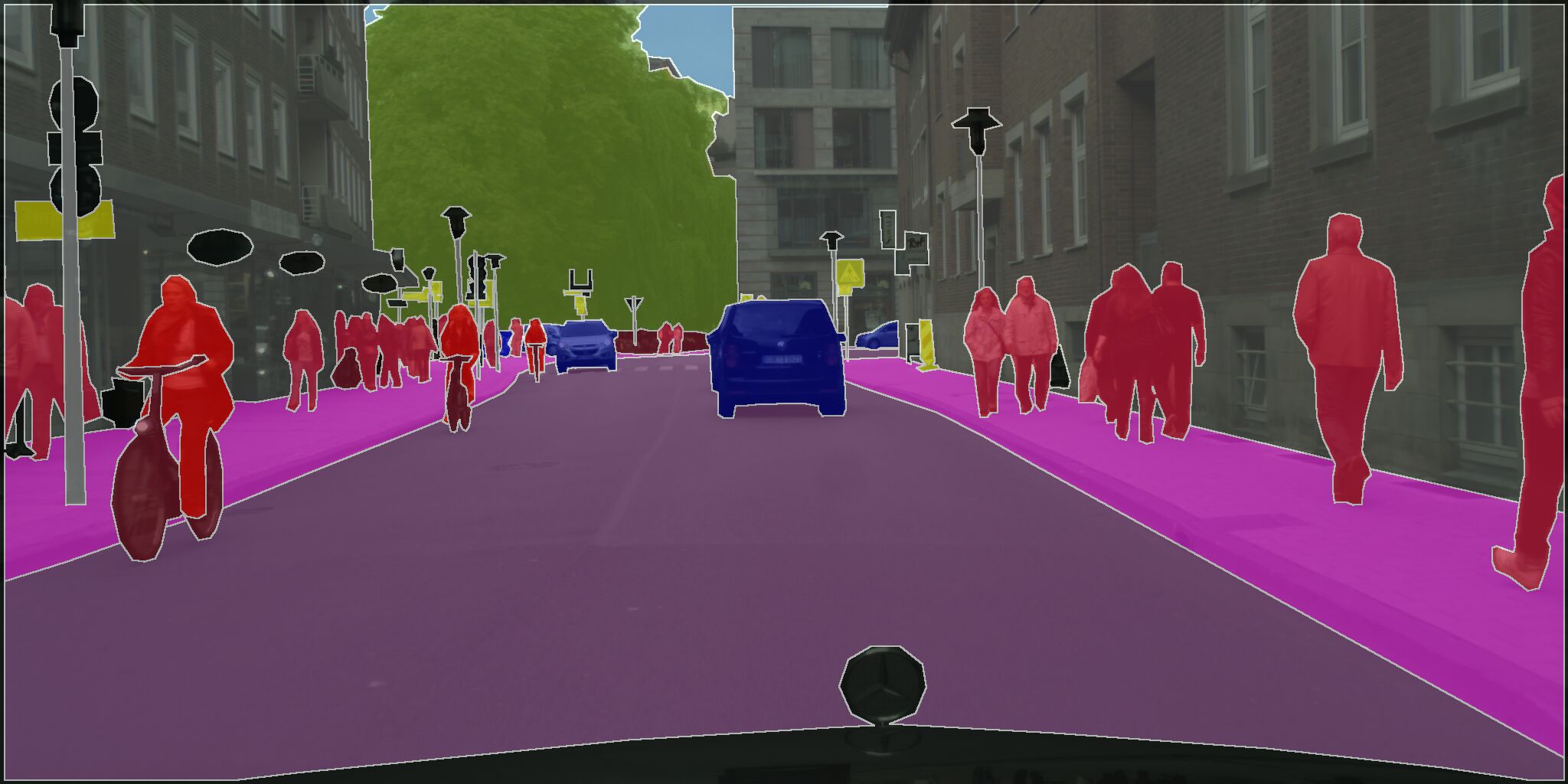} &
        \includegraphics[width=\linewidth]{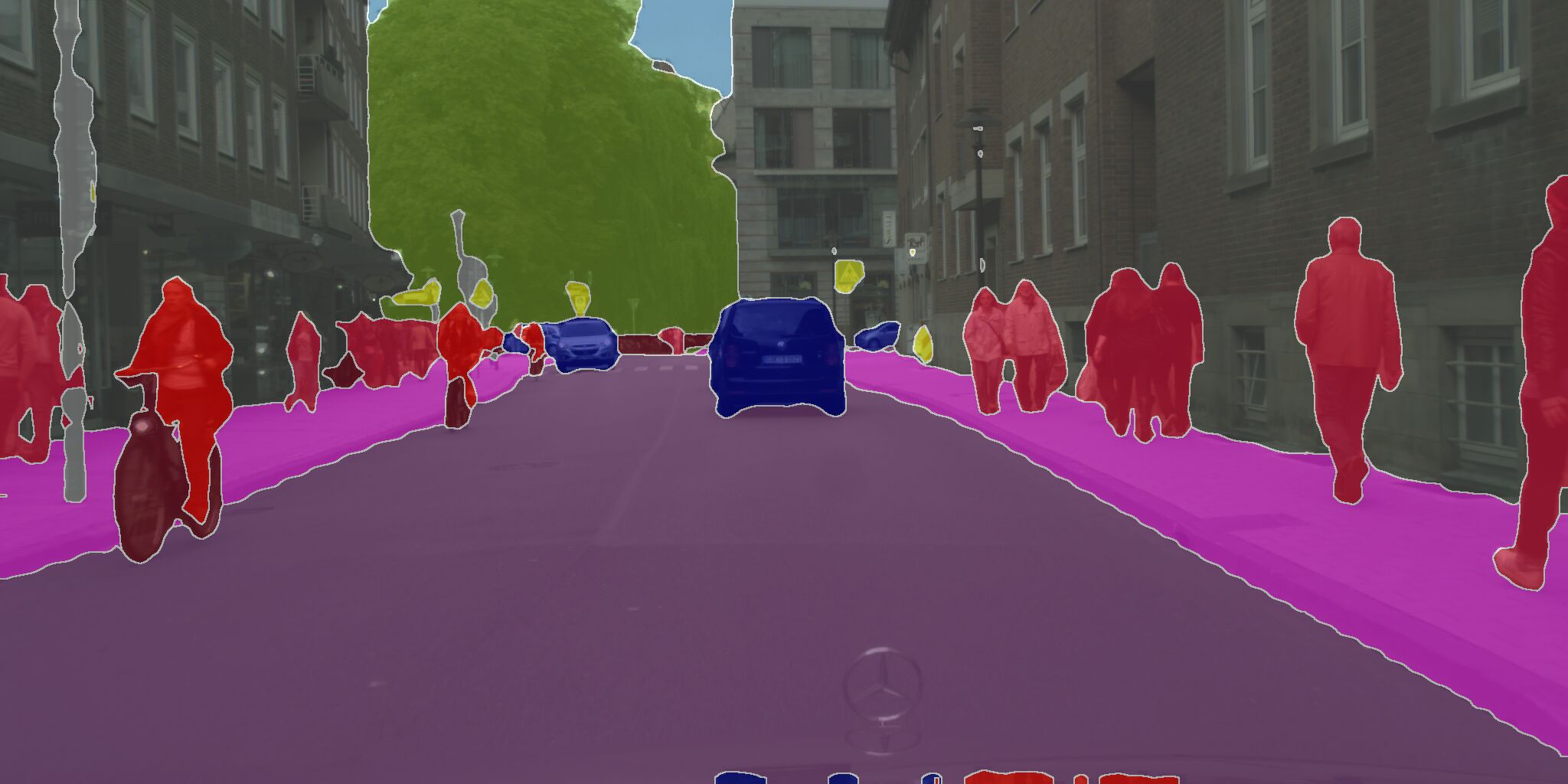} &
        \includegraphics[width=\linewidth]{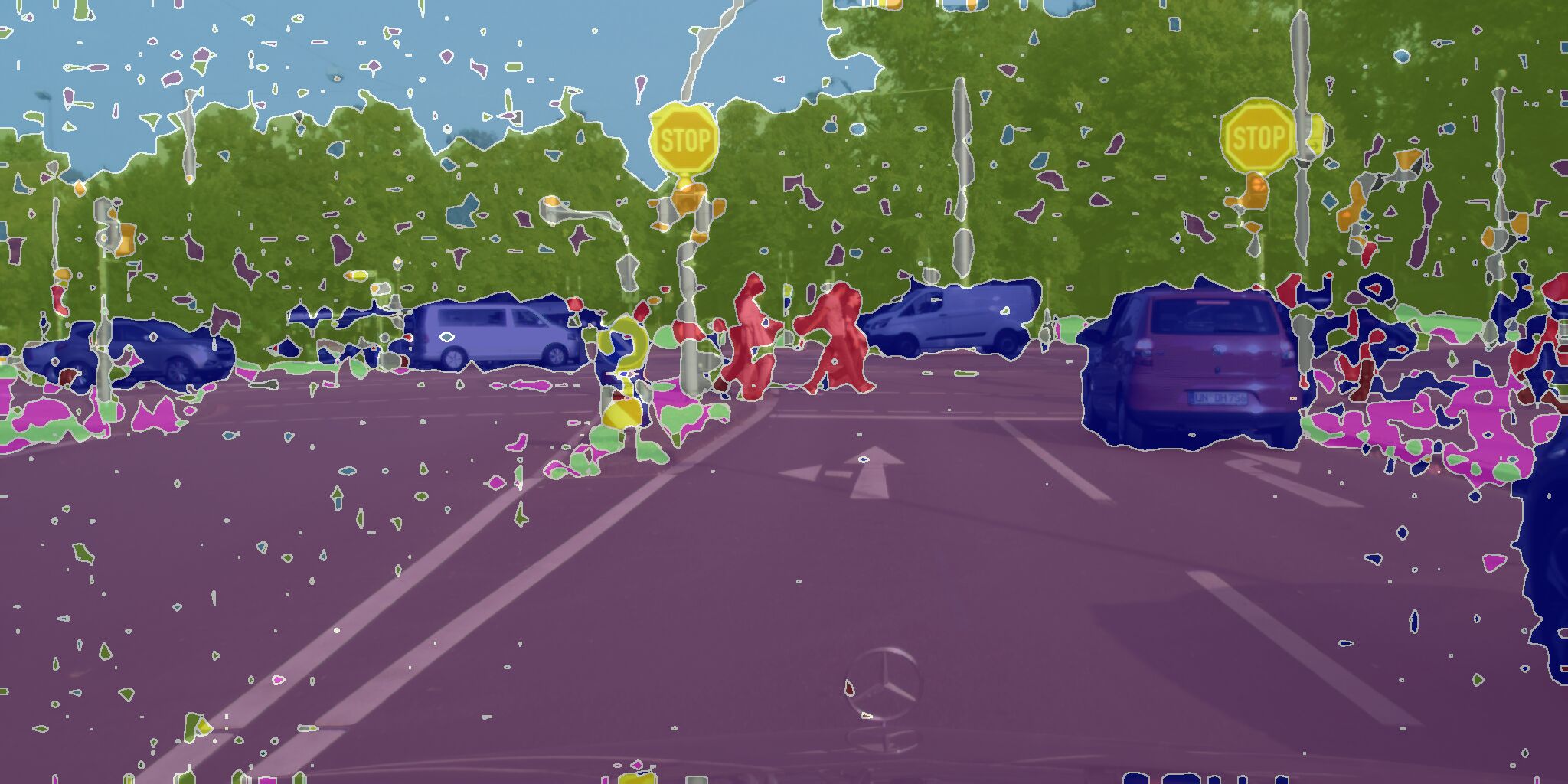} &
        \includegraphics[width=\linewidth]{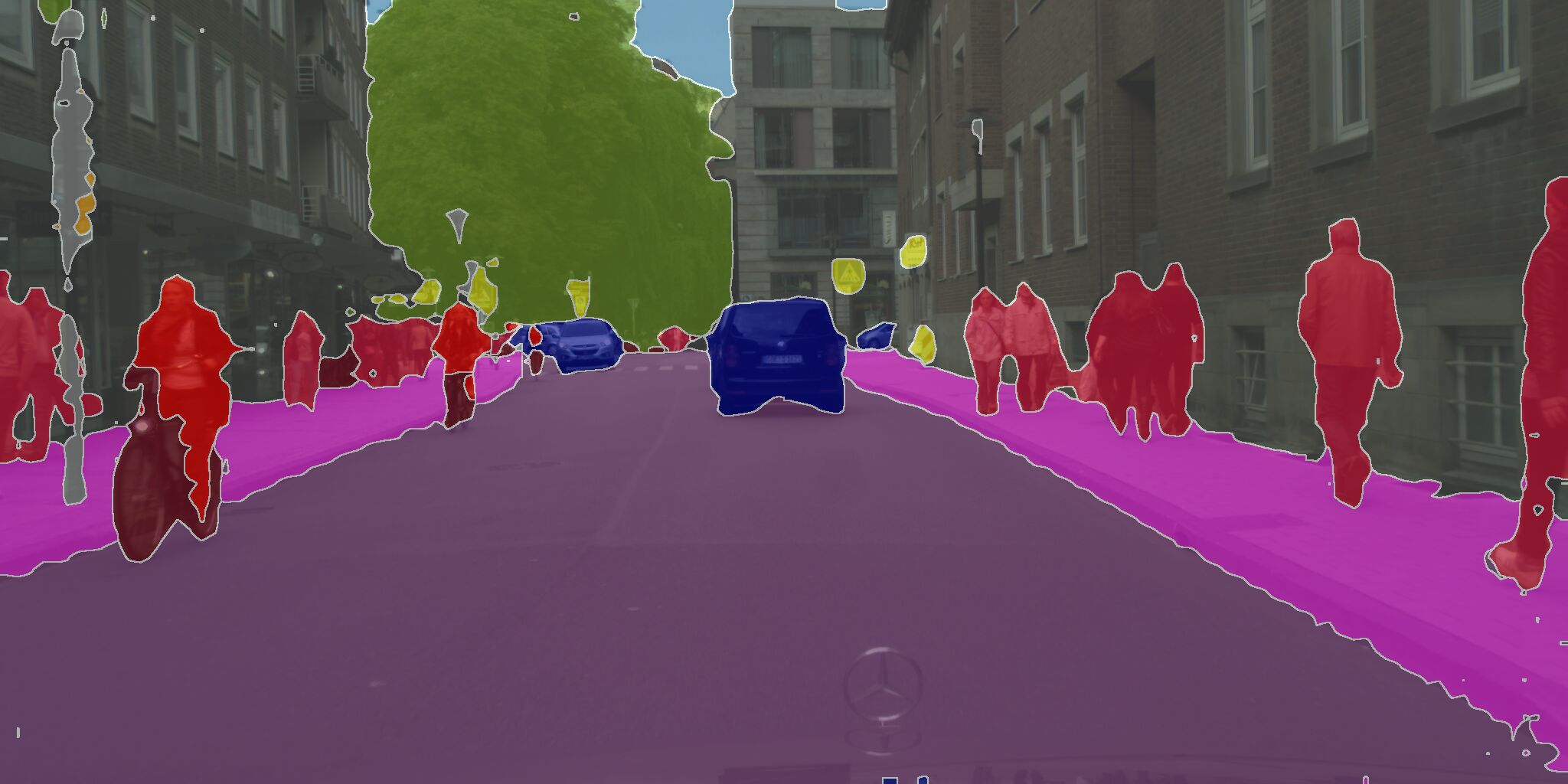} \\

        \includegraphics[width=\linewidth]{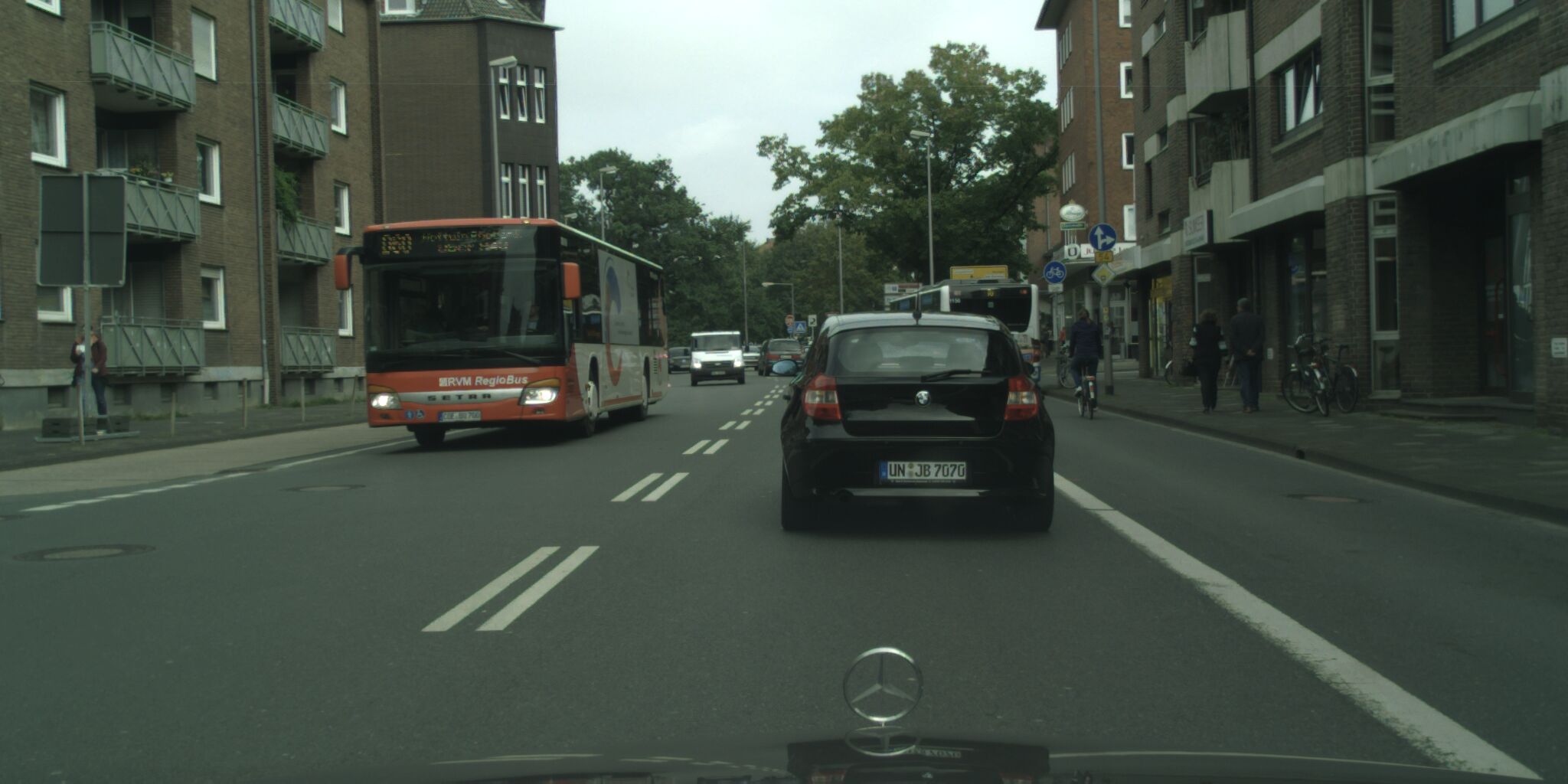} &
        \includegraphics[width=\linewidth]{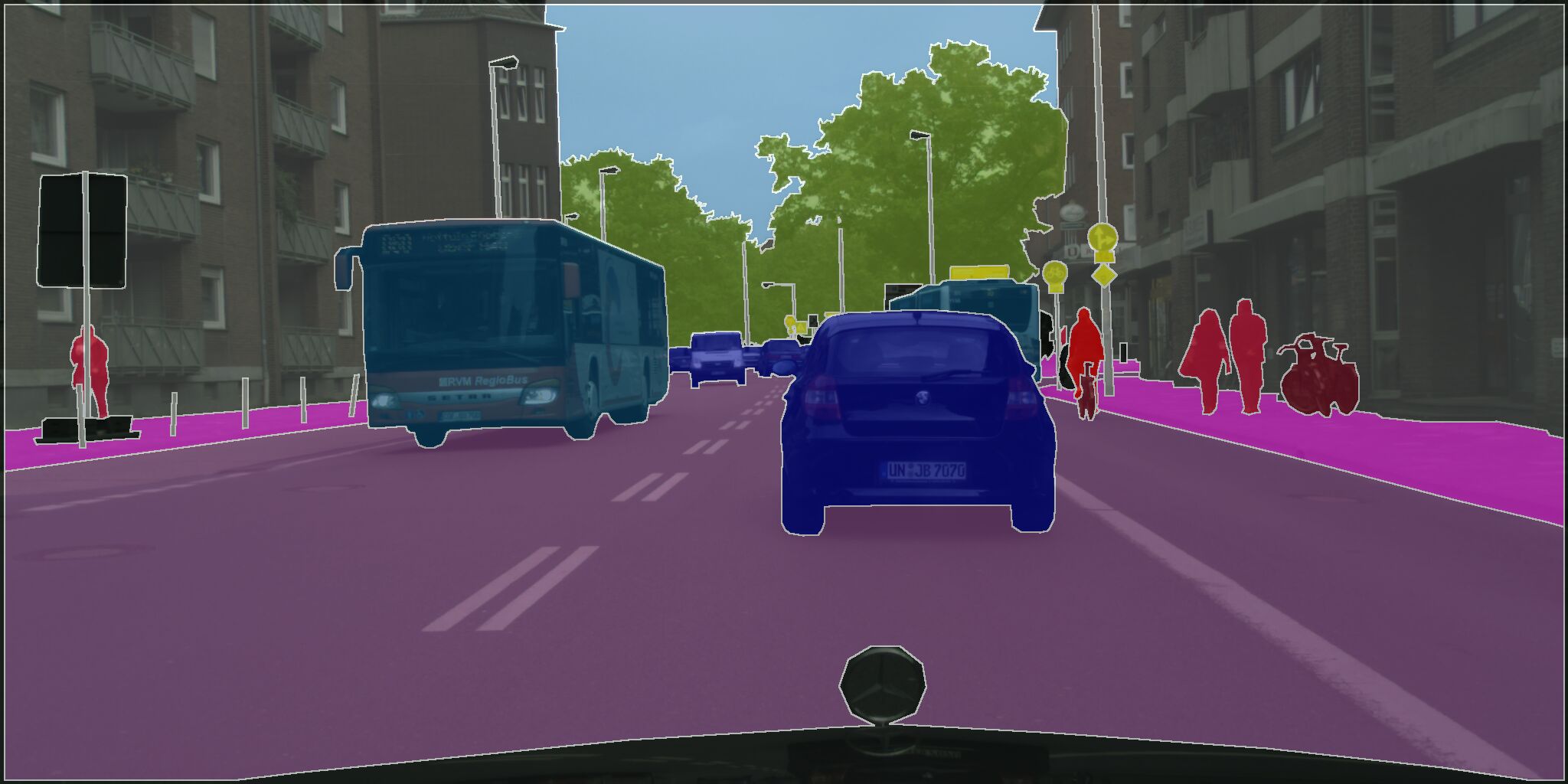} &
        \includegraphics[width=\linewidth]{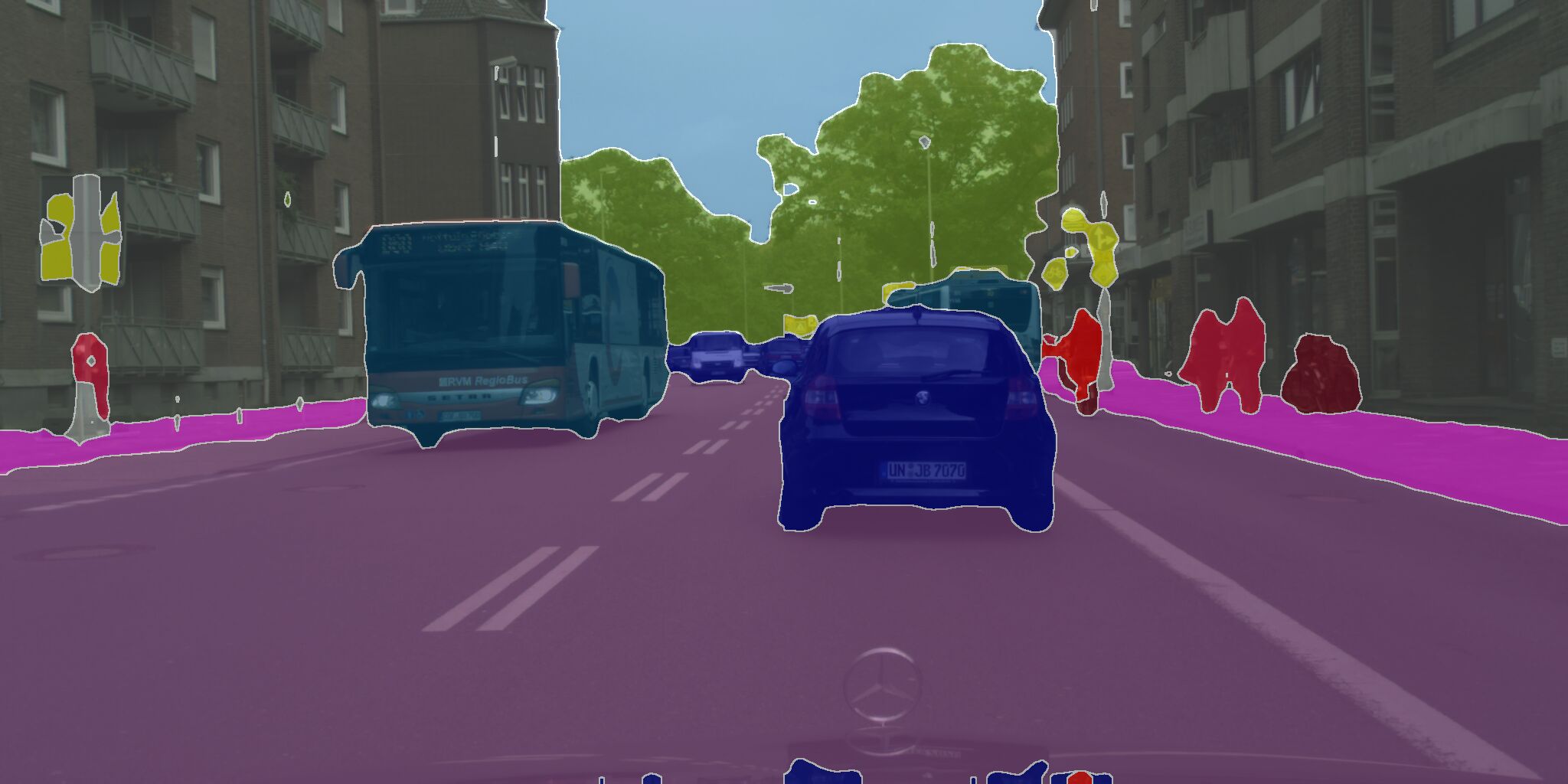} &
        \includegraphics[width=\linewidth]{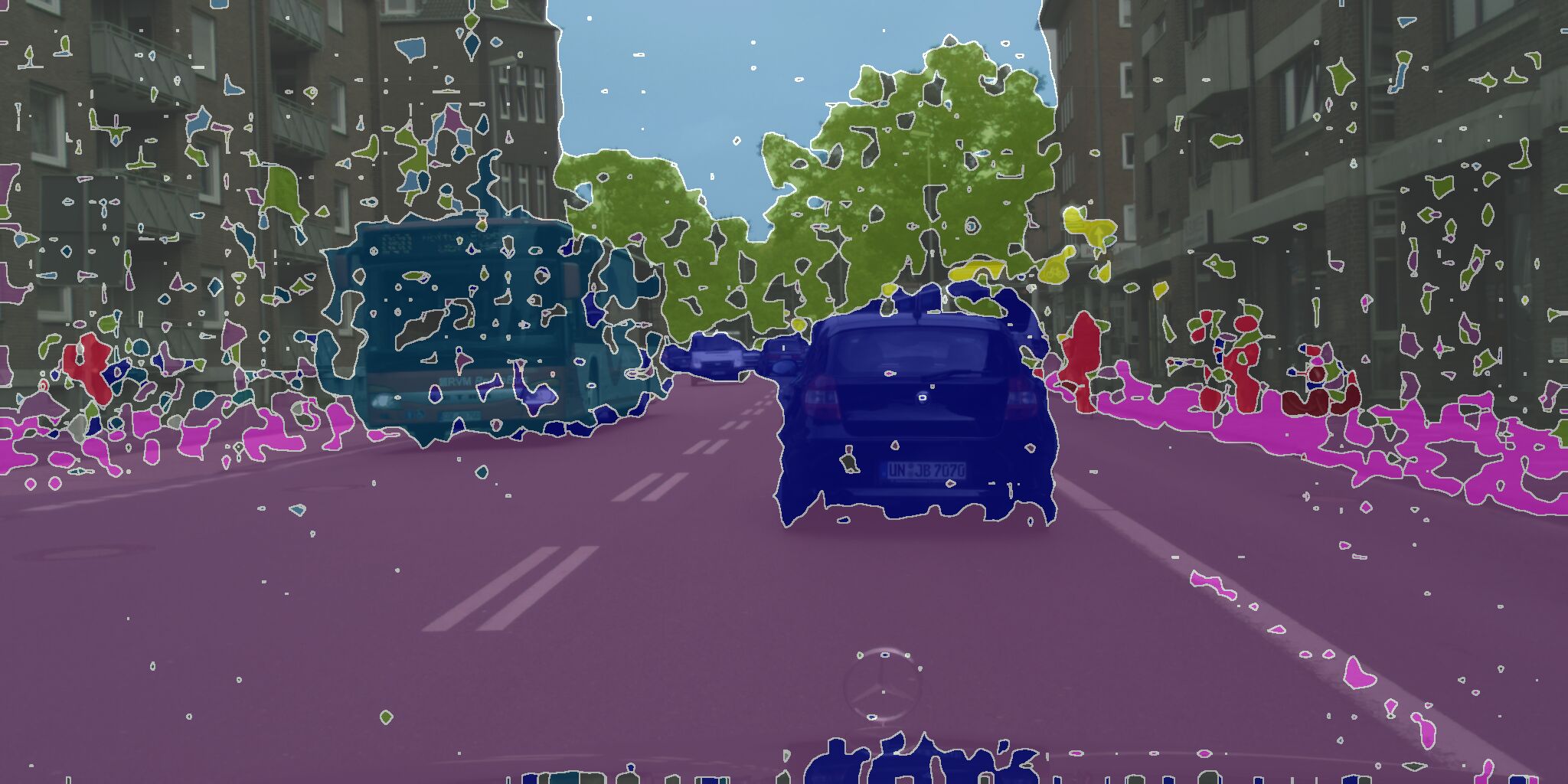} &
        \includegraphics[width=\linewidth]{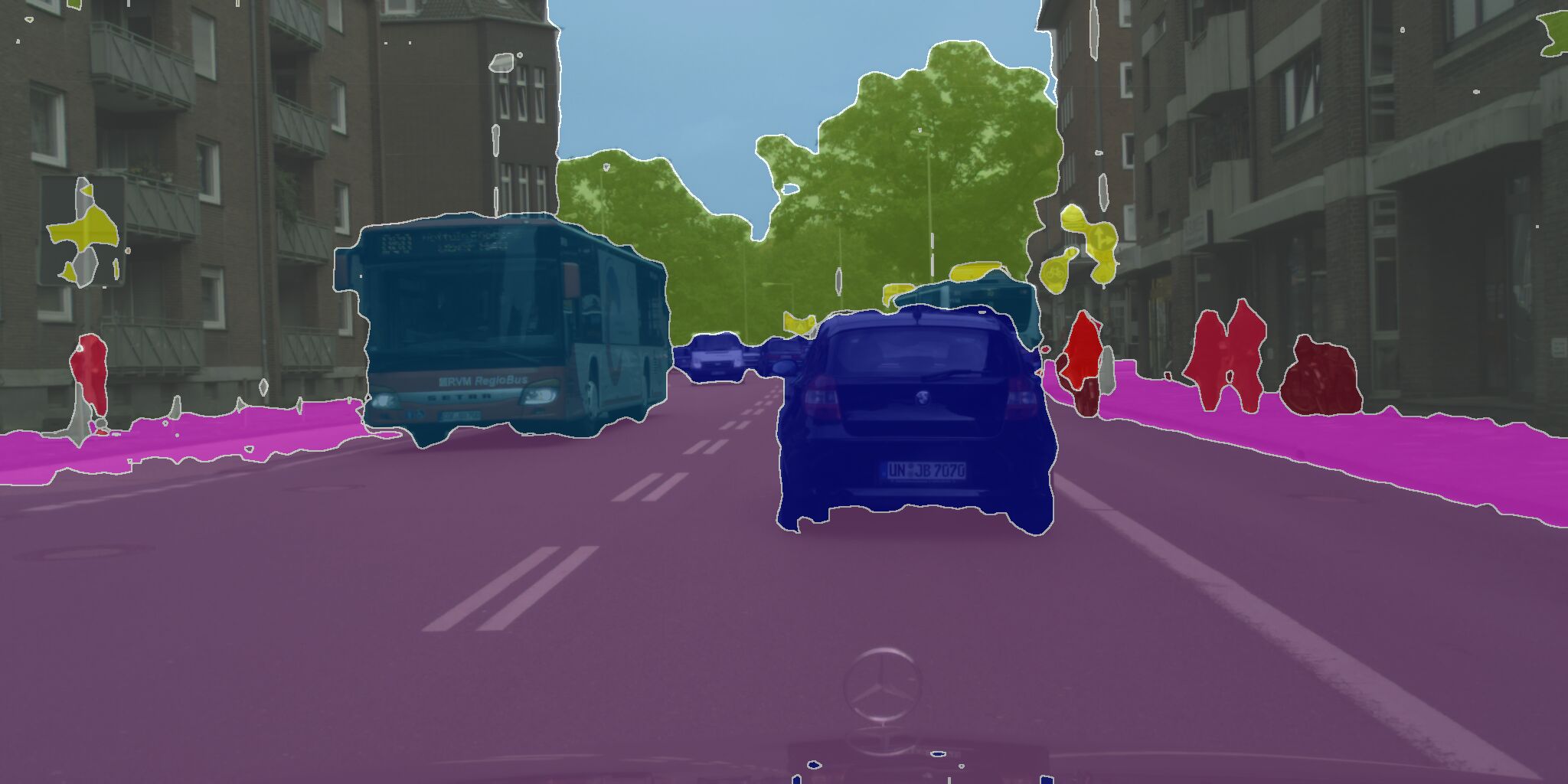} \\

    \end{tabularx}
    \caption{\textbf{Semantic segmentation qualitative results in VOC12 and Cityscapes datasets.} \normalfont The previous version, V-JEPA 2, produced sparse semantic masks due to the presence noisy feature maps. However, V-JEPA 2.1 achieves competitive performance with state of the art methods such as DINOv3 ViT-H+.}
    \label{fig:semantic_segm_examples}
\end{figure}

\subsection{Depth Estimation and Semantic Segmentation} \label{sec:depth_segmentation}

We evaluate the quality of the learned representations on semantic segmentation and depth estimation, two tasks that require fine-grained understanding of the image spatial structure.

\paragraph{\bf{Task.}} 
Semantic segmentation probes the model’s ability to capture category-level semantics and object boundaries. We evaluate our model on PASCAL VOC12 \citep{everingham2015pascal}, ADE20K \citep{zhou2017scene}, and Cityscapes \citep{cordts2016cityscapes}, reporting the mean Intersection over Union (mIoU). As in \citet{simeoni2025dinov3}, the input resolution of the images is adapted to 1024 patch tokens ($i.e.$, $512 \times 512$ for patch size 16, $448 \times 448$ for patch size 14).
On the other hand, depth estimation evaluates the model’s understanding of the scene’s geometric structure. For this task, we use the NYUv2 \citep{silberman2012indoor} and KITTI \citep{geiger2013vision} benchmarks and report the Root Mean Squared Error (RMSE).

\begin{wrapfigure}{r}{0.49\textwidth}
    \centering
    \vspace{-3mm}
    \includegraphics[width=\linewidth]{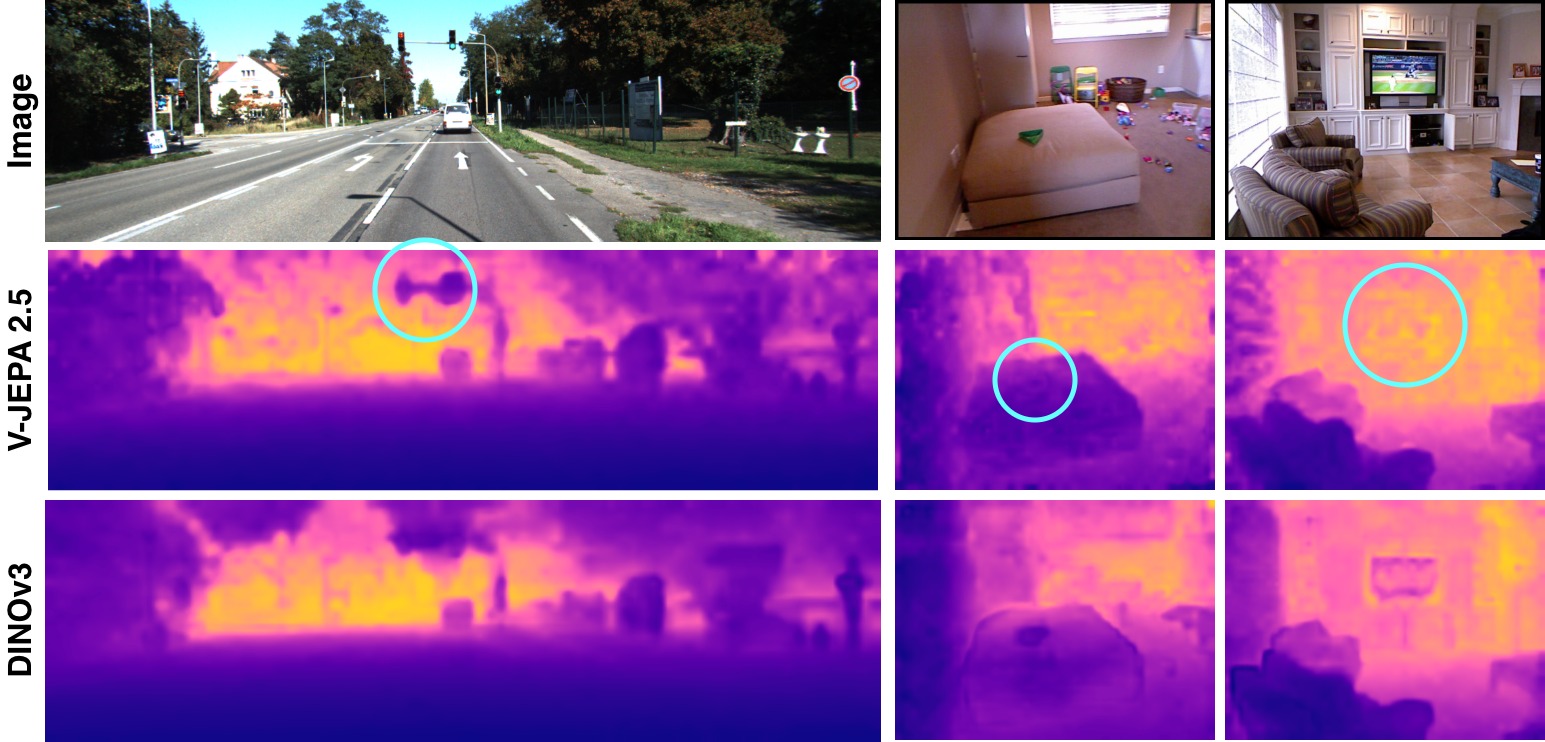}
    \caption{\textbf{Qualitative comparative of V-JEPA 2.1 ViT-G vs DINOv3 ViT-H$+$.} \normalfont V-JEPA 2.1 captures fine-grained details (i.e the traffic light contour) and obtains better scale consistency. In the two rightmost columns, DINOv3 mislabel the depth of the green object on the bed, and the TV, and detect them closer to the camera than they are.}
    % \nicolas{not sure if the correct depth is V-JEPA 2.1 or DINOv3 on the rightmost two columns.}}
    \label{fig:depth_comp}
    \vspace{-5mm}
\end{wrapfigure}

\paragraph{\bf{Evaluation protocol.}}
Following the protocol of DINOv3 \citep{simeoni2025dinov3}, we train a dense linear projection on top of the frozen final-layer features of the V-JEPA 2.1 encoder using the image patchifier. 
Importantly, we do not utilize intermediate layers from the encoder. 
As V-JEPA 2.1 adopts a 3D RoPe for positional encoding, the frequencies are interpolated according to the image resolution.
%Both training and evaluation of the linear layers are performed exclusively using task-spec images from the datasets.

\paragraph{\bf{Results.}} 
Table~\ref{tab:segmentation_depth_extended} summarizes the performance of V-JEPA 2.1 and other visual encoders on depth estimation and single-image semantic segmentation.
V-JEPA 2.1 ViT-G achieves state-of-the-art results on linear-probe monocular depth estimation, reaching 0.307 RMSE on NYUv2 and strong performance on KITTI with 2.461 RMSE, performing best across models that have less than 2 billion parameter. 
On NYUv2, V-JEPA 2.1 surpasses prior image encoders, including DINOv3 ViT-7B (0.309 RMSE on NYU) and PE\textsubscript{spatial}-G (0.362 RMSE on NYU), and it represents a significant improvement over video encoders such as InternVideo2-1B (0.471 RMSE) and V-JEPA 2 (0.642 RMSE).
Figure~\ref{fig:depth_qual} provides a qualitative comparison of the depth maps predicted by V-JEPA 2 and V-JEPA 2.1. While V-JEPA 2 captures the overall scene geometry, the inconsistencies of its local features lead to noisy depth maps. In contrast, our novel V-JEPA 2.1 produces sharper and more coherent depth maps with well-defined object boundaries. 
We also visualize a more detailed qualitative comparative with DINOv3 ViT-H+ in Figure~\ref{fig:depth_comp}.% , where V-JEPA 2.1 ViT-G produces fewer depth hallucinations and produces improved scale consistency.

In semantic segmentation, V-JEPA 2.1 is also highly competitive with the state-of-the-art models, obtaining 85.0 mIoU on VOC12, 73.5 mIoU on Cityscapes and 47.9 mIoU on ADE20K. Compared with its previous version V-JEPA 2, the gains are remarkable across all datasets: +23.4 points in ADE20K, +27.6 on Cityscapes, and + 20.7 on VOC12; showing the benefits of explicitly supervising the context tokens and incorporating the multi-level predictor.

Performance on ADE20K and Cityscapes remains slightly behind the best image encoders. These datasets contain numerous object classes spanning large scale variations, with cluttered scene layouts that impose strict demands on fine-grained segmentation. We hypothesize that VisionMix contains comparatively fewer highly cluttered scenes, limiting exposure to the level of granularity required by such benchmarks.
Figure~\ref{fig:semantic_segm_examples} illustrates segmentation predictions on Cityscapes and VOC12. V-JEPA 2.1 yields detailed and spatially accurate masks, capturing multi-scale structures and fine object contours with high fidelity, showing competitive performance with state-of-the-art methods such as DINOv3 ViT-H+ \citep{simeoni2025dinov3}.

%Relative to video encoders such as InternVideo, V-JEPA 2.1 also shows consistently superior performance \TODO{END THIS PART}.

\subsection{Video Object Segmentation} \label{sec:video_segmentation}

We evaluate V-JEPA 2.1 on Video Object Segmentation (VOS) to assess the temporal consistency of its learned features.

\begin{figure}[t]
    \centering
    \begin{tabularx}{\textwidth}{Y Y Y Y Y Y}
        \textbf{Initial GT frame} & \multicolumn{5}{c}{\textbf{Frames sequence}} \\
        \includegraphics[width=\linewidth]{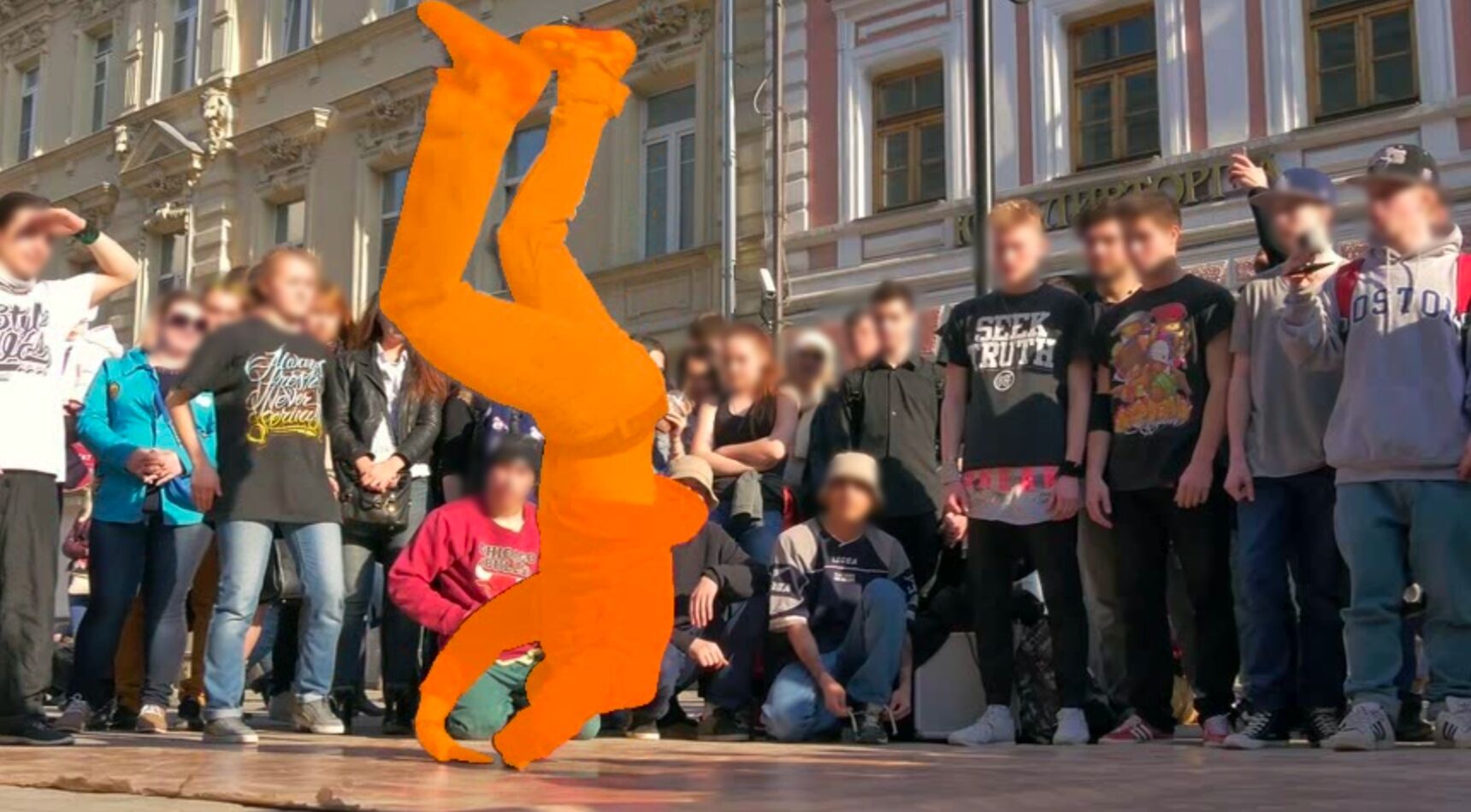} &
        \includegraphics[width=\linewidth]{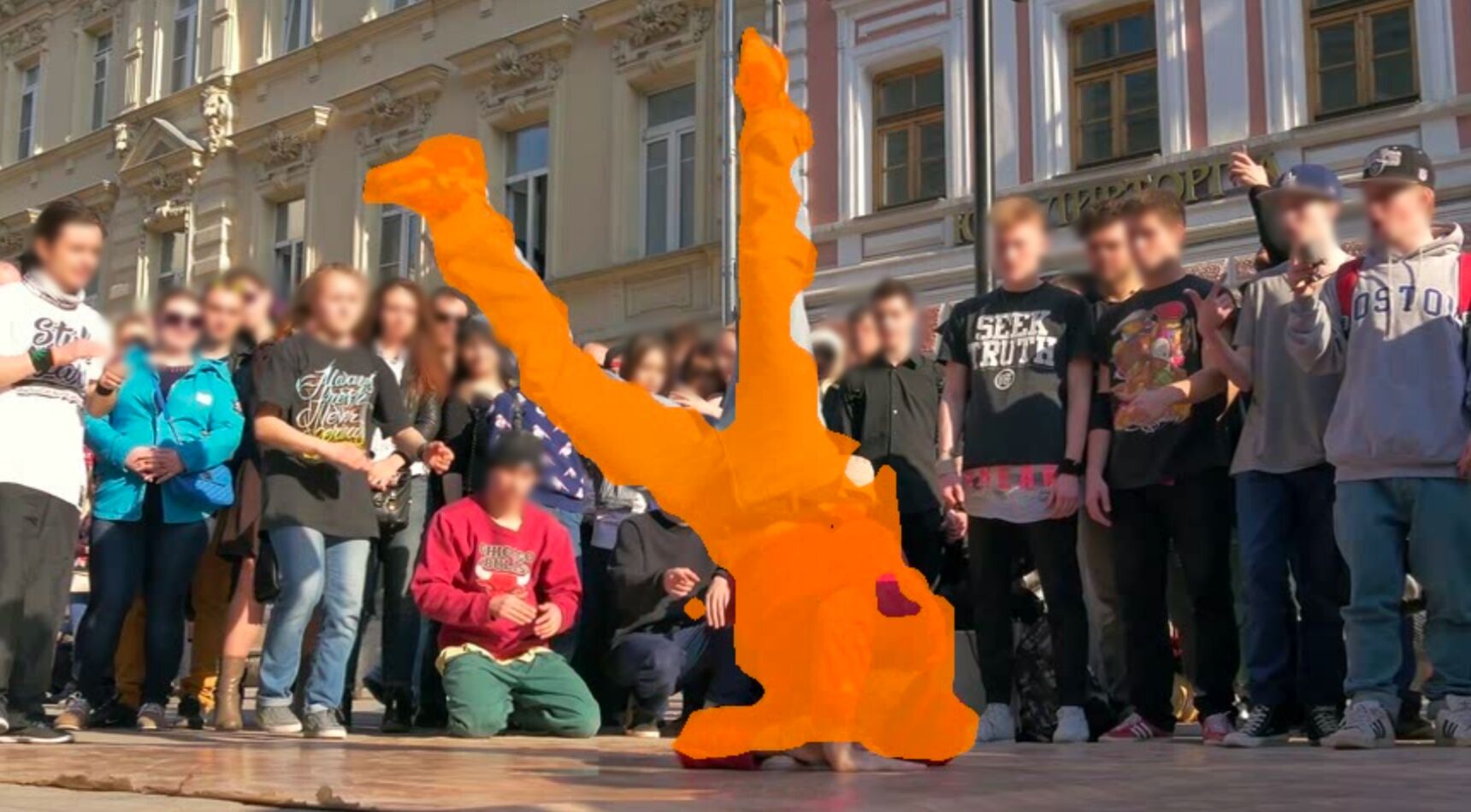} &
        \includegraphics[width=\linewidth]{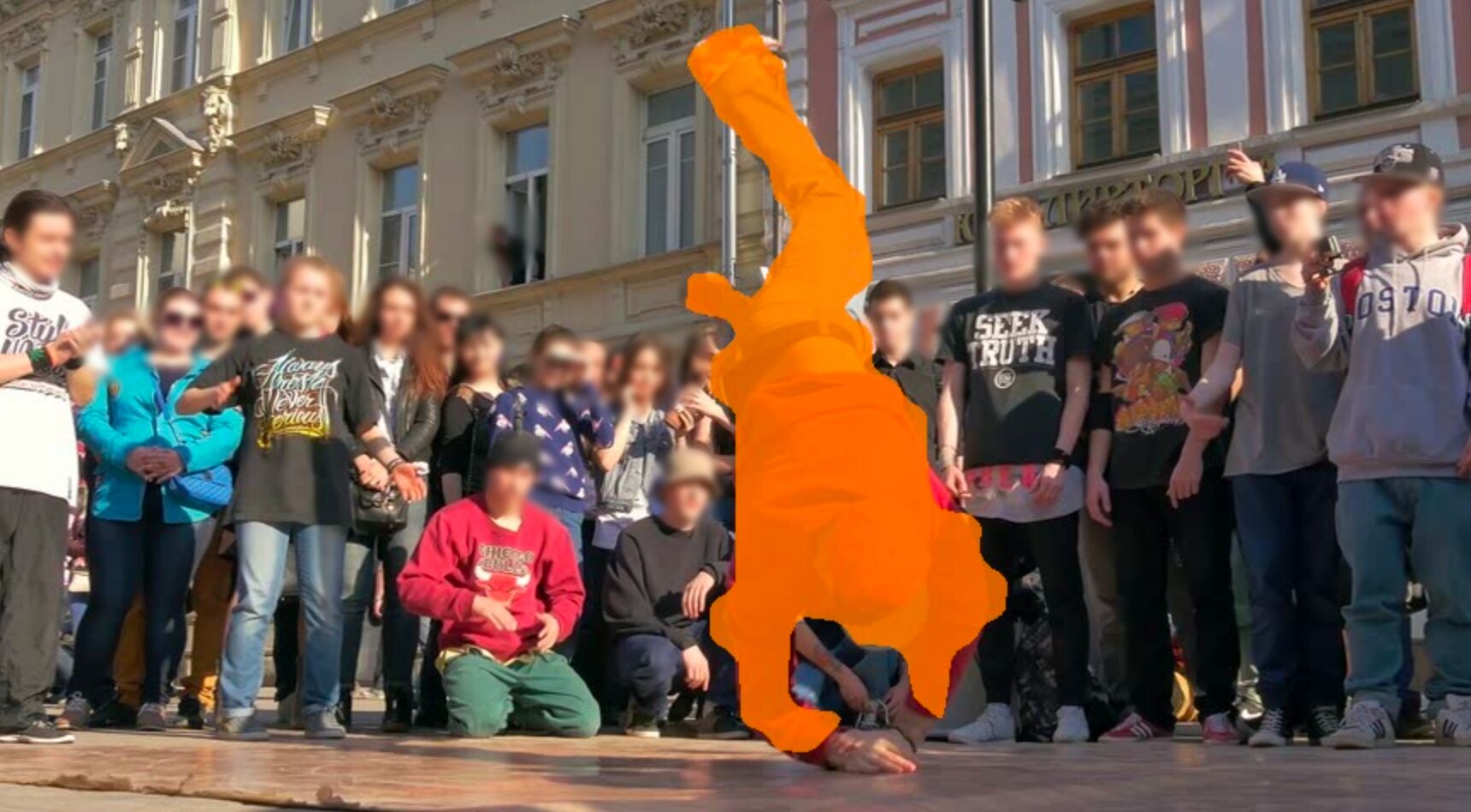} &
        \includegraphics[width=\linewidth]{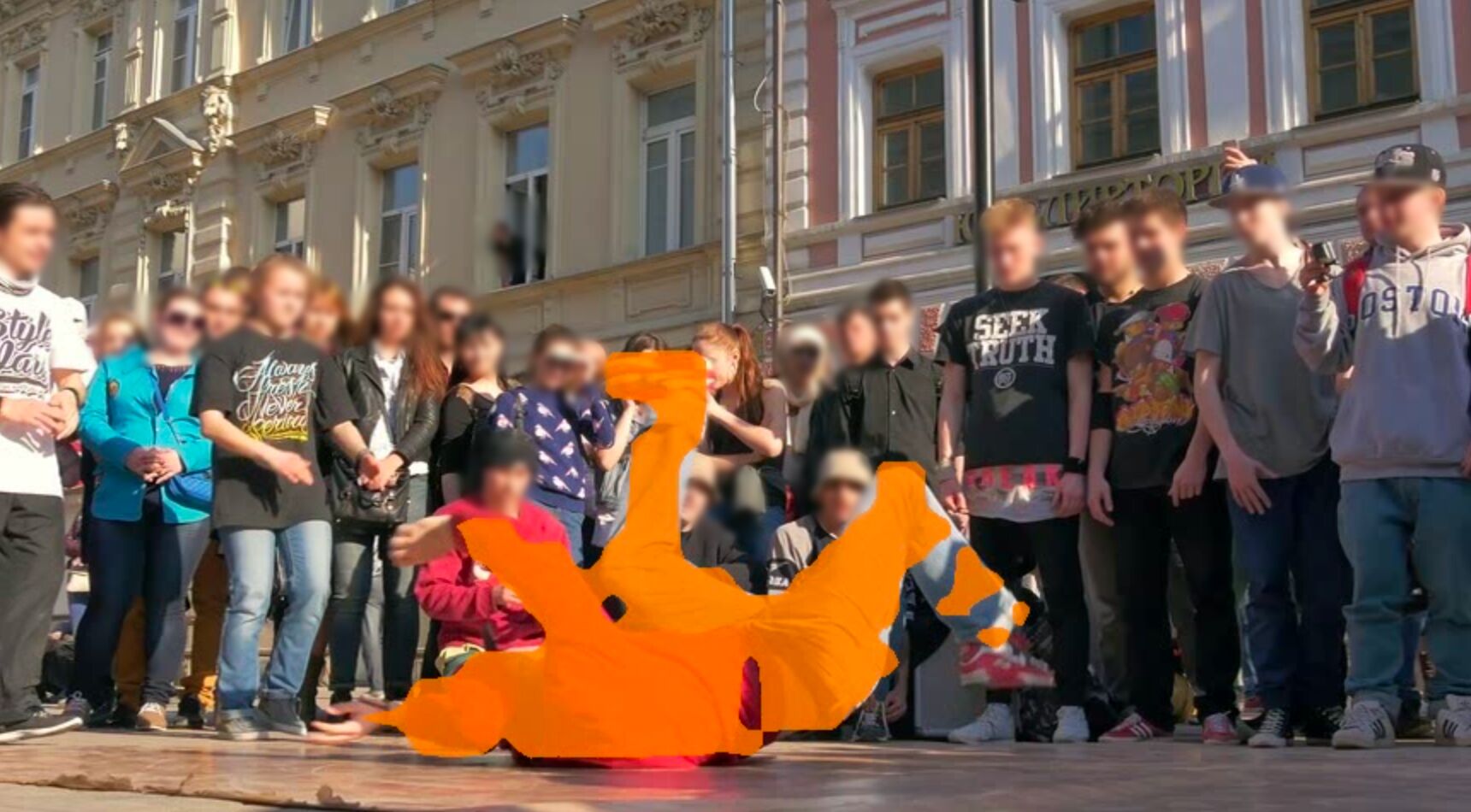} &
        \includegraphics[width=\linewidth]{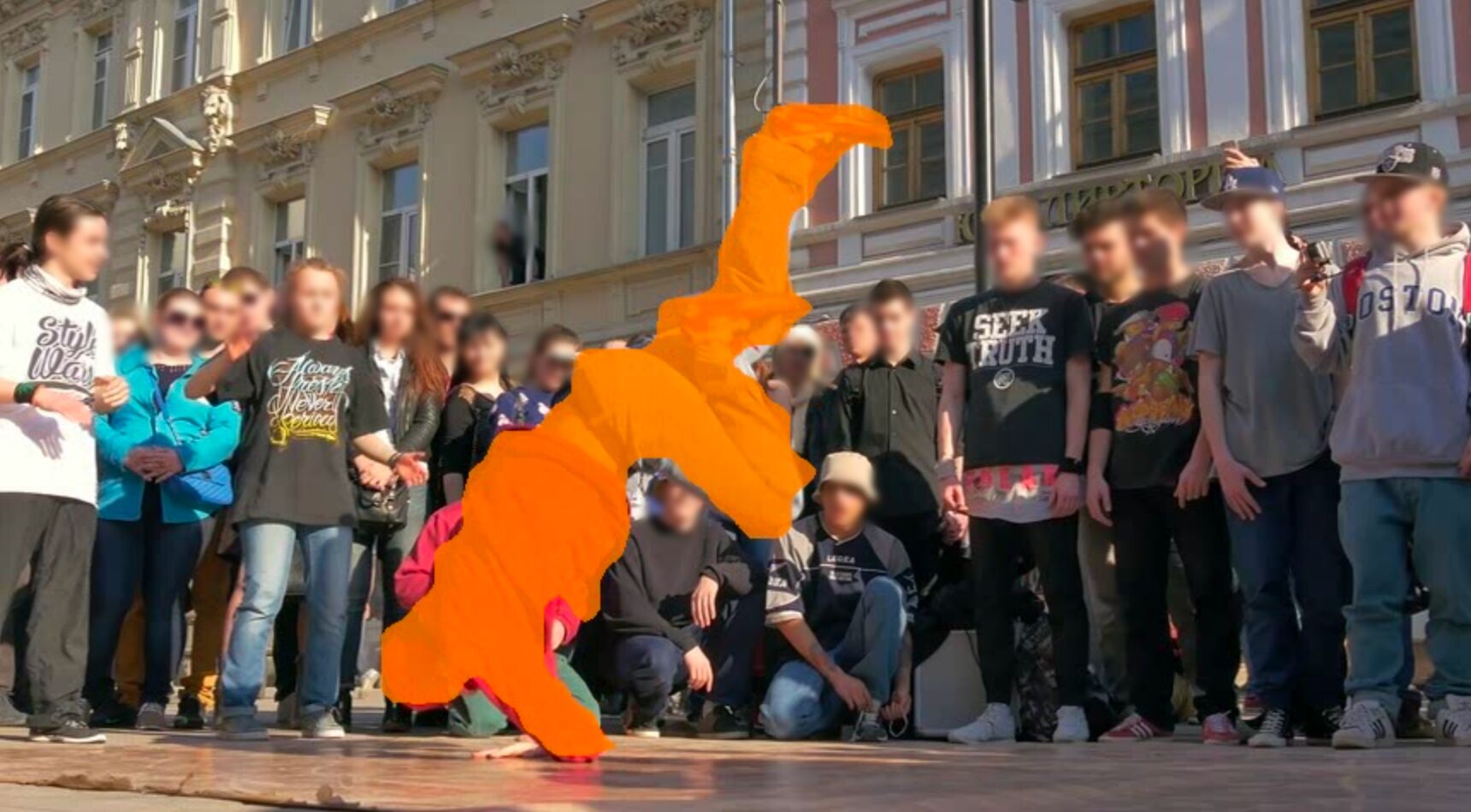} &
        \includegraphics[width=\linewidth]{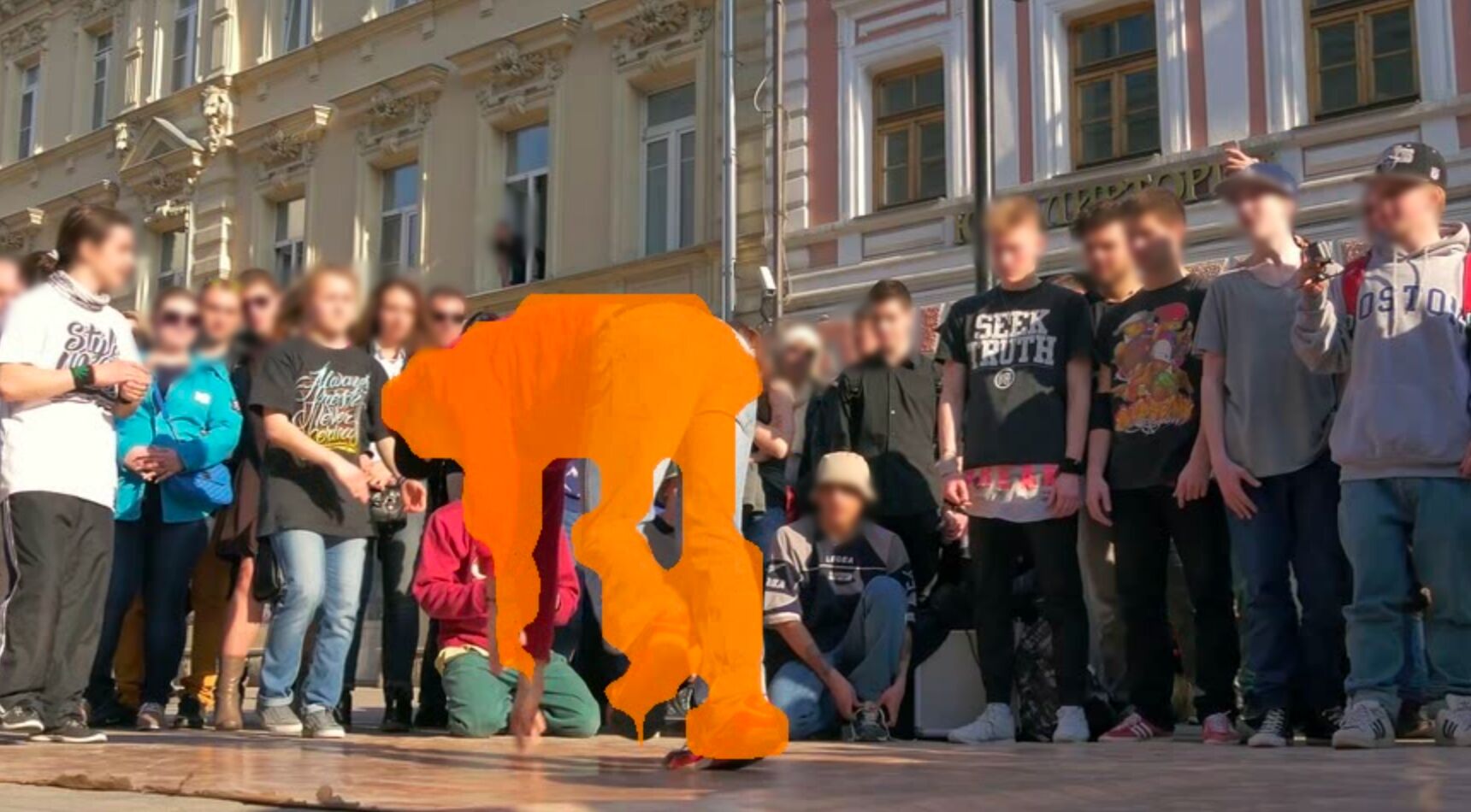} \\
        \includegraphics[width=\linewidth]{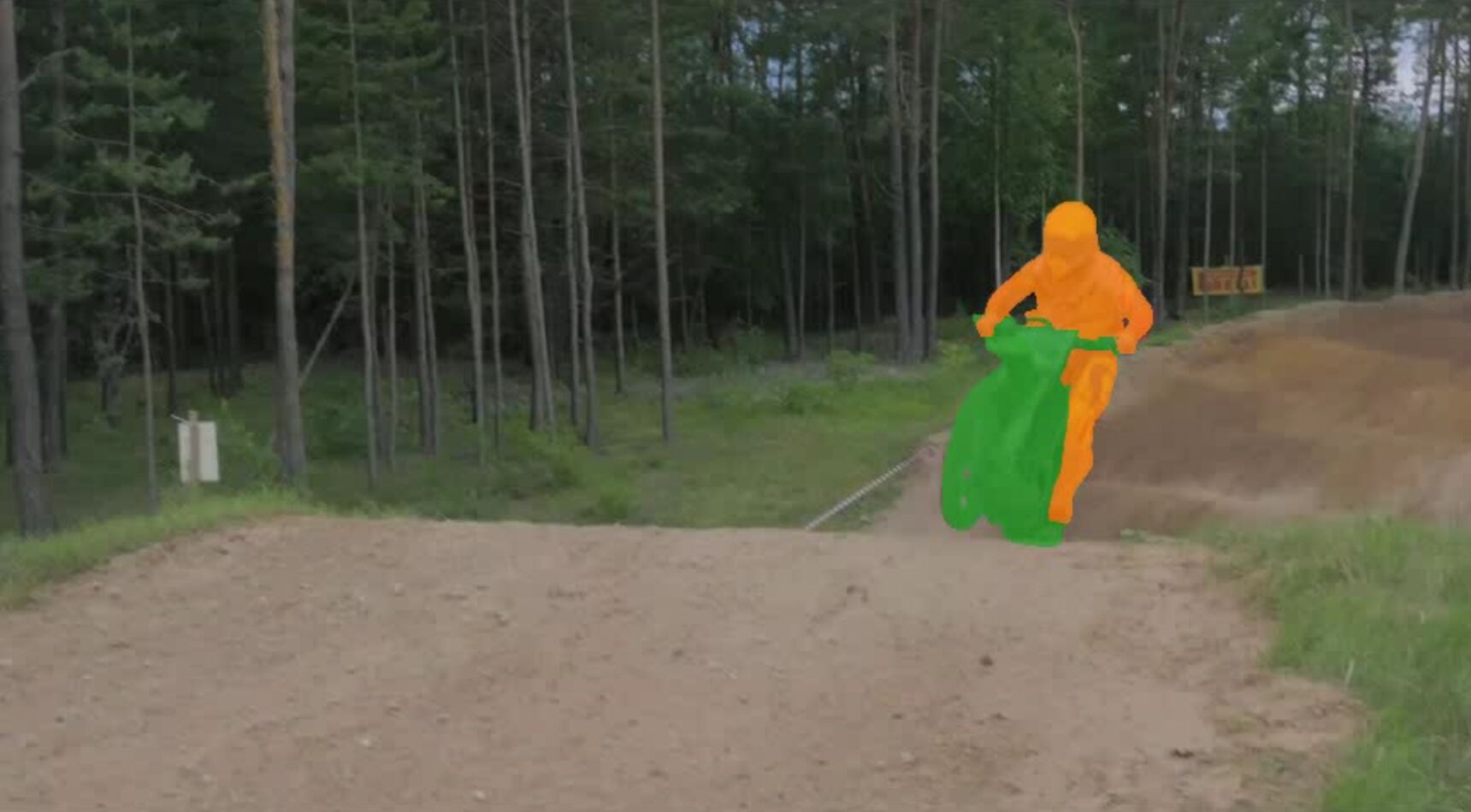} &
        \includegraphics[width=\linewidth]{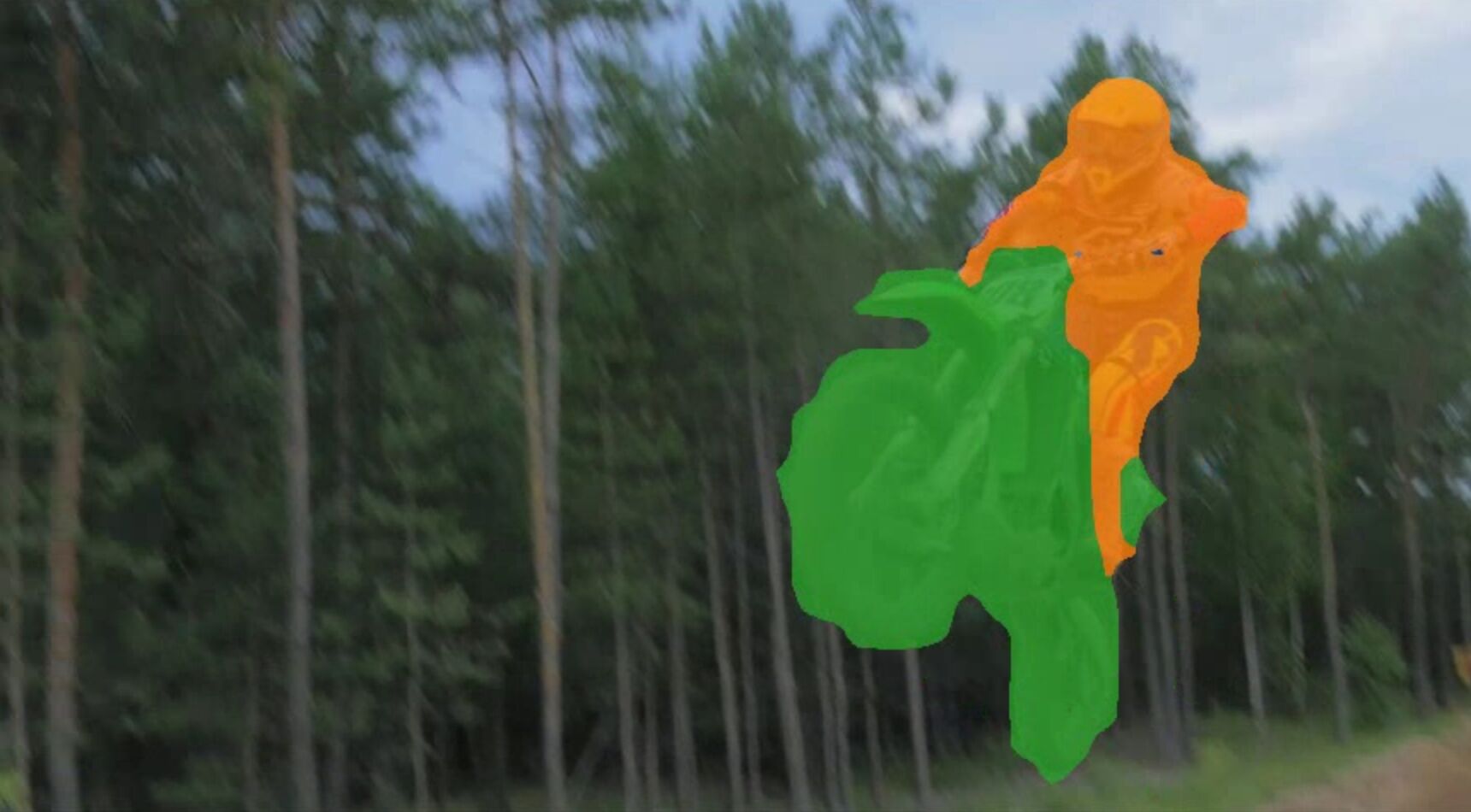} &
        \includegraphics[width=\linewidth]{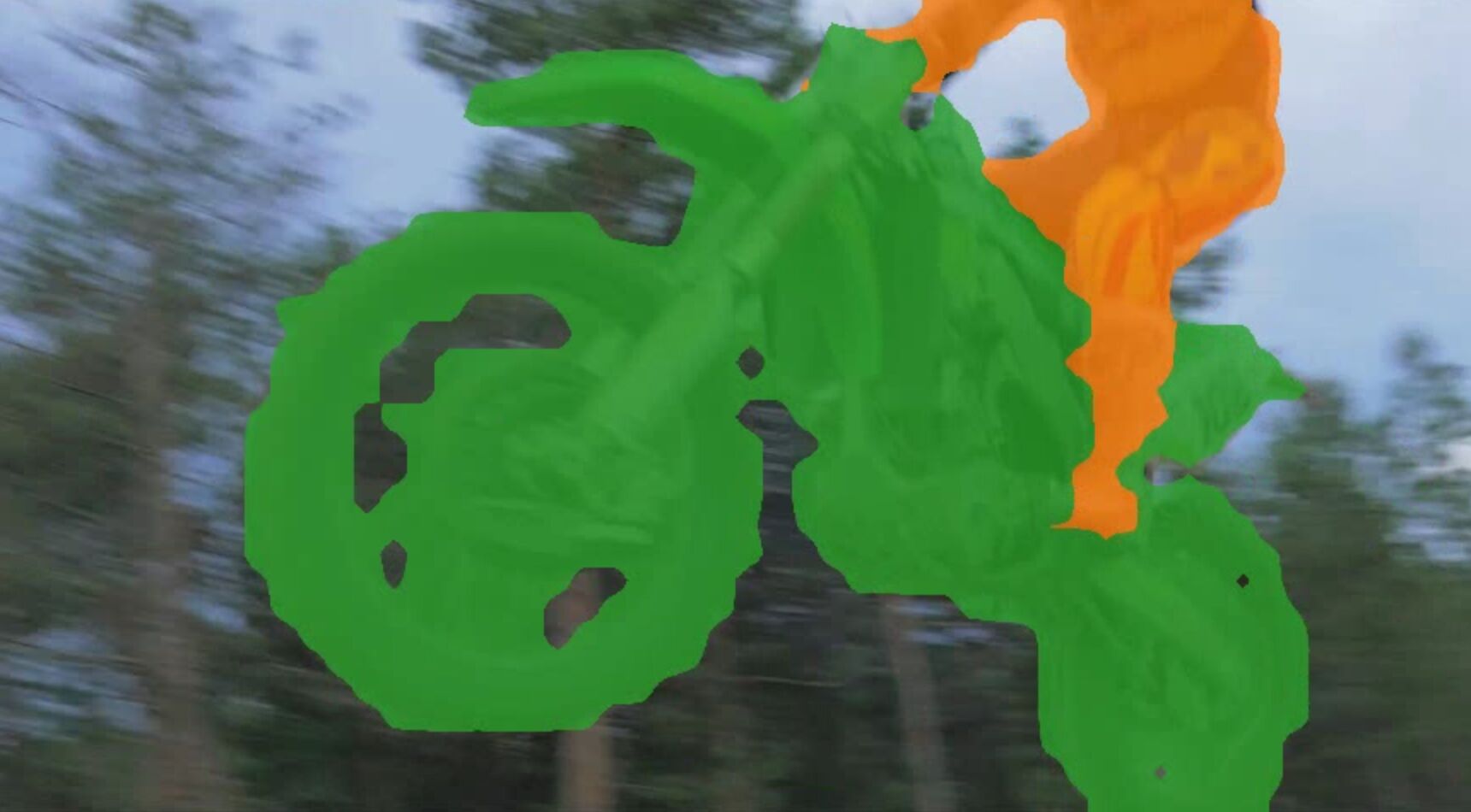} &
        \includegraphics[width=\linewidth]{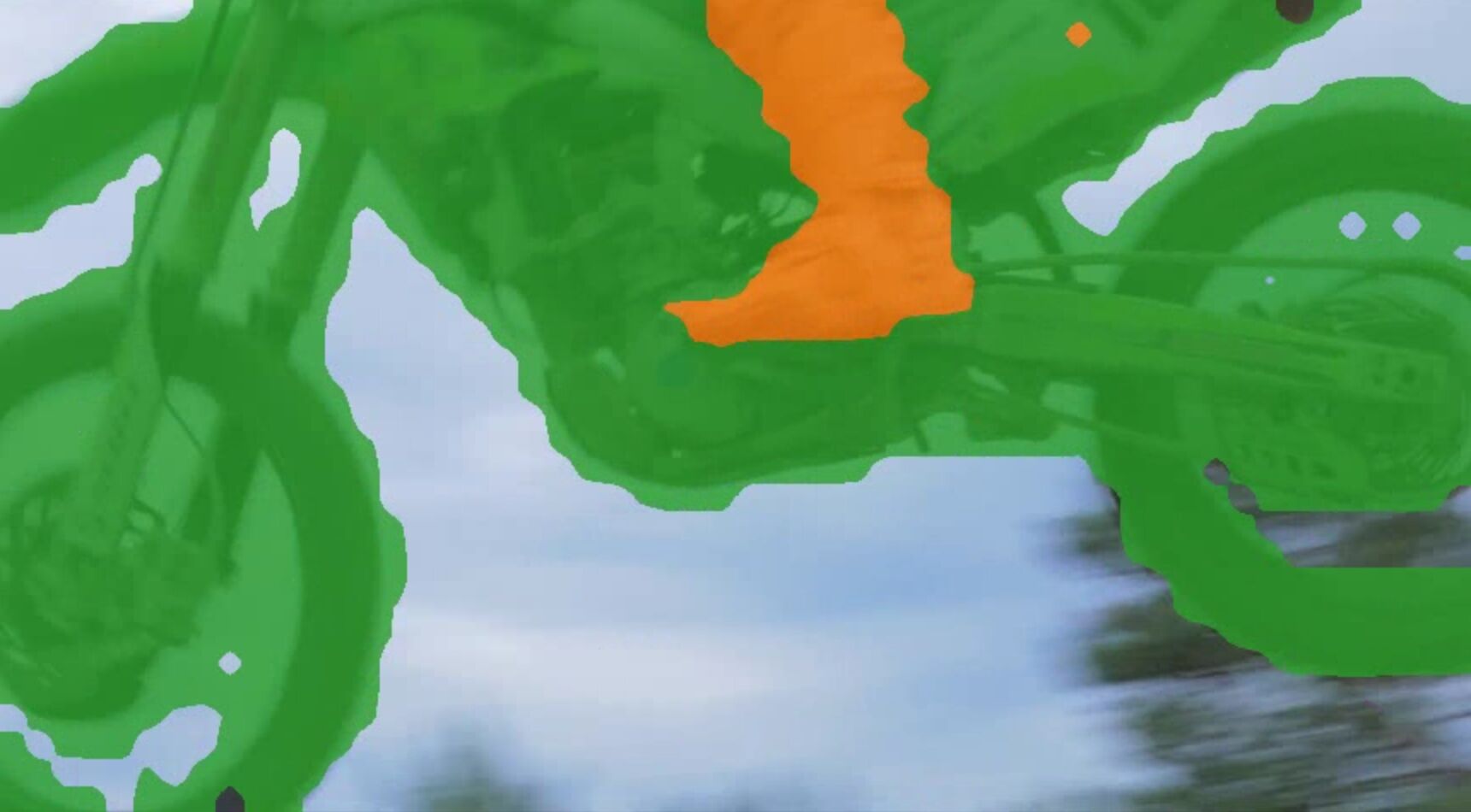} &
        \includegraphics[width=\linewidth]{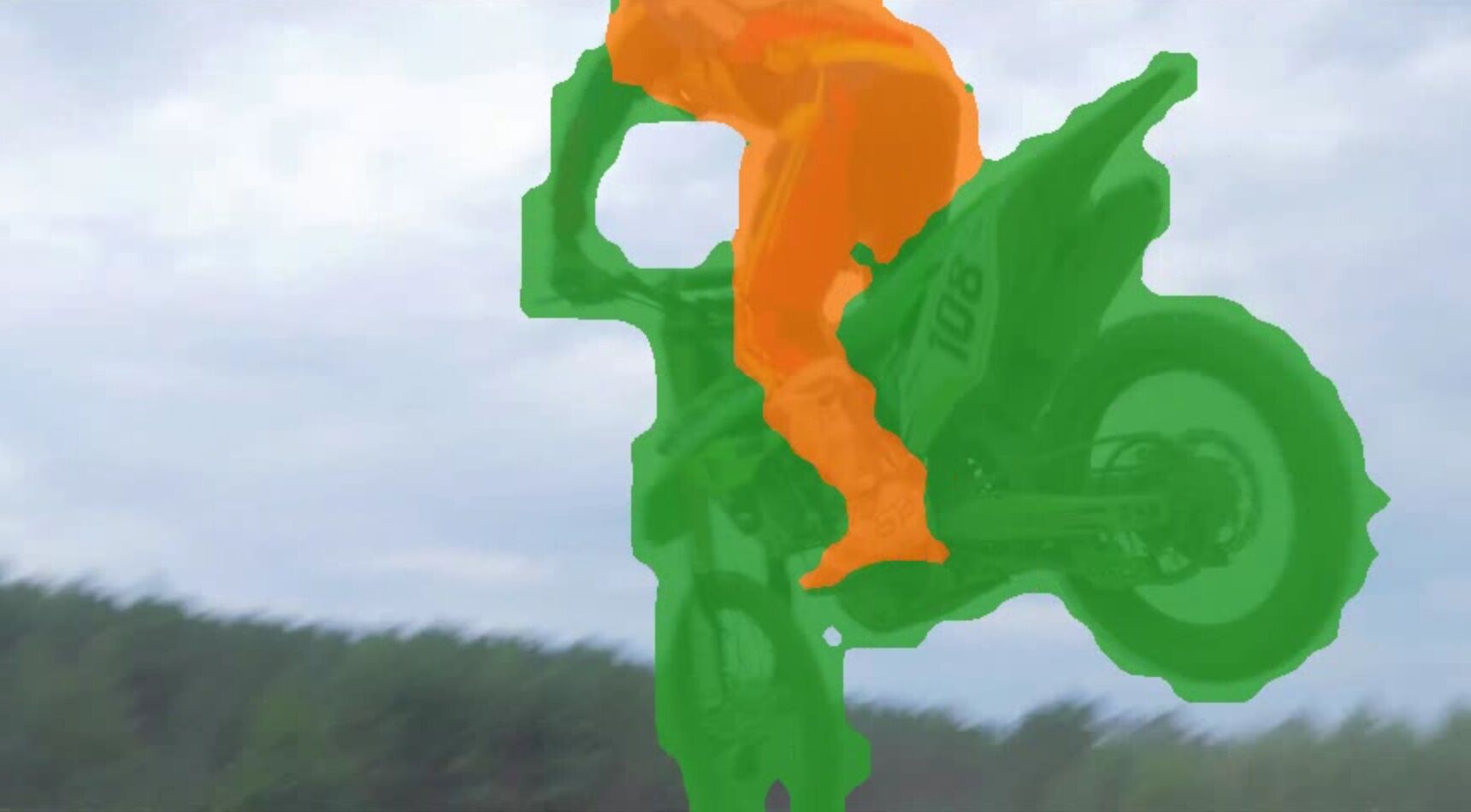} &
        \includegraphics[width=\linewidth]{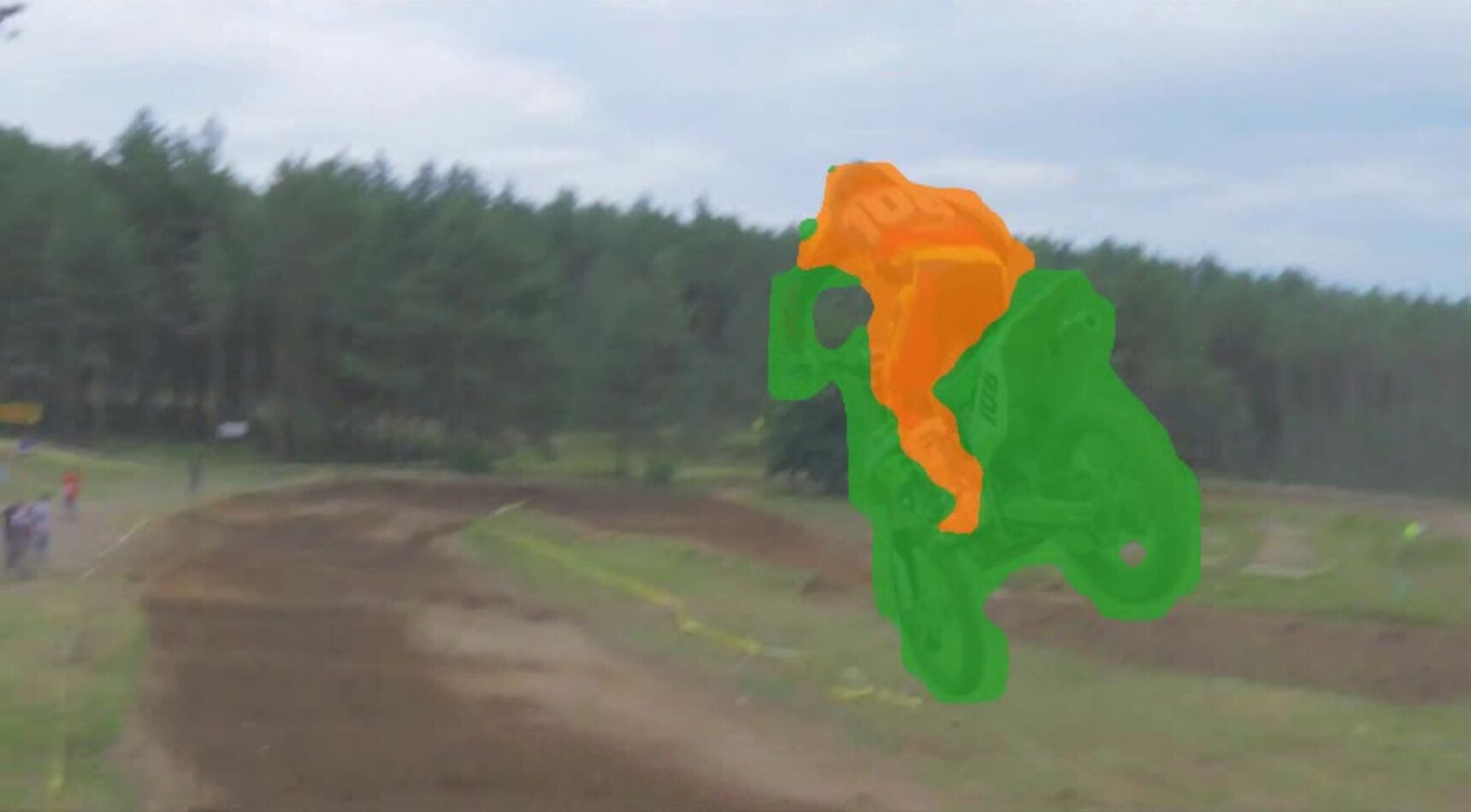} \\
    \end{tabularx}
    \caption{\textbf{Video Object Segmentation qualitative examples.} \normalfont Given the ground-truth object masks for the initial frame, we propagate the instance masks to subsequent frames based on patch similarity within the V-JEPA 2.1 embedding space. All videos are processed at a resolution where the shorter side is set to 480 pixels.}
    \label{fig:vos_examples}
\end{figure}

\paragraph{\bf{Task.}} 
Video Object Segmentation consists in, given the ground-truth object mask in the first frame, propagating this mask accurately across all subsequent frames. This task evaluates the model’s ability to preserve long-range correspondences while remaining robust to camera motion, object deformation, and visual distractions.
We evaluate on DAVIS 2017 \citep{pont20172017} and YouTube-VOS \citep{xu2018youtube} datasets, reporting the standard $\mathcal{J}\&\mathcal{F}$-mean metric \citep{perazzi2016benchmark} that jointly measures the region similarity and contour accuracy.  

\paragraph{\bf{Evaluation protocol.}}
We adopt a non-parametric label propagation approach \citep{jabri2020space} that matches local patch features across frames using cosine similarity in the embedding space. This procedure introduces no learnable parameters, making it a direct probe of the representation’s temporal stability.
Following \cite{simeoni2025dinov3}, input videos are resized to produce a consistent number of patch tokens (short side of 420 px for patch size 14, and 480 px for patch size 16).
On the training split of each dataset, we conduct a systematic search of the best hyper-parameters: maximum context length, neighborhood mask size, number of top-$K$ nearest neighbors, and the temperature parameter used in similarity computation. The best configuration in DAVIS-S (15 context frames, circle mask of size 12, top-5 neighbors and temperature = 0.2), was applied to all the validation sets.

\paragraph{\bf{Results.}}

V-JEPA 2.1 achieves 69.0 $\mathcal{J}\&\mathcal{F}$ on DAVIS-17 and 72.7 $\mathcal{J}\&\mathcal{F}$ on YouTube-VOS datasets, obtaining the second best performance and surpassing all prior encoders except DINOv3, which attains a slightly higher score.
These results highlight the temporal consistency of the V-JEPA 2.1 features, which enable stable object tracking despite significant appearance changes across frames. 
Figure~\ref{fig:vos_examples} illustrates two qualitative examples: even under fast motion and substantial visual variations, V-JEPA 2.1 maintains consistent object segmentation masks throughout the sequence.

%\begin{table}[t]
%\centering
%\setlength{\tabcolsep}{6pt}
%\renewcommand{\arraystretch}{1.2}
%\begin{tabular}{lccccc}
%\toprule
%\bf{Model} &  &\bf{mAP N} & \bf{mAP N + V} & \bf{mAP N + ttc} & \bf{Overall} \\
%\midrule
%StillFast \citep{ragusa2023stillfast} &  &25.01& 13.29& 9.14& 5.12\\
%GANO v2 \citep{thakur2023guided} &  &25.67& 13.60& 9.02&5.16\\
%STAformer \citep{mur2024aff} &  &33.50& 17.25& 11.77& 6.75\\
%EgoVideo \citep{pei2024egovideo} &  &31.08& 16.18& 12.41&7.21\\
%\midrule
%V-JEPA 2.1 ViT-G\textsubscript{384} &   &&  &  & \\
%\bottomrule
%\end{tabular}
%\caption{\bf{Comparative of model performance on the Ego4D STA test split.} The test split is in a private server.}
%\label{tab:staformer_results_extended}
%\end{table}

\subsection{Video and Image Classification} \label{sec:probe_classification}

\begin{table*}[t]
\centering

\caption{\textbf{Performance on global tasks.} \normalfont
We report Top-1 classification accuracy on action recognition benchmarks (SSv2, Diving-48, and K400) and image classification (IN1K). 
Our protocol uses a resolution of $256 \times 256$ by default, and $384 \times 384$ for V-JEPA 2 ViT-g and V-JEPA 2.1; and uses $16 \times 2 \times 3$ inputs for SSv2 (16-frame clips, 2 temporal crops, 3 spatial crops), $16 \times 8 \times 3$ for K400, and $32 \times 4 \times 3$ for Diving-48. V-JEPA 2.1 VIT-G achieves state-of-art performance on the SSv2 action recognition asks while being competitive on appearance understanding tasks.}

\resizebox{\columnwidth}{!}{%
\begin{tabular}{lccccc}
\toprule
\bf{Method} & \bf{Param.} &
\multicolumn{2}{c}{\bf{Motion Understanding}}&
\multicolumn{2}{c}{\bf{Appearance Understanding}}\\
\cmidrule(lr){3-4} \cmidrule(lr){5-6}
& & SSv2 & Diving-48 & K400 & IN1K \\
\midrule
\multicolumn{6}{l}{\textit{Results Reported in the Literature}}\\
VideoMAEv2~\citep{wang2023videomae} & 1B & 56.1 & -- & 82.8 & 71.4 \\
InternVideo2-1B \citep{wang2024internvideo2} & 1B & 67.3 & -- & 87.9 & -- \\
%\bf{InternVideo2-6B} \citep{wang2024internvideo2} & 6B & -- & 67.7 & -- & -- & 88.8 & -- & -- \\
VideoPrism \citep{zhao2024videoprism} & 1B & 68.5 & 71.3 & 87.6 & -- \\
DINOv3 ViT-H+~\citep{simeoni2025dinov3} & 0.8B & 69.8& --& 86.7& 87.9\\
DINOv3~\citep{simeoni2025dinov3} & 7B & 70.1& -- & 87.8& \bf{88.4} \\
\midrule

\multicolumn{6}{l}{\textit{Image Encoders Evaluated Using the Same Protocol}}\\
DINOv2~\citep{darcet2023vision} & 1.1B & 50.7 & 82.5 & 83.6 & 86.1 \\
PE\textsubscript{core} G \citep{bolya2025perception} & 1.9B & 55.4 & 76.9 & 88.5 & 87.6$^*$ \\
SigLIP2~\citep{tschannen2025siglip} & 1.2B & 49.9 & 75.3 & 87.3 & 88.0 \\
%\bf{DINOv3 ViT-H+}\citep{simeoni2025dinov3} & 0.8B & 57.2& 68.3& 84.9  & 88.1\\
\midrule

\multicolumn{6}{l}{\textit{Video Encoders Evaluated Using the Same Protocol}}\\
InternVideo2\textsubscript{2s}-1B \citep{wang2024internvideo2} & 1B & 69.7 & 86.4 & \bf{89.4} & 85.8 \\
V-JEPA ViT-H \citep{bardes2024revisiting} & 600M & 74.3 & 87.9 & 84.5 & 80.0 \\
% V-JEPA 2 ViT-g \citep{assran2025v} & 1B & 75.3 & 90.1 & 86.6 & 84.6 \\
V-JEPA 2 ViT-g \citep{assran2025v} & 1B & 77.3 & \bf{90.2} & 87.3 & 85.1 \\
%\bf{V-JEPA 2.1 ViT-L} & 300M & 84.2 & 76.1 & 90.1 & 86.6 & 84.1 \\
%\bf{V-JEPA 2.1 ViT-g\textsubscript{384}} & 1B & 85.3 & \bf{77.4} & \bf{90.6} & 88.1 & 85.3 \\
\hdashline
%\bf{V-JEPA 2.1 ViT-L} & 1B & 74.2 & -- & -- & -- & 81.8 \\
%\bf{V-JEPA 2.1 ViT-g} & 1B & 76.7& 88.6& 86.2 & 84.1 \\
V-JEPA 2.1 ViT-g & 1B & 76.9& 89.0& 87.0& 84.8 \\
%\bf{V-JEPA 2.1 ViT-G} & 2B & 77.1& 89.5& 87.3 & 84.8 \\
V-JEPA 2.1 ViT-G & 2B & \bf{77.7}& 89.2 & 87.7& 85.5 \\
\bottomrule
\end{tabular}
}

\label{tab:motion_appearance_understanding}
\end{table*}

Real-world video reasoning requires recognizing static visual cues (i.e, objects, textures, scene layouts) as well as dynamic patterns (i.e, gestures, hand-object interactions, camera motion, long-term temporal evolution). 
To capture this dual nature, we evaluate the representations learned during pretraining on four complementary classification datasets. 
Something-Something v2 \citep{goyal2017something} and Diving-48 \citep{li2018resound} assess motion-centric understanding, as their labels require modeling multi-frame dynamics to correctly identify the action.
In contrast, Kinetics-400 \citep{kay2017kinetics} and ImageNet-1K \citep{deng2009imagenet} primarily measure appearance-based recognition, since many of their classes can be predicted from a single frame, even when the label describes an action.

\paragraph{\bf{Task.}}
We compare the video classification performance of V-JEPA 2.1 against a broad set of visual encoders. 
For image-based self-supervised models, we include DINOv2 with registers \citep{darcet2023vision} and the more recent DINOv3 \citep{simeoni2025dinov3}, both representing state-of-the-art in image-only pretraining. 
We further evaluate against leading image–text contrastive approaches, including SigLIP2 \citep{tschannen2025siglip} and the Perception Encoder PE\textsubscript{core}G \citep{bolya2025perception}. For video models, we benchmark against the previous V-JEPA \citep{bardes2024revisiting} and V-JEPA 2 \citep{assran2025v}, as well as InternVideo2s2-1B \citep{wang2024internvideo2}, a state-of-the-art video encoder trained primarily through vision–text contrastive objectives.

In addition, we report comparisons with results available in the literature for VideoMAEv2 \citep{wang2023videomae}, InternVideo2-1B \citep{wang2024internvideo2}, VideoPrism \citep{zhang2024video}, and DINOv3 \citep{simeoni2025dinov3}. Note that these models are evaluated under similar frozen-probe protocols, but their attentive heads differ from ours. For instance, DINOv3 augments its probe with explicit spatial and temporal positional embeddings and applies a 3D factorized RoPE across its attention blocks. In contrast, our V-JEPA 2.1 encoder already encodes spatio–temporal structure in its features, allowing our probe to operate without such additional components.

\paragraph{\bf{Evaluation protocol.}} 
Following the protocol proposed by \cite{assran2025v}, we train a 4-layers attentive probe on top of the frozen encoder features using the respective training data from each task. 
The attentive probe consists of four transformer blocks followed by a final cross-attention mechanism that employs a learnable query token.
At inference, we sampled several clips with a fixed number of frames from the video, and we averaged the classification logits across clips.

\paragraph{\bf{Results.}}
Table~\ref{tab:motion_appearance_understanding} summarizes the global video understanding performance of V-JEPA 2.1, previous V-JEPA models, and other strong visual encoders. V-JEPA 2.1 ViT-G achieves a top-1 accuracy of 77.7 on SSv2, setting a new absolute state-of-the-art on the task compared to 77.5 for InternVideo2 full fine-tuning, 77.3 for V-JEPA 2, 70.1 for DINOv3 ViT-7B and 69.7 for InternVideo2 using the same protocol, demonstrating a strong understanding of video dynamics. Our model is also highly-competitive in appearance-based tasks, reaching 87.7 on K400 and 85.5 on IN1K. 
These results show that properly designing the loss over context tokens, together with deep self-supervision, enables fine-grained dense features that also capture strong global scene understanding.
We also observe consistent improvements with model scaling from ViT-g to ViT-G with an average improvement of +0.6 points.

\subsection{Video Question Answering} 
\label{sec:vqa}

In this section, we evaluate V-JEPA 2.1 on Video Question Answering (VidQA), by training a Video Large Language Model (VidLLM) using V-JEPA 2.1 encoder and Llama 3.1 8B LLM as the backbone, following the recipe proposed by \citet{assran2025v}.

\begin{table}[t]
    \centering
    \small
    \caption{\small \textbf{VideoQA results.} We report the performance of VJEPA 2.1 trained with Llama 3.1 8B backbone on several popular temporal video understanding datasets. We train on a modified subset of the PerceptionLM \citep{cho2025perceptionlm} data, and compare to VJEPA 2 \citep{assran2025v} by training it on the same data regime. VJEPA 2.1 achieves comparatively better performance than VJEPA 2 on several video understanding tasks which require both rich visual information and temporal understanding. For PerceptionTest \citep{patraucean2023perception}, we report the validation accuracy, as the test server to compute the test accuracy (indicated by *)  is no longer active. }
    %\footnote{\href{https://eval.ai/web/challenges/challenge-page/2091/overview}{https://eval.ai/web/challenges/challenge-page/2091/overview}}. 
    
    \label{tab:llm_sota}
    {\fontsize{8pt}{8pt}\selectfont
    \begin{tabular}{l c  c c c c c c c c}
      \bf Method & \tiny \parbox{2cm}{\centering{\bf Params} \\ Enc / LLM} & \tiny Avg. & \tiny \rotatebox[]{90}{\parbox{2cm}{\centering {\bf PerceptionTest} \\ Test Acc}} & \tiny \rotatebox[]{90}{\parbox{2cm}{\centering {\bf MVP} \\ Paired-Acc}}  & \tiny \rotatebox[]{90}{\parbox{2cm}{\centering {\bf TempCompass} \\ multi-choice}}  & \tiny \rotatebox[]{90}{\parbox{2cm}{\centering{\bf TemporalBench} \\ (MBA-short QA) }}  & \tiny \rotatebox[]{90}{\parbox{2cm}{\centering{\bf TOMATO} \\ Acc}} & \tiny \rotatebox[]{90}{\parbox{2cm}{\centering{\bf TVBench} \\ Acc}} & \tiny \rotatebox[]{90}{\parbox{2cm}{\centering{\bf MVBench} \\ Acc}} \\
        \toprule
         \multicolumn{8}{l}{\bf\it $\leq 8B$ Video Language Models Results Reported in the Literature }\\[1ex]
         InternVL-2.5~{\tiny\citep{chen2024expanding}}  & 300M/7B & 52.1 & 68.9 & 39.9 & 68.3 & 24.3 & 29.4 & 61.6 & 72.6\\
         Qwen2VL~{\tiny\citep{wang2024qwen2}}  & 675M/7B & 47.0 & 66.9 & 29.2 & 67.9 & 20.4 & 31.5 & 46.0 & 67.0\\
         Qwen2.5VL~{\tiny\citep{bai2025qwen2}}  & 1B/7B & 49.7 & 70.5 & 36.7 & 71.7 & 24.5 & 24.6 & 50.5 & 69.6 \\
         PLM 8B~{\tiny{\citep{cho2025perceptionlm}}} & 1B/8B & 56.7 & 82.7* & 39.7 & 72.7 & 28.3 & 33.2 &  \bf{63.5} & \bf{77.1} \\
        VJEPA 2 ViT-g$_{384}$ {\tiny\citep{assran2025v}} & 1B/8B & \bf 59.5 & \bf{84.0}*  & \bf{44.5} & \bf{76.9} & \bf 36.7 & 40.3 & 60.6 & 73.5\\ \midrule 
        \multicolumn{8}{l}{\bf\it Models trained in this paper using filtered Perception LM data \citep{cho2025perceptionlm}}\\[1ex]
        VJEPA 2 ViT-g$_{384}$       & 1B/8B &  57.8   &  80.1  & 40.9 & \underline{74.1} & \underline{32.3} & \bf \underline{41.4}  & 62.6 & 73.3 \\
        \bf VJEPA 2.1 ViT-G$_{384}$ & 2B/8B &  \underline{57.9}   &  \underline{83.1} & \underline{43.2} & 74  & 28.5 & 38  &  \underline{62.8} & \underline{75.4} \\
        \bottomrule
    \end{tabular}
     }
\end{table}

\paragraph{\bf{Task.}} 
We compare the VidQA performance of V-JEPA 2.1 with popular open-source multimodal models reported in the literature, where the LLM backbone size is $\leq$ 8B. Following the protocol set of \citet{assran2025v}, we report performance on video understanding benchmarks which require rich visual and temporal understanding, such as PerceptionTest \citep{patraucean2023perception}, Minimal Video Pairs (MVP) \citep{benno2024}, TempCompass \citep{liu2024tempcompass}, TemporalBench \citep{cai2024temporalbench}, TOMATO \citep{shangguan2024tomato} and MVBench \citep{li2024mvbench}.
% For reference, we also include closed source models performance wherever the evaluation data is used in the model reports / ran by community.
% This sentence reads weird givent hat the authors overalp betwee 2. and 2.1
%Since model comparisons are often obfuscated by the underlying training data, we report comparisons with our own reproduction of \citet{assran2025v} for an apples-to-apples comparison.

\paragraph{\bf{Evaluation protocol.}}
We train our Video-LLM model using a subset of PerceptionLM that is publicly available~\citep{cho2025perceptionlm}.
Starting with this data, we perform data quality filtering using domain filters, vision-text misalignment filters, and an external multimodal LLM, Qwen3VL, as a data-quality judge, assigning scores to each data samples. This process resulted us to filter out and remove \textbf{18\%} of the training data. 
It is important to note that we do not rephrase any data from PerceptionLM \citep{cho2025perceptionlm} using this setup - we only filter the data based on the model's judgement. 

We reproduced LLM alignment of VJEPA 2 on this new data, and compare our model, VJEPA 2.1, with it. We follow the same three-stage training procedure from \citet{assran2025v} to train VJEPA 2.1 LLM, starting from image-text captioning to progressively training on higher quality image and video-text captioning and QA data. We further add a fourth stage of alignment, where we expand the context length of the base LLM to support 64 frames of video input, compared to 32 frames supported by VJEPA 2 \citep{assran2025v}.
We primarily use VJEPA 2.1 in video-only mode in our experiments, to better compare with VJEPA 2. Interestingly, we find adding a final post-training stage using PerceptionTest training set further improves performance on all downstream tasks.% We therefore compare VJEPA 2 and our model, VJEPA 2.1 using this additional training step.

\paragraph{\bf{Results.}}
Table \ref{tab:llm_sota} summarizes the results on VidQA evaluation tasks. Comparing VJEPA 2 and VJEPA 2.1 on the same data, we find on average VJEPA 2.1 to be slightly better, with significant improvements on PerceptionTest, Minimal Video Pairs, and crucially, improvement on datasets requiring rich visual semantics - MVBench and TVBench. We observe VJEPA 2.1 to be worse on TemporalBench and TOMATO - both datasets requiring understanding of the motion events, but on shorter videos.% Our hypothesis is that since VJEPA 2.1 LLM alignment was trained on 64 frames, it loses some of the  capability on shorter video sequences, which are typical in these datasets.

Compared to results reported in the literature,  VJEPA 2.1 slightly underperforms results reported~\citet{assran2025v} which uses significantly alignment data (88.5 million samples in~\citep{assran2023self} vs 72.5 million samples in VJEPA 2.1) 
Nevertheless,  VJEPA 2.1  is competitive with state-of-the-art, outperforming most open-source multimodal LLMs with similar model size. 
% \subsection{Dense visual features}

\begin{figure}[]
    \centering
    \renewcommand{\arraystretch}{0.2} % Espaciado entre filas
    \setlength{\tabcolsep}{1pt}       % Espaciado entre columnas
    %\begin{tabular}{ccccc}
    \begin{tabularx}{\textwidth}{CCCCC}
        \textbf{Image} & \textbf{DINOv2} & \textbf{DINOv3} & \textbf{V-JEPA 2} & \textbf{V-JEPA 2.1} \\
        \includegraphics[width=\linewidth]{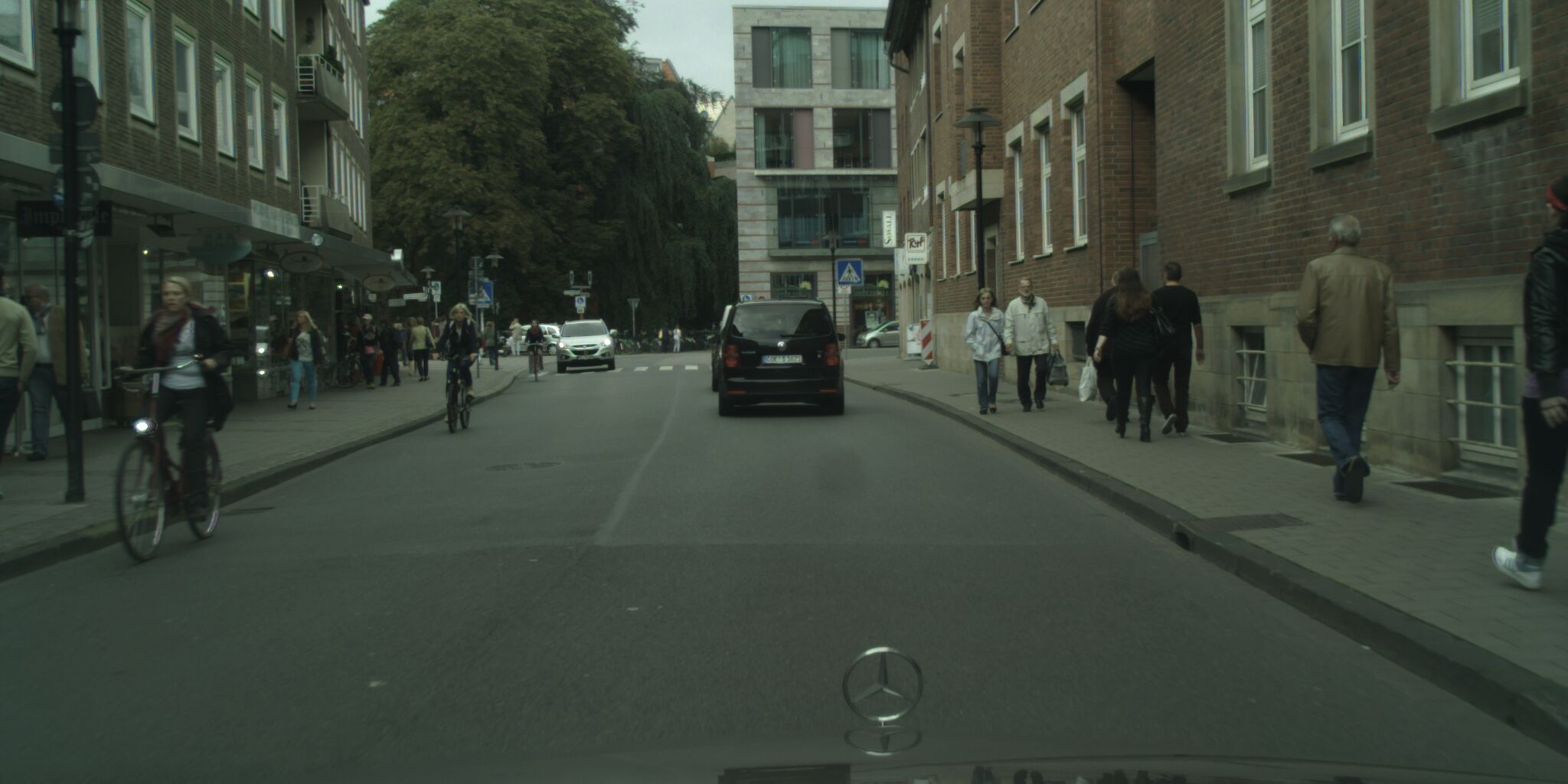} &
        \includegraphics[width=\linewidth]{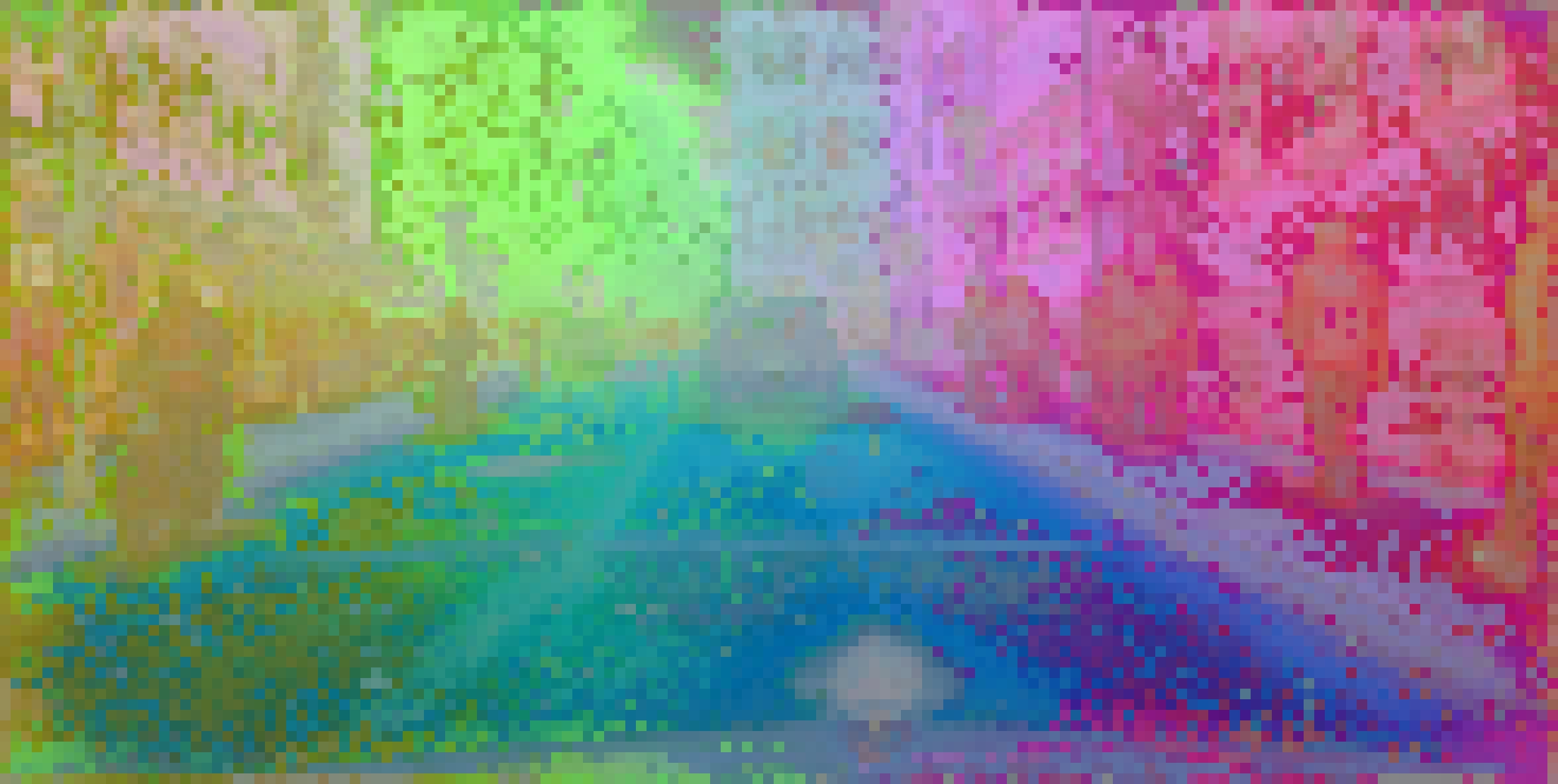} &
        \includegraphics[width=\linewidth]{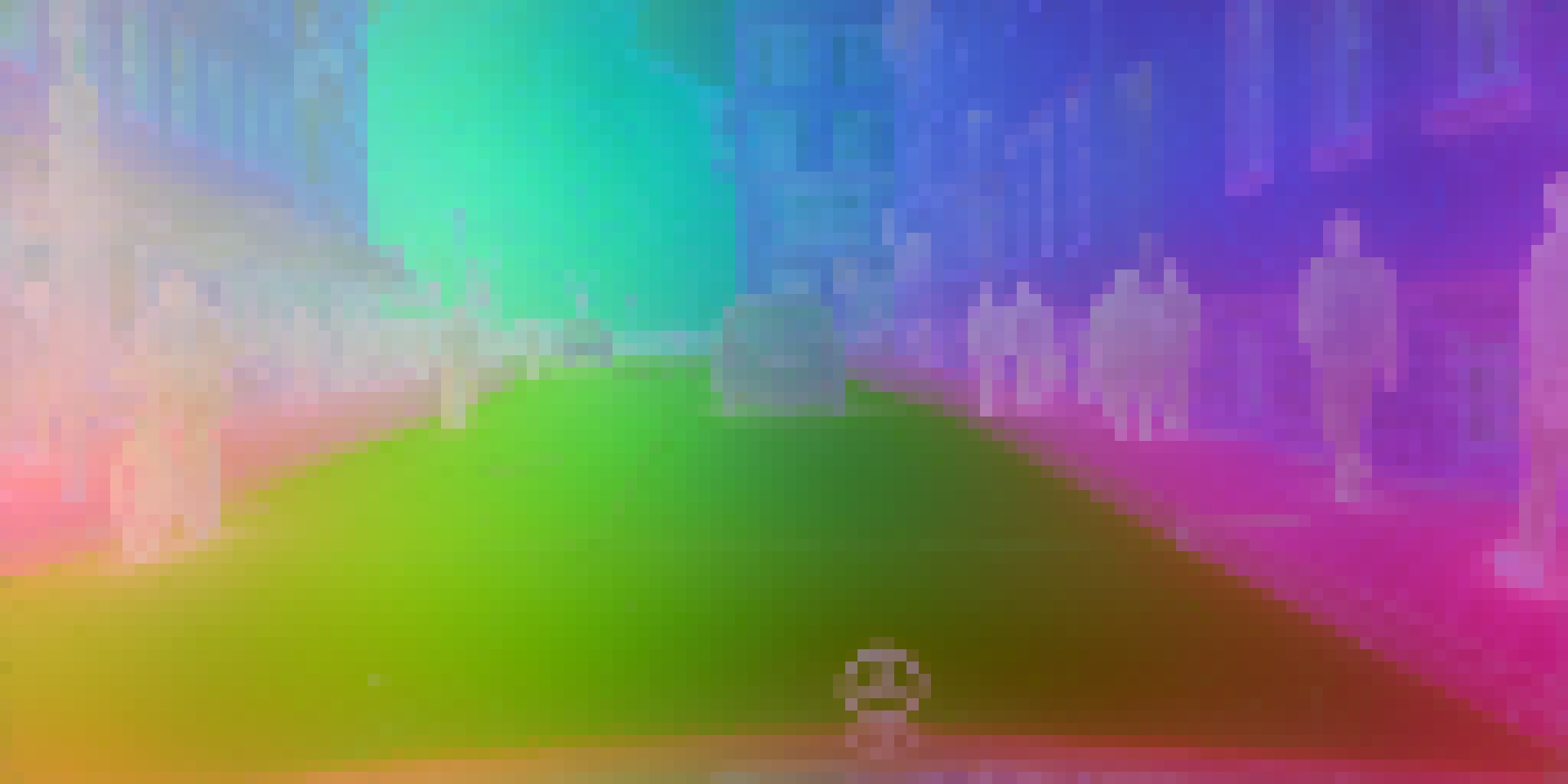} &
        \includegraphics[width=\linewidth]{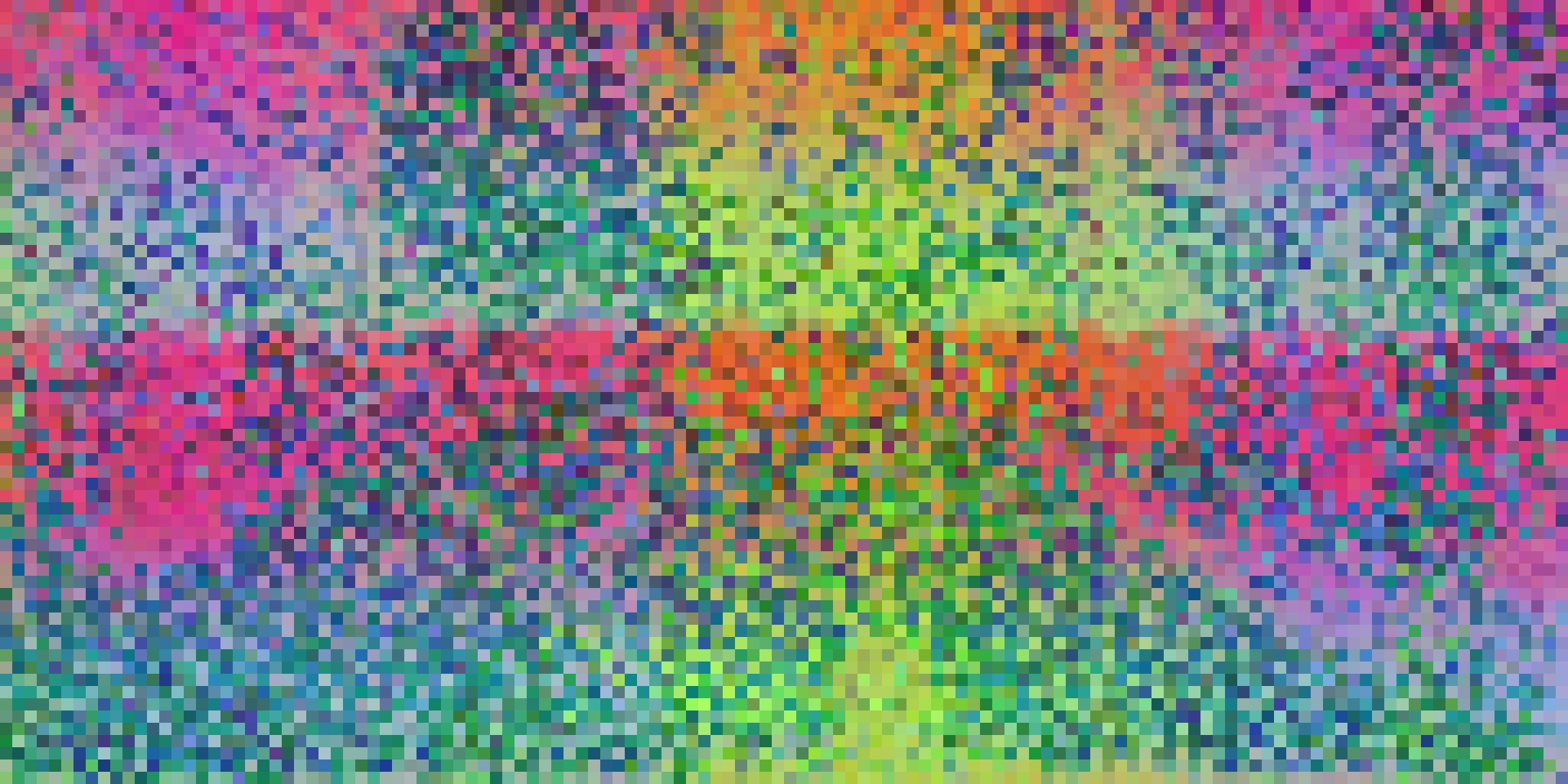} &
        \includegraphics[width=\linewidth]{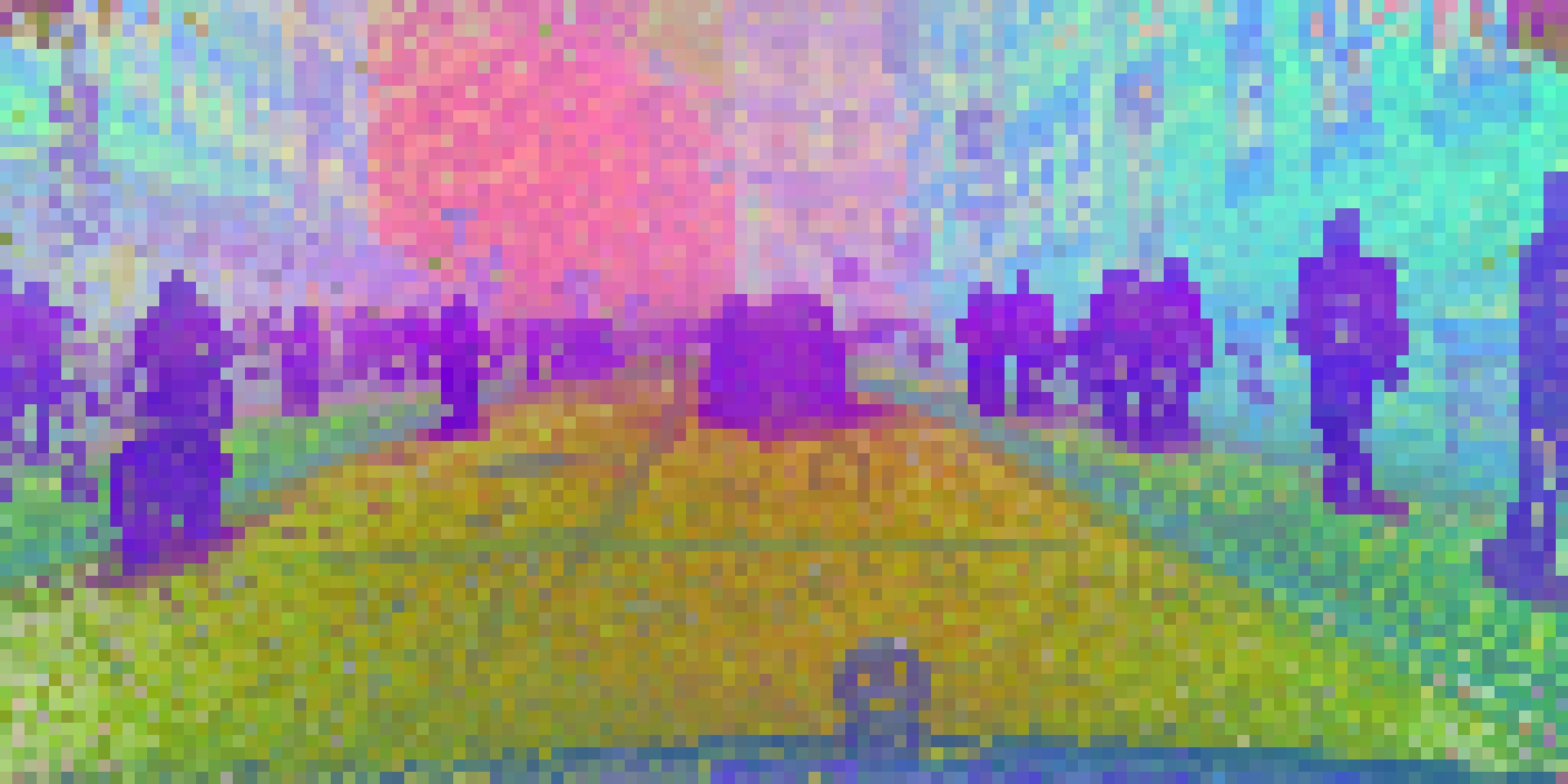} \\
        
        \includegraphics[width=\linewidth]{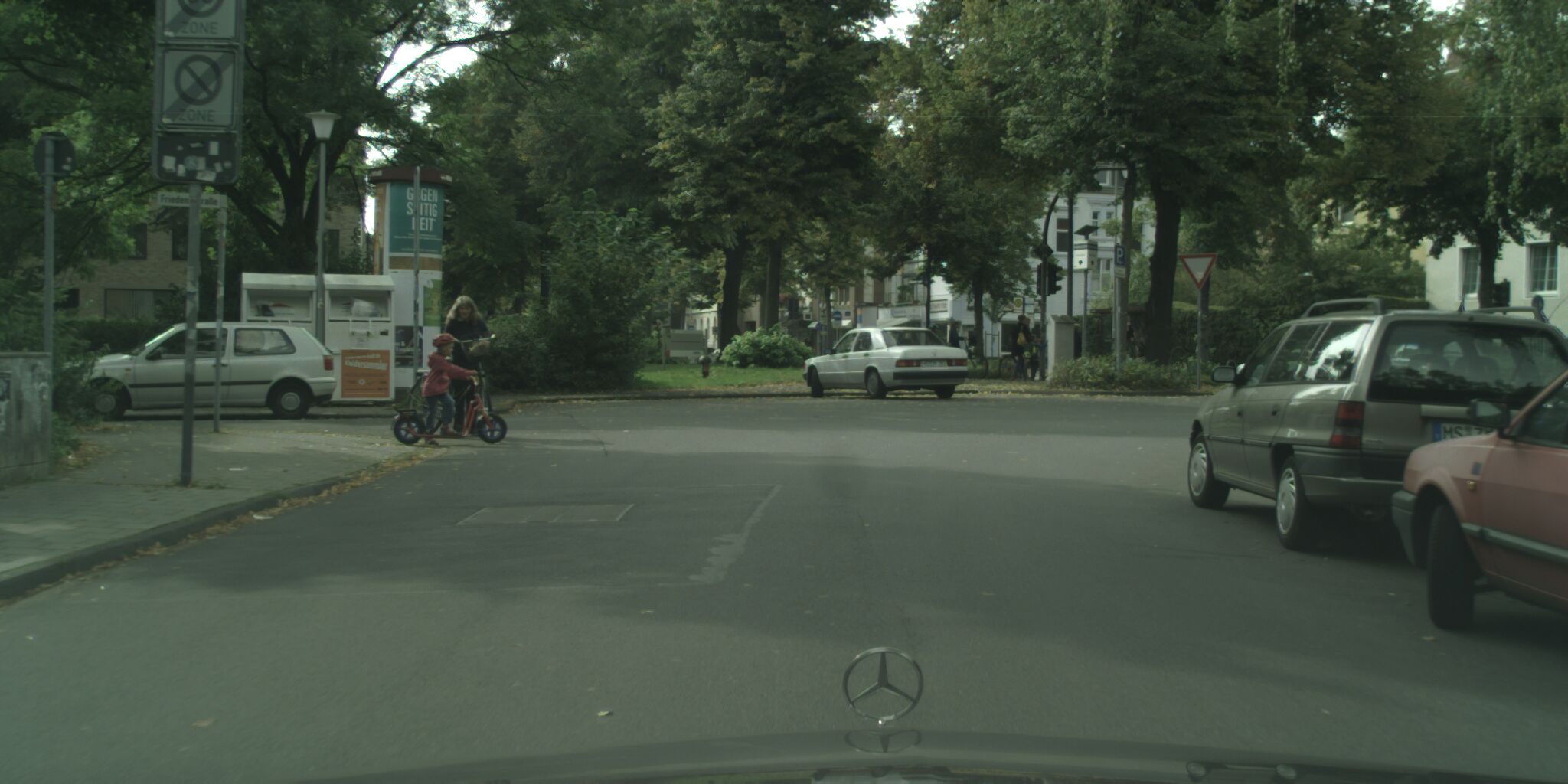} &
        \includegraphics[width=\linewidth]{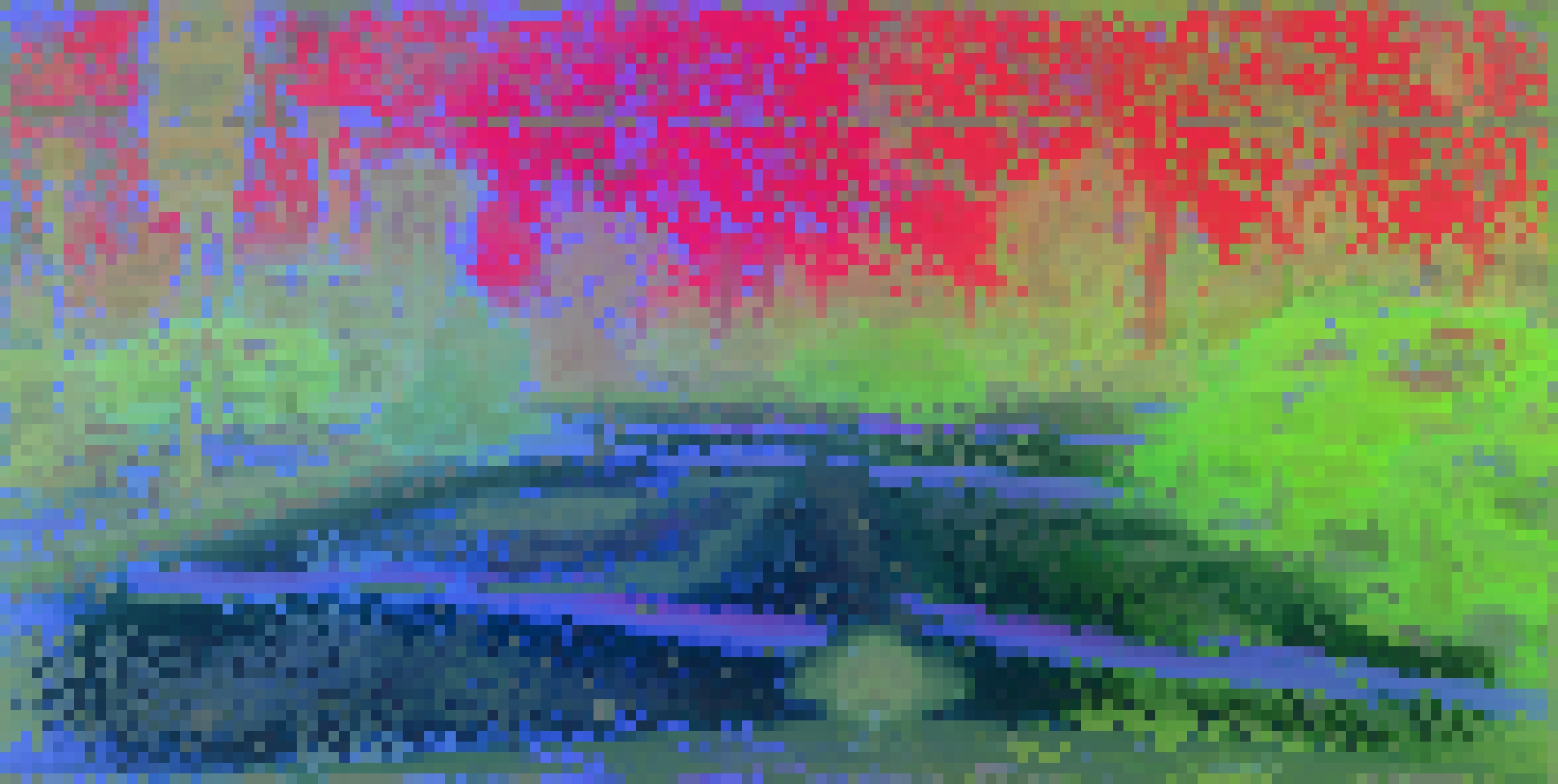} &
        \includegraphics[width=\linewidth]{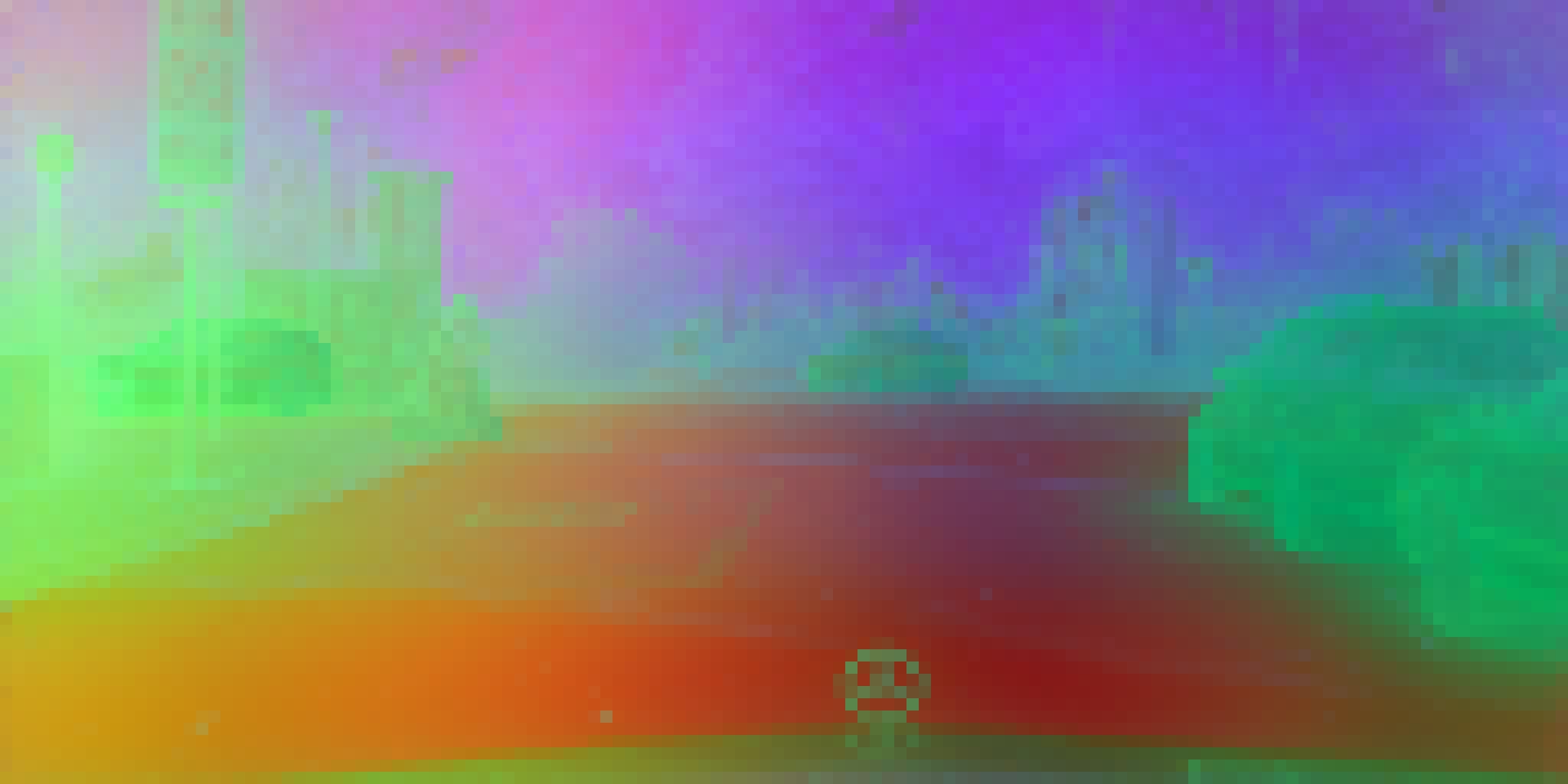} &
        \includegraphics[width=\linewidth]{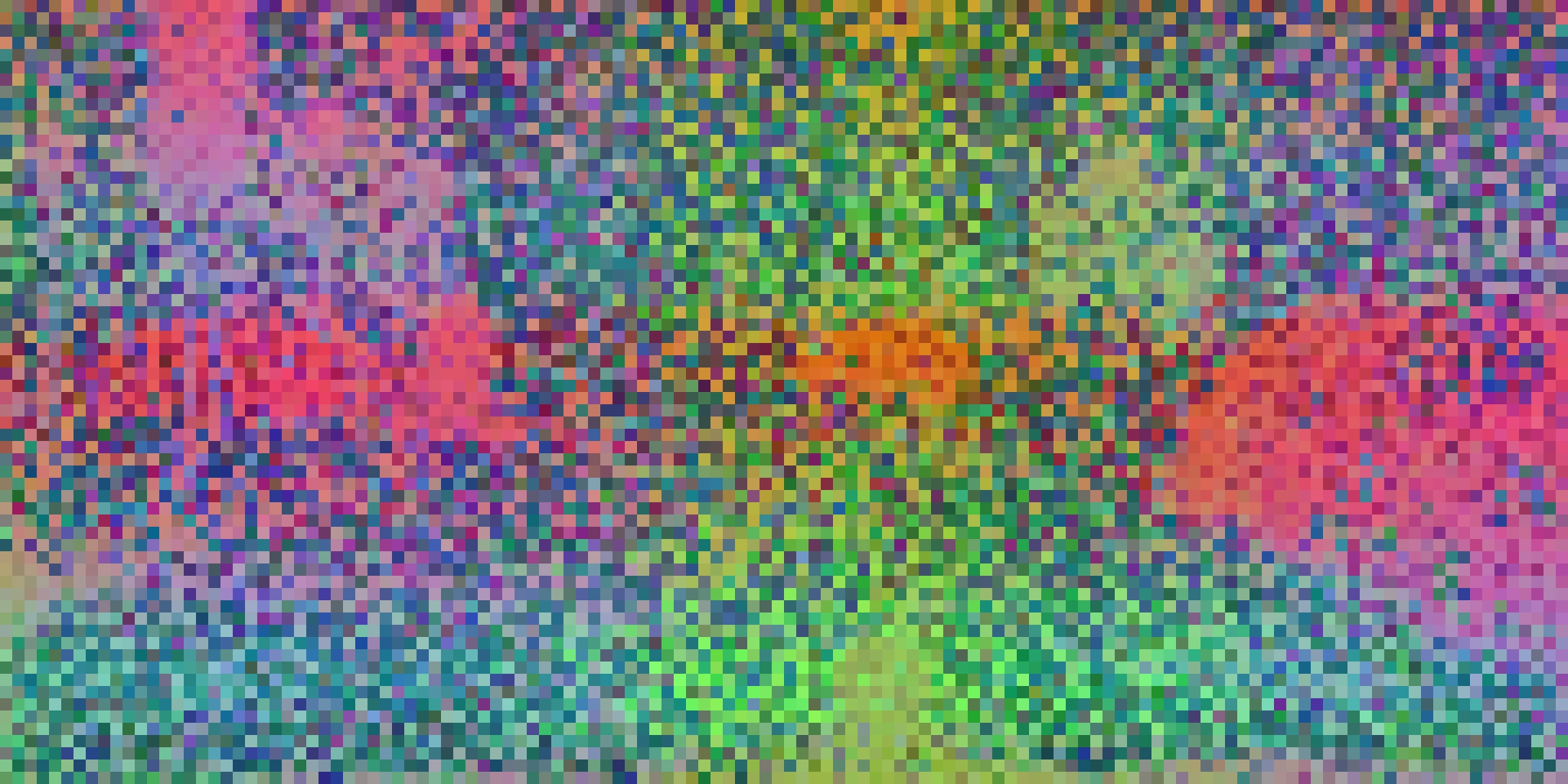} &
        \includegraphics[width=\linewidth]{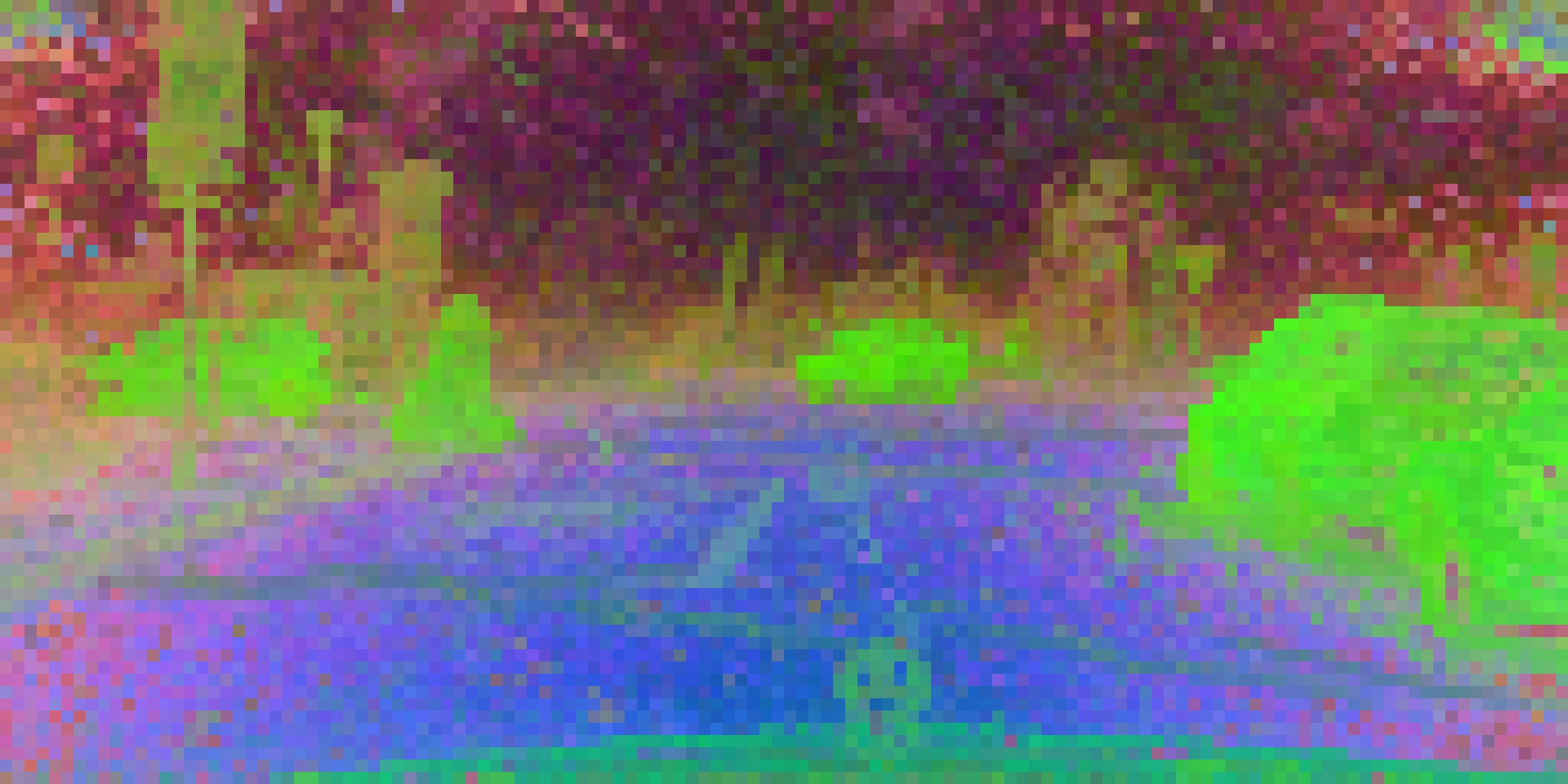} \\
        
        \includegraphics[width=\linewidth]{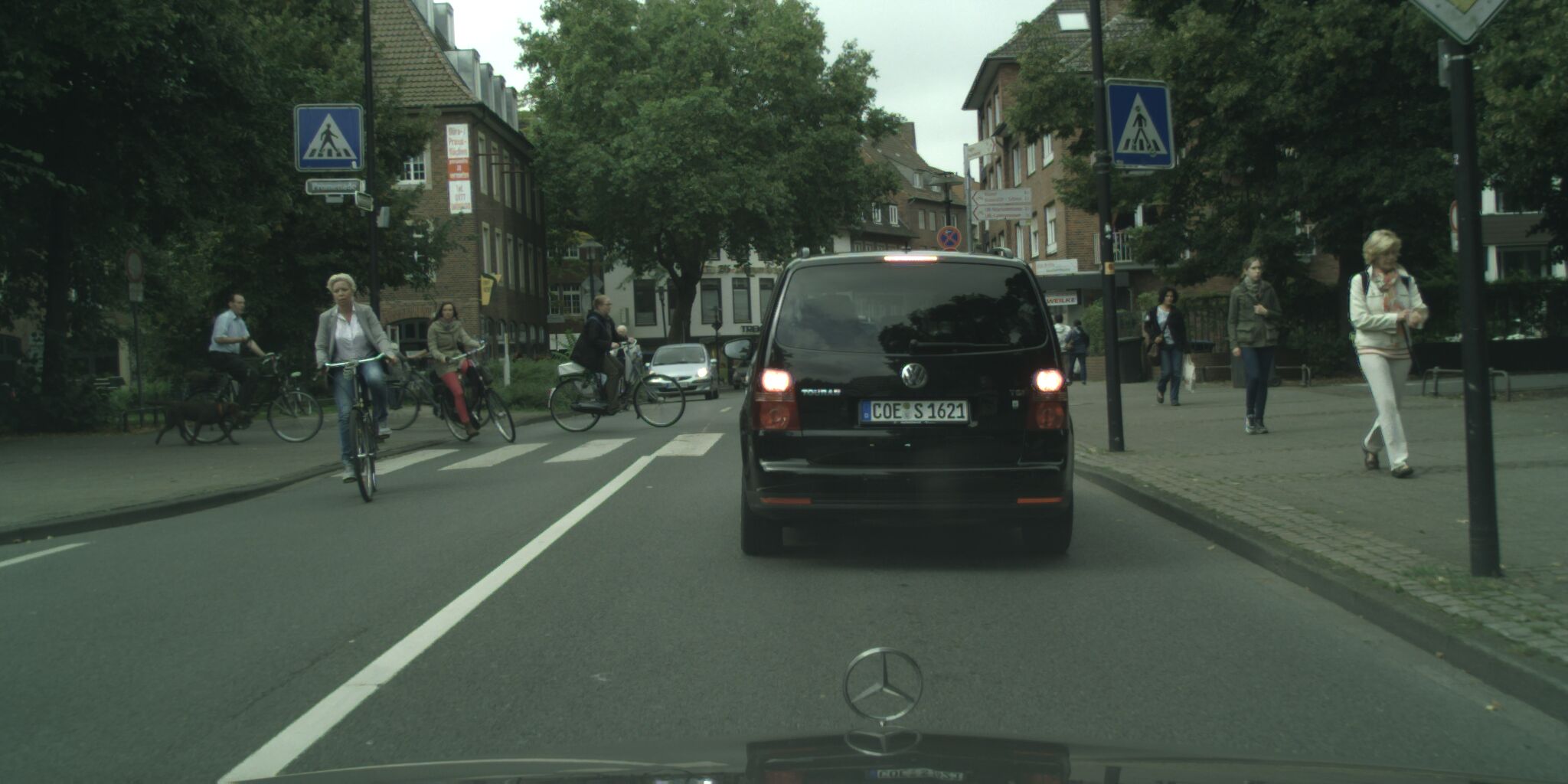} &
        \includegraphics[width=\linewidth]{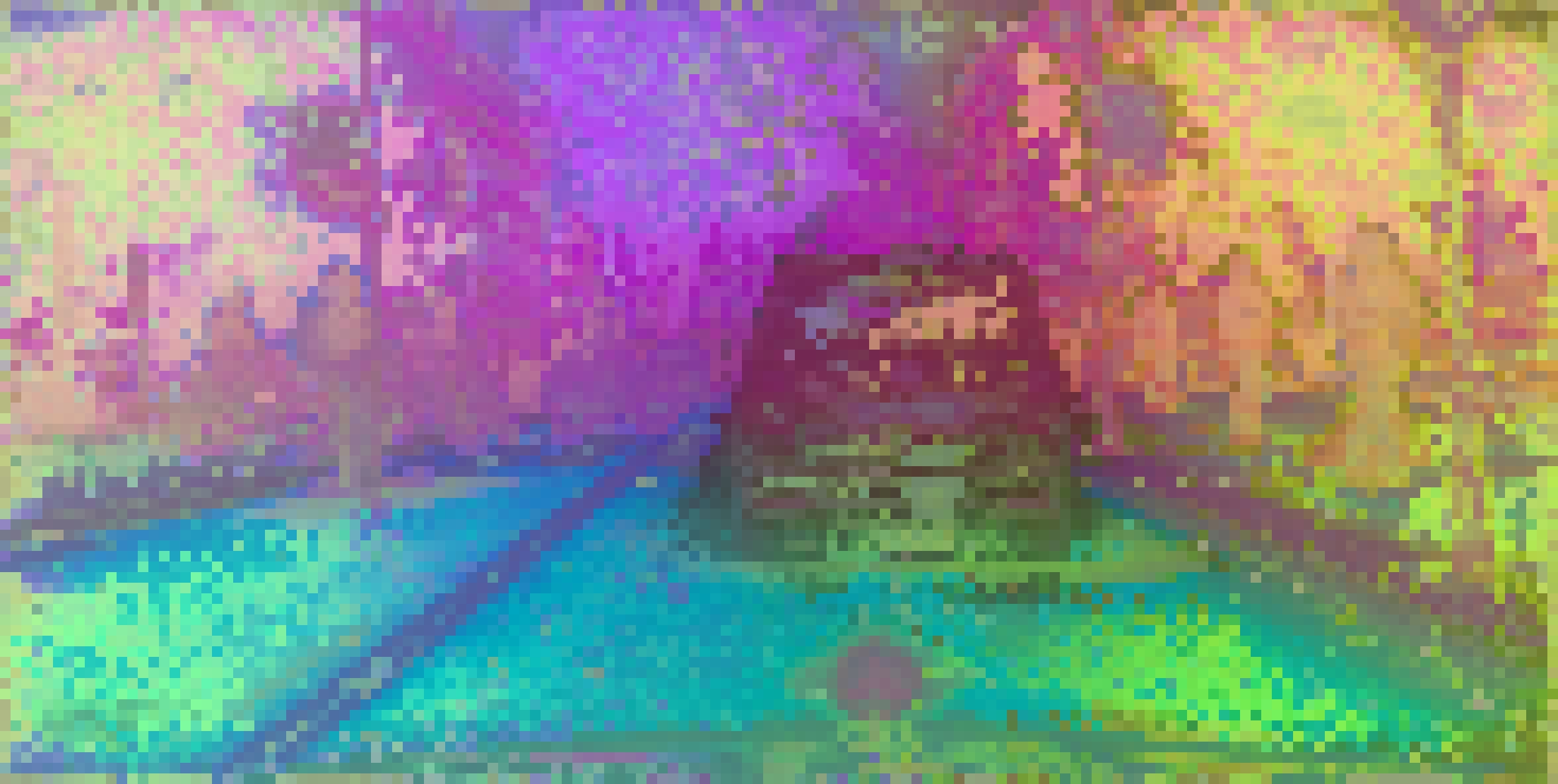} &
        \includegraphics[width=\linewidth]{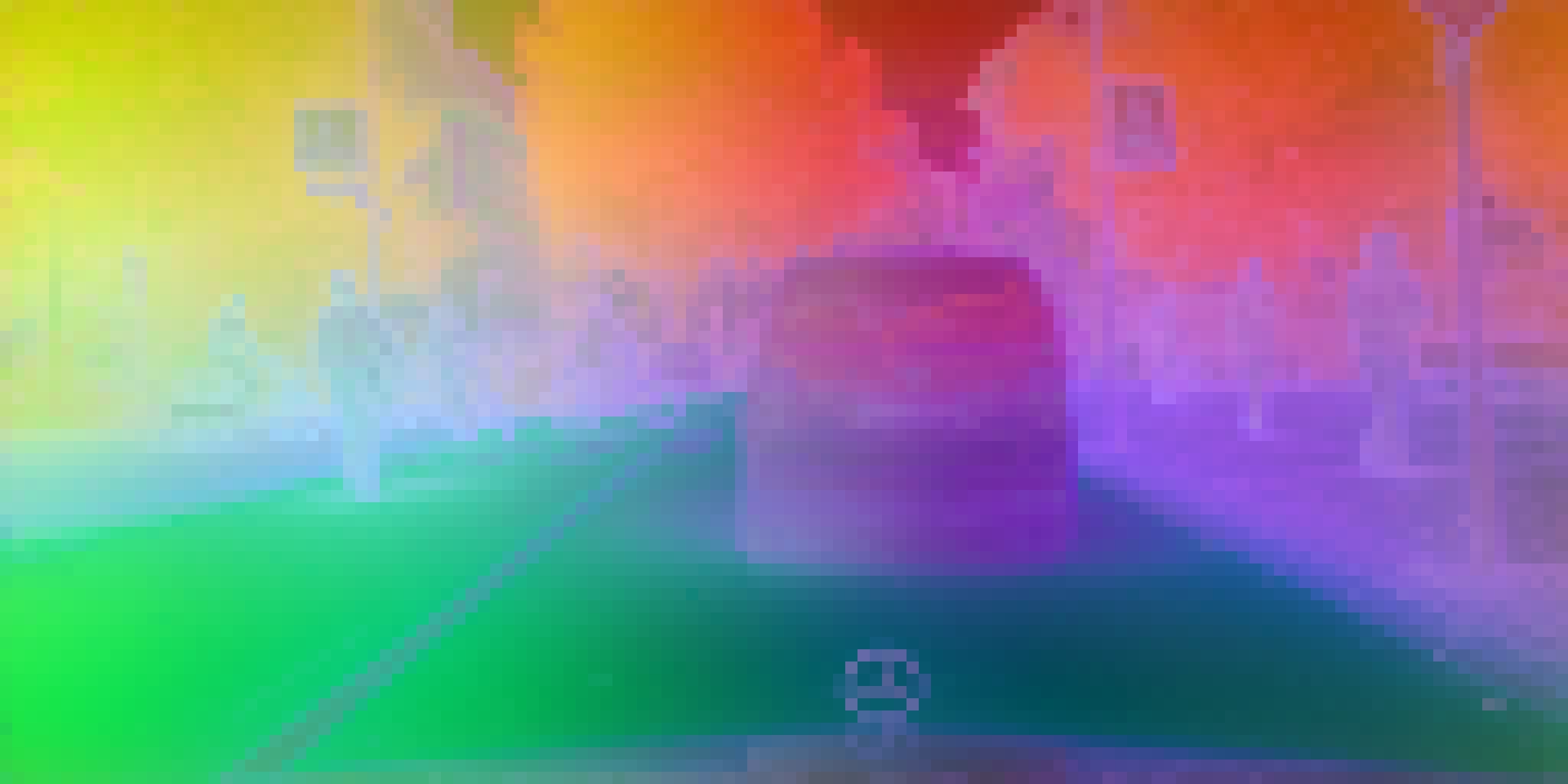} &
        \includegraphics[width=\linewidth]{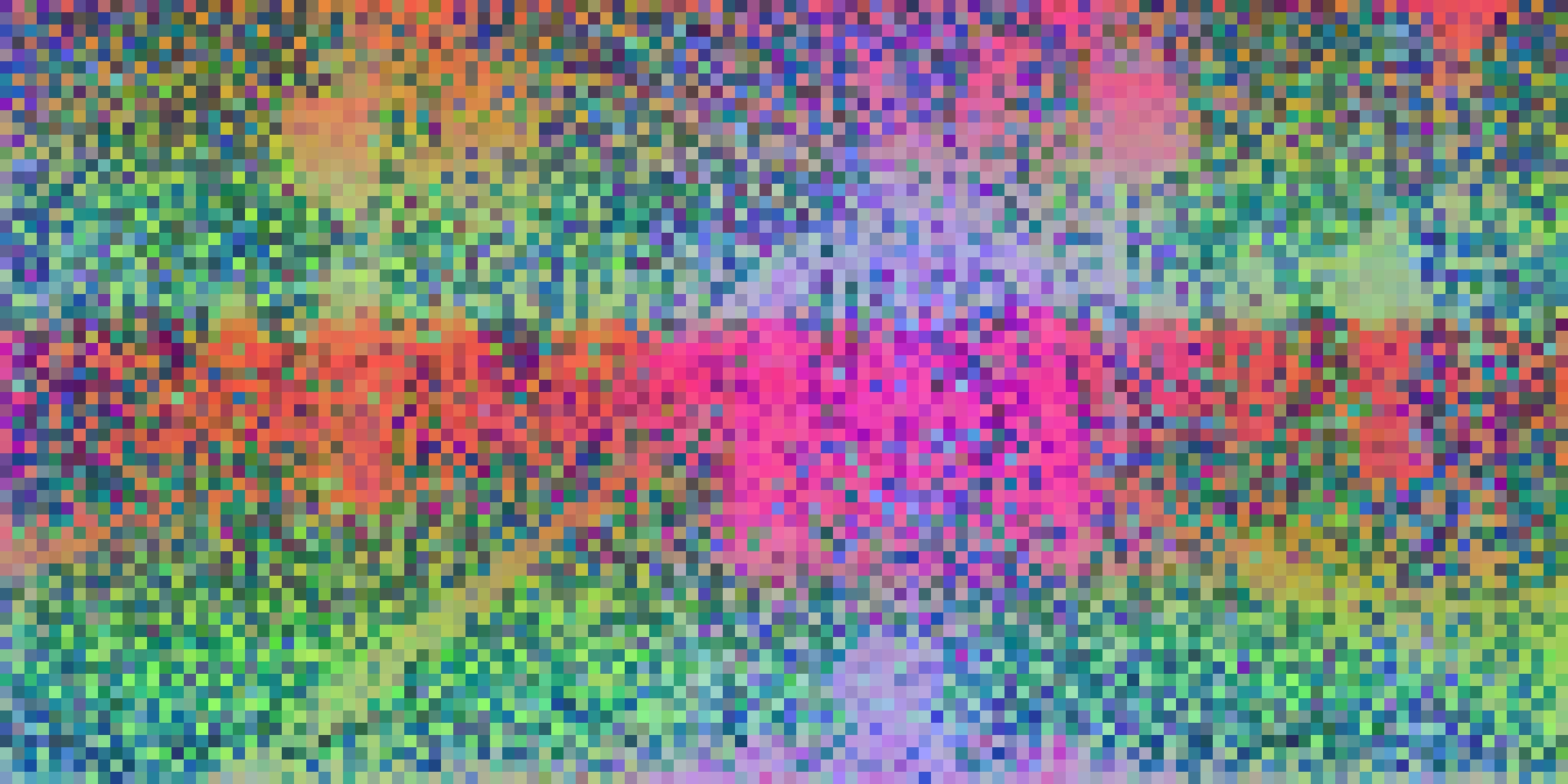} &
        \includegraphics[width=\linewidth]{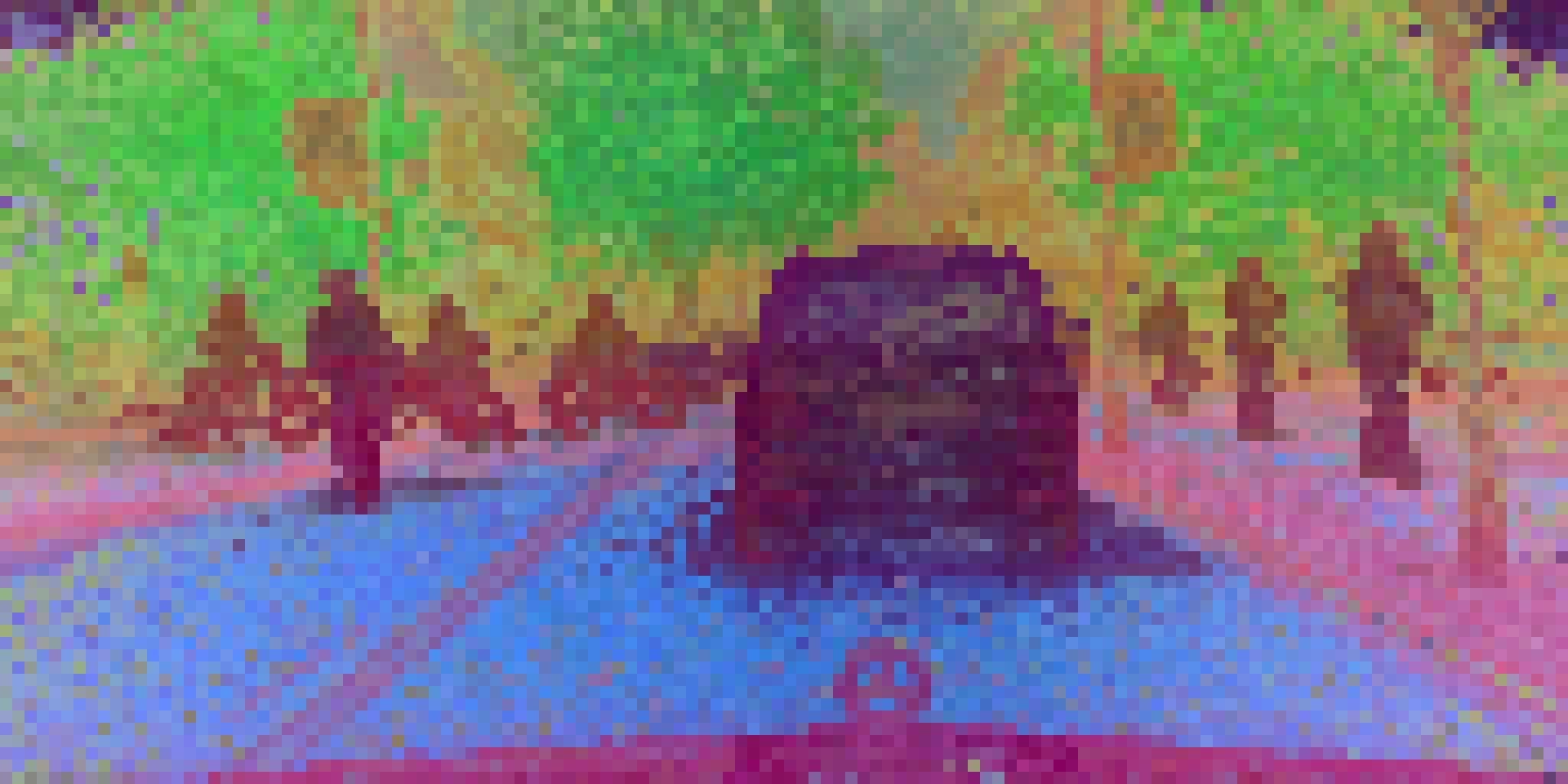} \\
        
        \includegraphics[width=\linewidth]{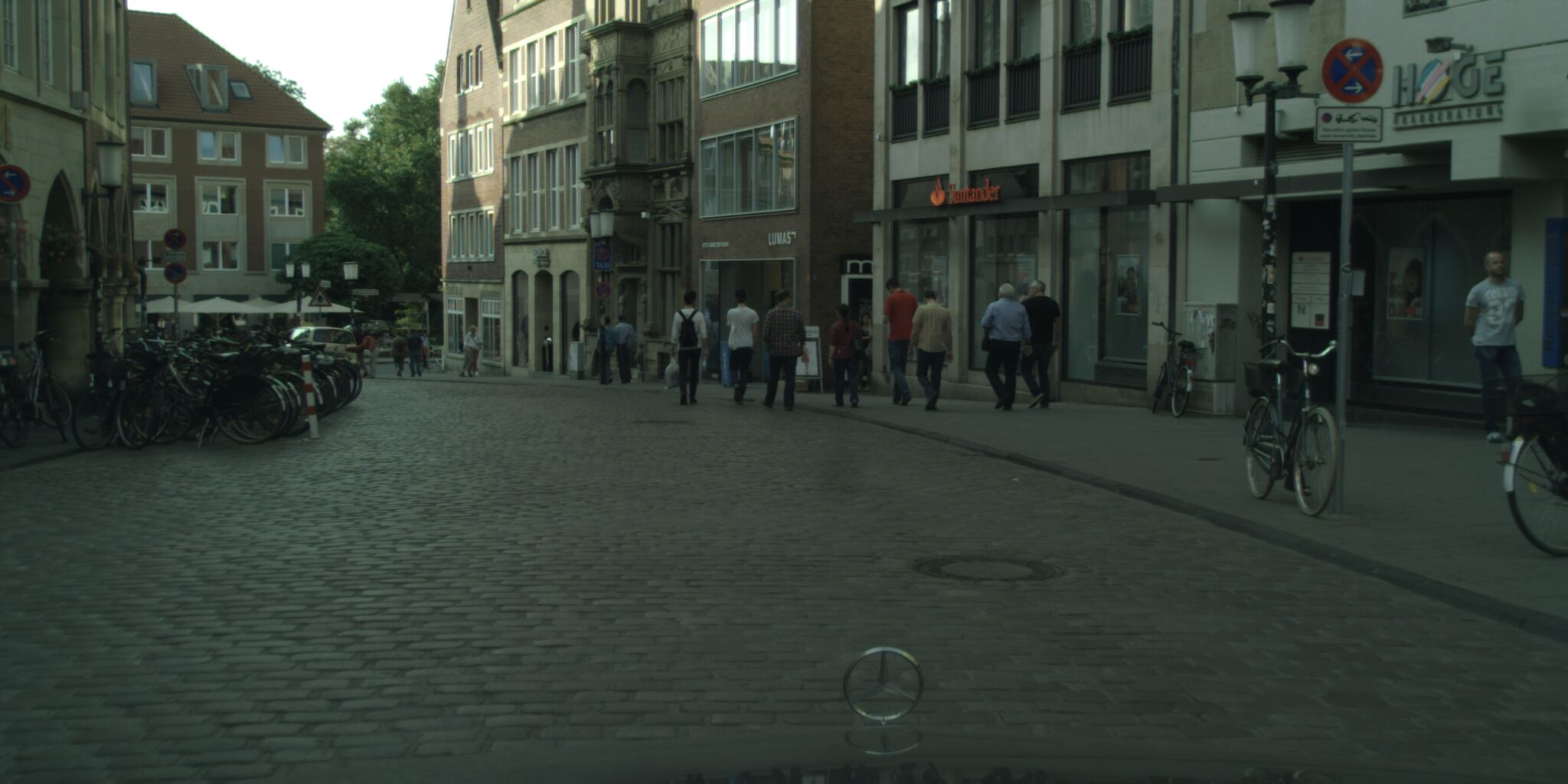} &
        \includegraphics[width=\linewidth]{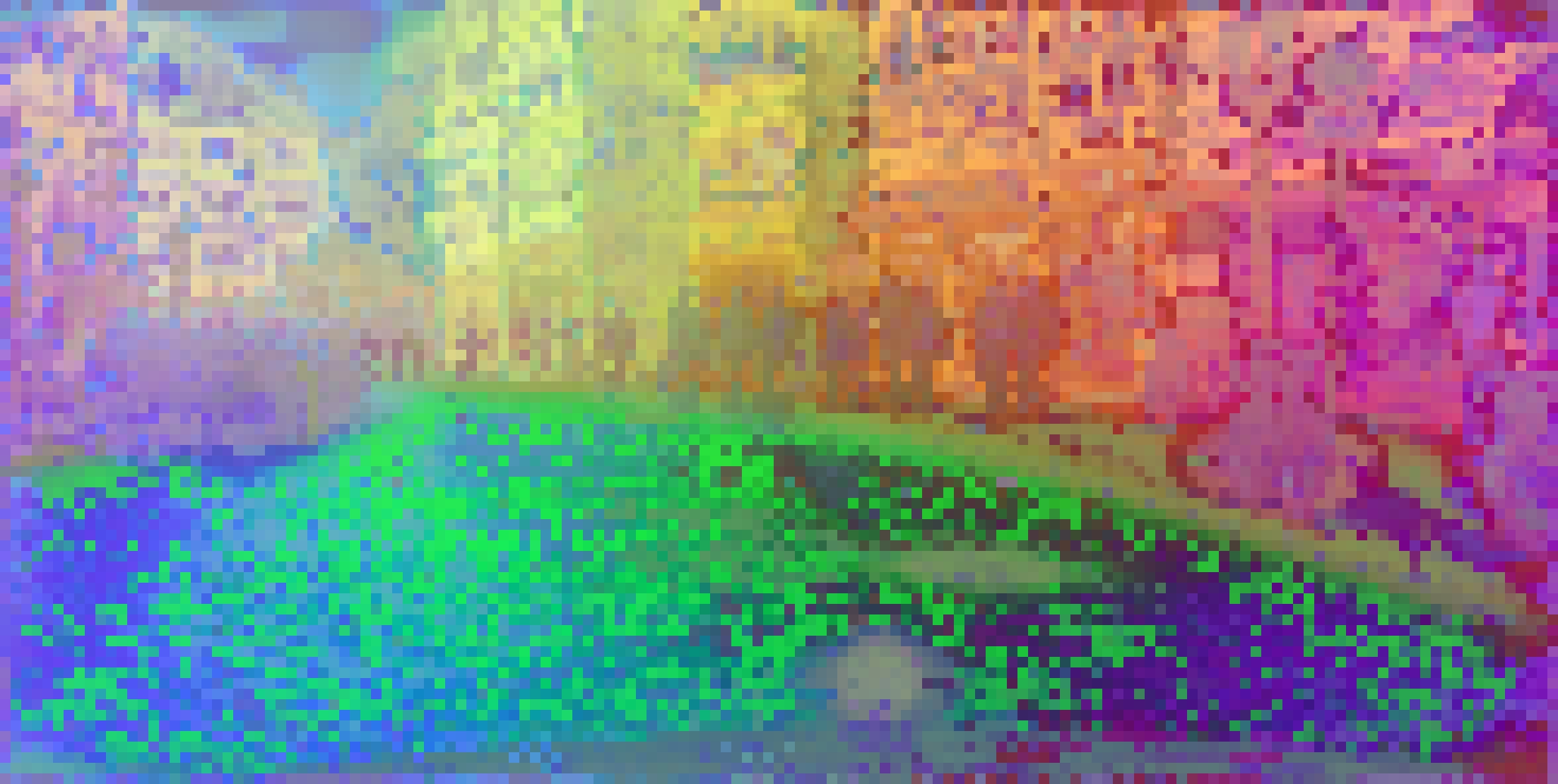} &
        \includegraphics[width=\linewidth]{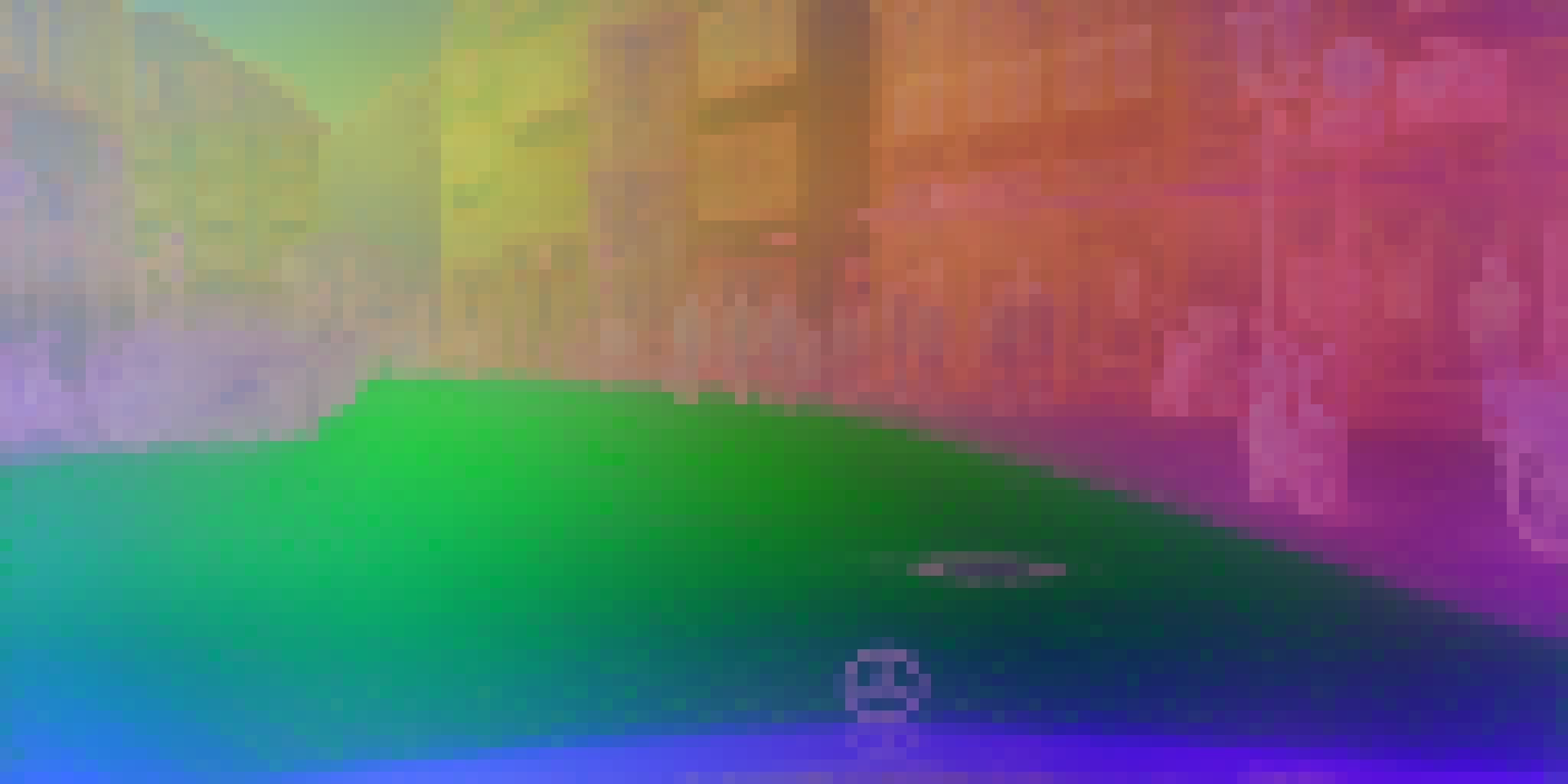} &
        \includegraphics[width=\linewidth]{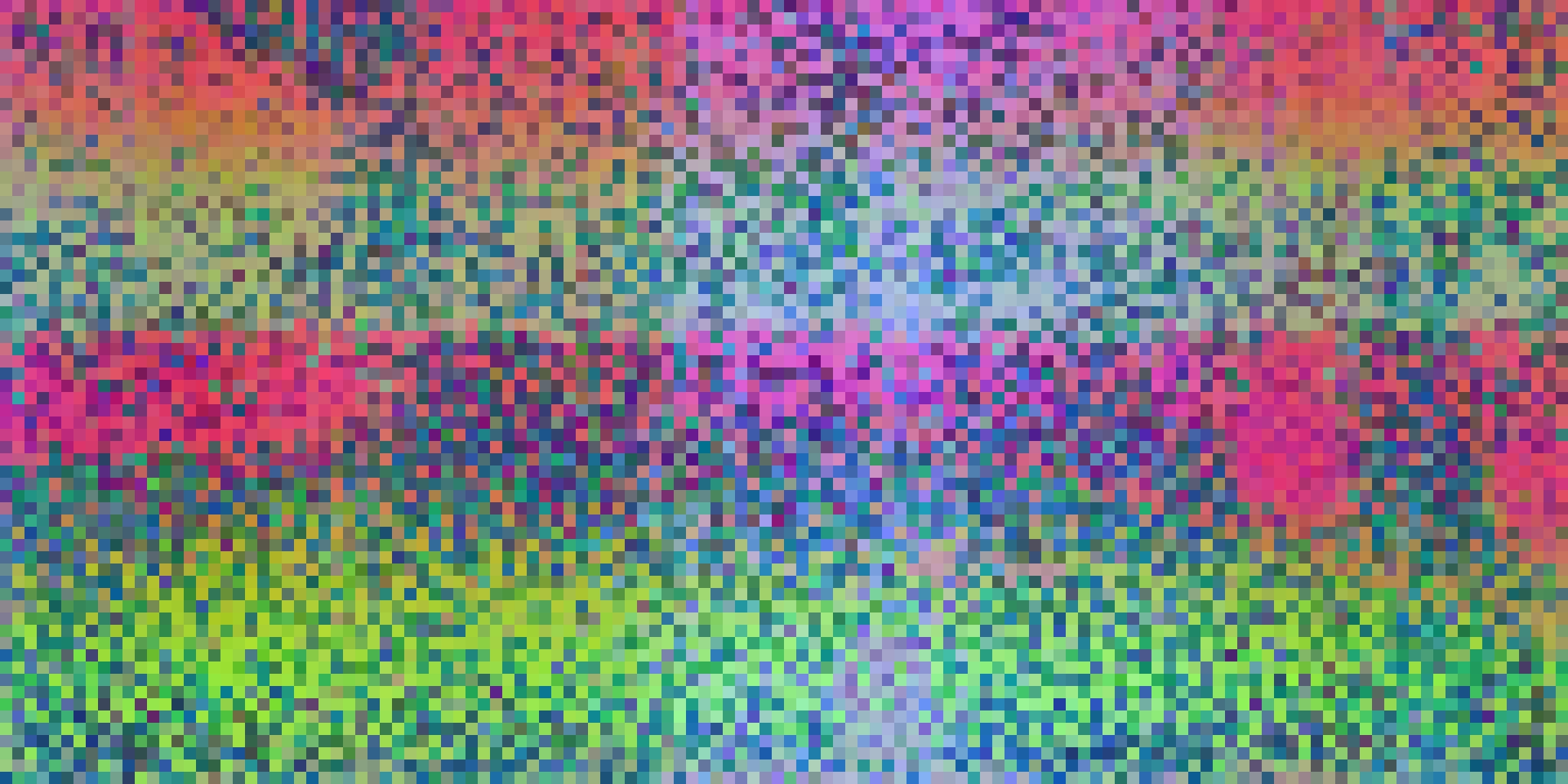} &
        \includegraphics[width=\linewidth]{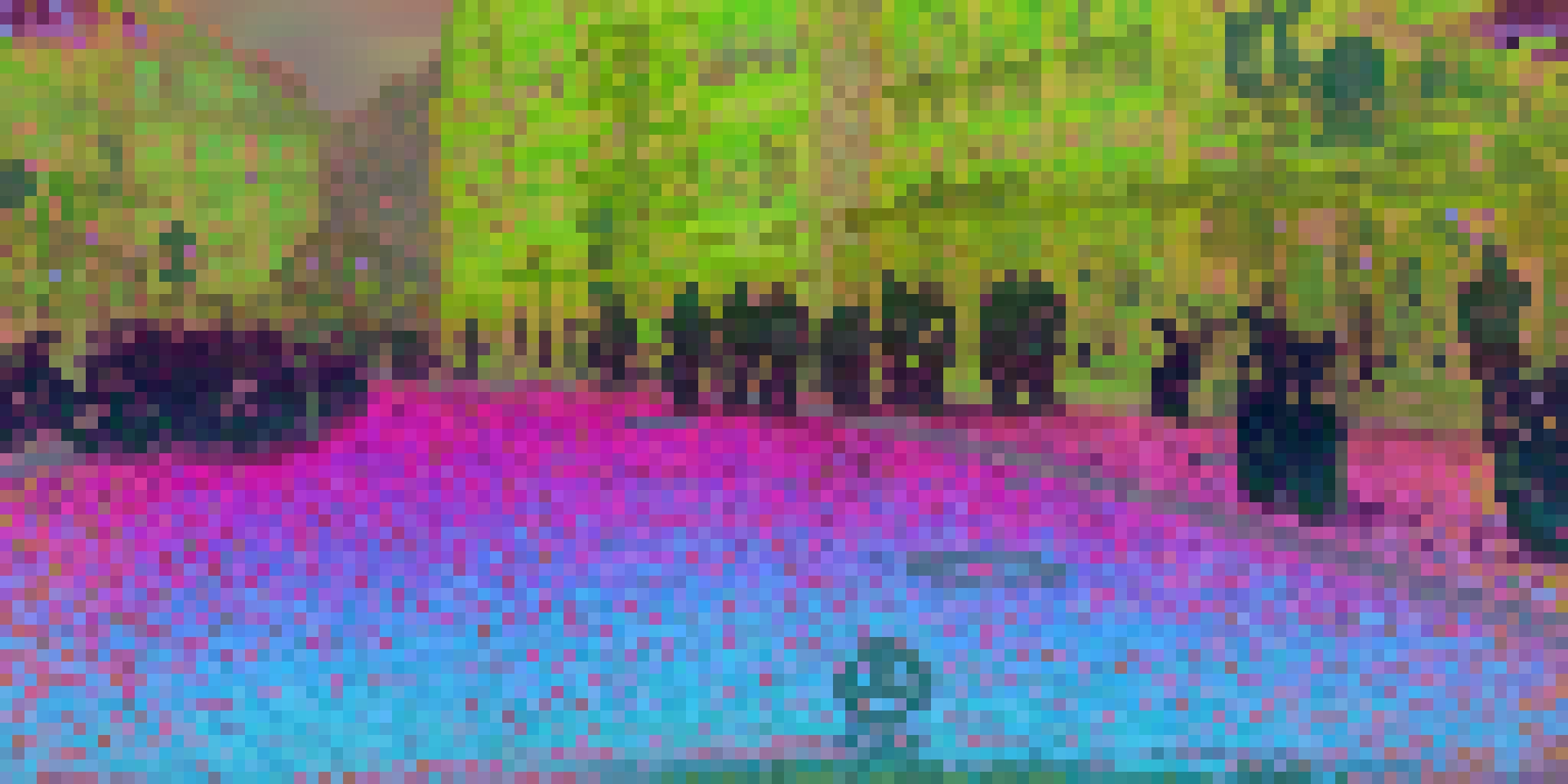} \\
        
        \includegraphics[width=\linewidth]{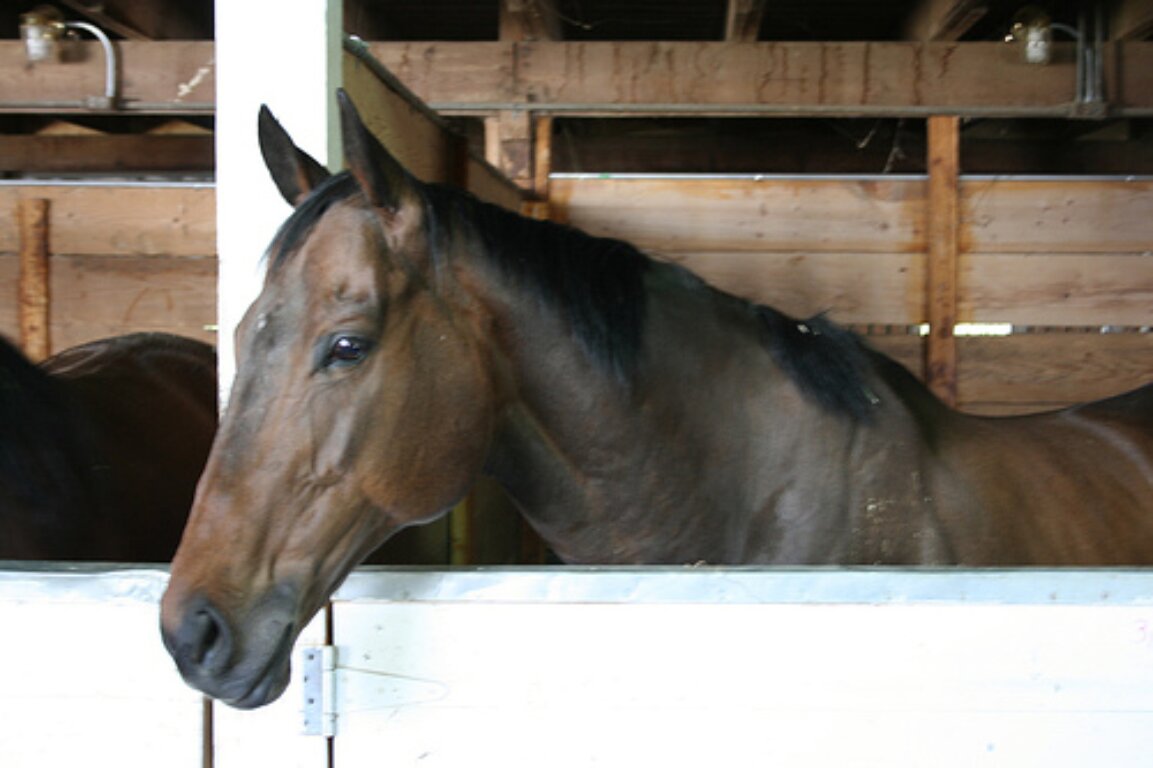} &
        \includegraphics[width=\linewidth]{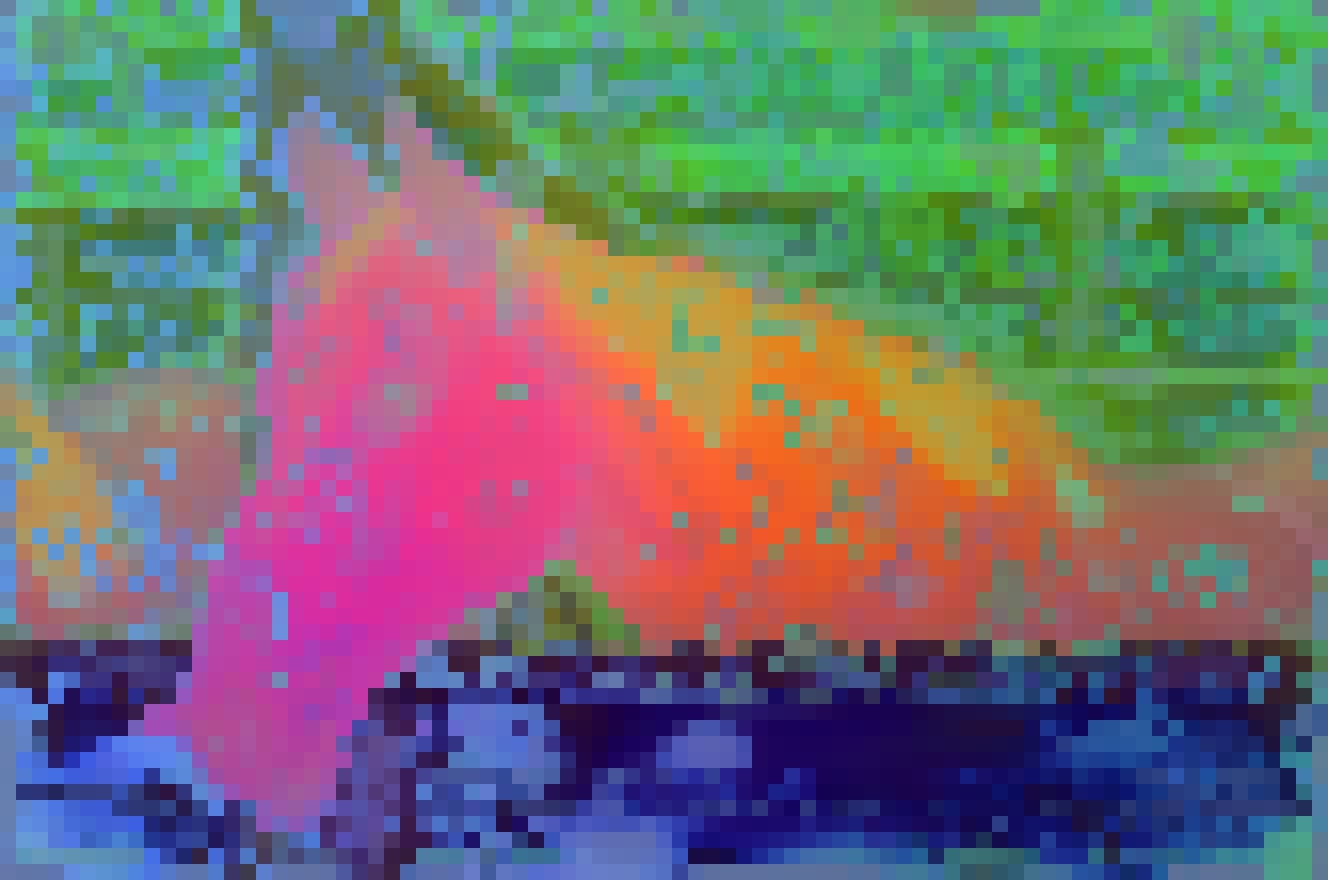} &
        \includegraphics[width=\linewidth]{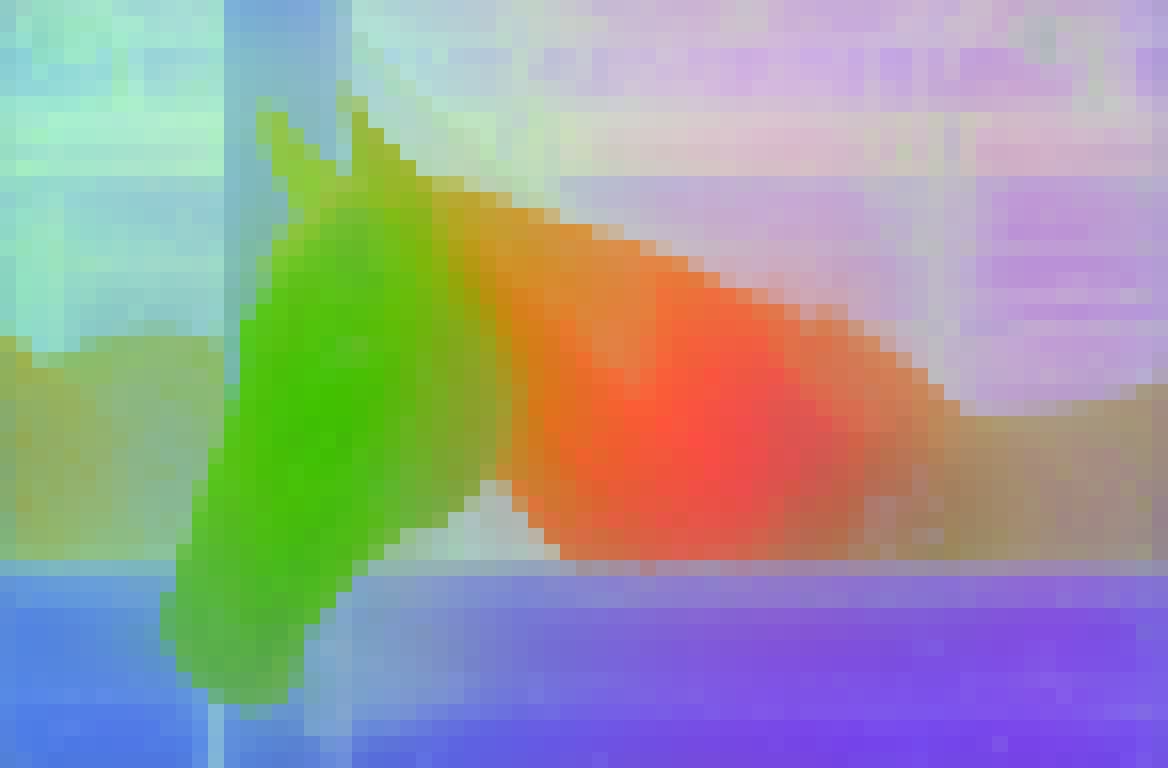} &
        \includegraphics[width=\linewidth]{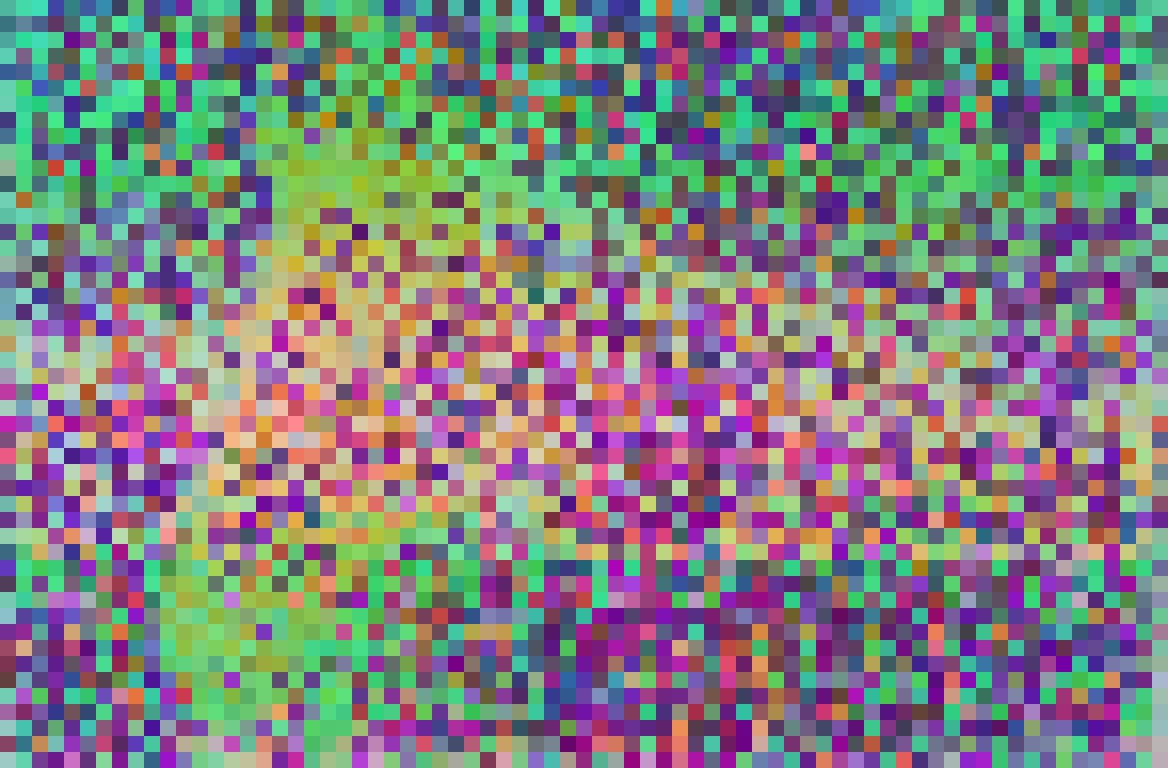} &
        \includegraphics[width=\linewidth]{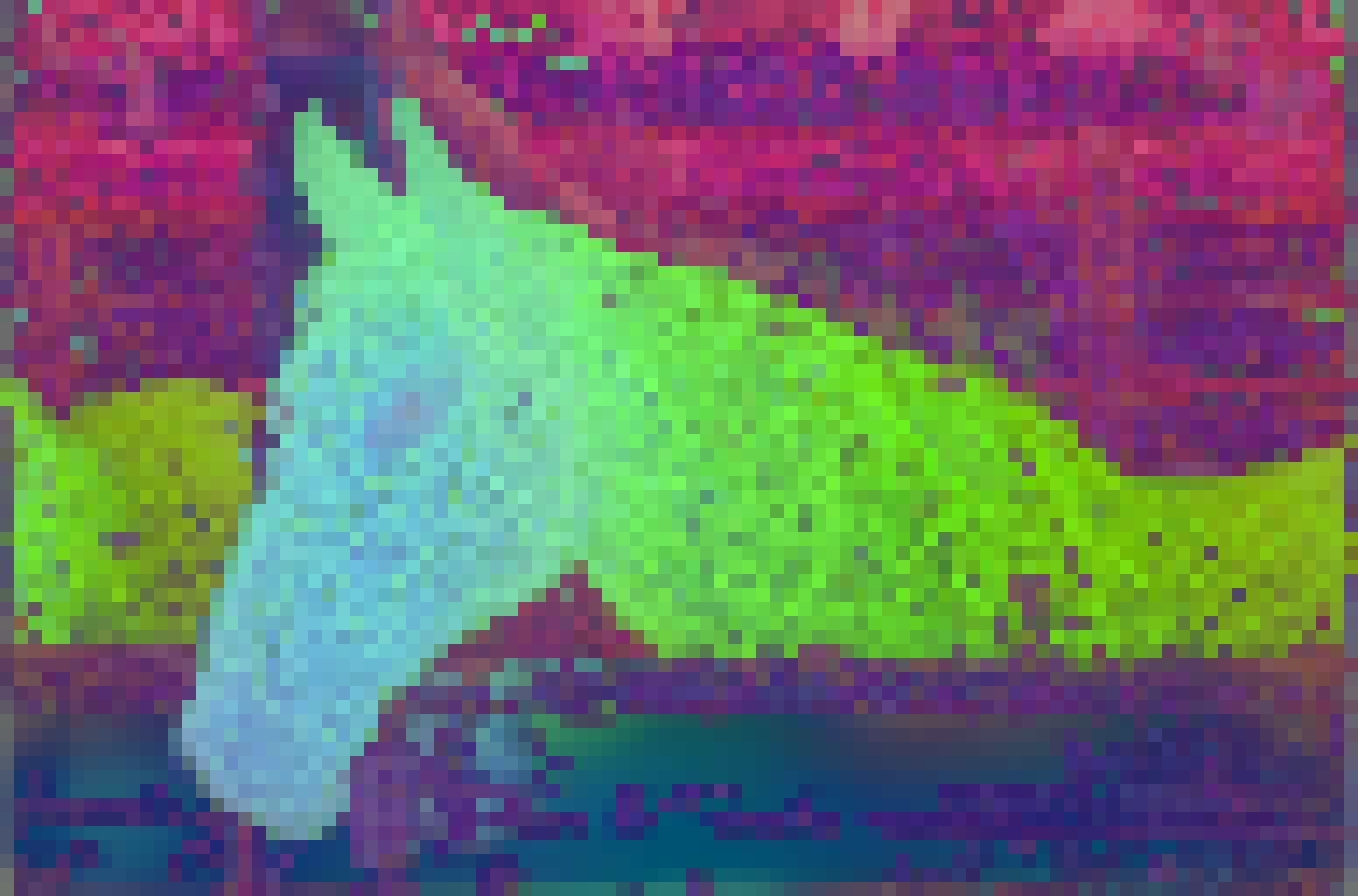} \\
        
        \includegraphics[width=\linewidth]{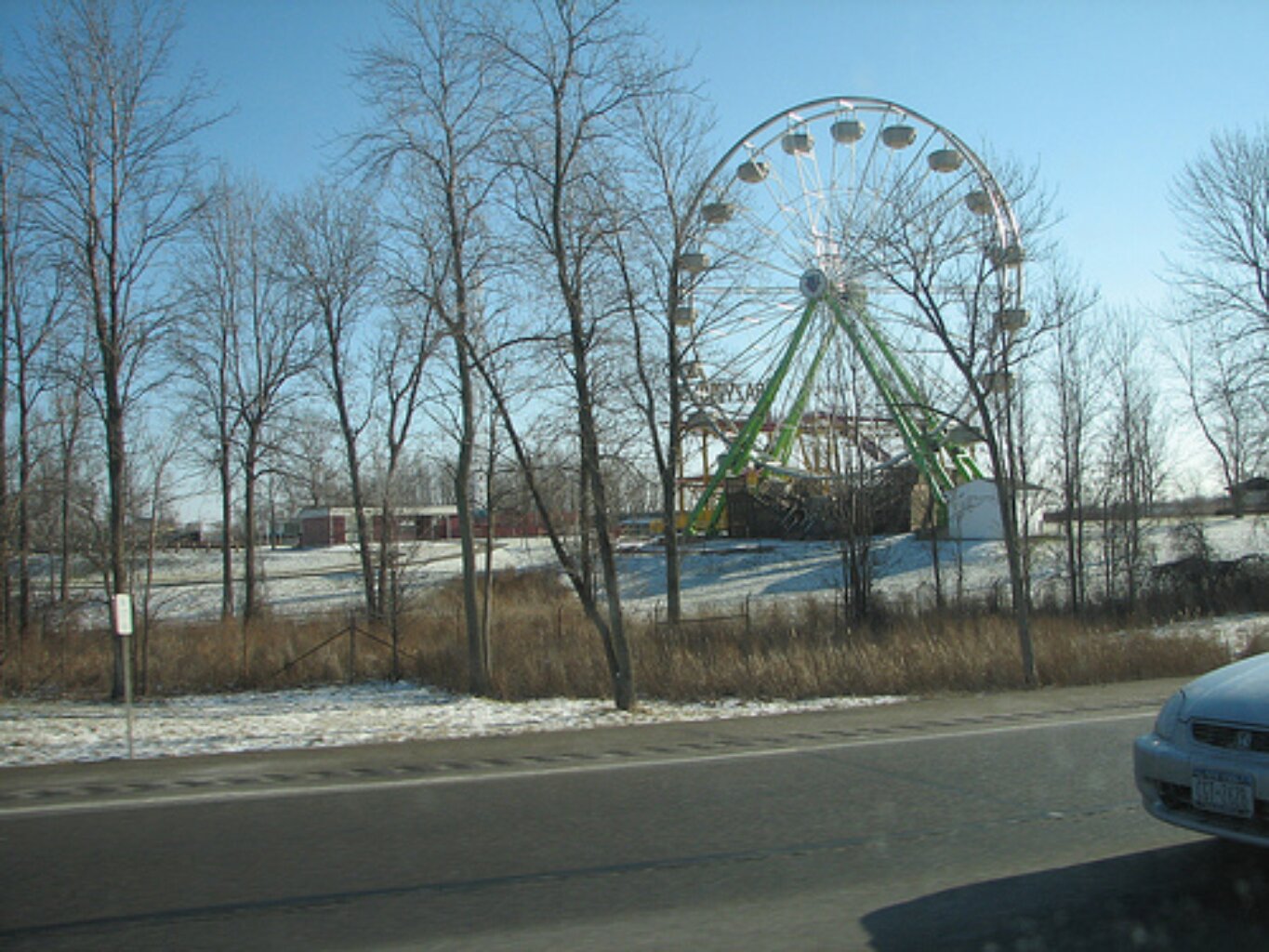} &
        \includegraphics[width=\linewidth]{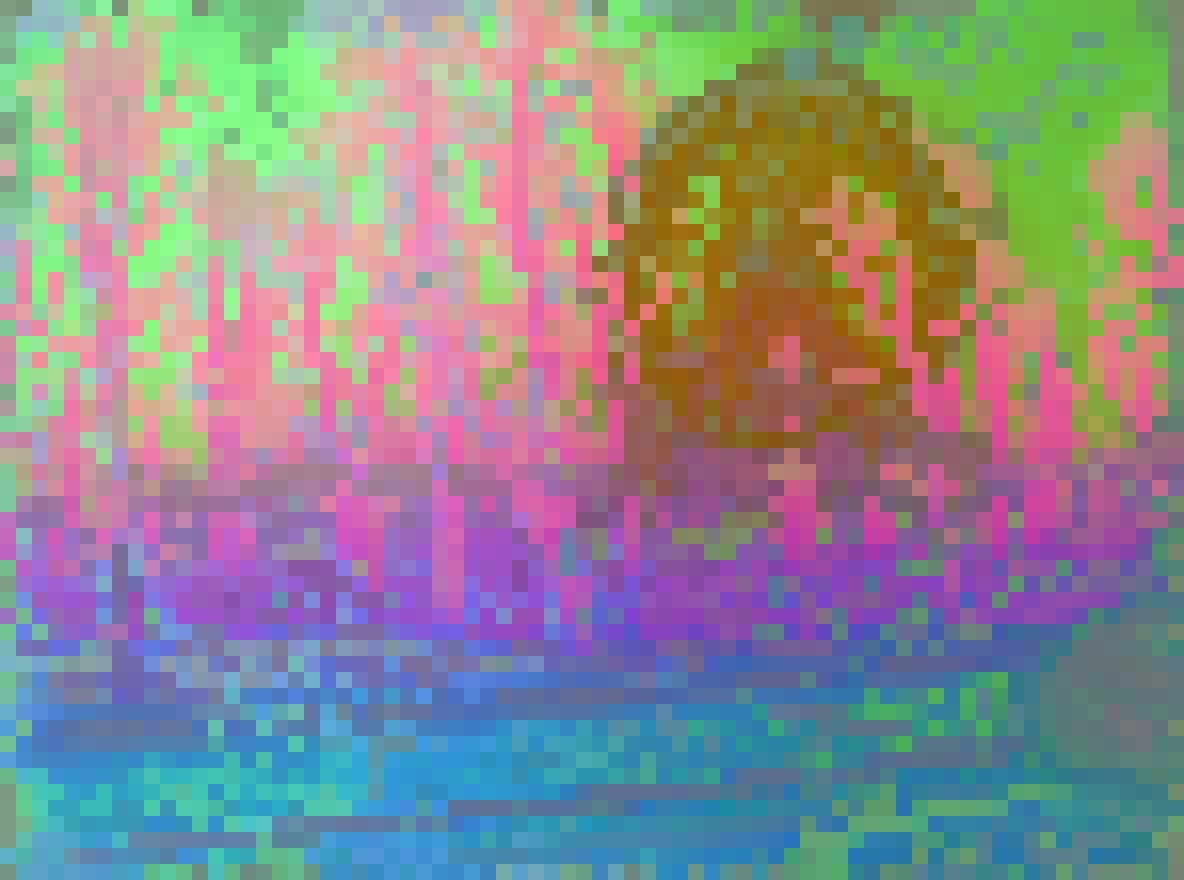} &
        \includegraphics[width=\linewidth]{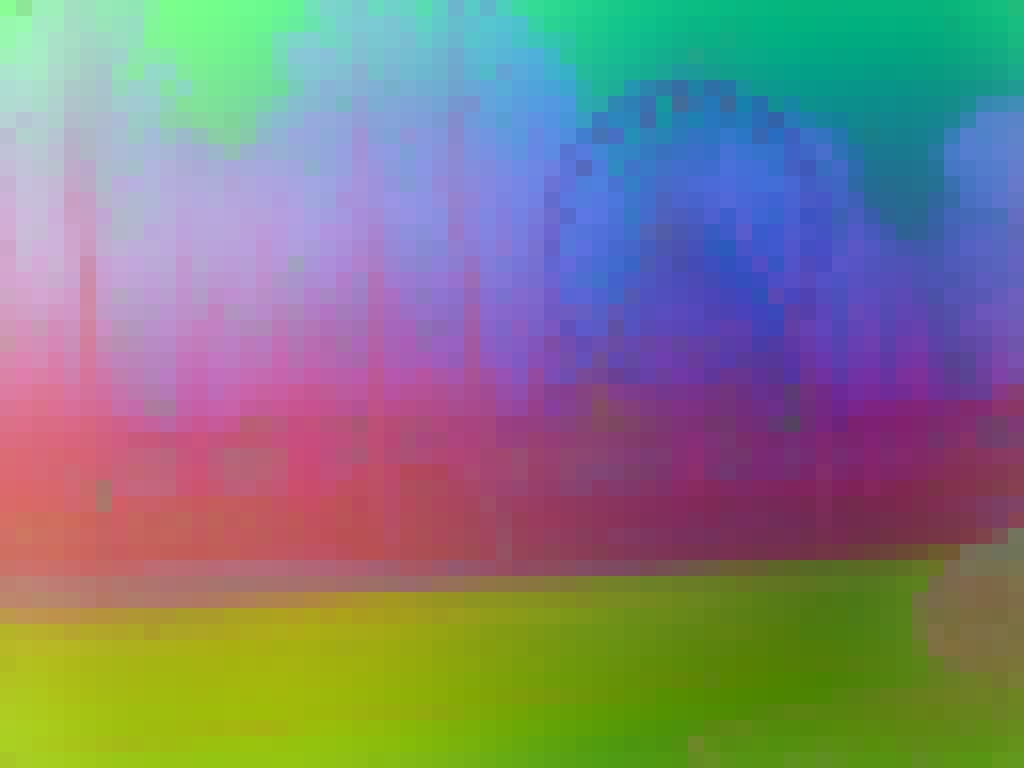} &
        \includegraphics[width=\linewidth]{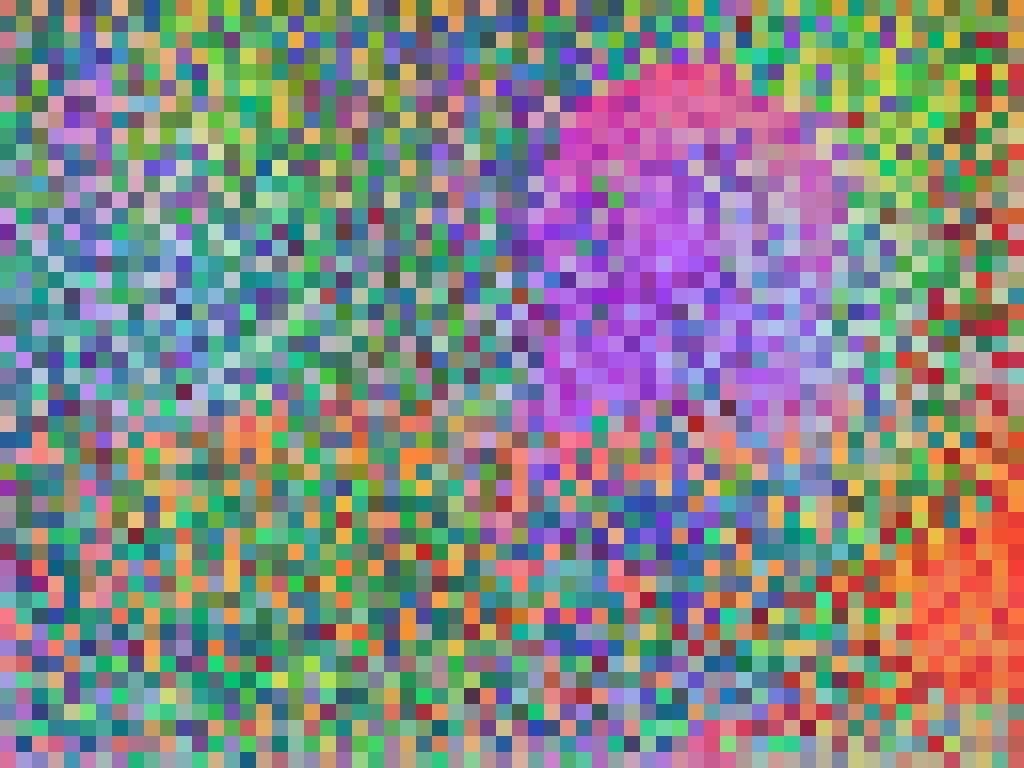} &
        \includegraphics[width=\linewidth]{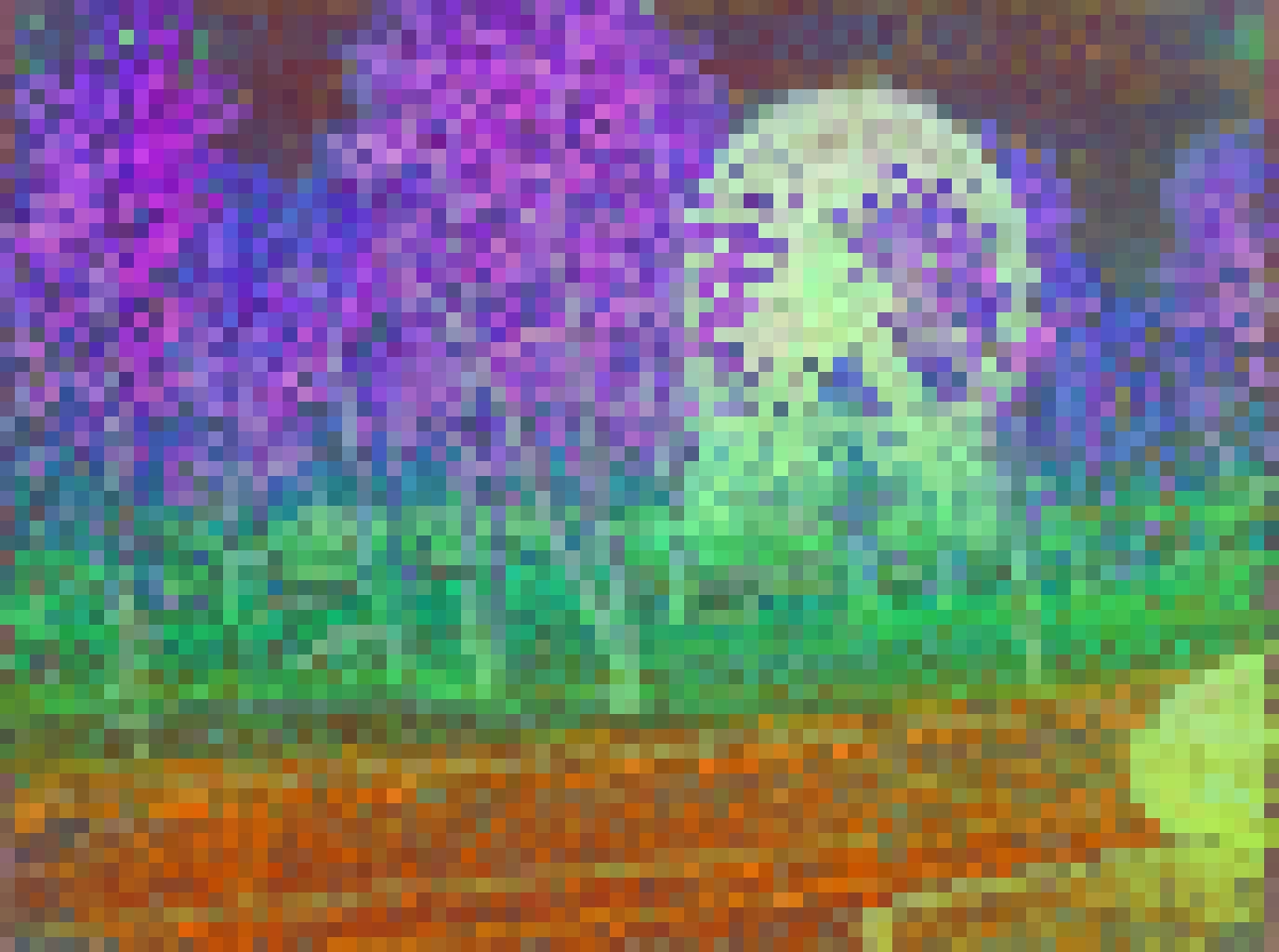} \\
        
        \includegraphics[width=\linewidth]{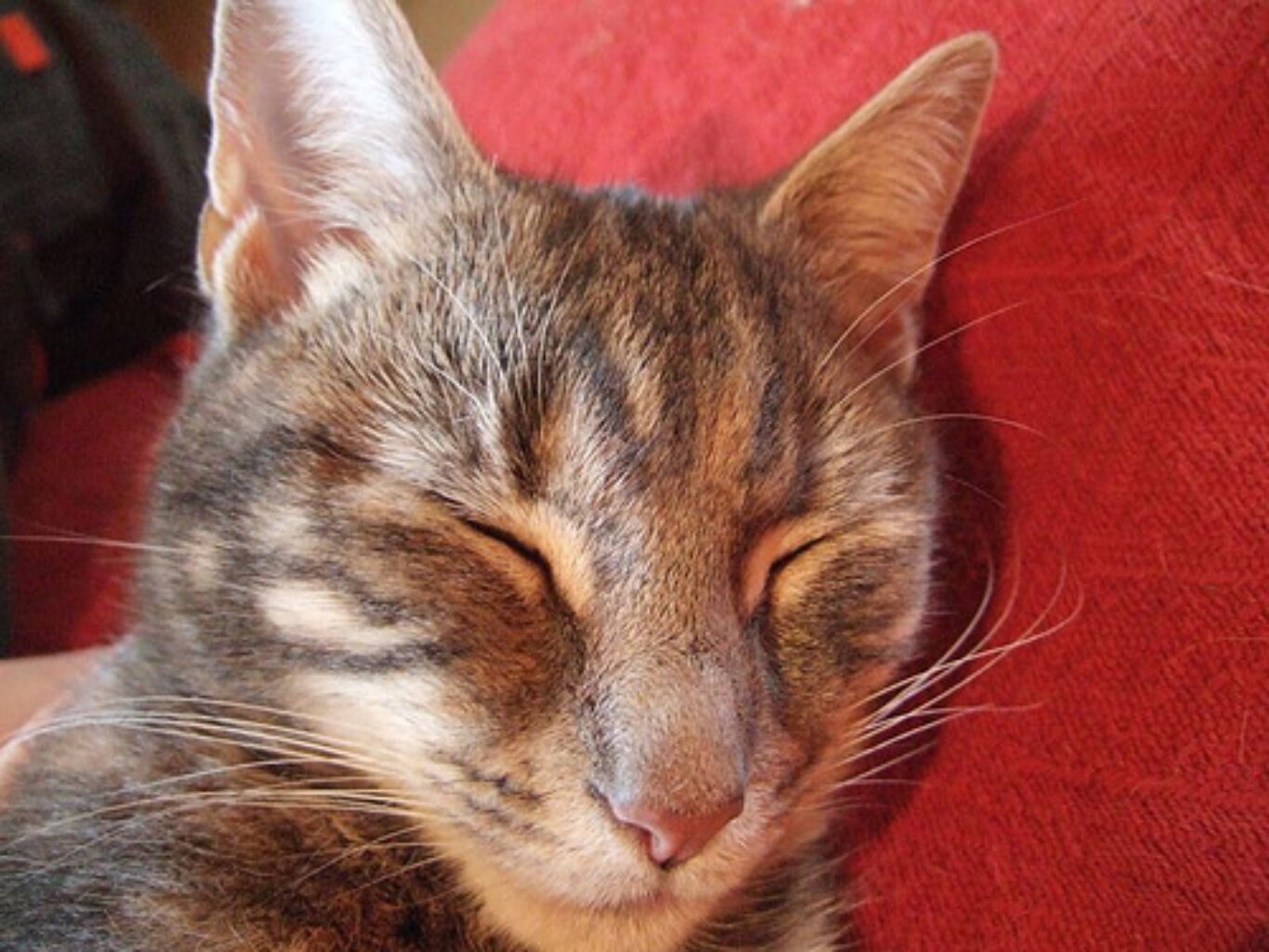} &
        \includegraphics[width=\linewidth]{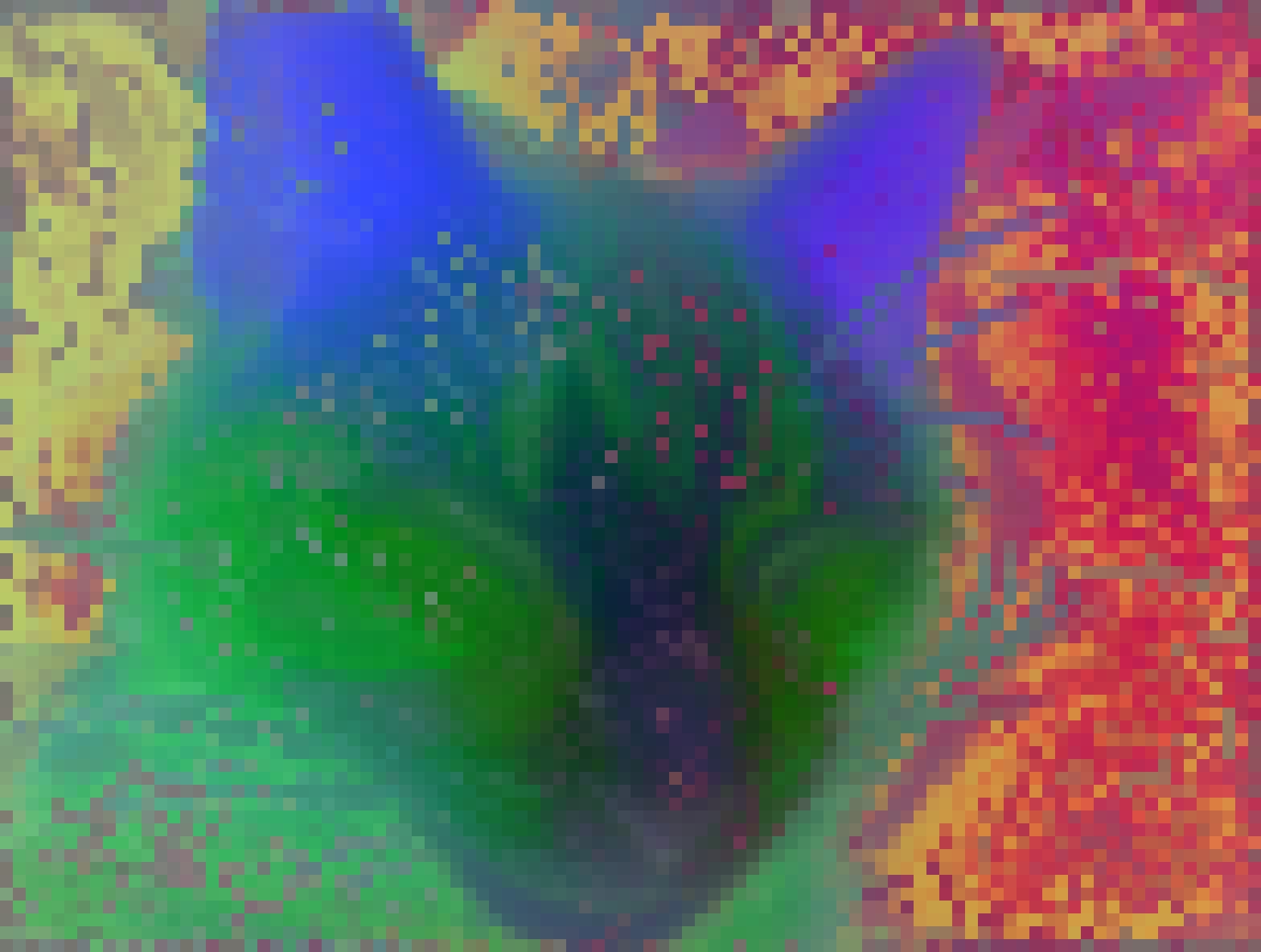} &
        \includegraphics[width=\linewidth]{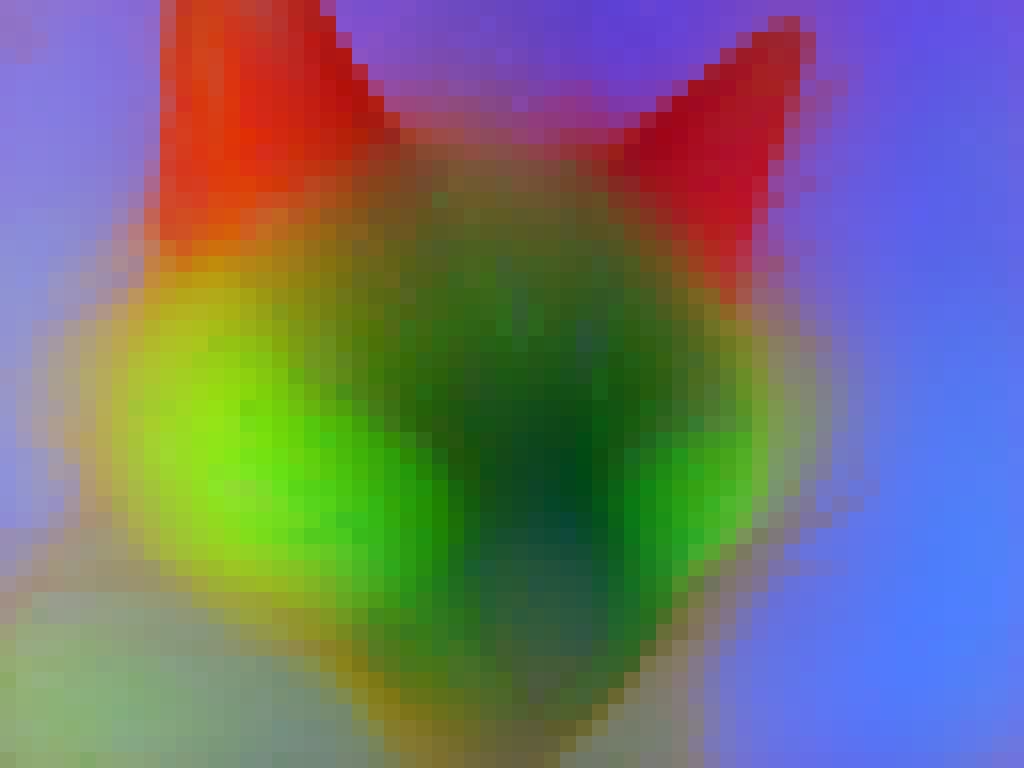} &
        \includegraphics[width=\linewidth]{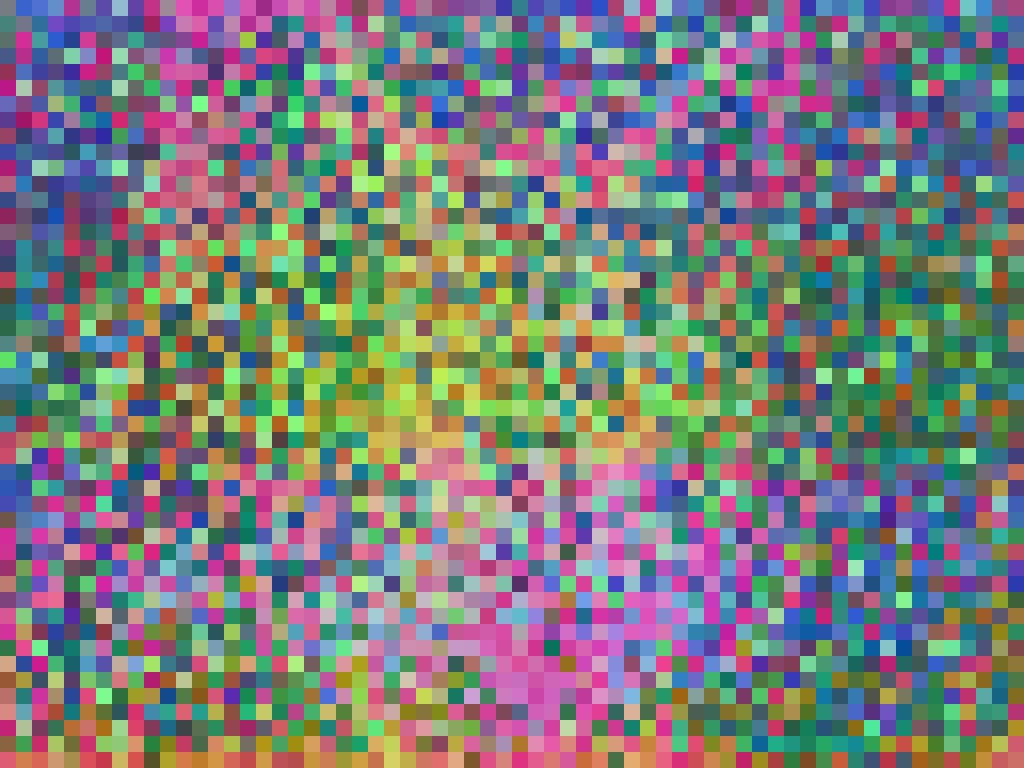} &
        \includegraphics[width=\linewidth]{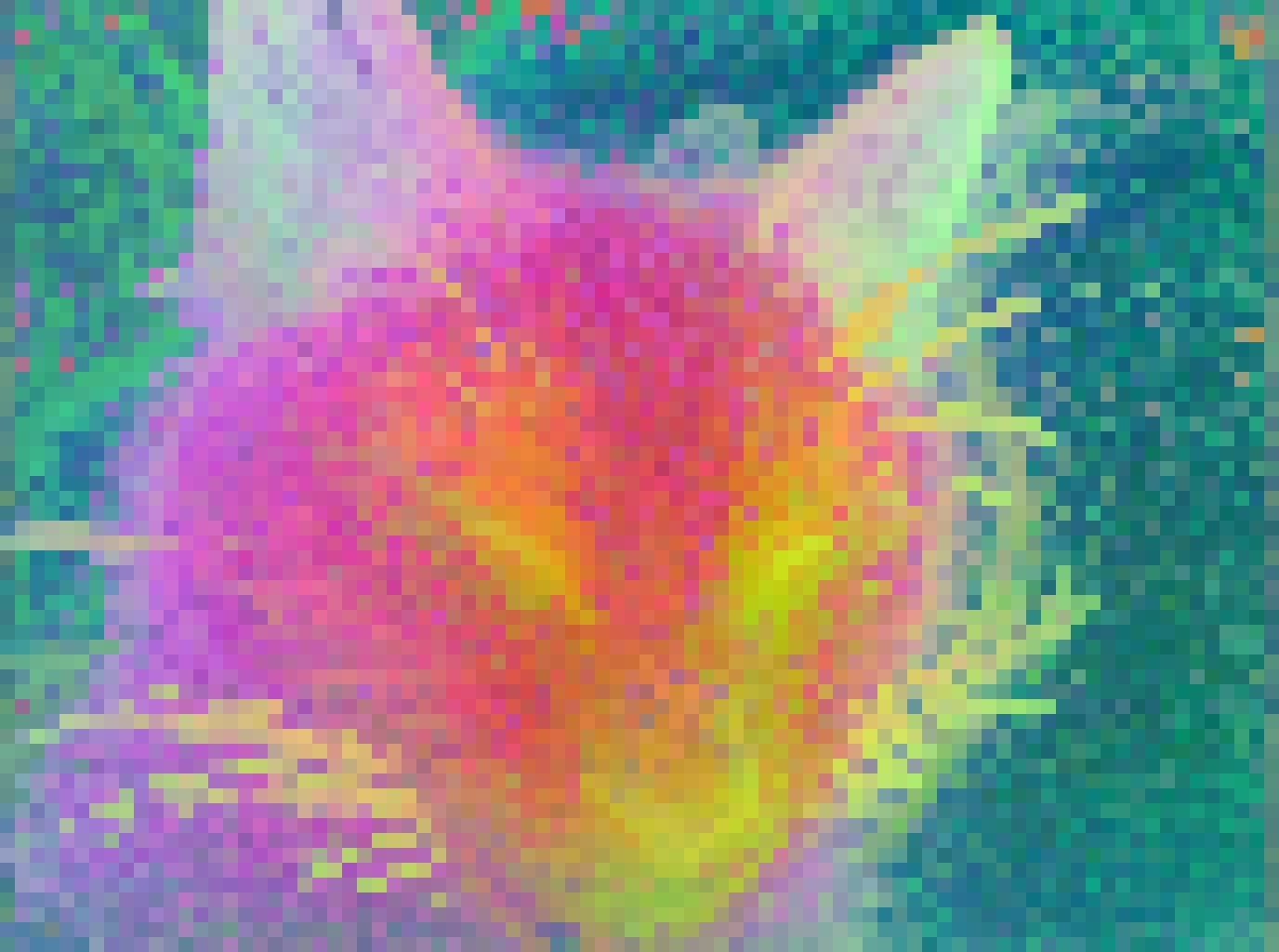} \\
        
        \includegraphics[width=\linewidth]{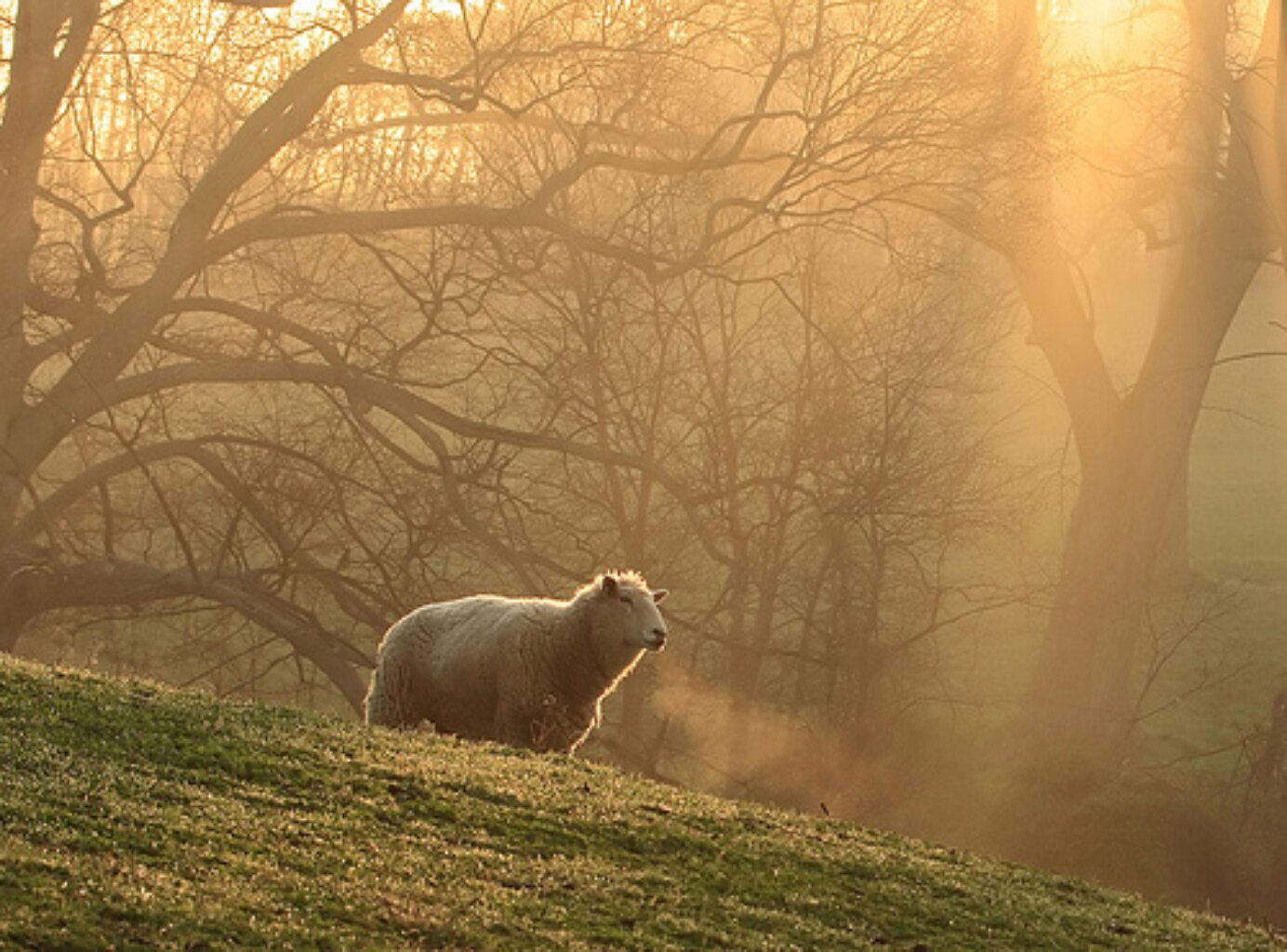} &
        \includegraphics[width=\linewidth]{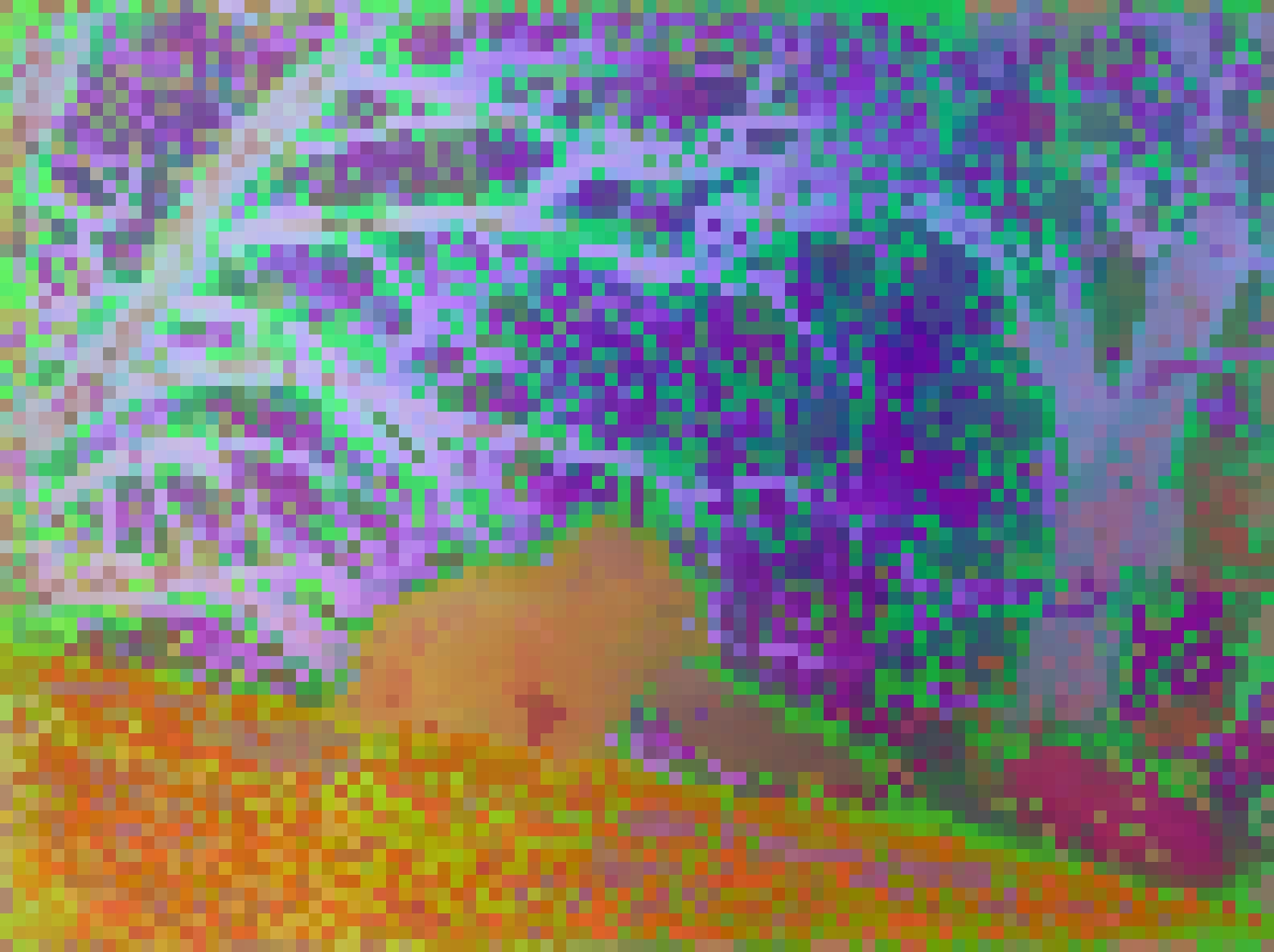} &
        \includegraphics[width=\linewidth]{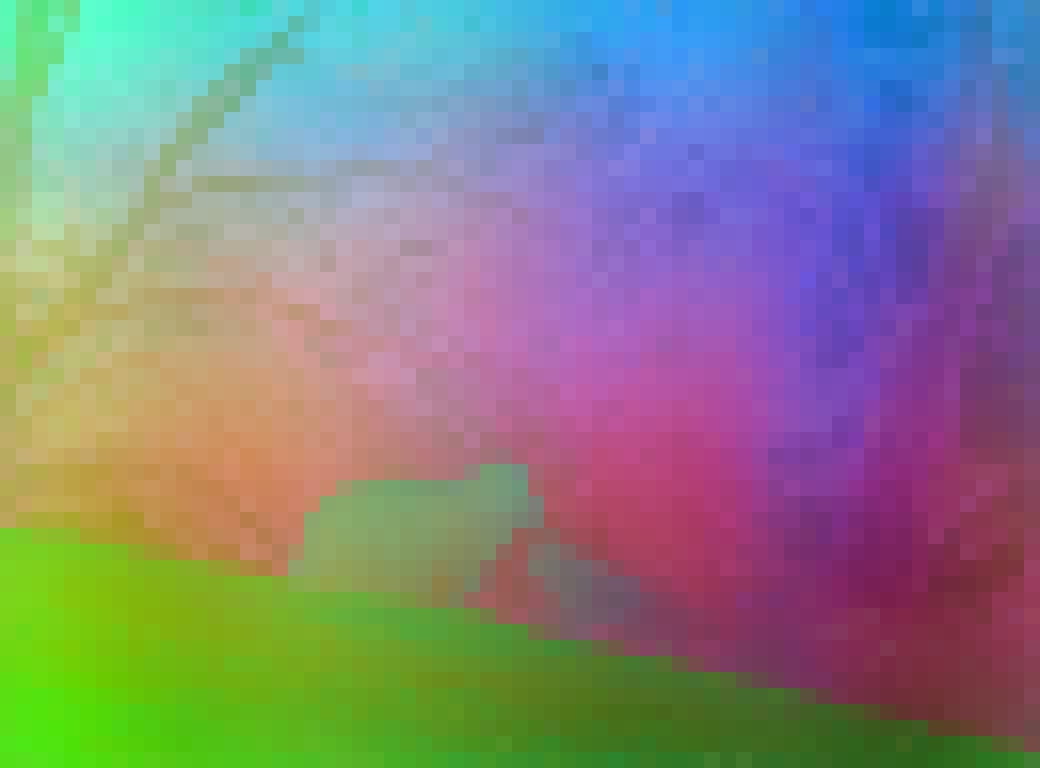} &
        \includegraphics[width=\linewidth]{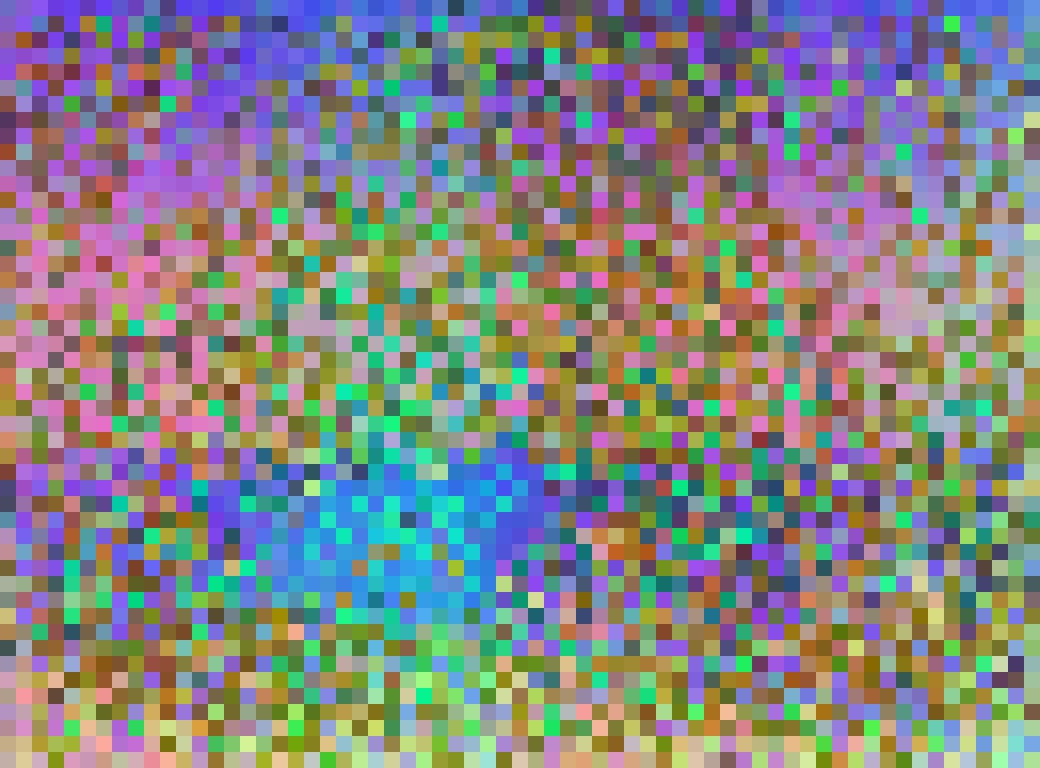} &
        \includegraphics[width=\linewidth]{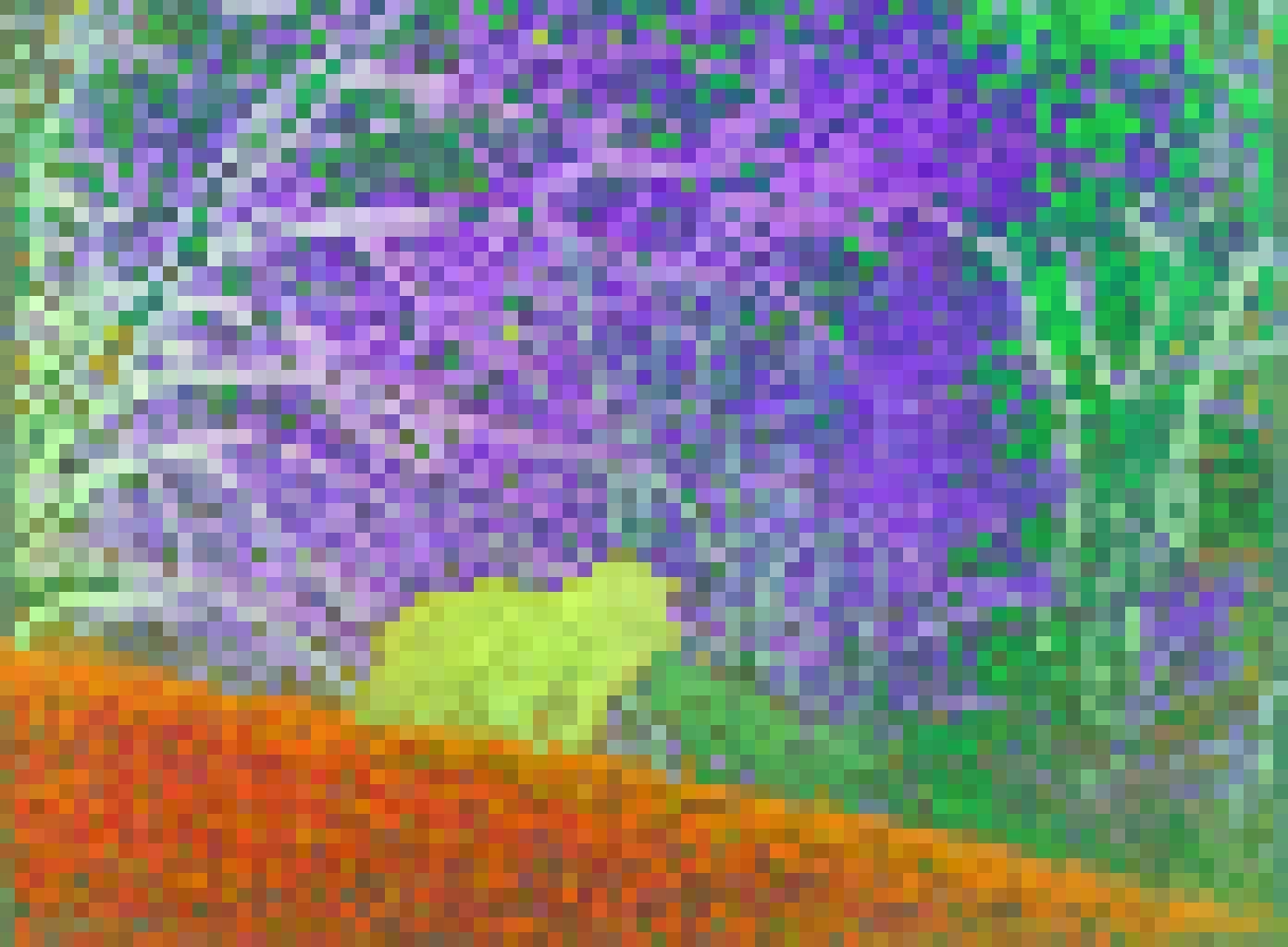} \\
        
        \includegraphics[width=\linewidth]{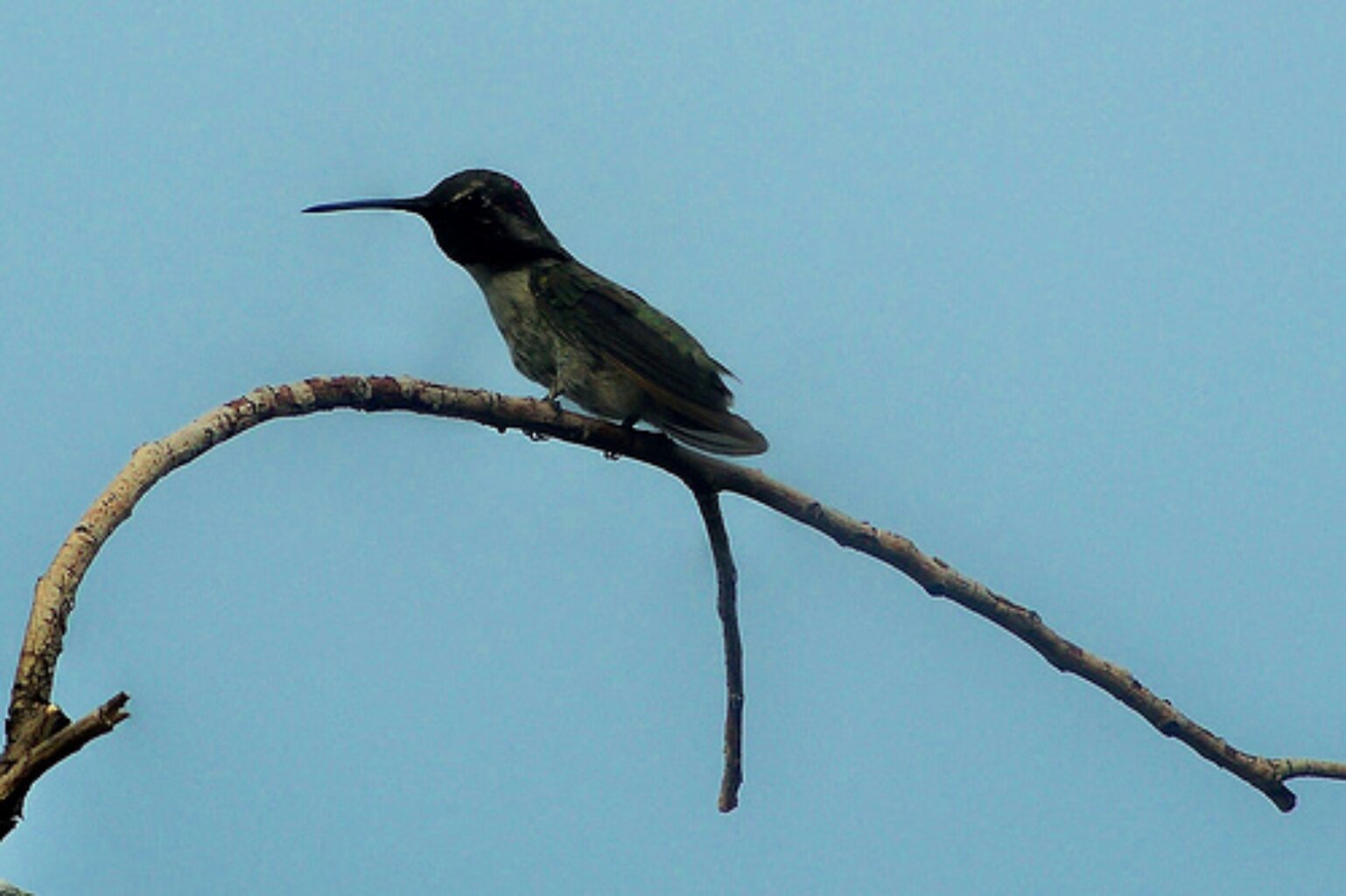} &
        \includegraphics[width=\linewidth]{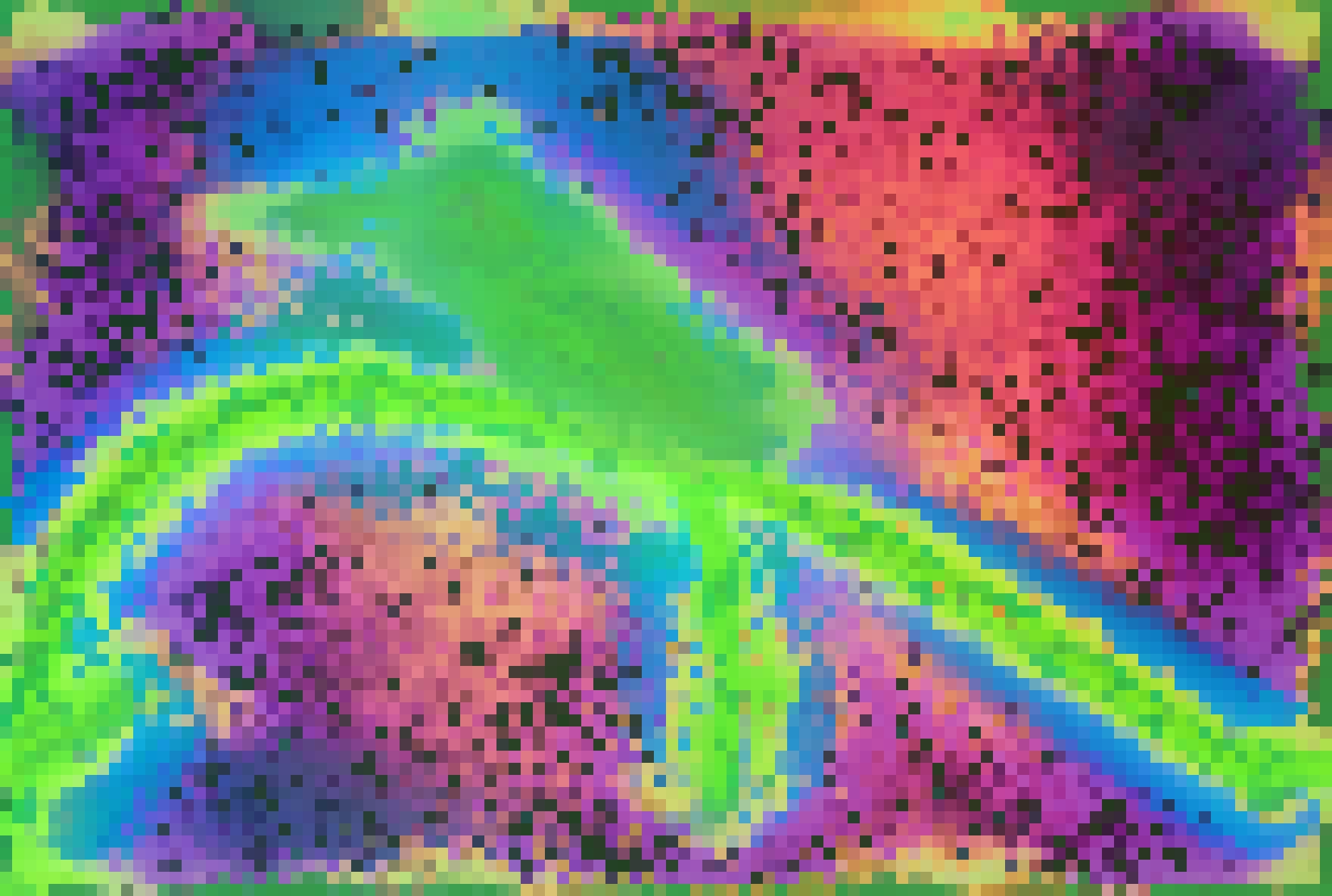} &
        \includegraphics[width=\linewidth]{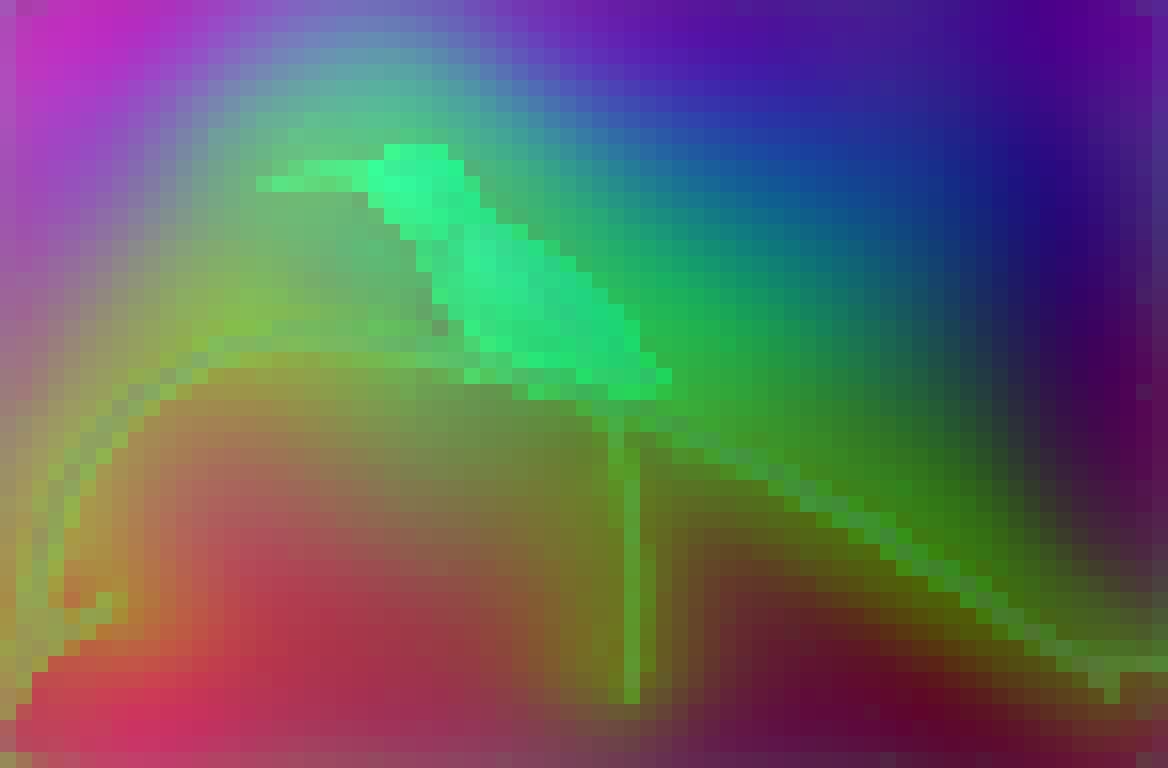} &
        \includegraphics[width=\linewidth]{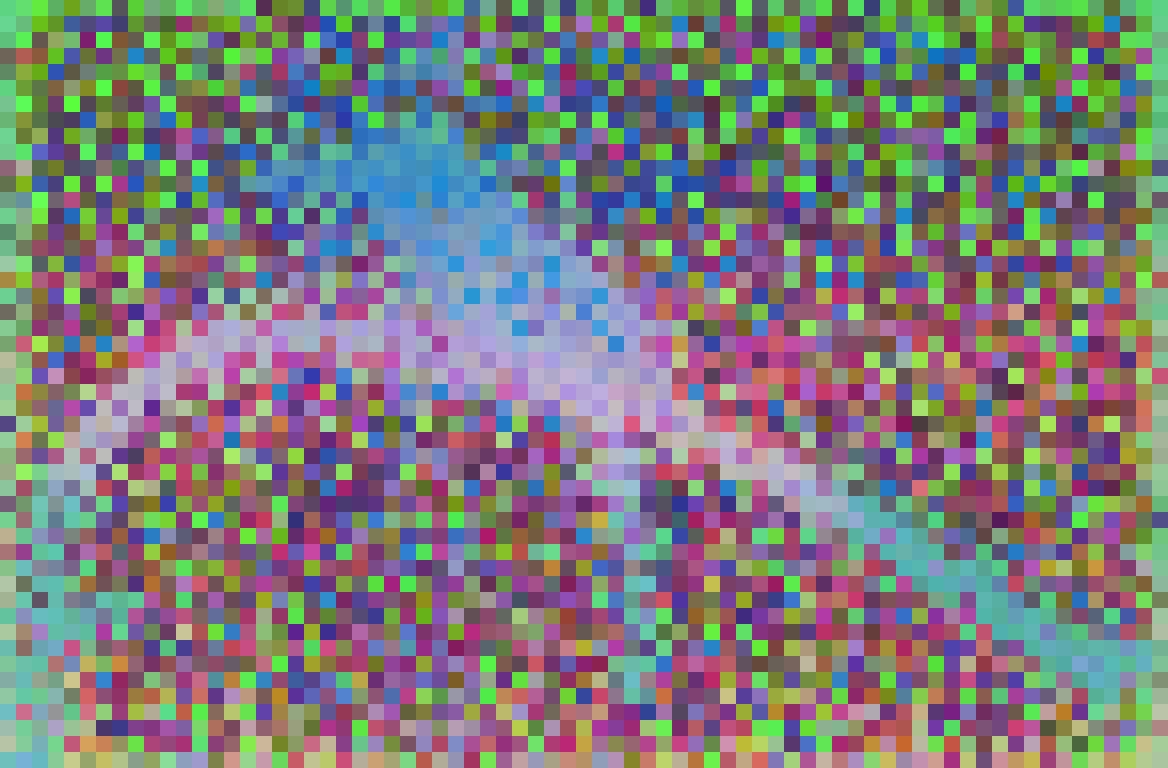} &
        \includegraphics[width=\linewidth]{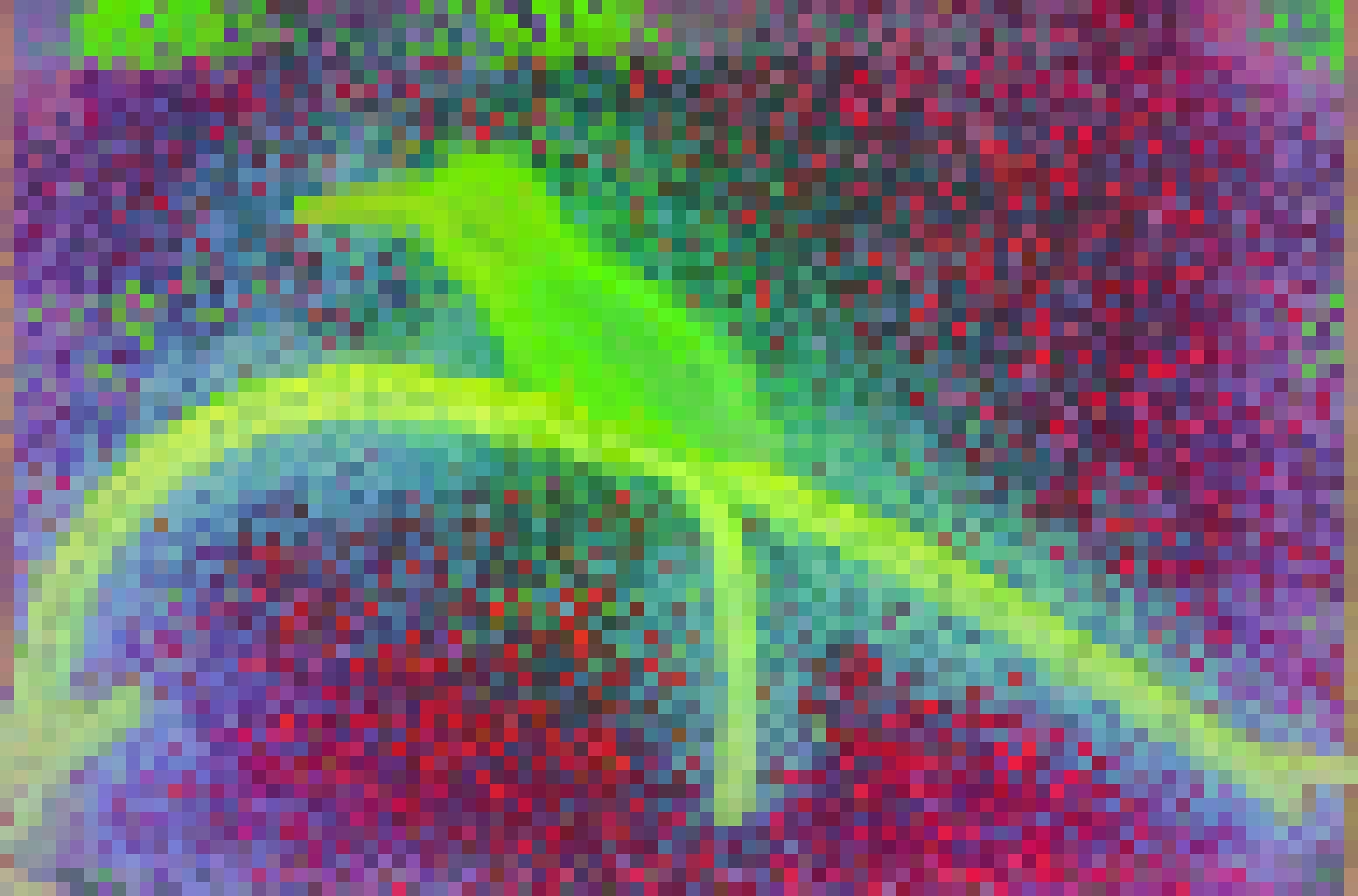} \\
        
        \includegraphics[width=\linewidth]{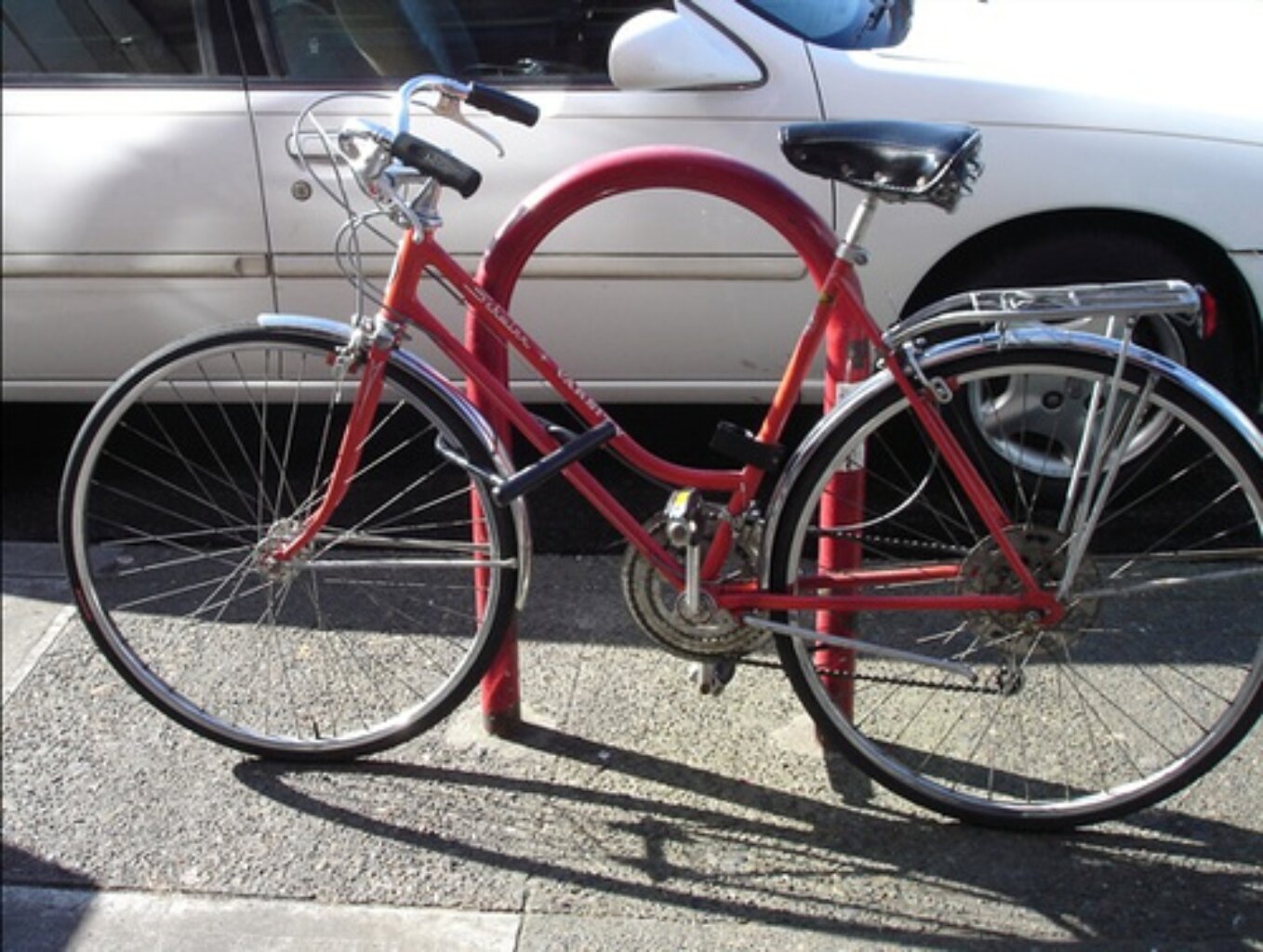} &
        \includegraphics[width=\linewidth]{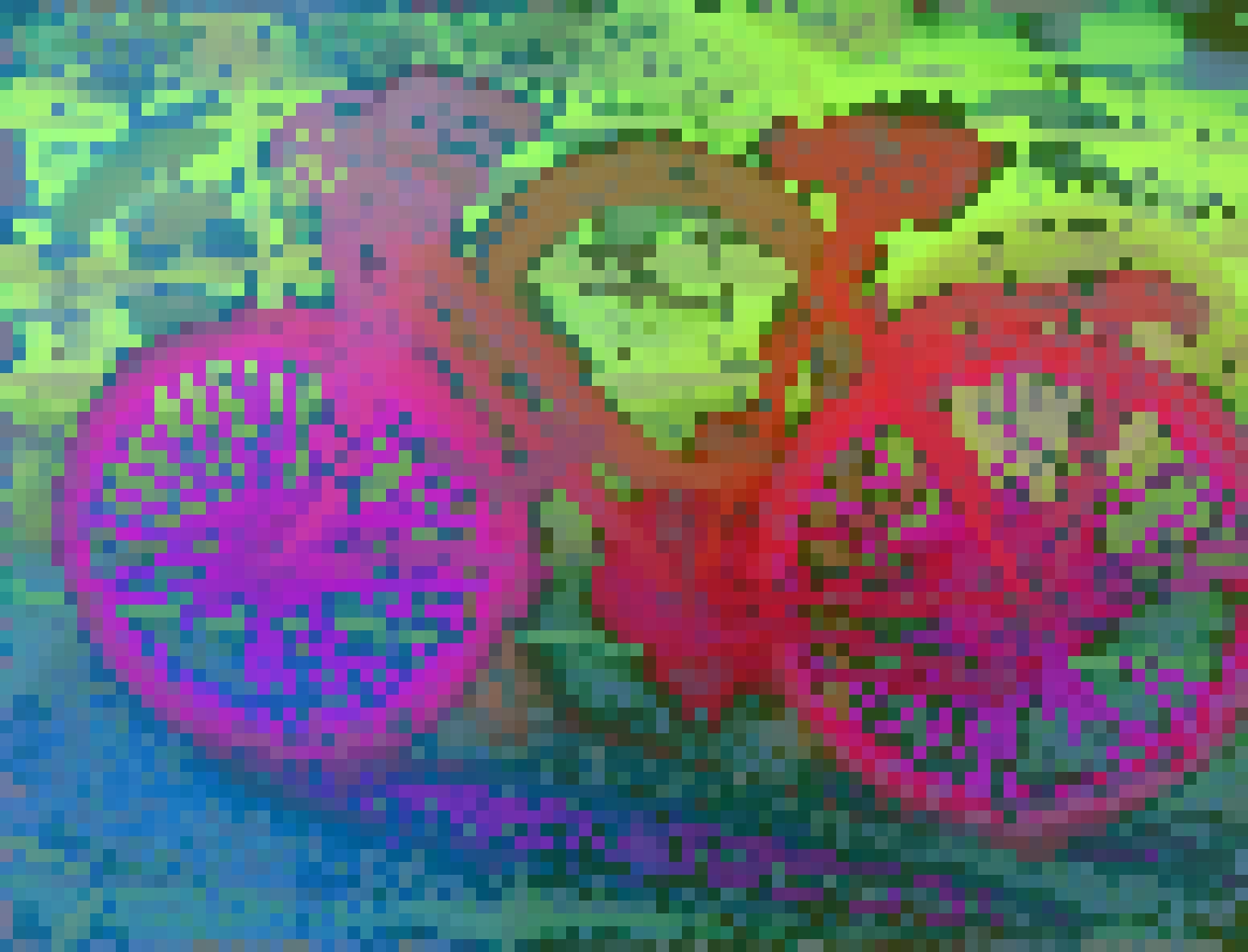} &
        \includegraphics[width=\linewidth]{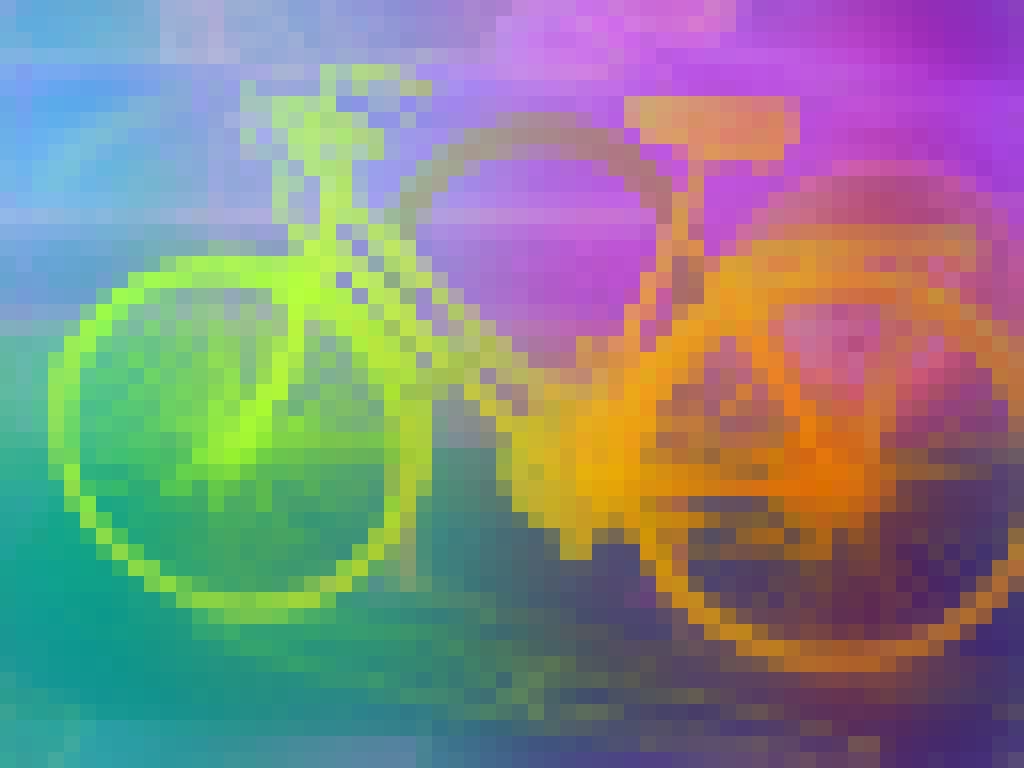} &
        \includegraphics[width=\linewidth]{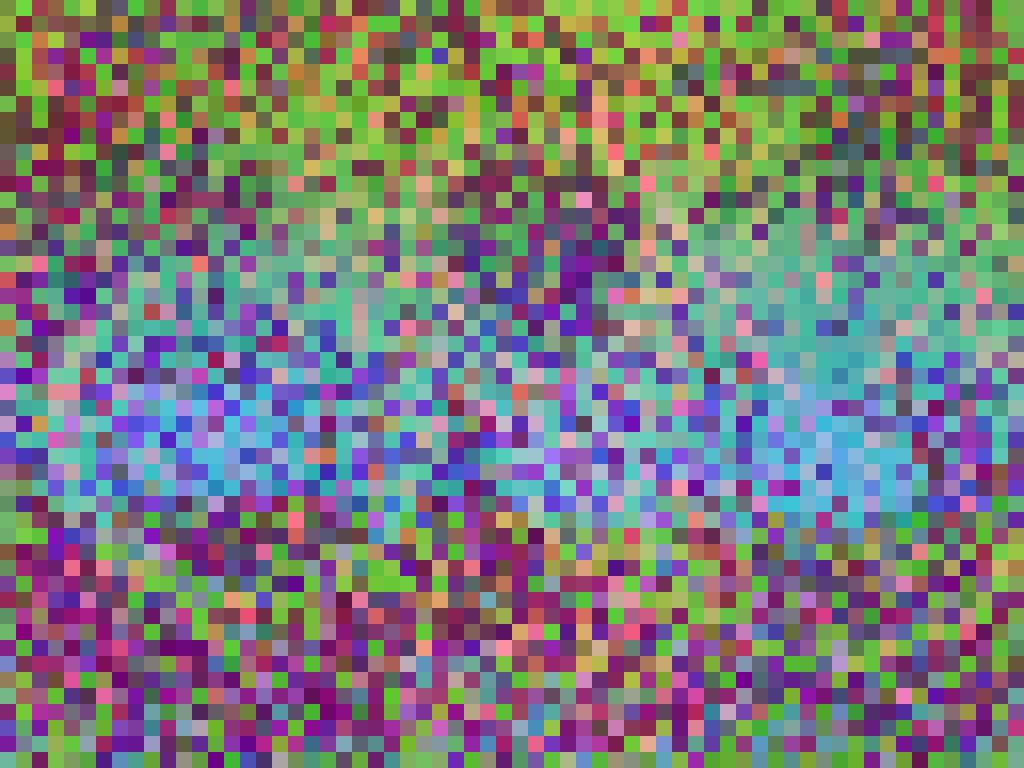} &
        \includegraphics[width=\linewidth]{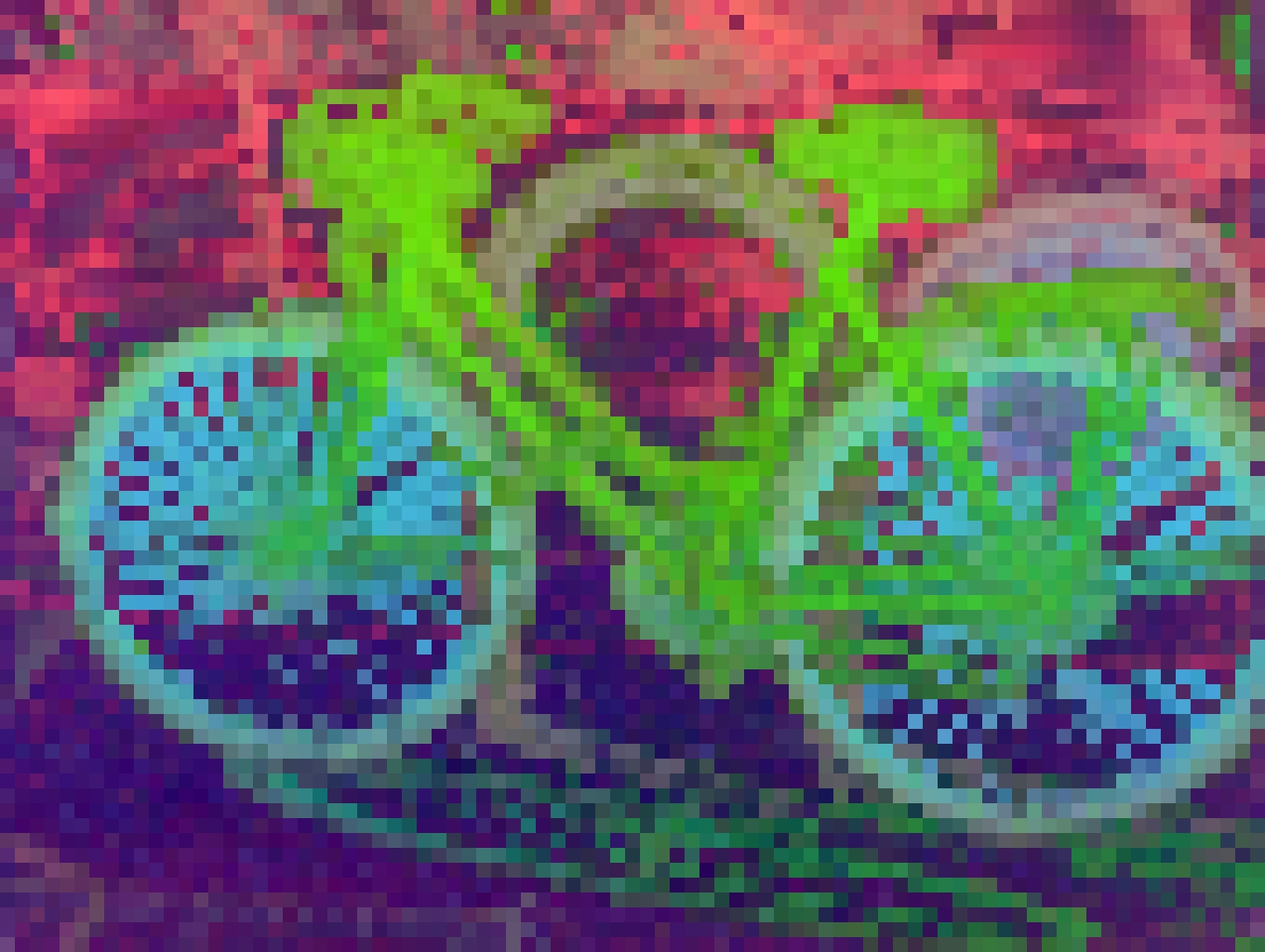} \\
    \end{tabularx}
    \caption{\textbf{Comparison of dense features produced by different SSL methods on a single image.} \normalfont We compare DINOv2 ViT-g \citep{oquab2023dinov2}, DINOv3 ViT-H+ \citep{simeoni2025dinov3} and V-JEPA 2 ViT-g \citep{assran2025v} against V-JEPA 2.1 ViT-G. Images are resized so that their shorter side is set to 1024 pixels, after which they are processed using the 2D convolution image tokenizer.}
    \label{fig: qualit_comparatie}
\end{figure}

\begin{figure}[]
    \centering
    
        \includegraphics[width=0.19\linewidth]{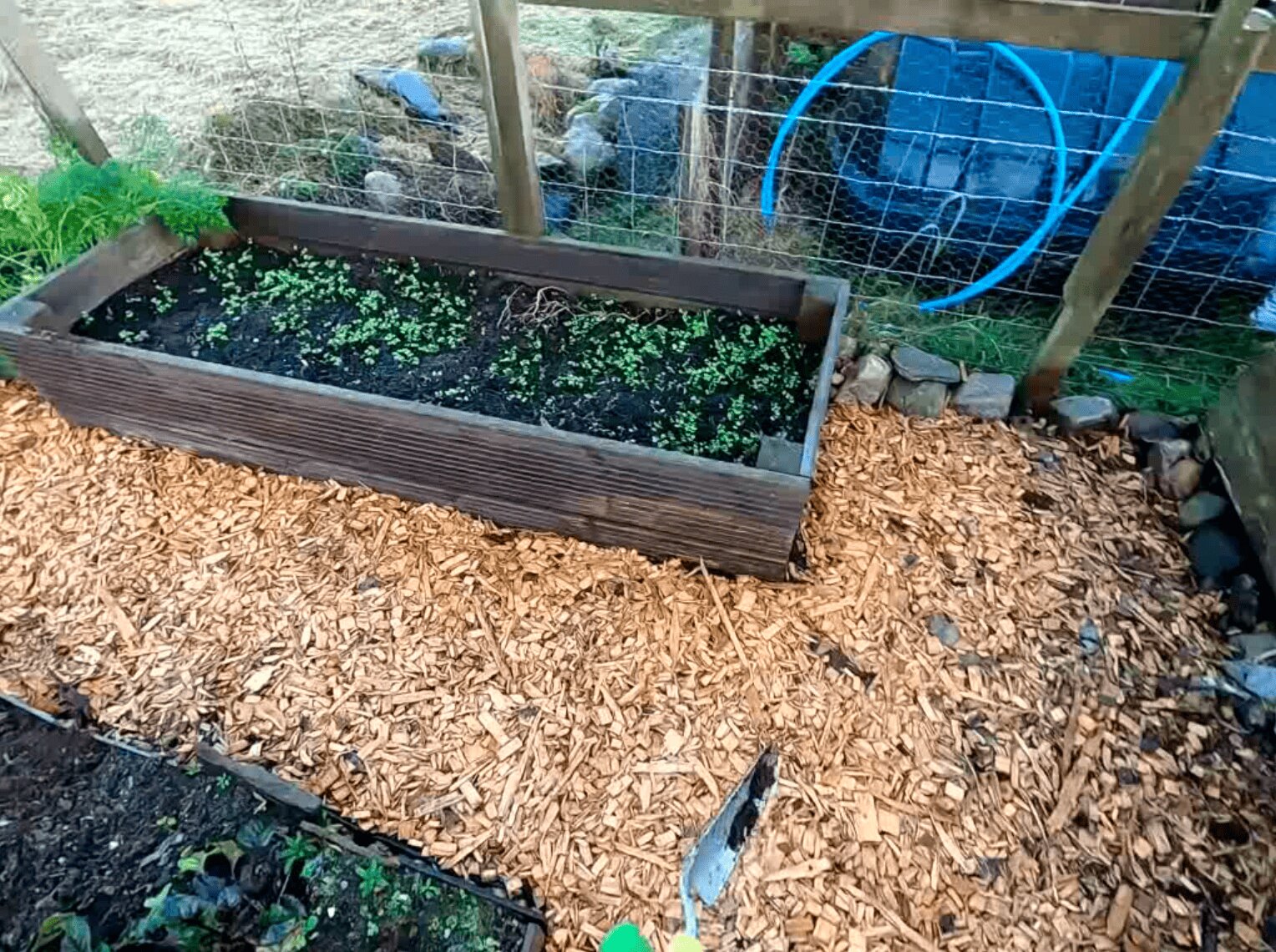}
        \includegraphics[width=0.19\linewidth]{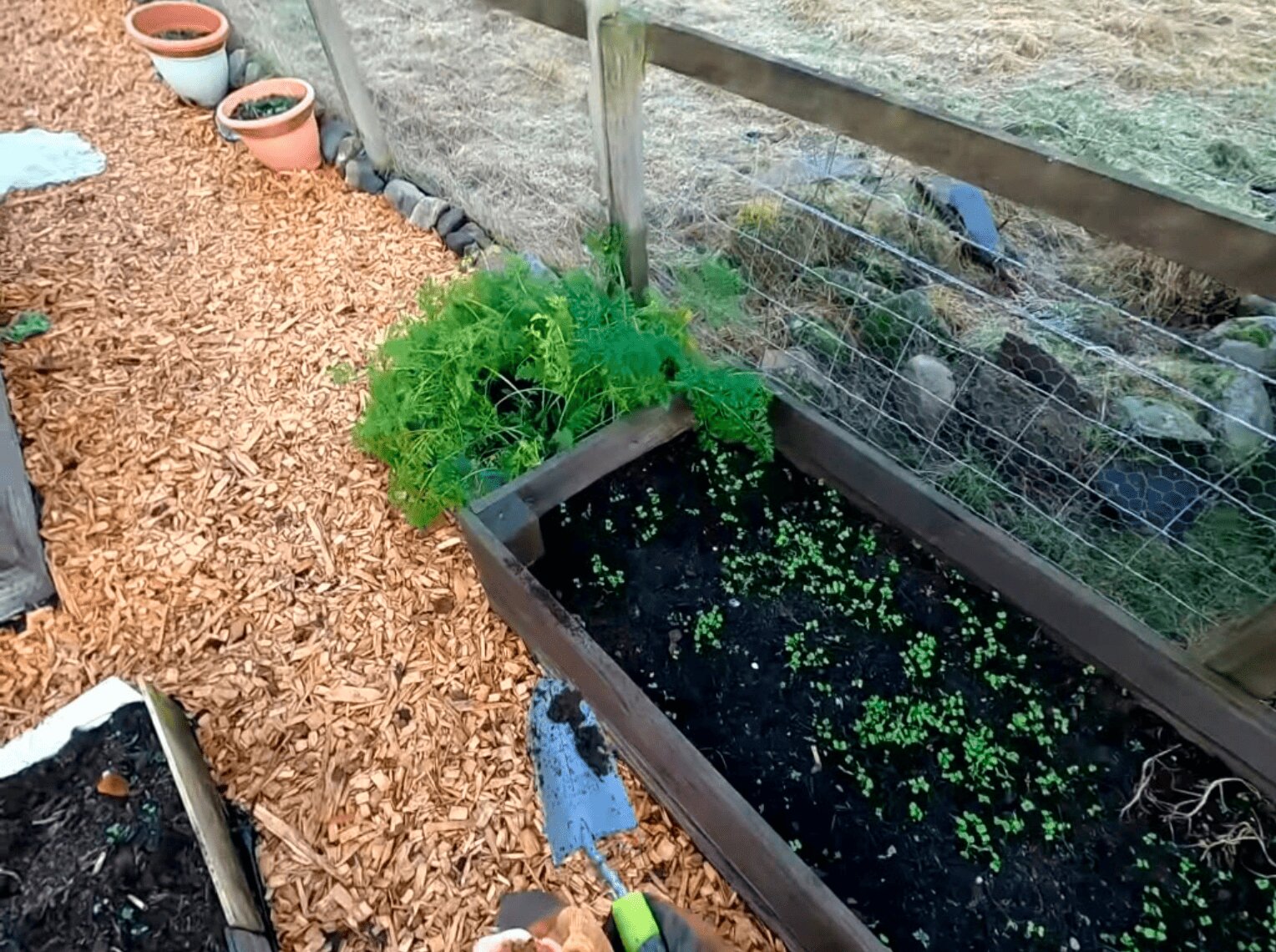}
        \includegraphics[width=0.19\linewidth]{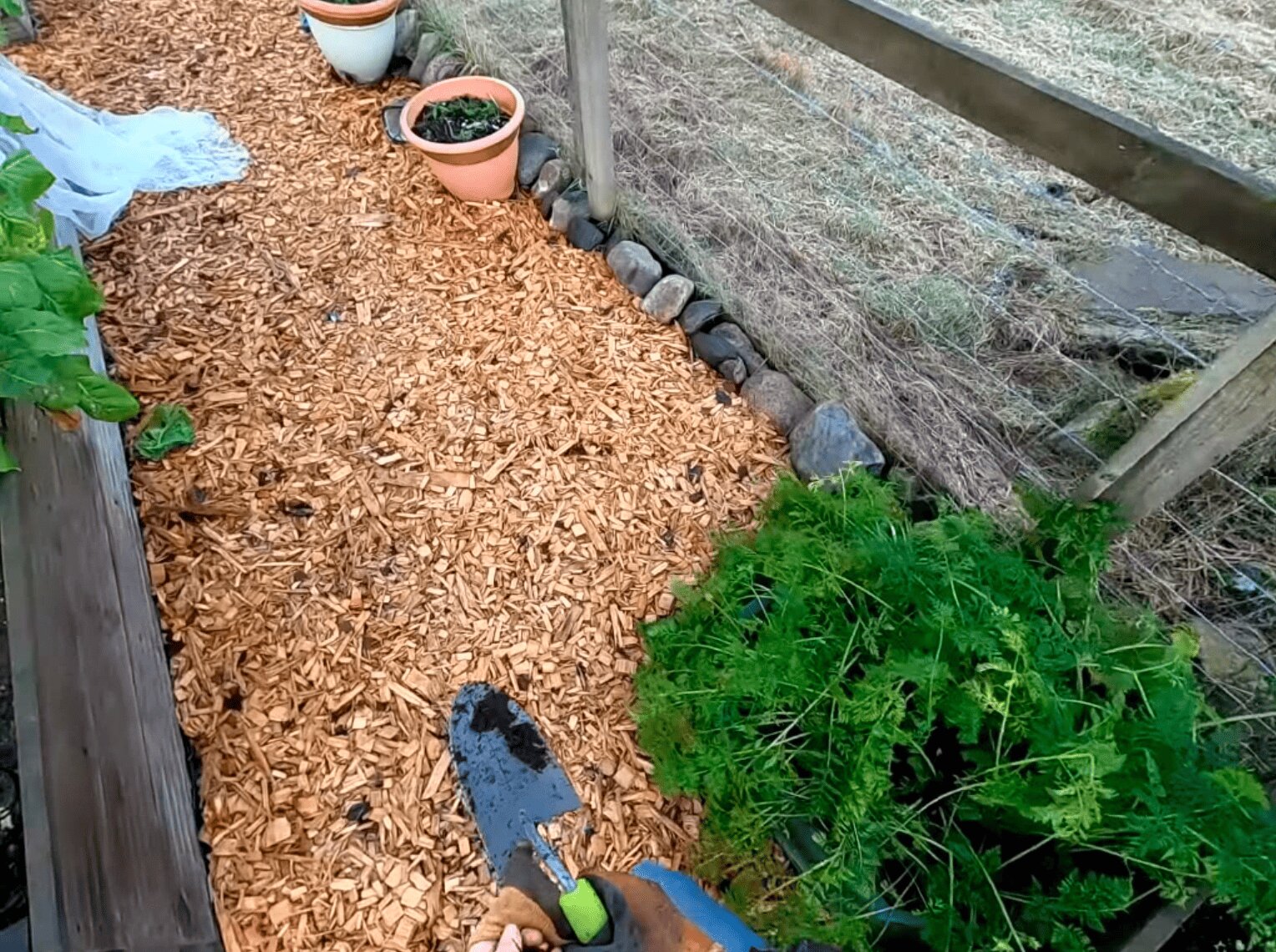}
        \includegraphics[width=0.19\linewidth]{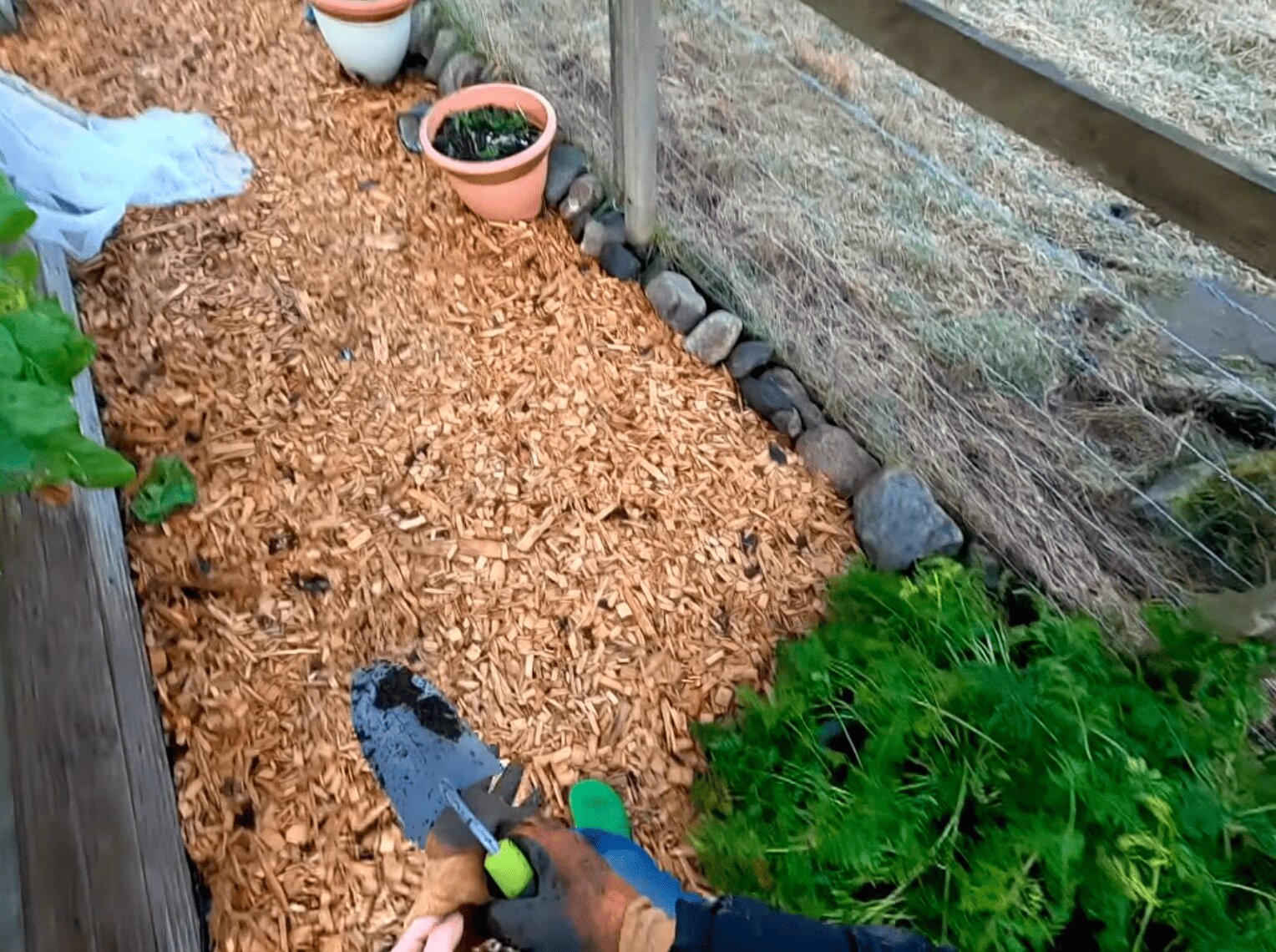}
        \includegraphics[width=0.19\linewidth]{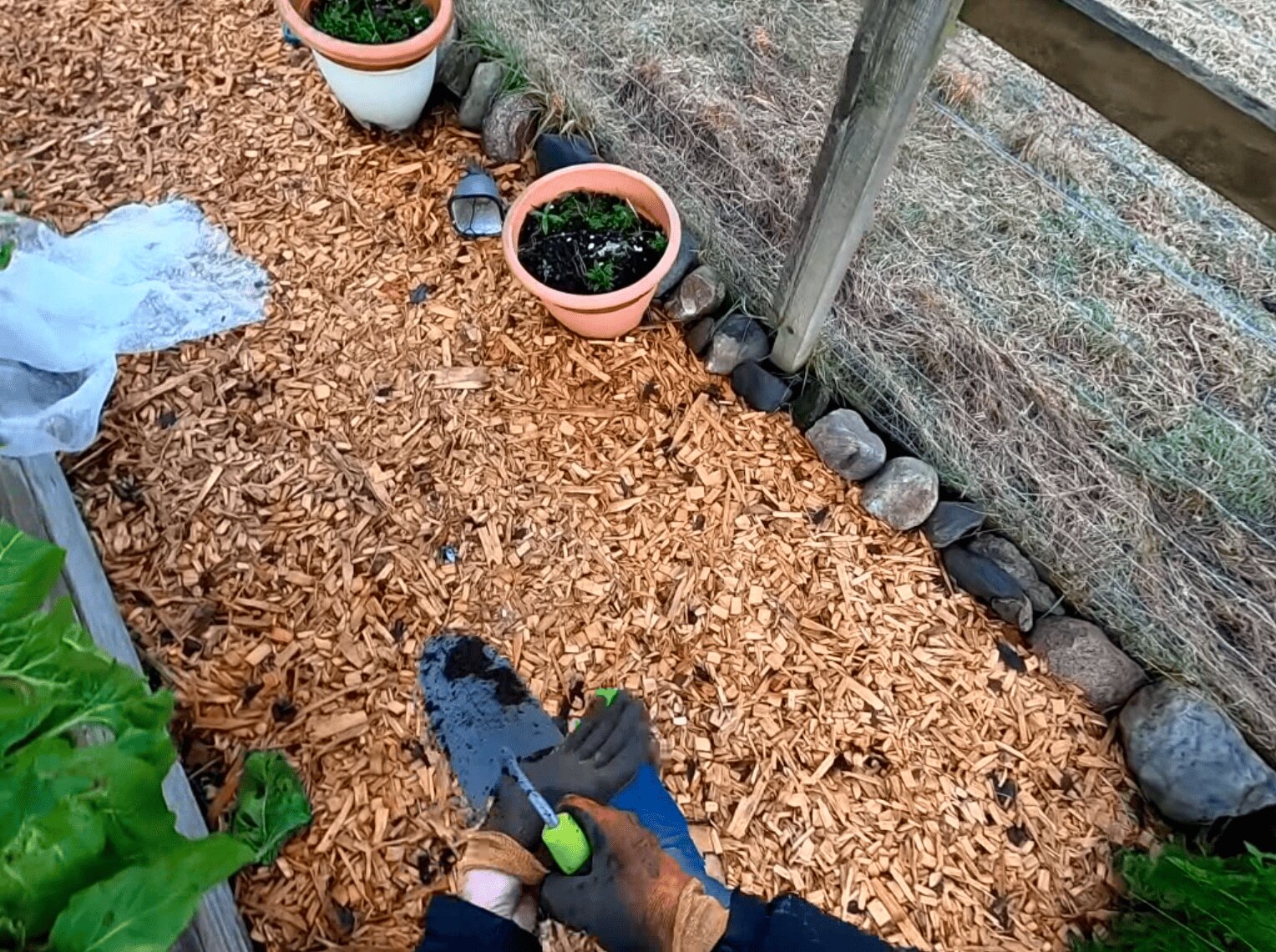} \\

    %\begin{minipage}{0.04\linewidth}
    %    \centering
    %    \rotatebox{90}{\bf{V-JEPA 2}}
    %\end{minipage}%
    %\begin{minipage}{0.95\linewidth}
    %    \includegraphics[width=0.19\linewidth]{video 101 vjepa/1.jpg}
    %    %\includegraphics[width=0.19\linewidth]{video 101 vjepa/2.jpg}
    %    \includegraphics[width=0.19\linewidth]{video 101 vjepa/3.jpg}
    %    %\includegraphics[width=0.19\linewidth]{video 101 vjepa/4.jpg}
    %    \includegraphics[width=0.19\linewidth]{video 101 vjepa/5.jpg}
    %    \includegraphics[width=0.19\linewidth]{video 101 vjepa/6.jpg}
    %    \includegraphics[width=0.19\linewidth]{video 101 vjepa/8.jpg}
    %\end{minipage} \\

        \includegraphics[width=0.19\linewidth]{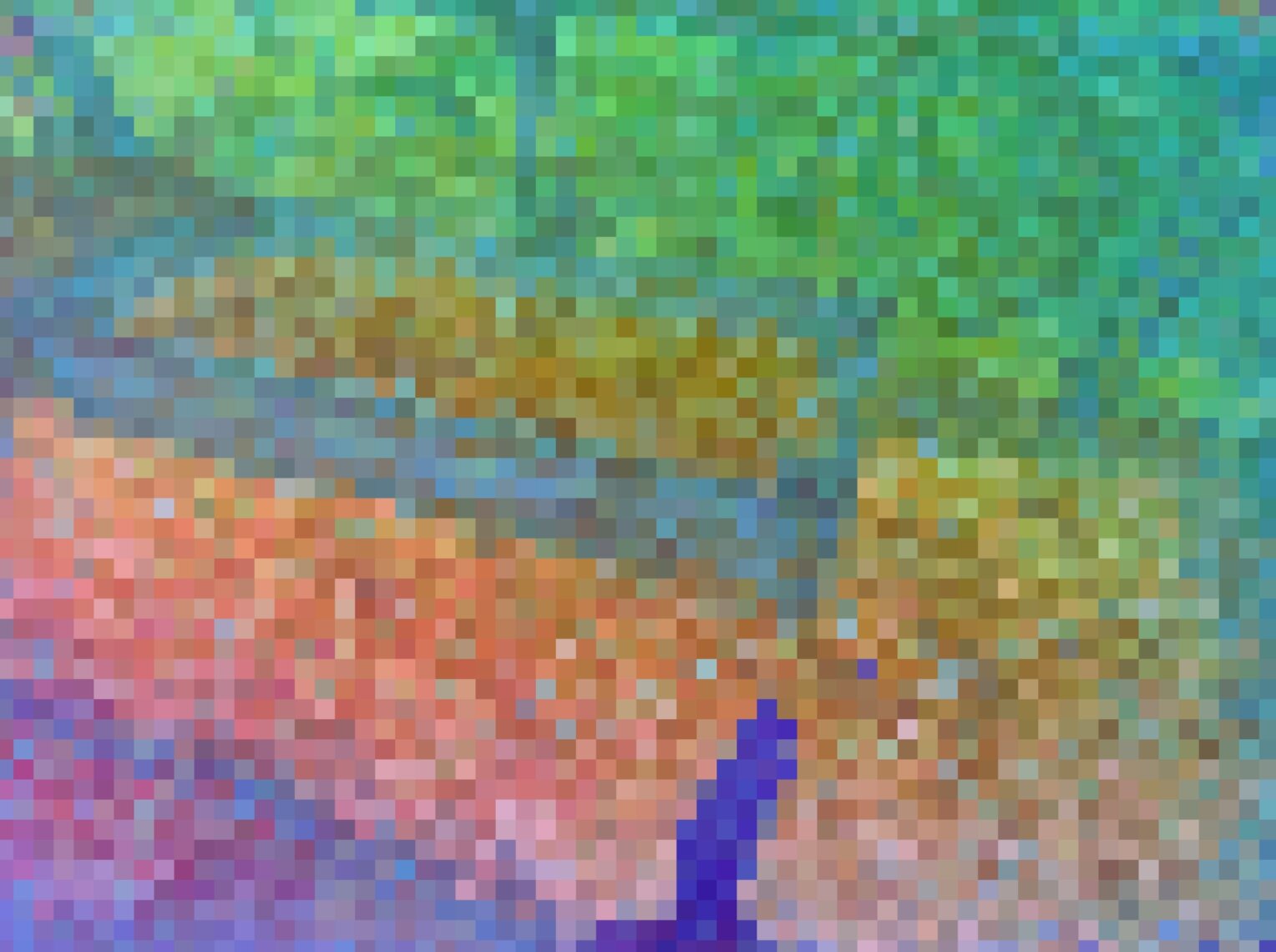}
        \includegraphics[width=0.19\linewidth]{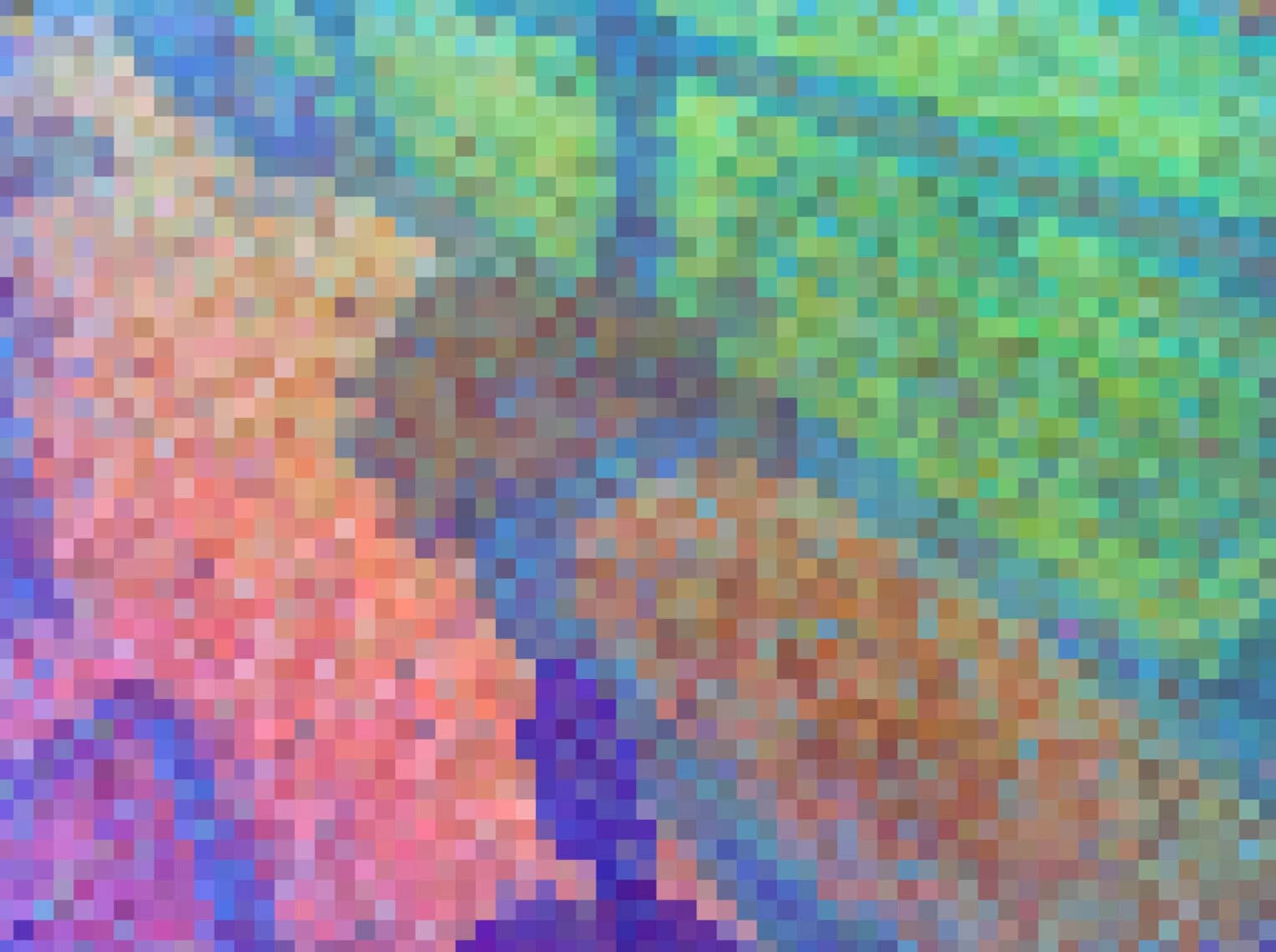}
        \includegraphics[width=0.19\linewidth]{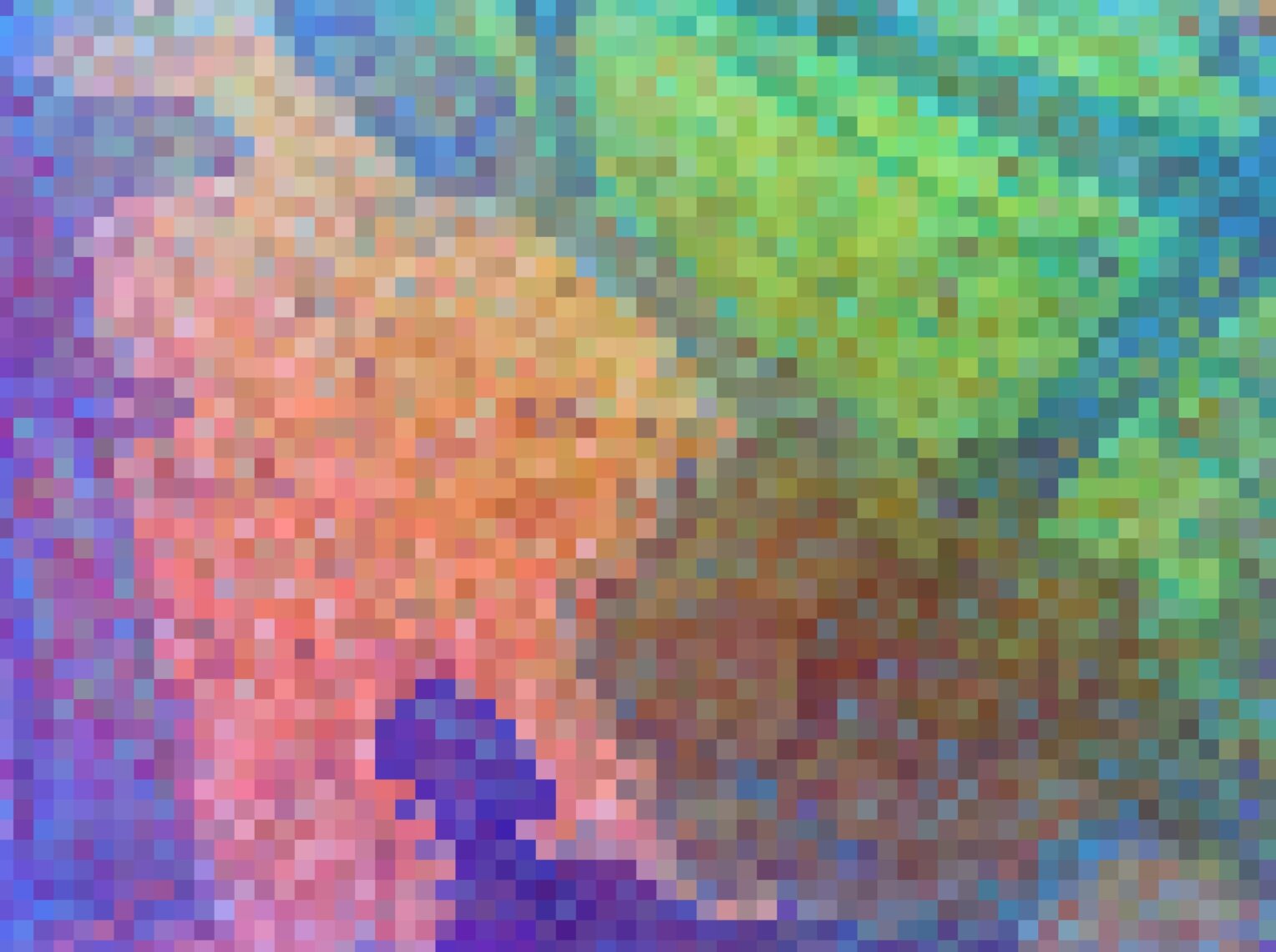}
        \includegraphics[width=0.19\linewidth]{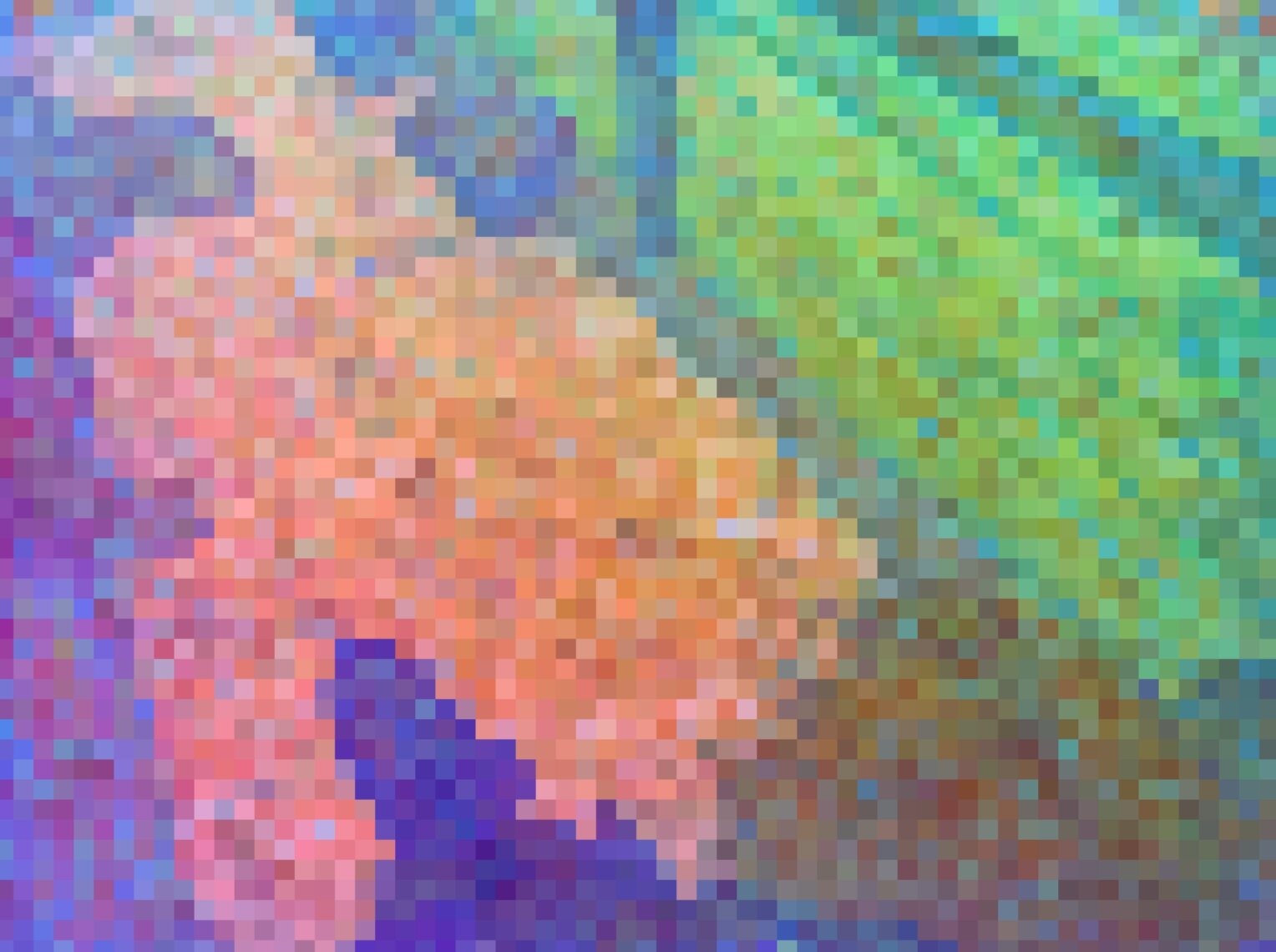}
        \includegraphics[width=0.19\linewidth]{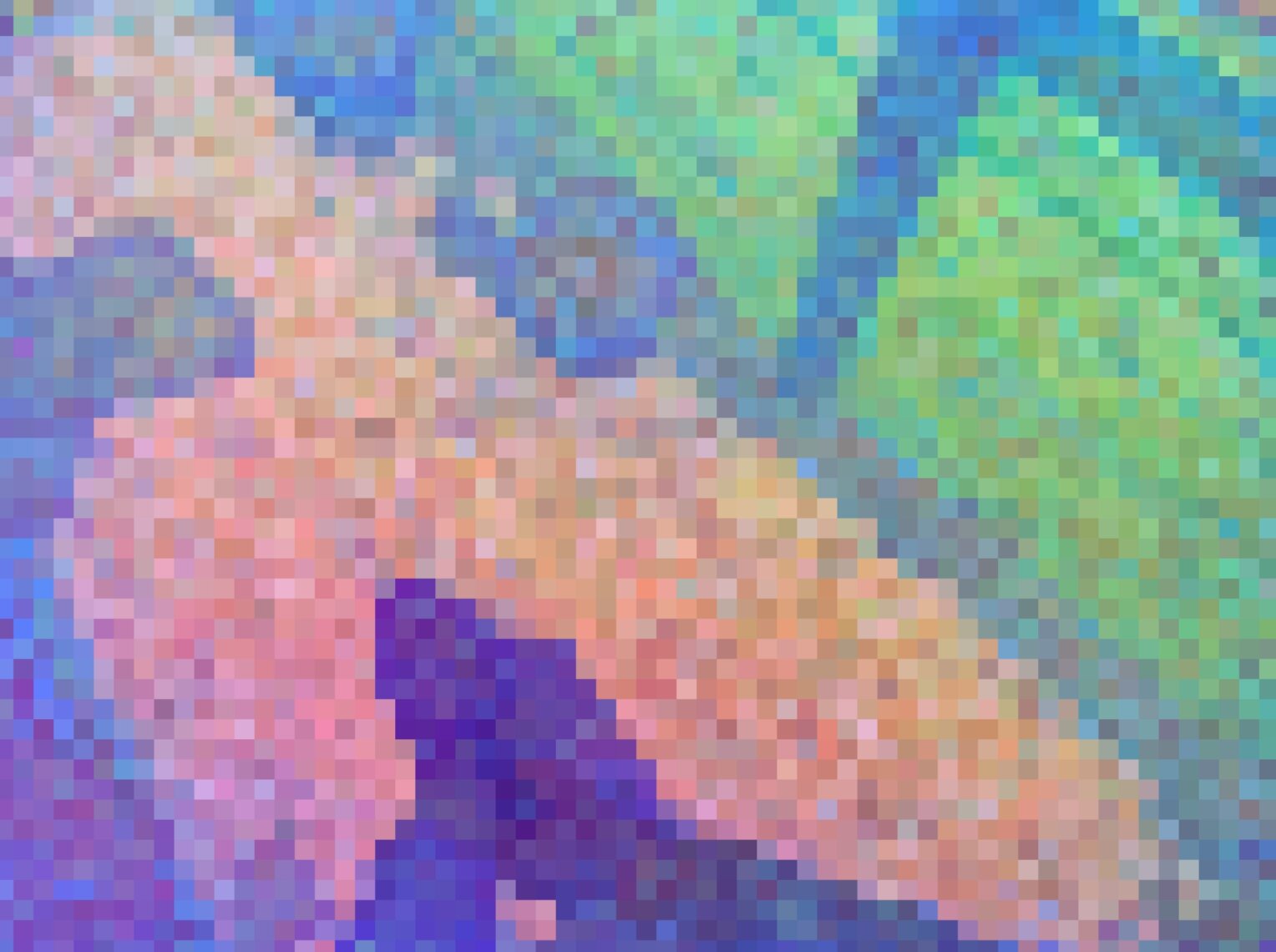} \\

        \vspace{0.1cm}

        \includegraphics[width=0.19\linewidth]{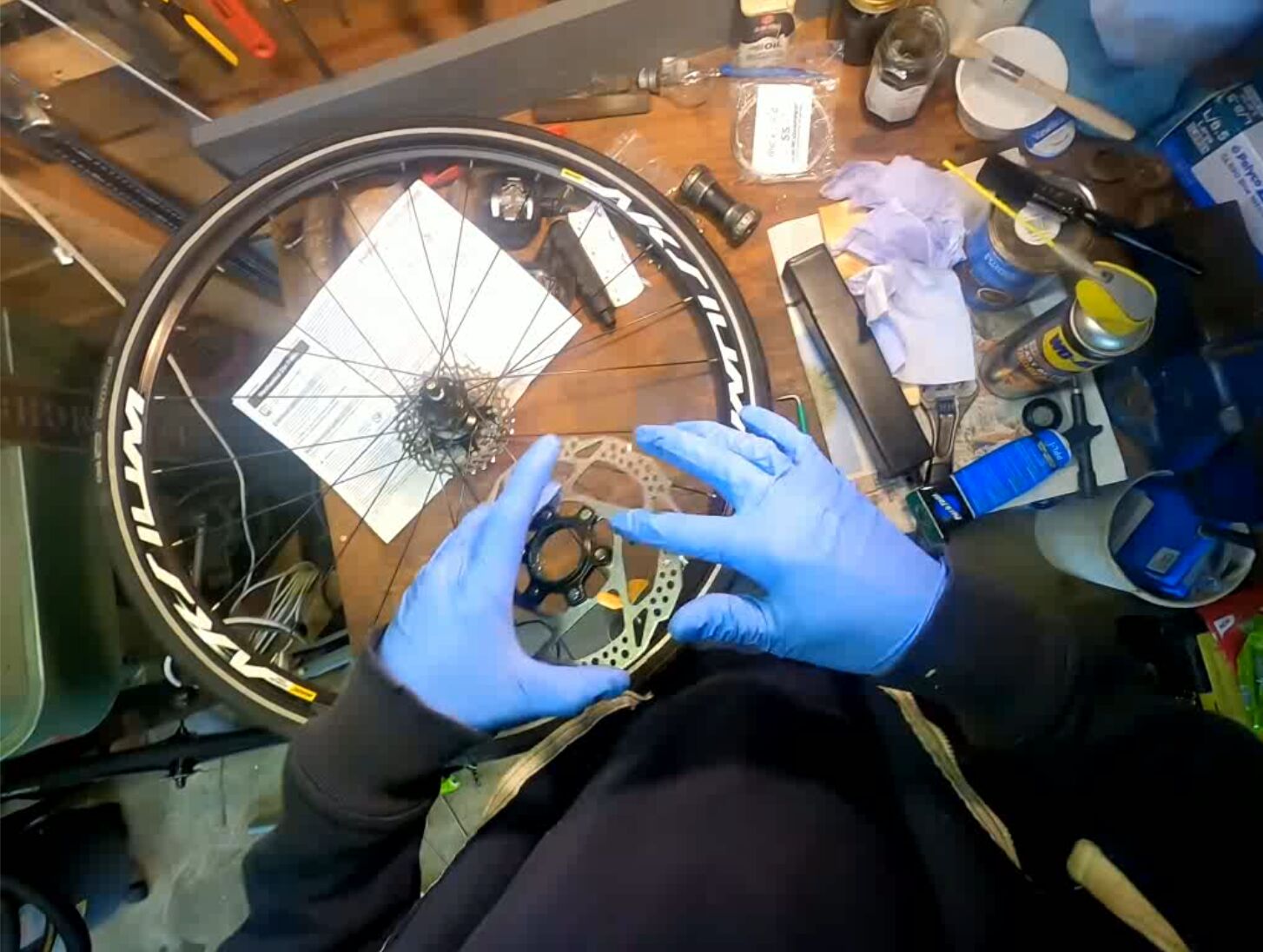}
        \includegraphics[width=0.19\linewidth]{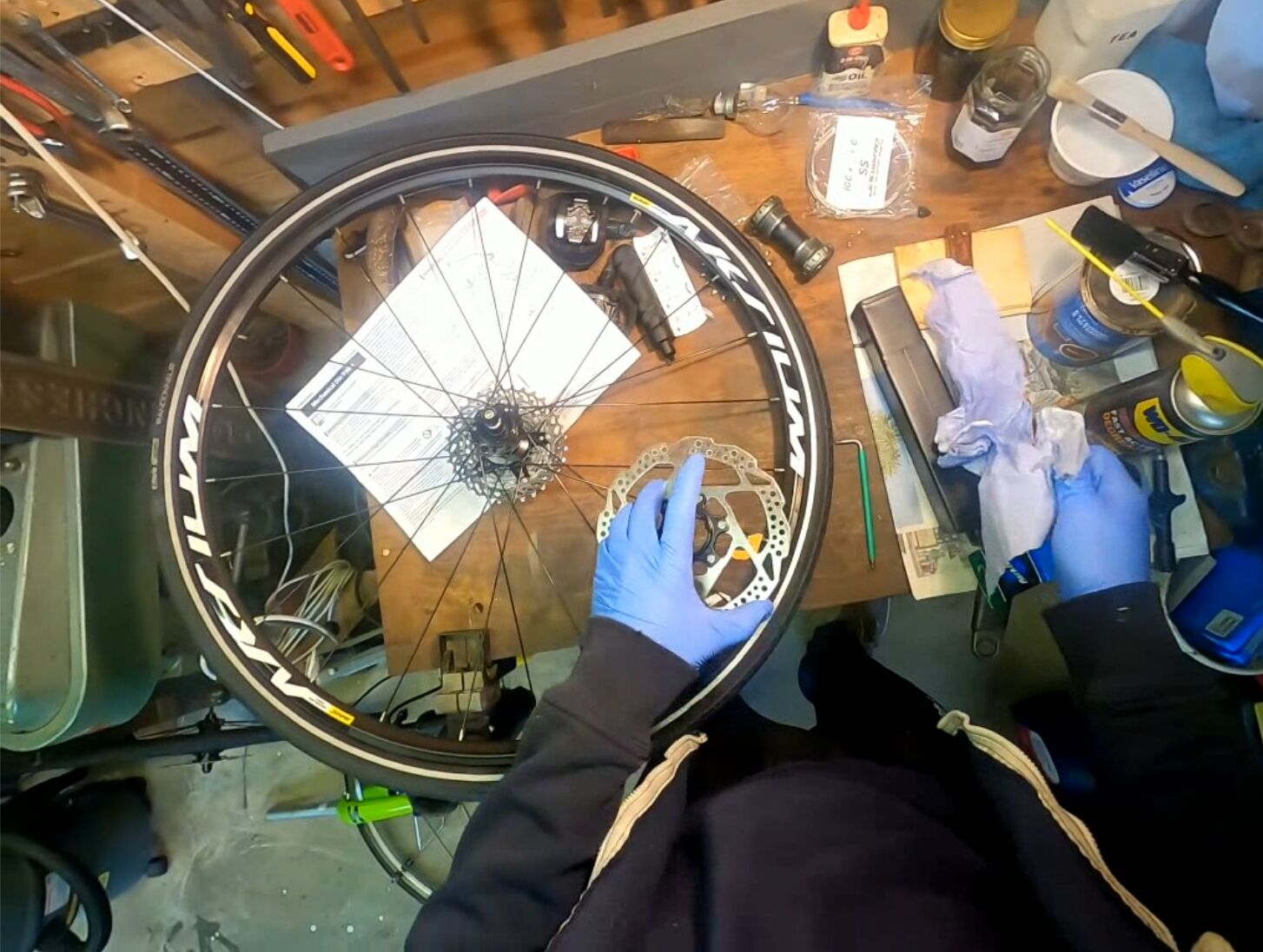}
        \includegraphics[width=0.19\linewidth]{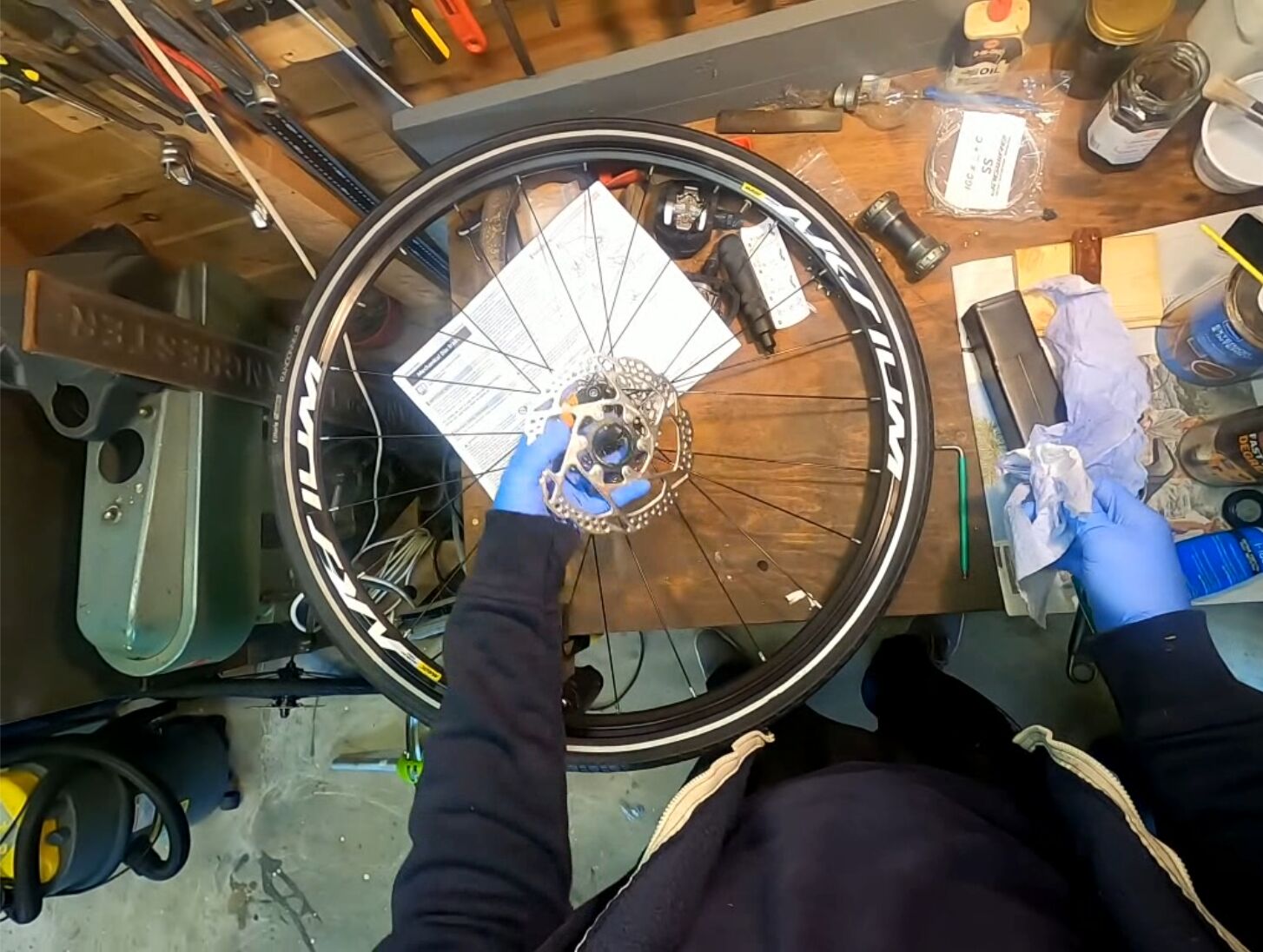}
        \includegraphics[width=0.19\linewidth]{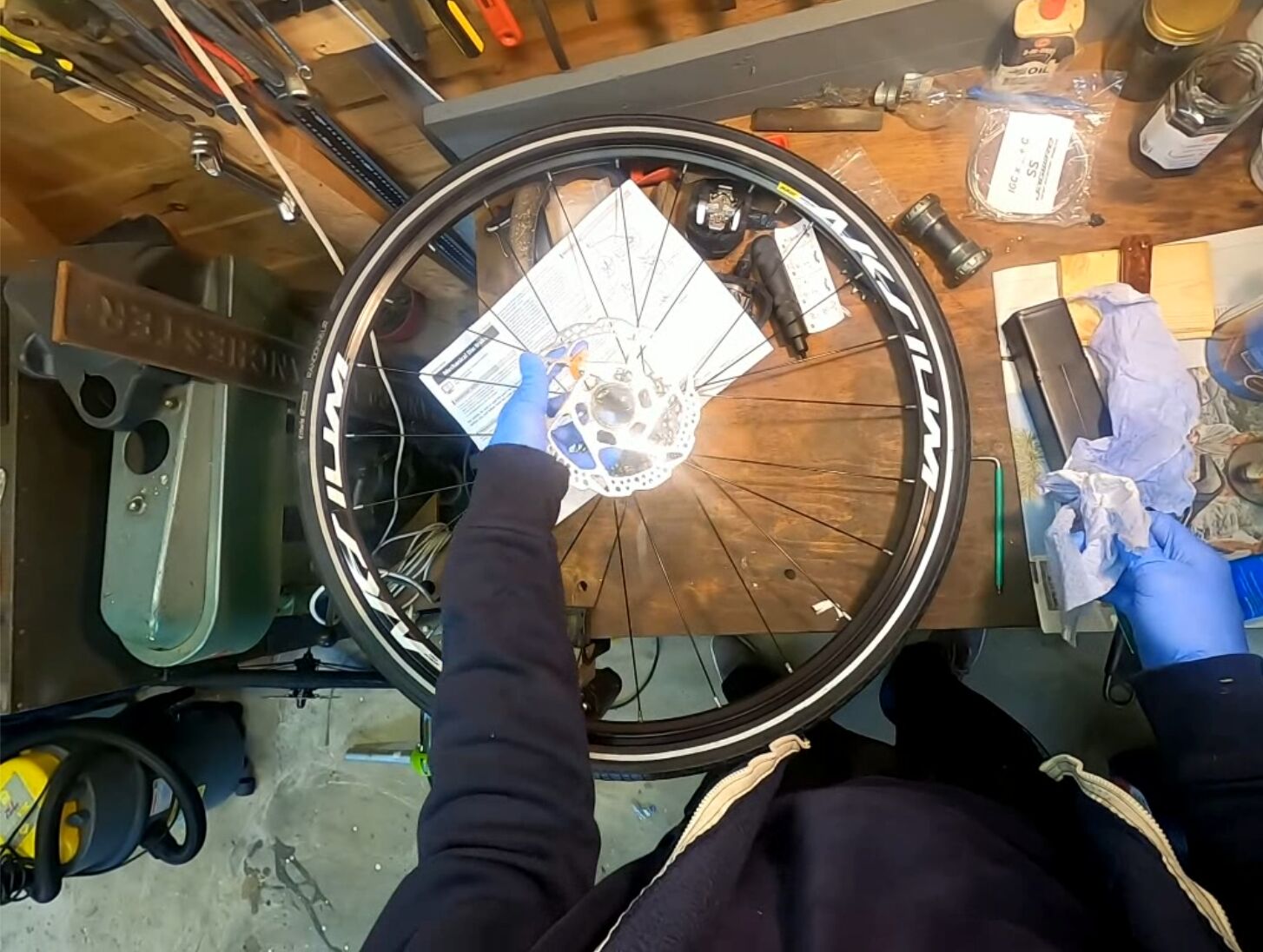}
        \includegraphics[width=0.19\linewidth]{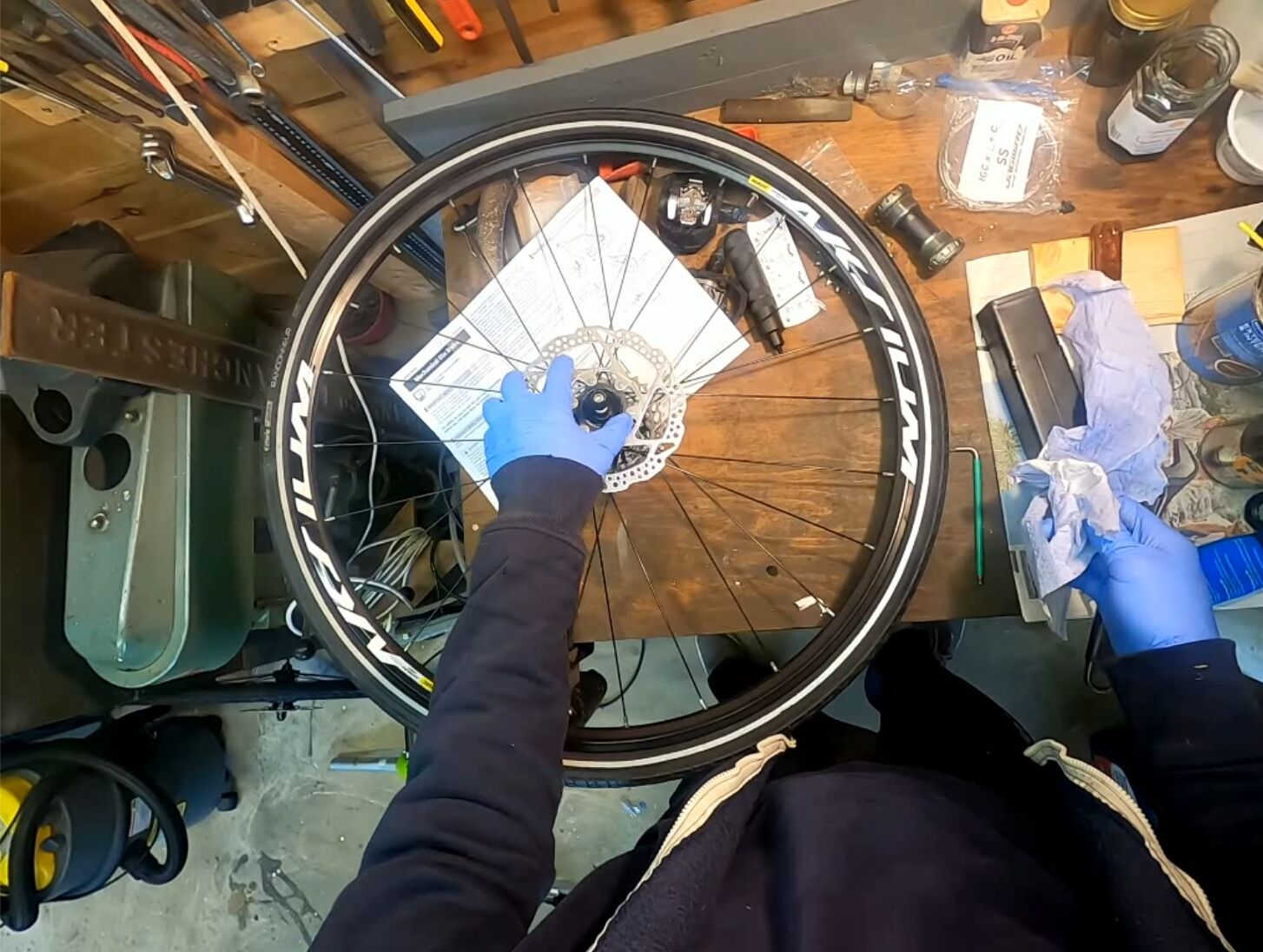} \\

    %\begin{minipage}{0.04\linewidth}
    %    \centering
    %    \rotatebox{90}{\bf{V-JEPA 2}}
    %\end{minipage}%
    %\begin{minipage}{0.95\linewidth}
    %    \includegraphics[width=0.19\linewidth]{video_202/1__.jpg}
    %    %\includegraphics[width=0.19\linewidth]{video_202/2__.jpg}
    %    \includegraphics[width=0.19\linewidth]{video_202/3__.jpg}
    %    %\includegraphics[width=0.19\linewidth]{video_202/4__.jpg}
    %    \includegraphics[width=0.19\linewidth]{video_202/5__.jpg}
    %    \includegraphics[width=0.19\linewidth]{video_202/6__.jpg}
    %    \includegraphics[width=0.19\linewidth]{video_202/8__.jpg}
    %\end{minipage} \\

        \includegraphics[width=0.19\linewidth]{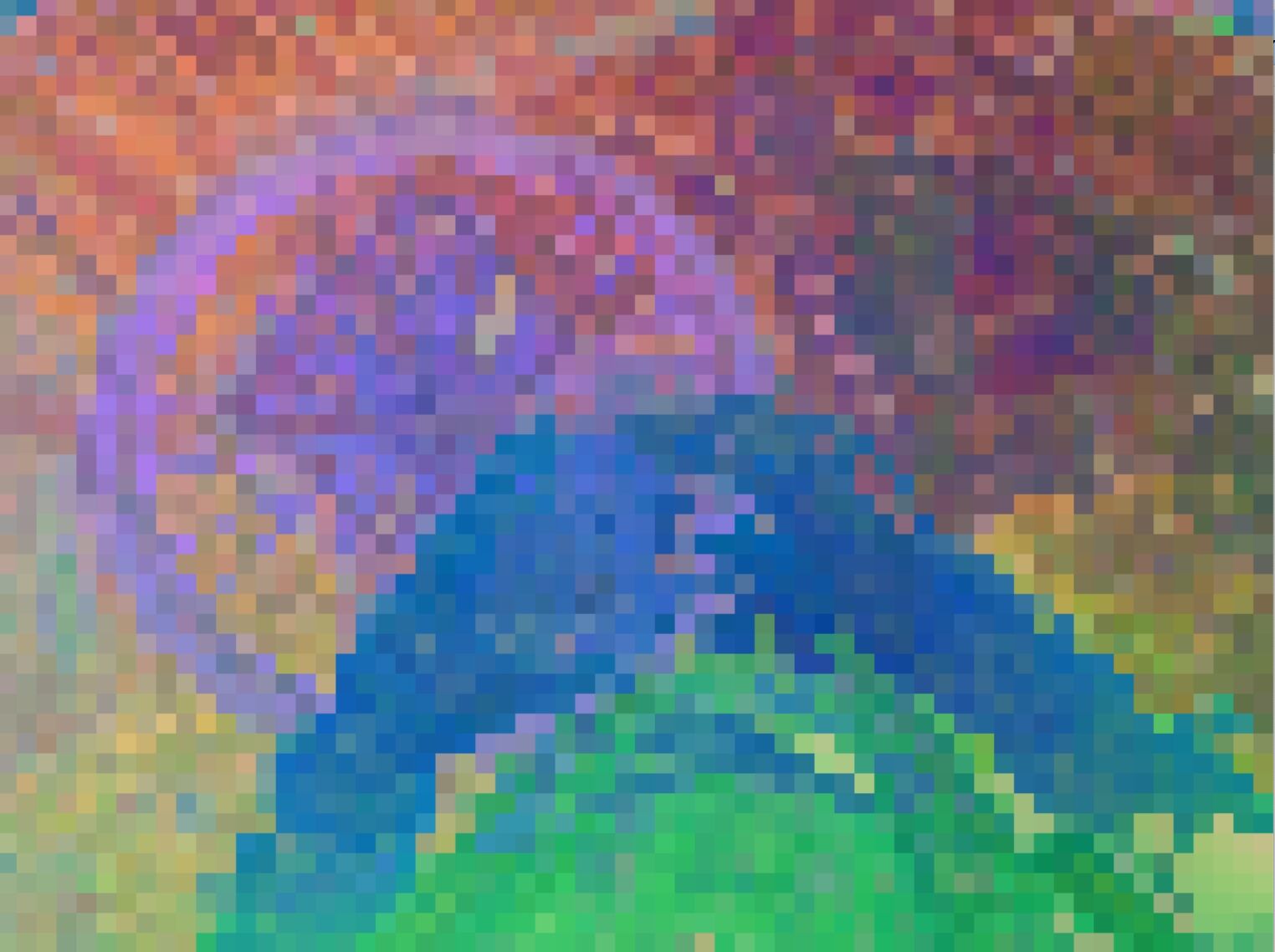}
        \includegraphics[width=0.19\linewidth]{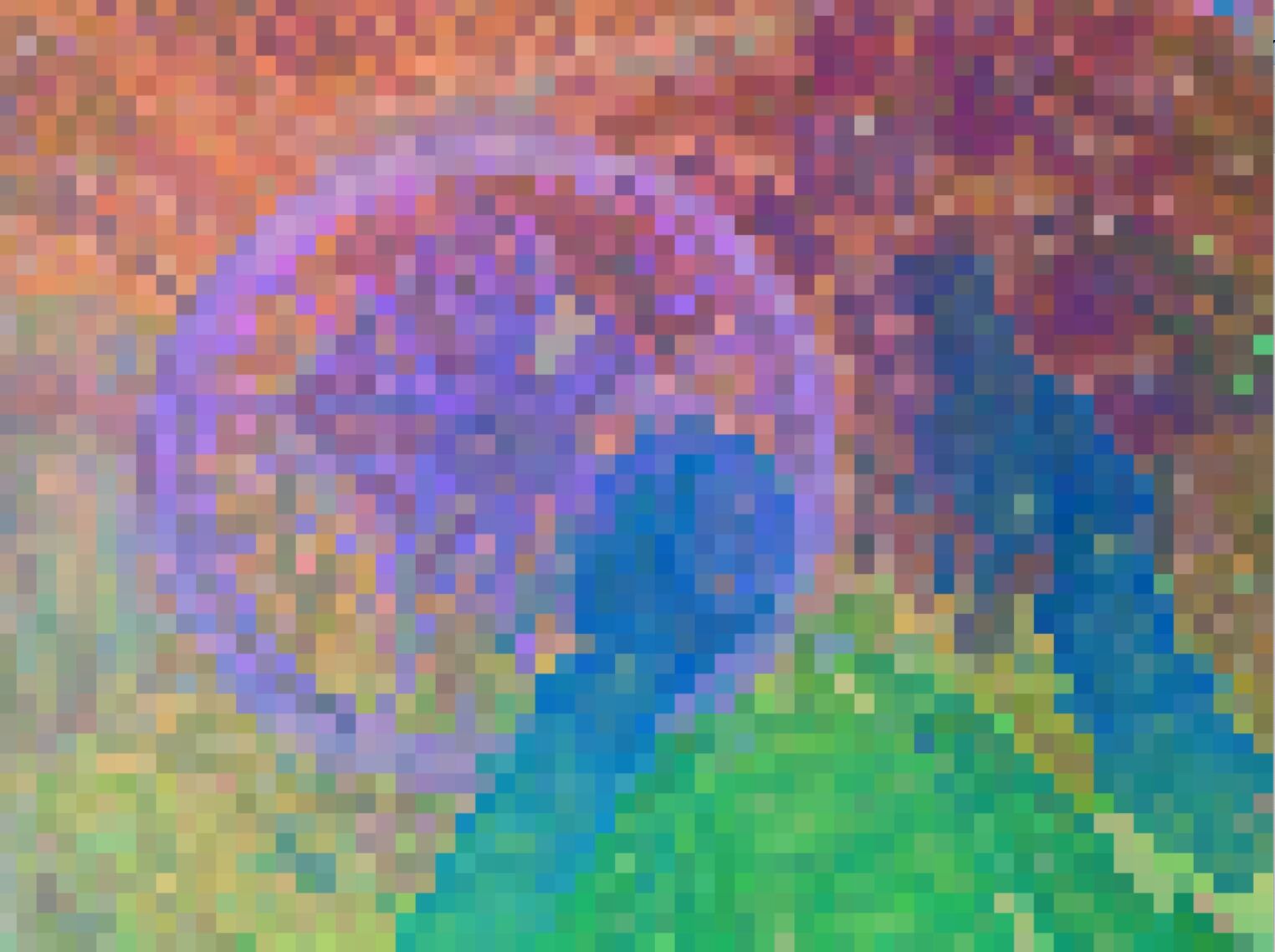}
        \includegraphics[width=0.19\linewidth]{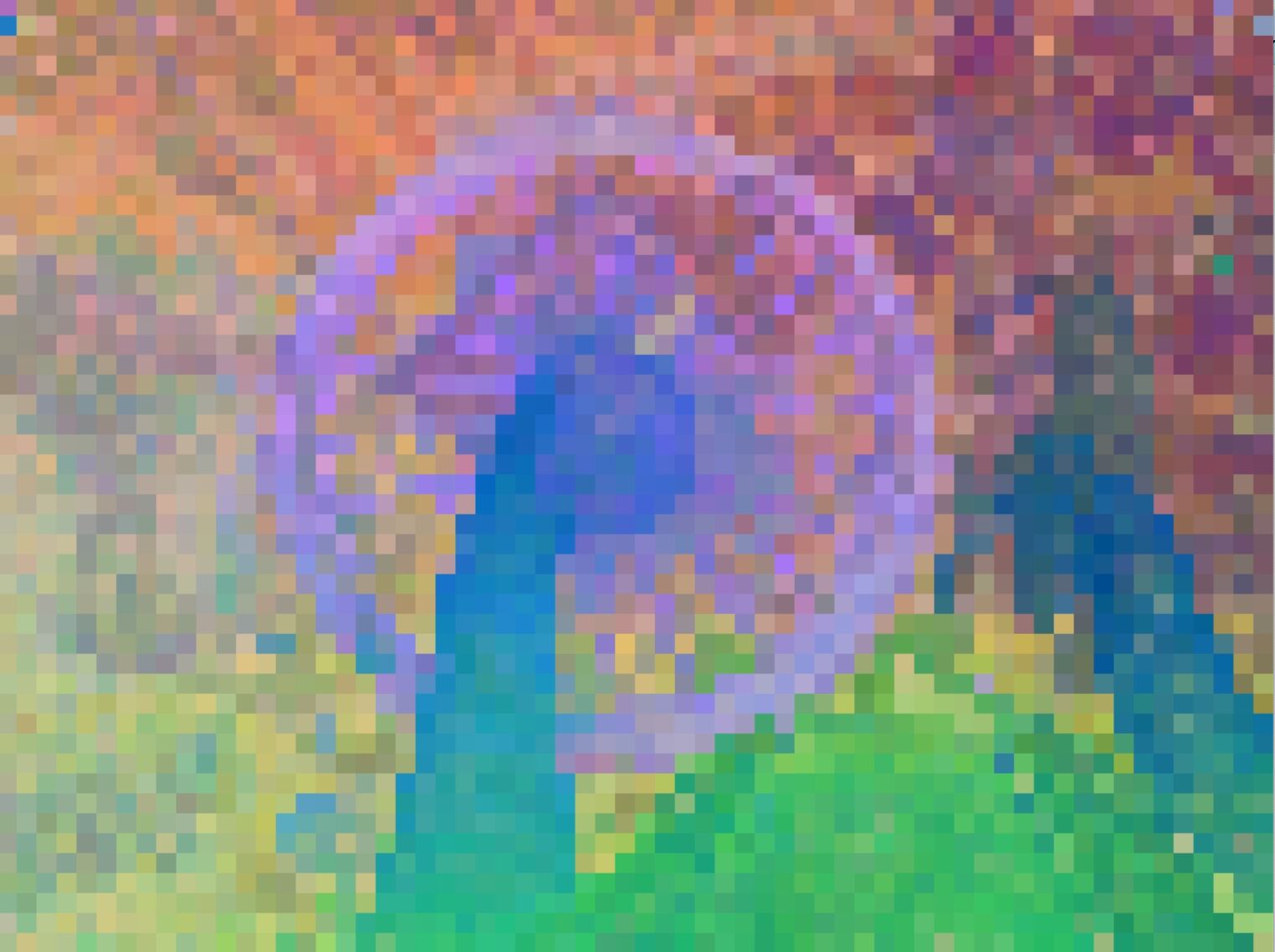}
        \includegraphics[width=0.19\linewidth]{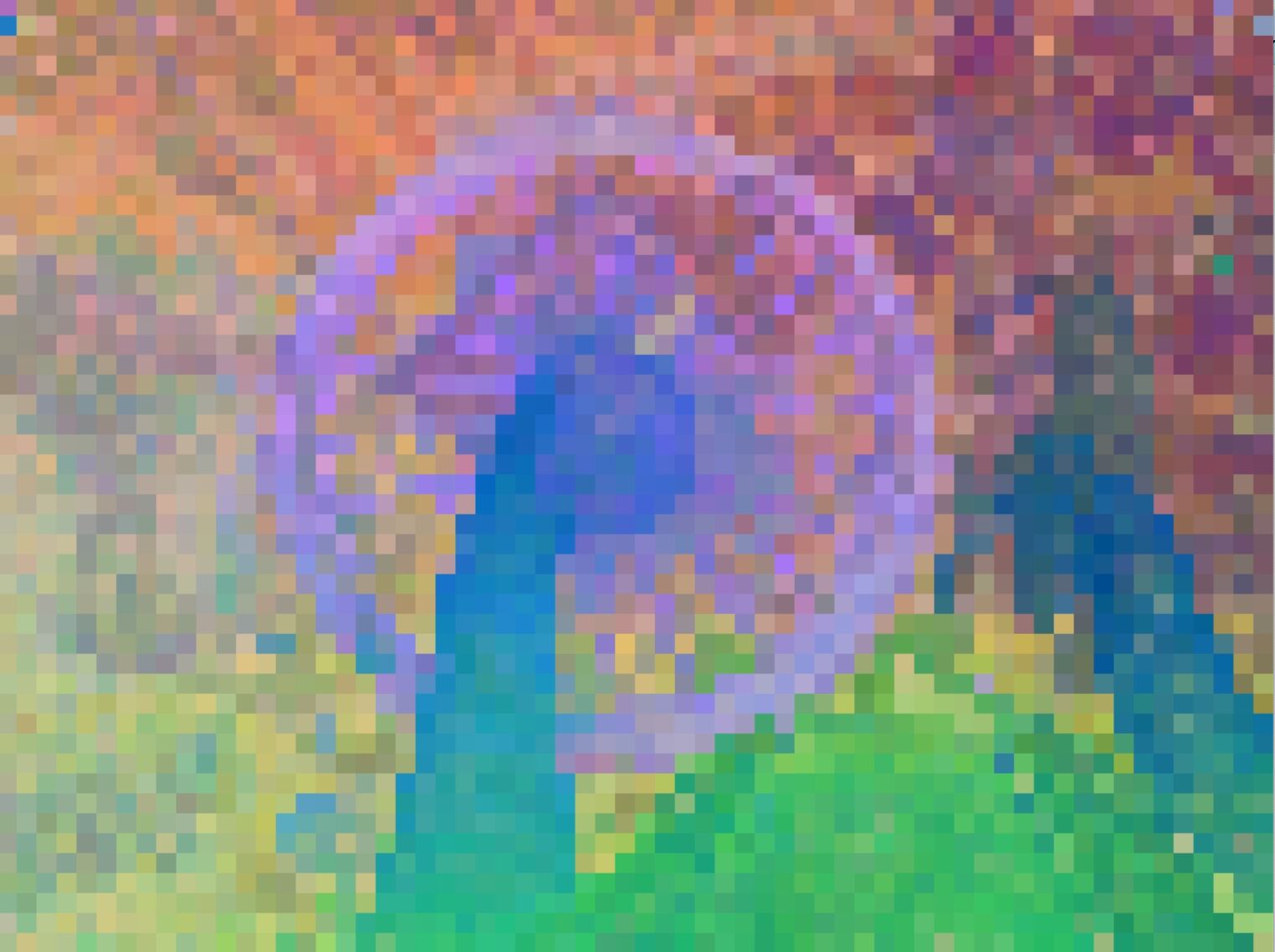}
        \includegraphics[width=0.19\linewidth]{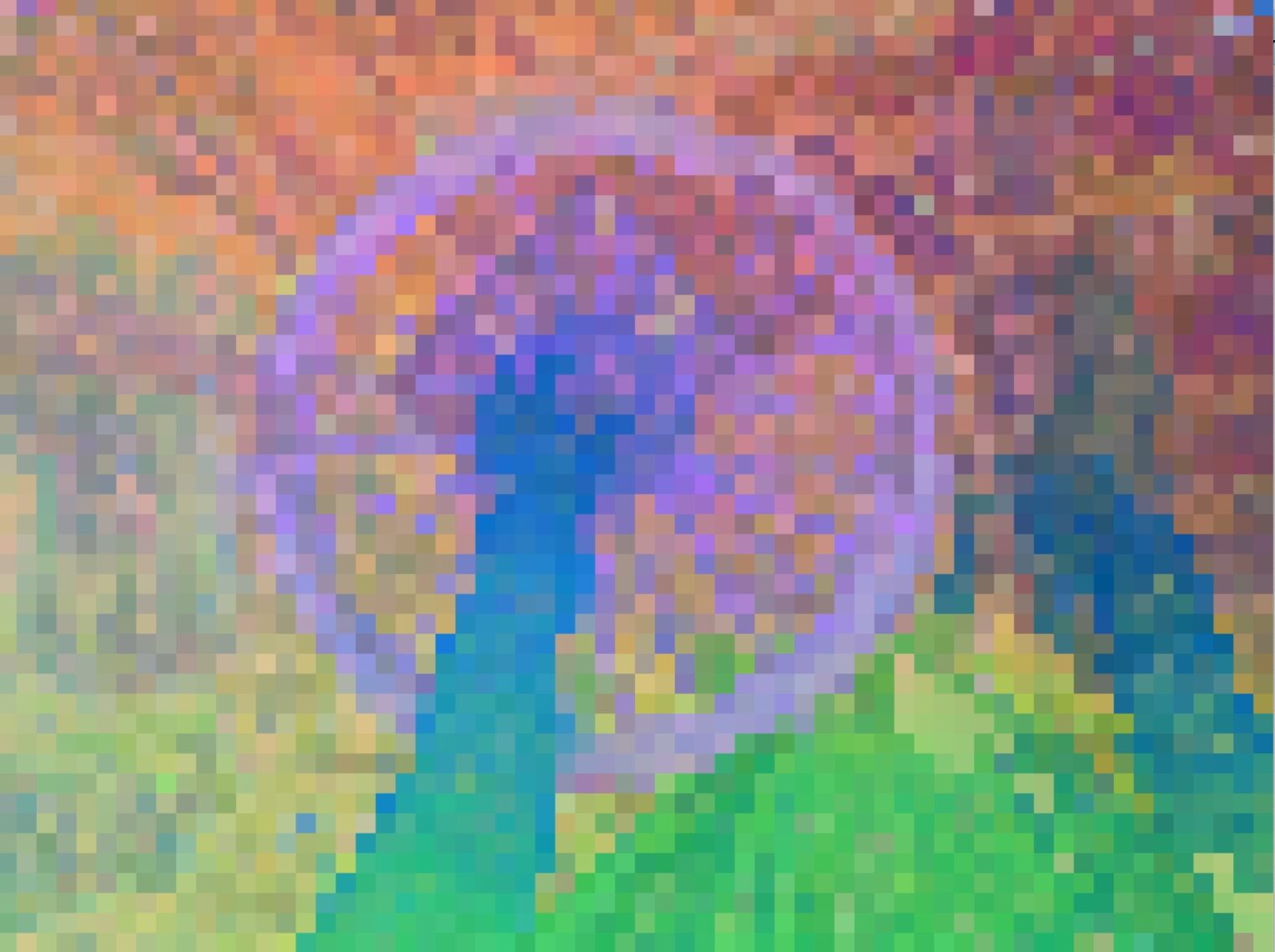} \\

        \vspace{0.1cm}

        \includegraphics[width=0.19\linewidth]{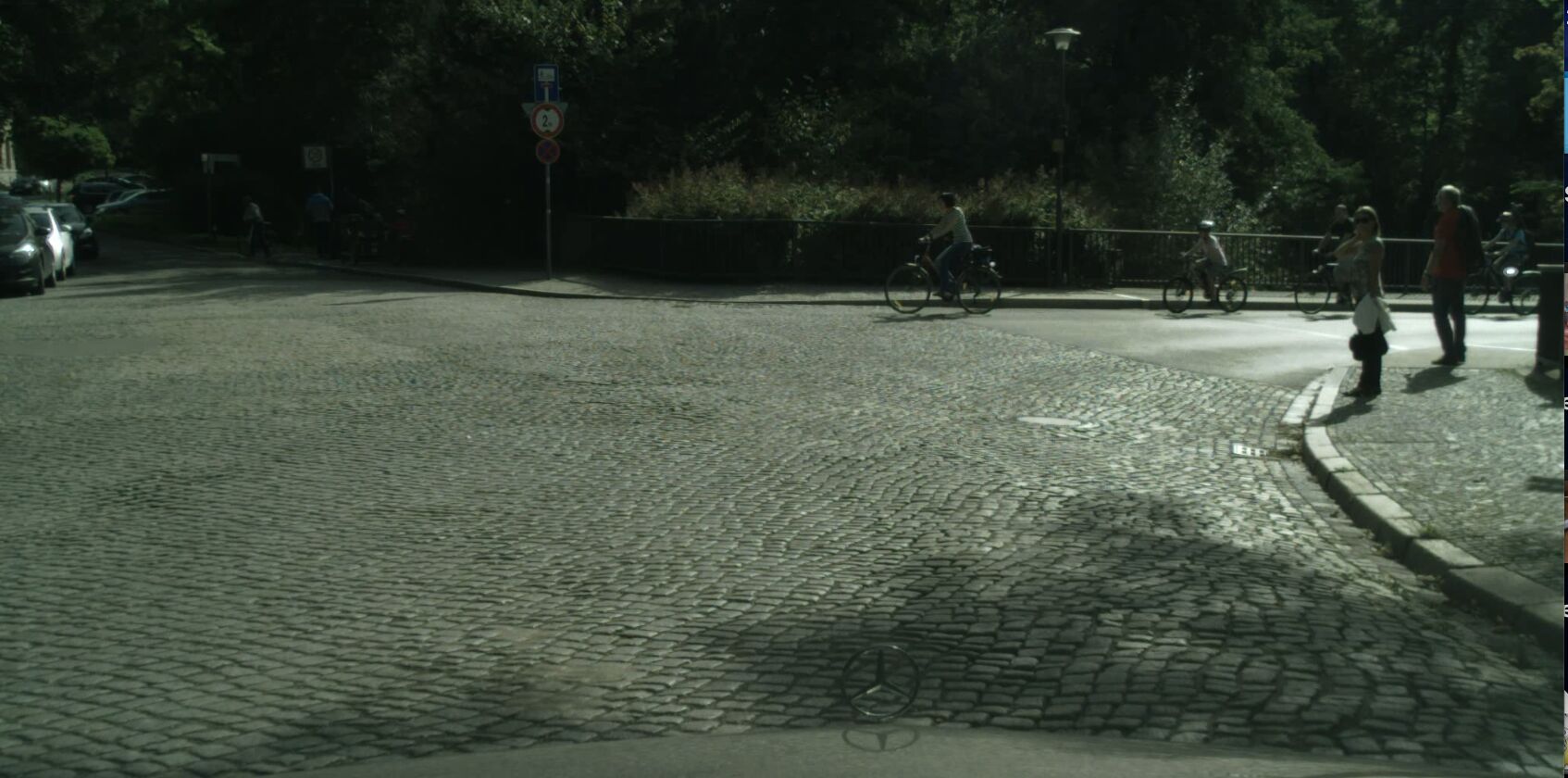}
        \includegraphics[width=0.19\linewidth]{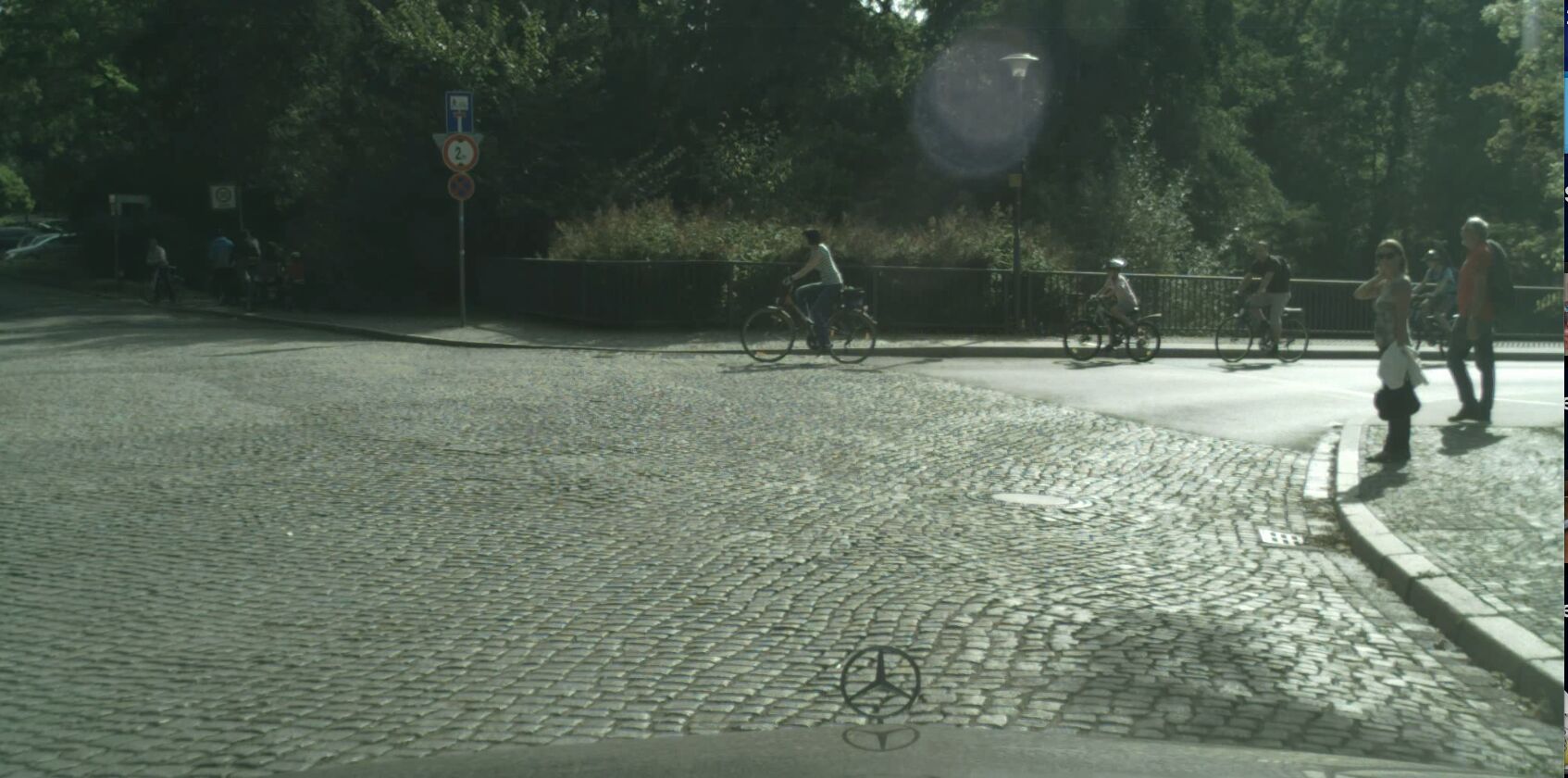}
        \includegraphics[width=0.19\linewidth]{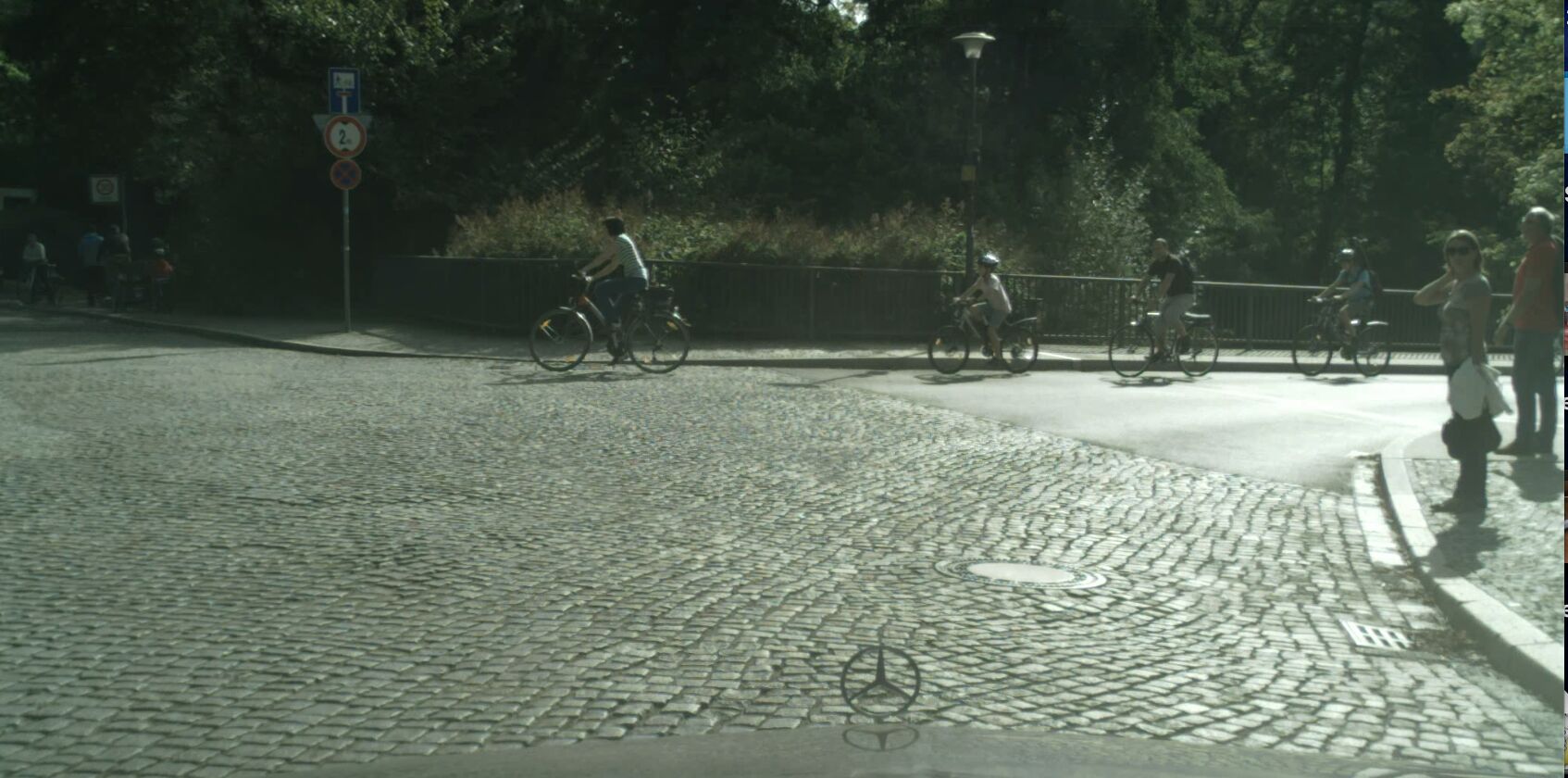}
        \includegraphics[width=0.19\linewidth]{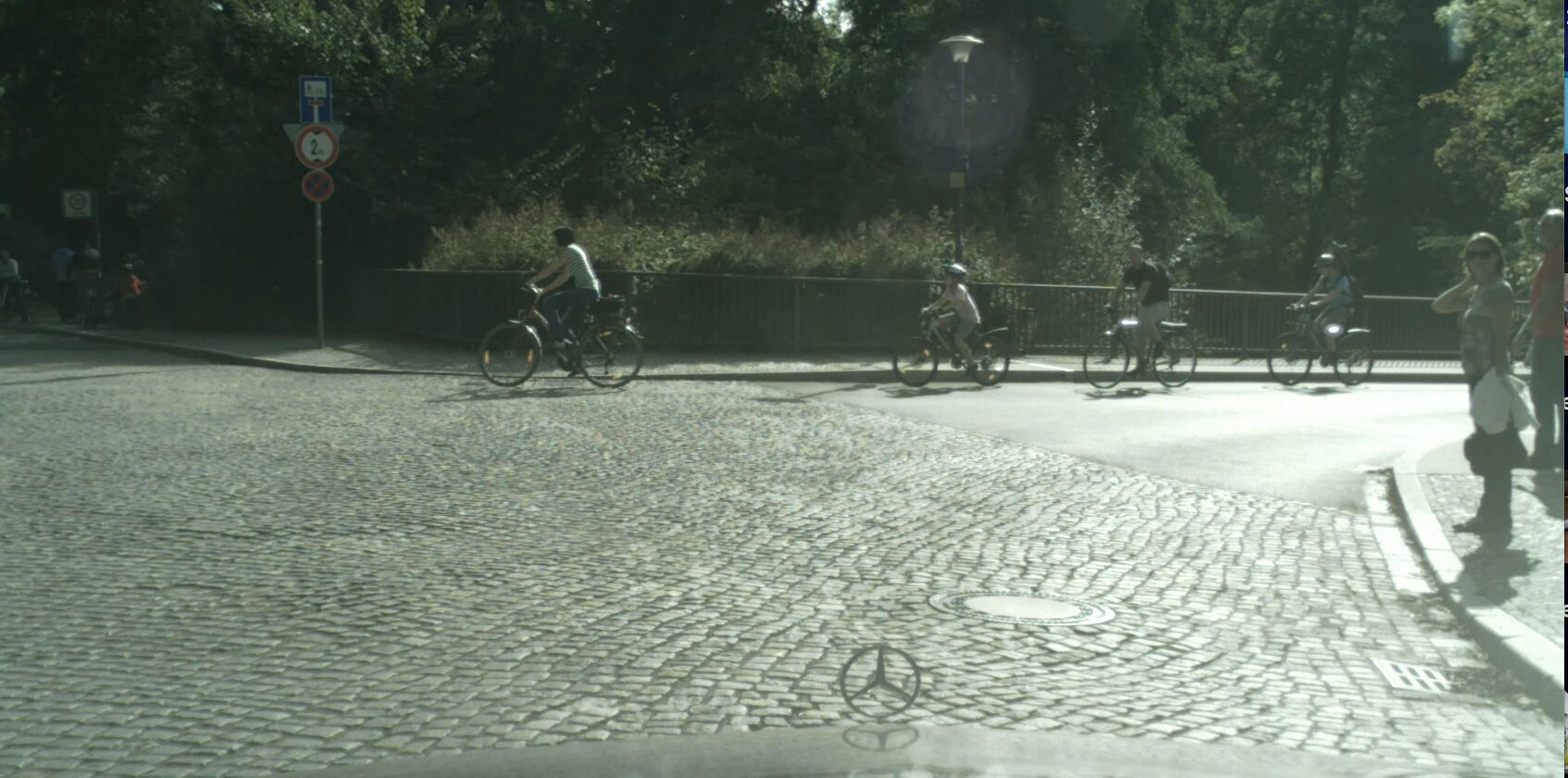}
        \includegraphics[width=0.19\linewidth]{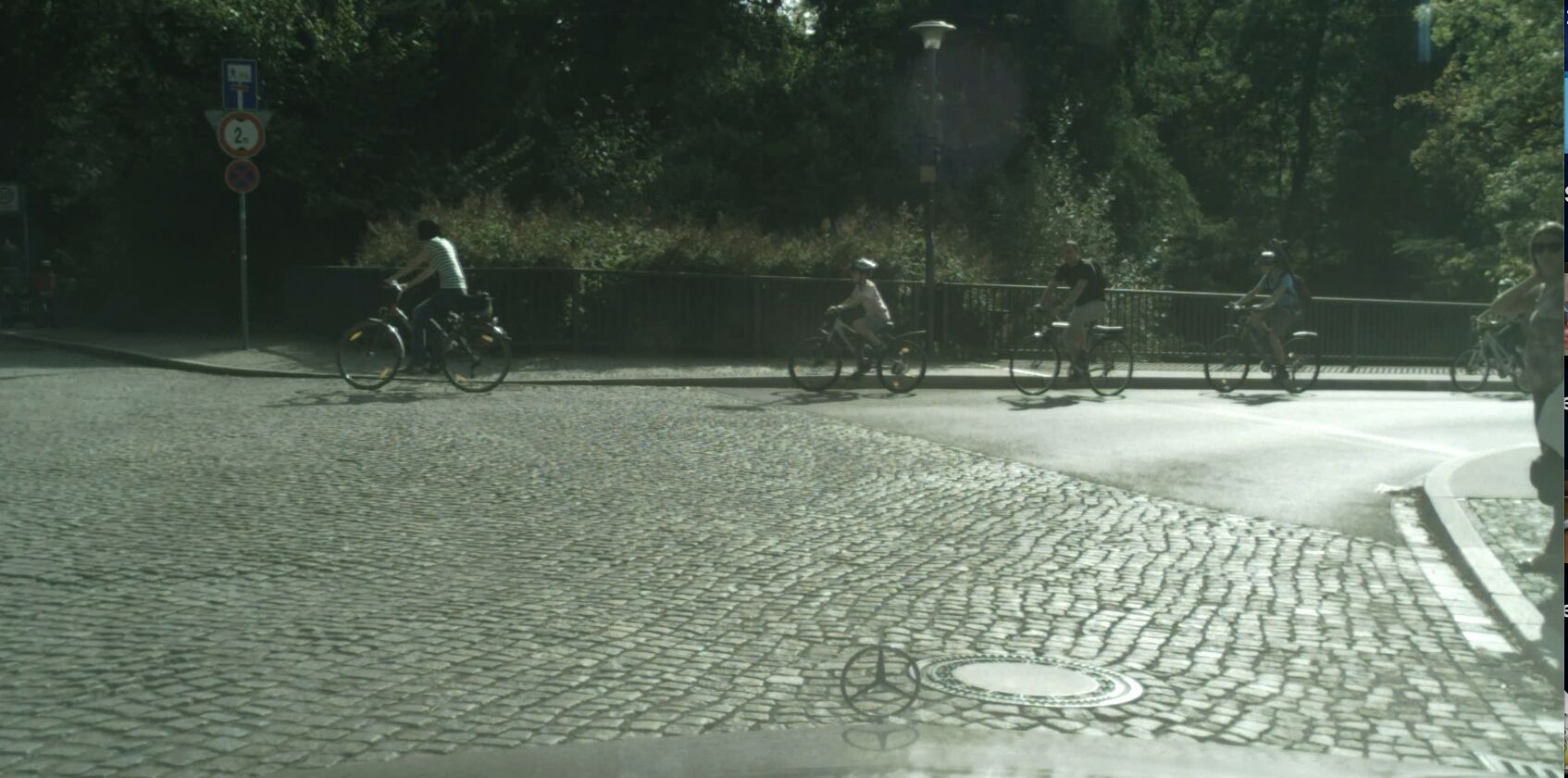} \\

    %\begin{minipage}{0.04\linewidth}
    %    \centering
    %    \rotatebox{90}{\bf{V-JEPA 2}}
    %\end{minipage}%
    %\begin{minipage}{0.95\linewidth}
    %    \includegraphics[width=0.19\linewidth]{weimar_13/vjepa21.jpg}
    %    %\includegraphics[width=0.19\linewidth]{weimar_13/vjepa22.jpg}
    %    \includegraphics[width=0.19\linewidth]{weimar_13/vjepa23.jpg}
    %    %\includegraphics[width=0.19\linewidth]{weimar_13/vjepa24.jpg}
    %    \includegraphics[width=0.19\linewidth]{weimar_13/vjepa25.jpg}
    %    \includegraphics[width=0.19\linewidth]{weimar_13/vjepa26.jpg}
    %    \includegraphics[width=0.19\linewidth]{weimar_13/vjepa27.jpg}
    %\end{minipage} \\

        \includegraphics[width=0.19\linewidth]{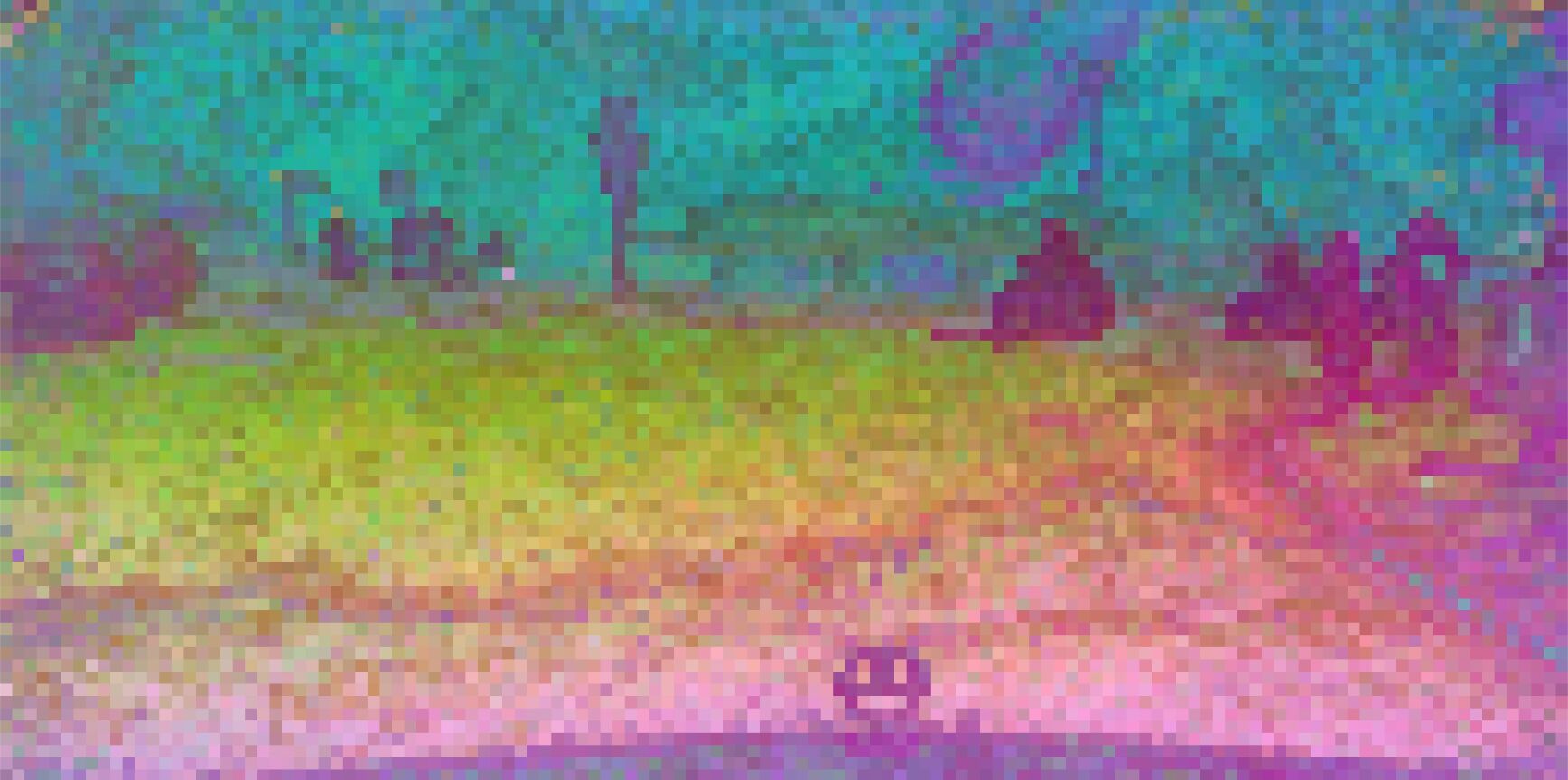}
        \includegraphics[width=0.19\linewidth]{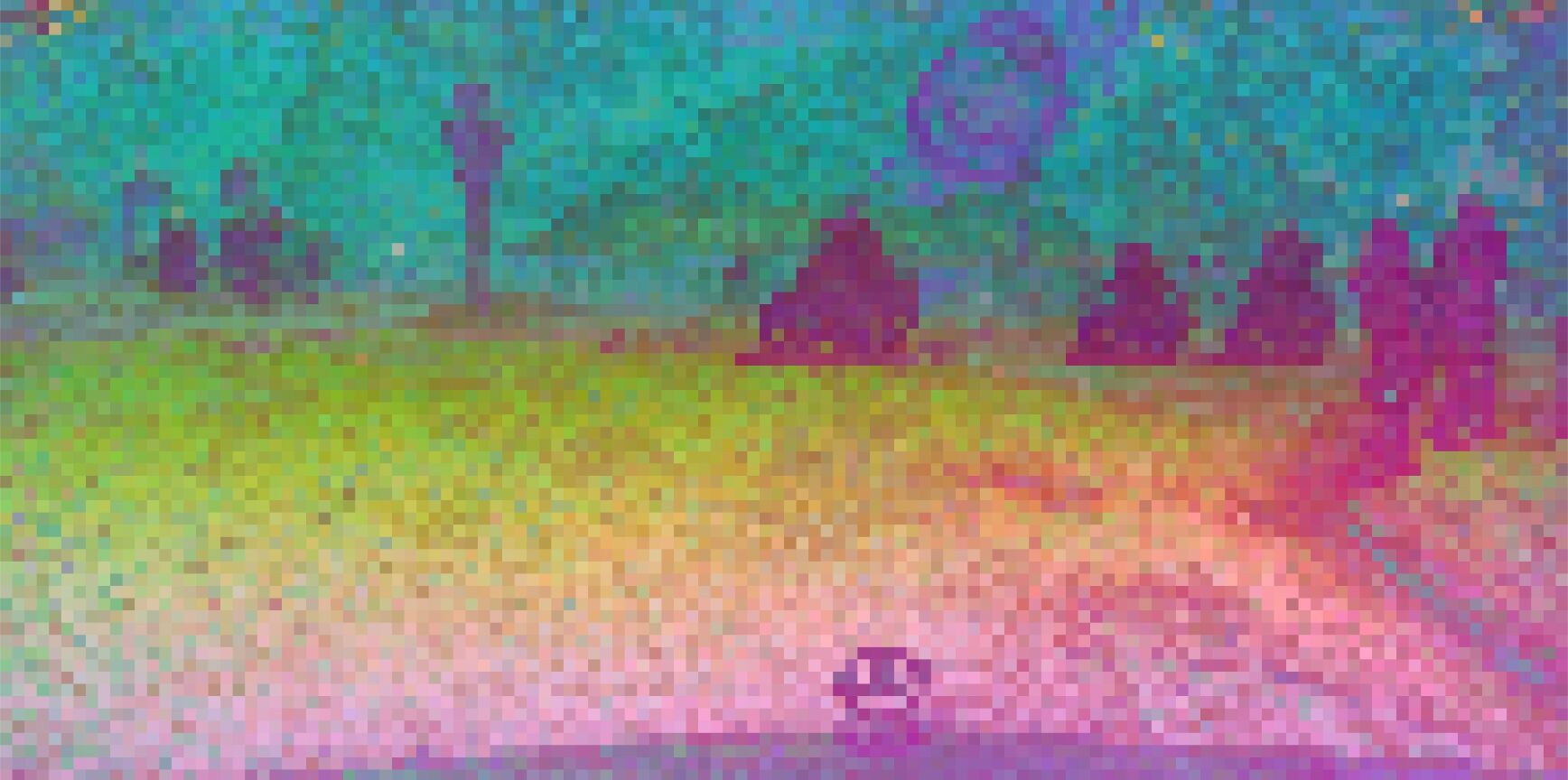}
        \includegraphics[width=0.19\linewidth]{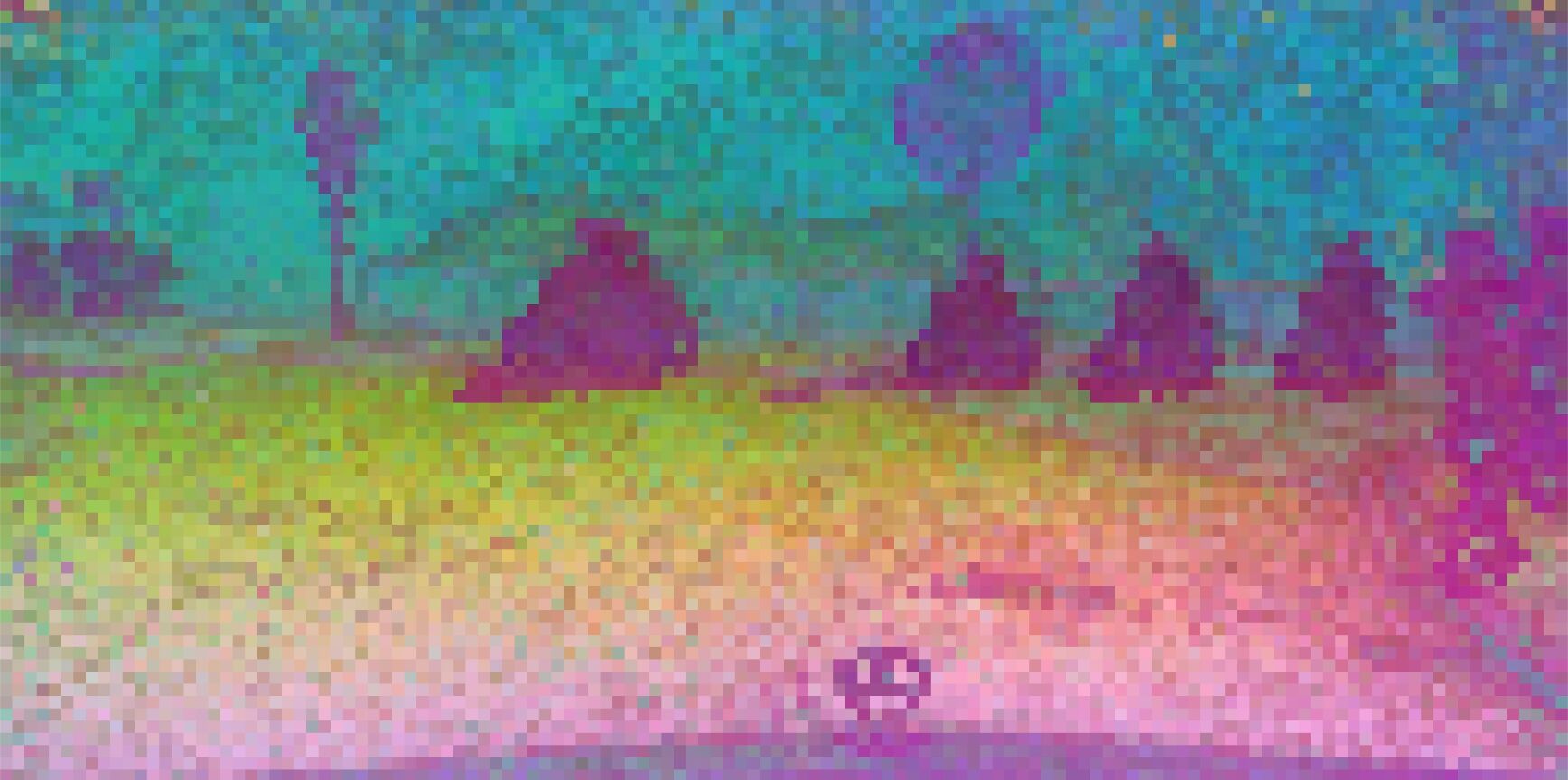}
        \includegraphics[width=0.19\linewidth]{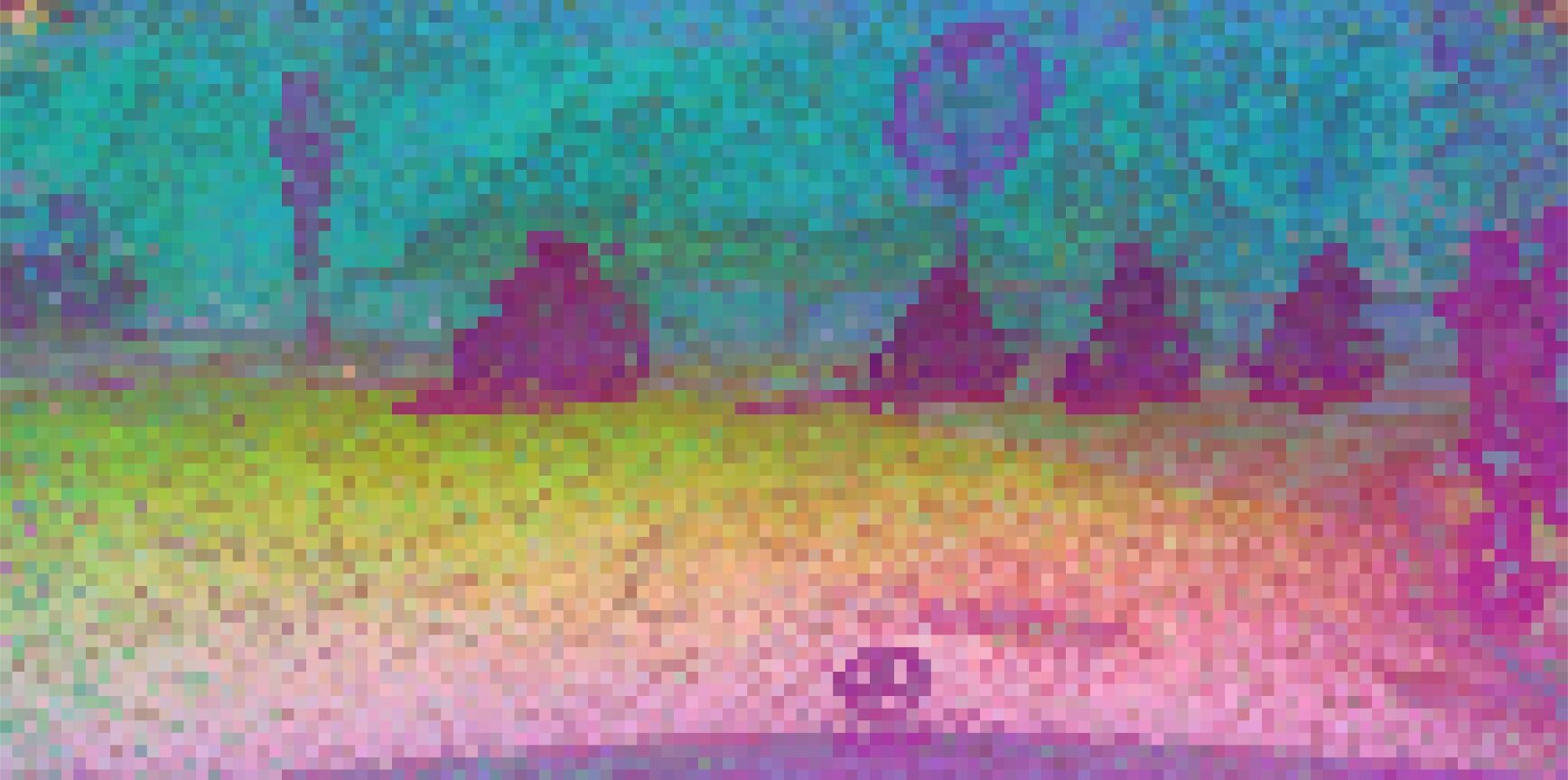}
        \includegraphics[width=0.19\linewidth]{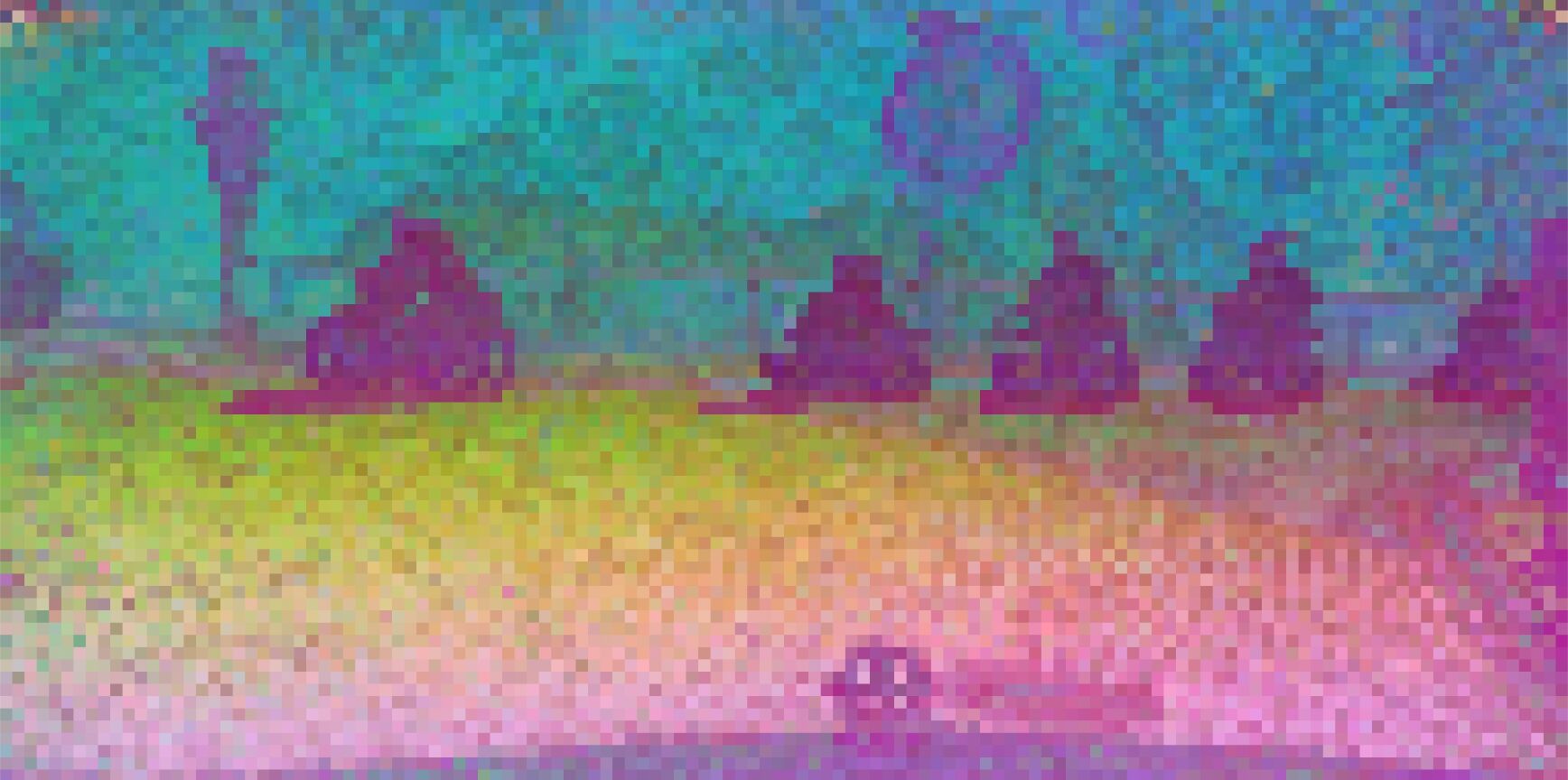} \\

        \vspace{0.1cm}

        \includegraphics[width=0.19\linewidth]{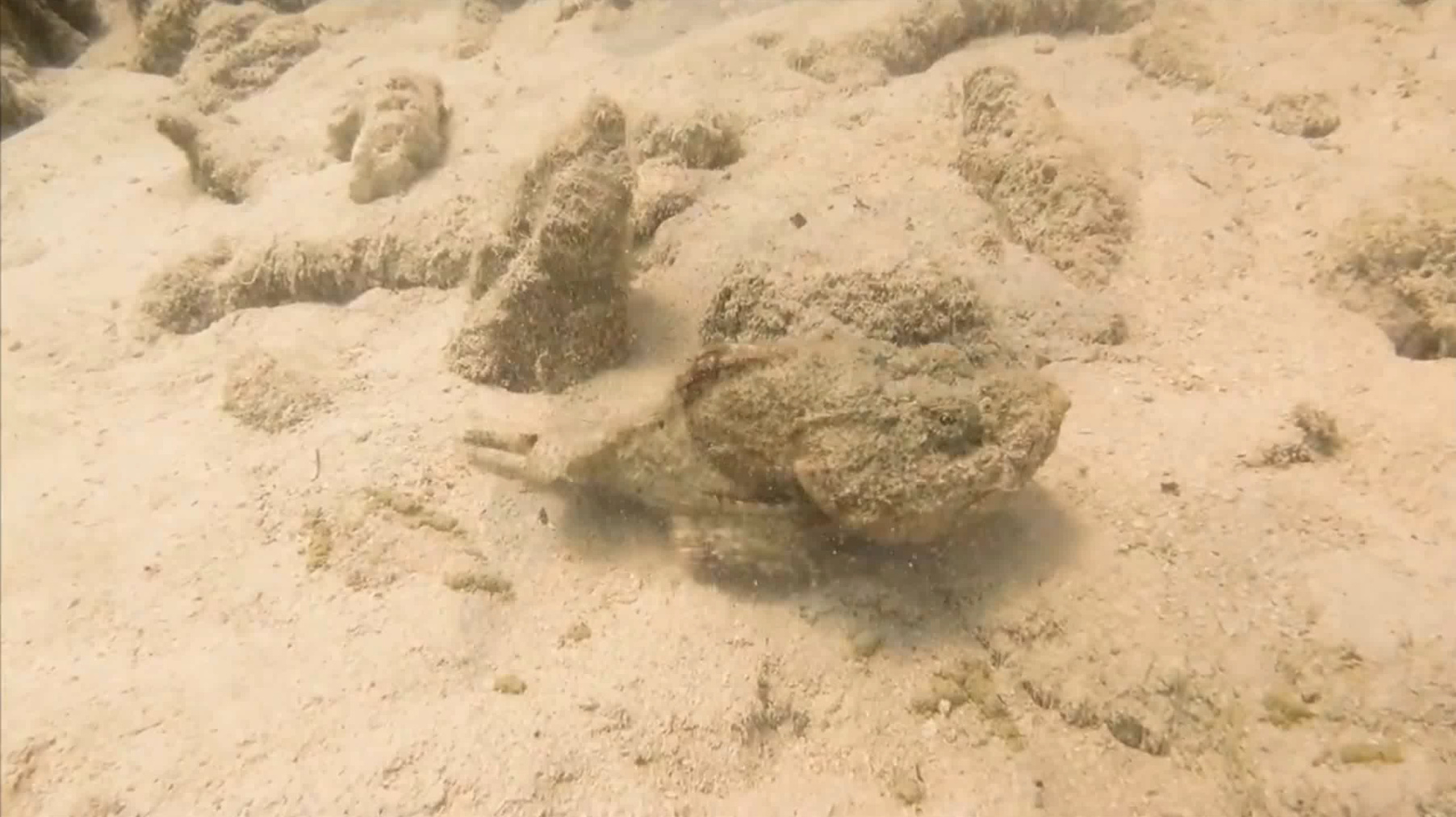}
        \includegraphics[width=0.19\linewidth]{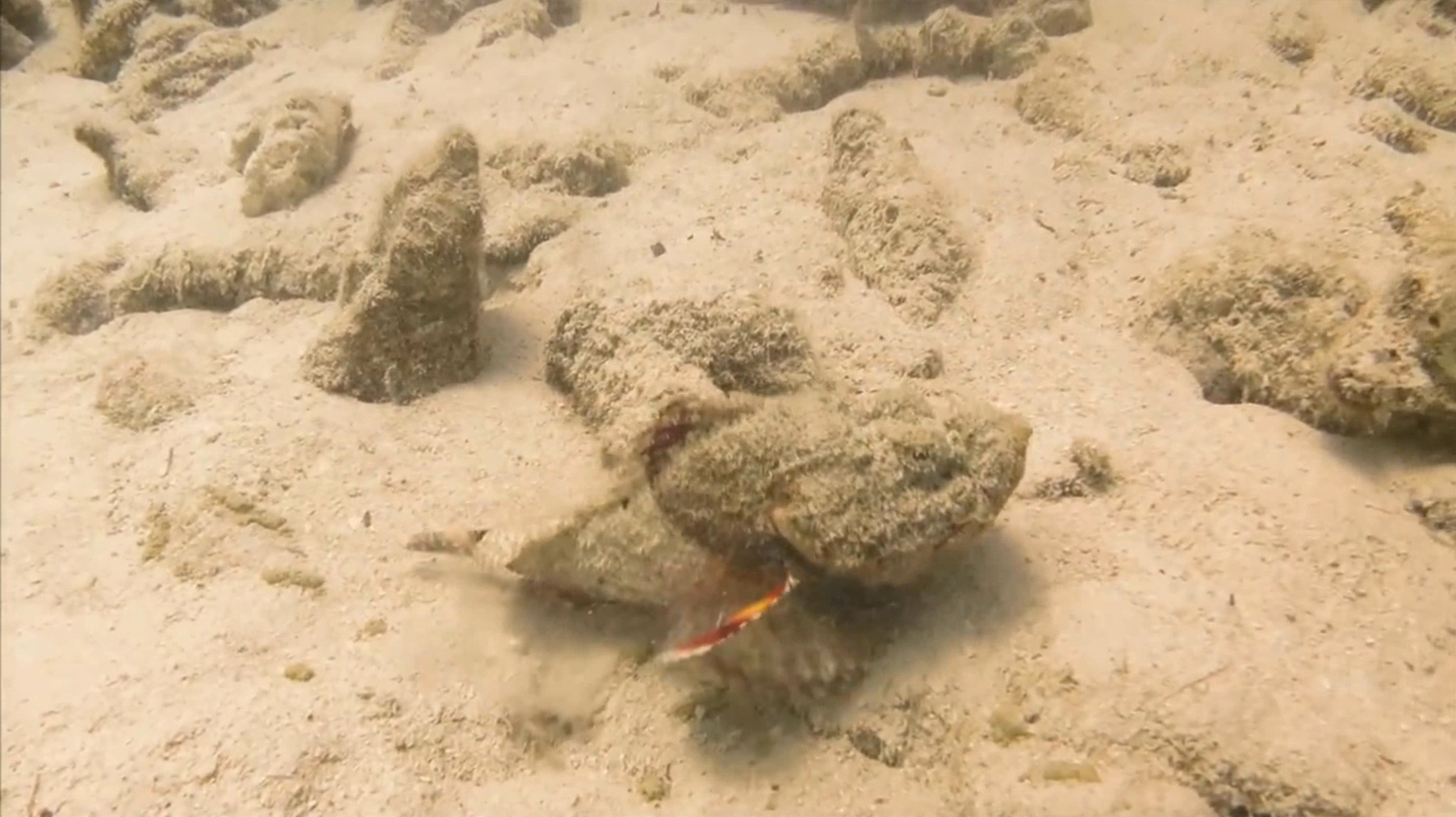}
        \includegraphics[width=0.19\linewidth]{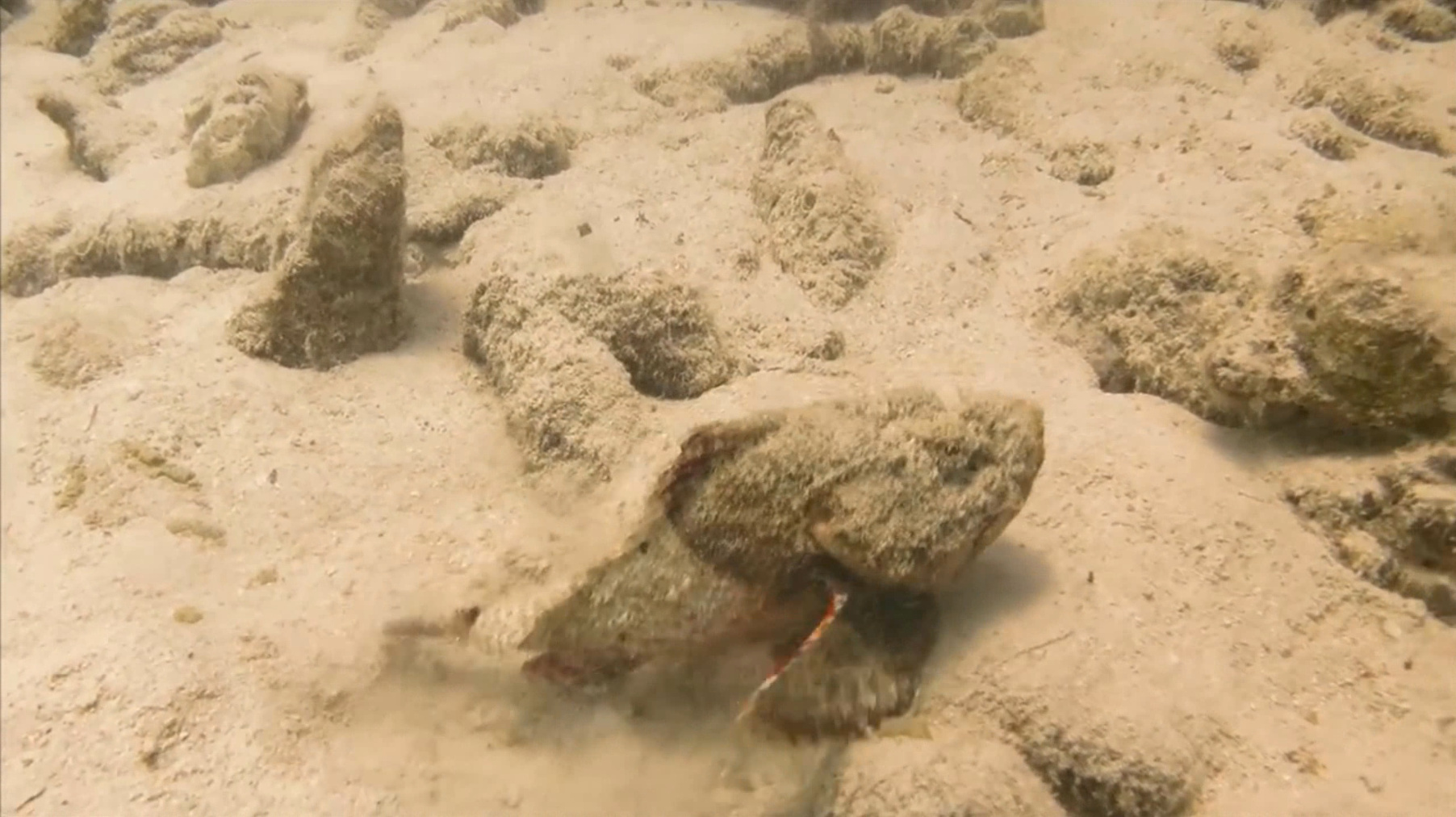}
        \includegraphics[width=0.19\linewidth]{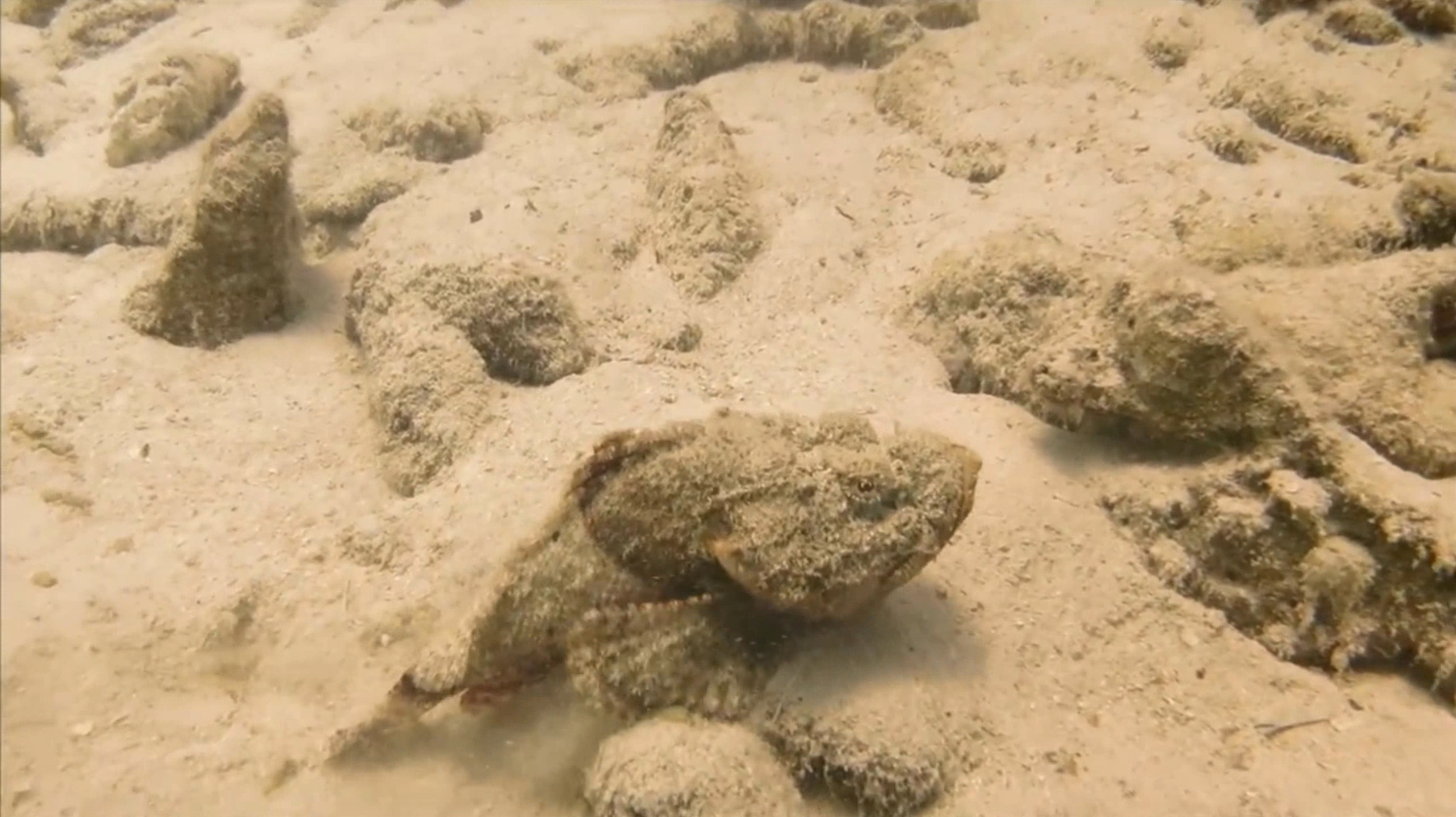}
        \includegraphics[width=0.19\linewidth]{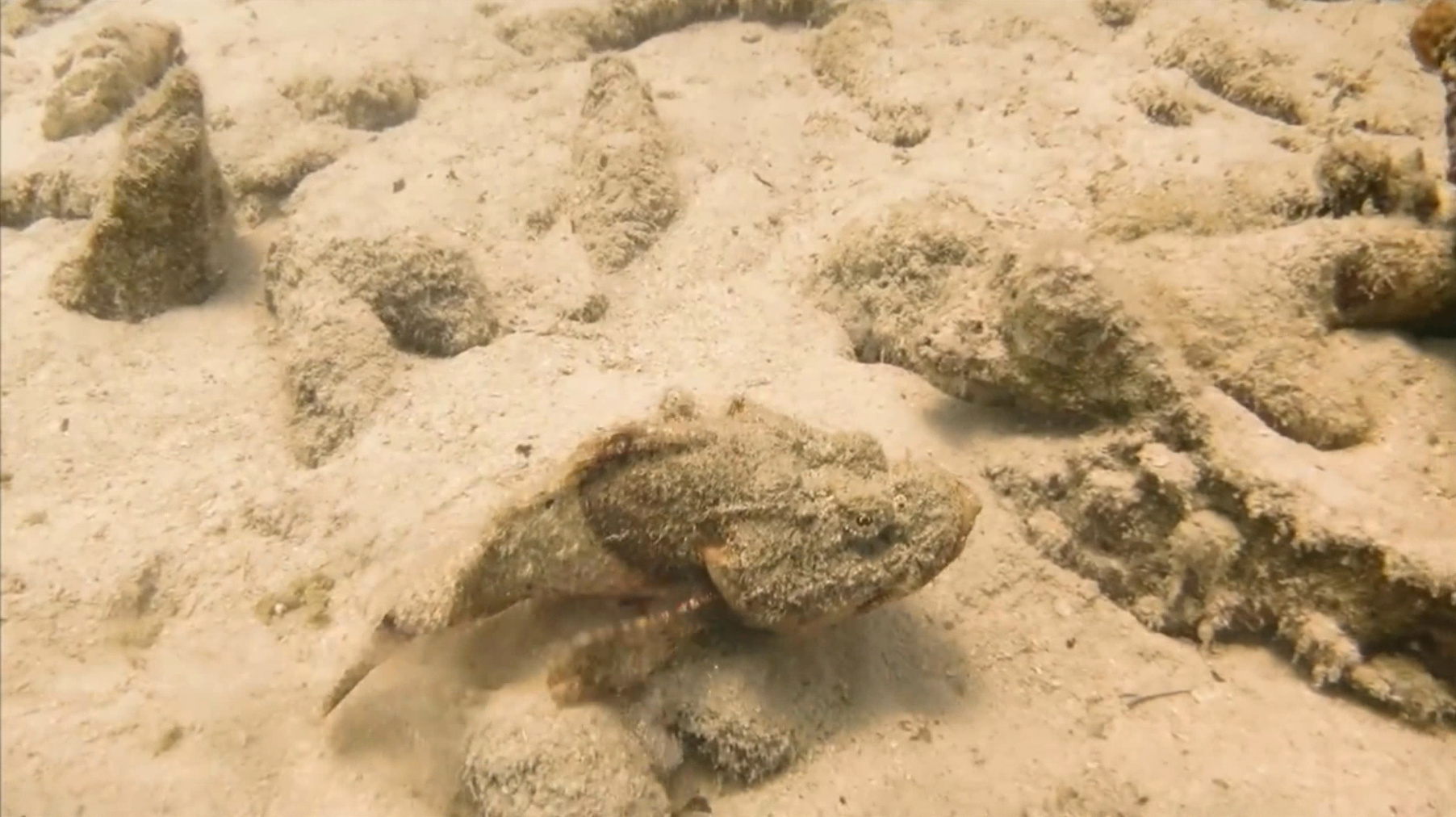} \\

         \includegraphics[width=0.19\linewidth]{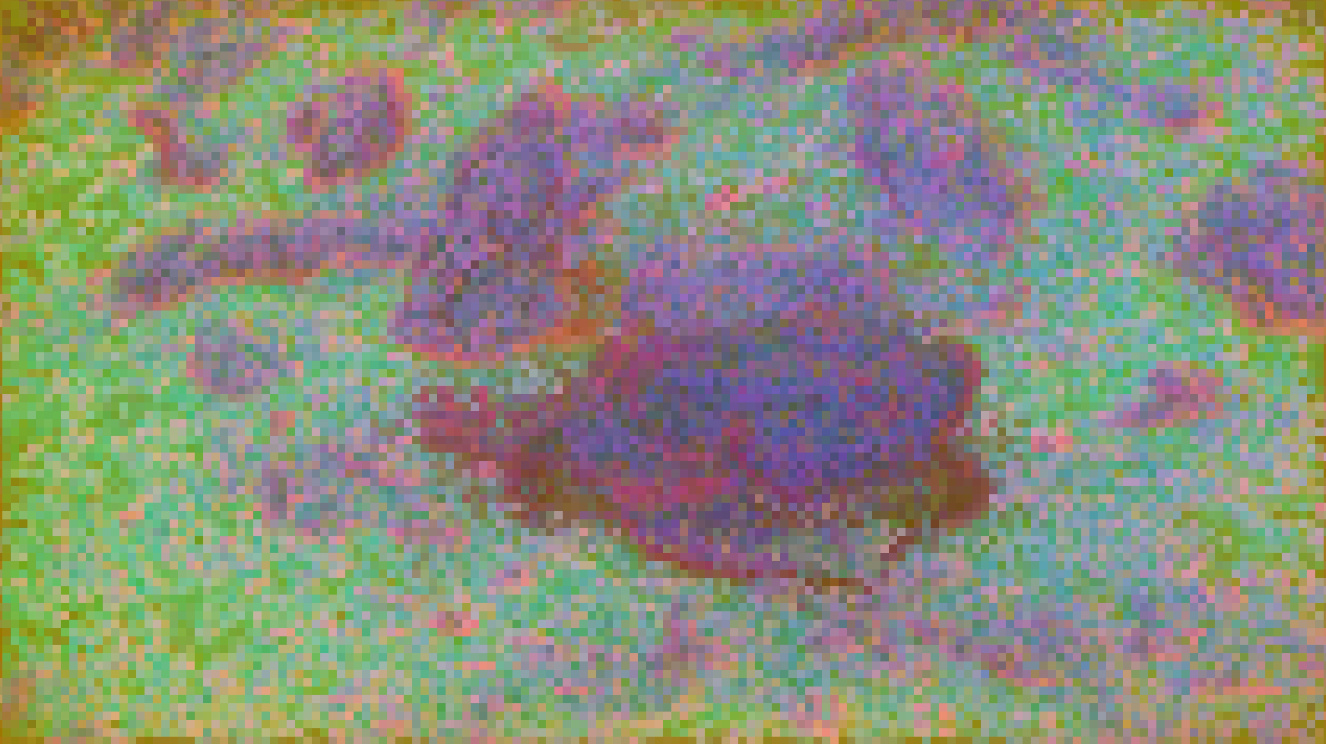}
        \includegraphics[width=0.19\linewidth]{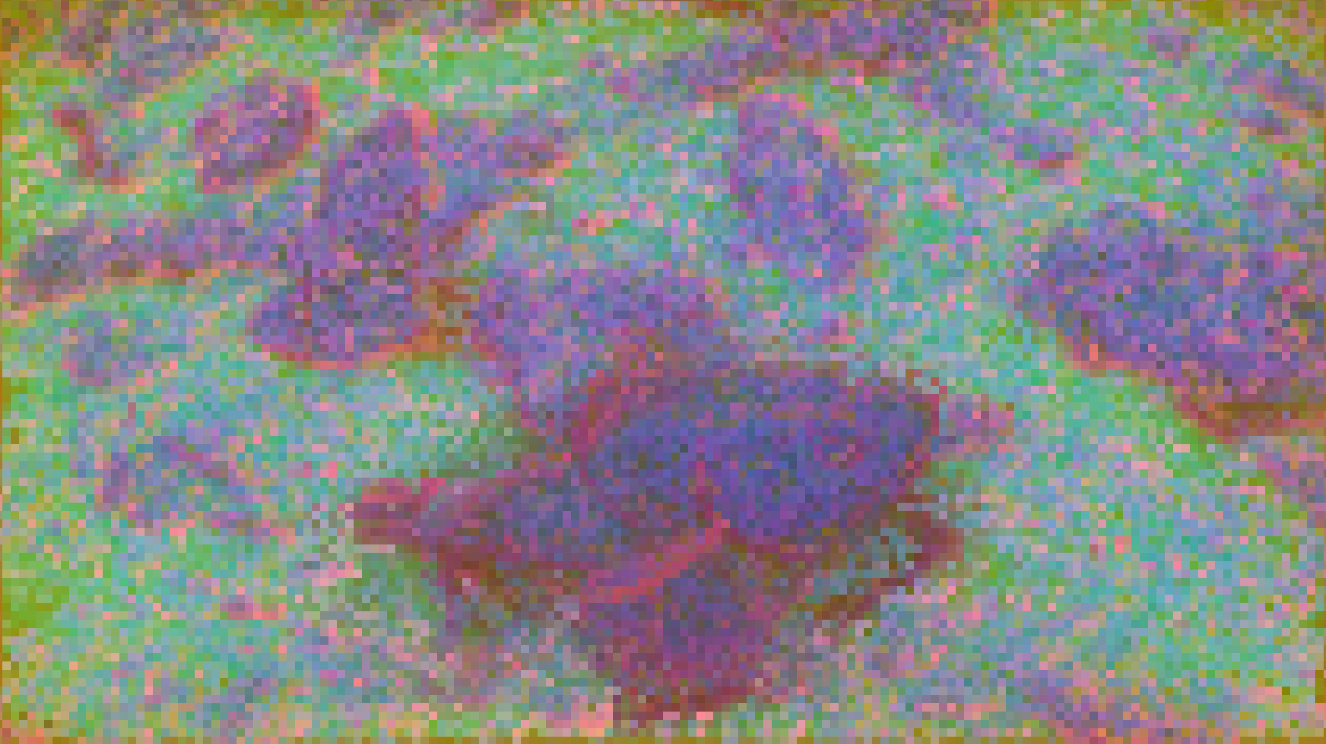}
        \includegraphics[width=0.19\linewidth]{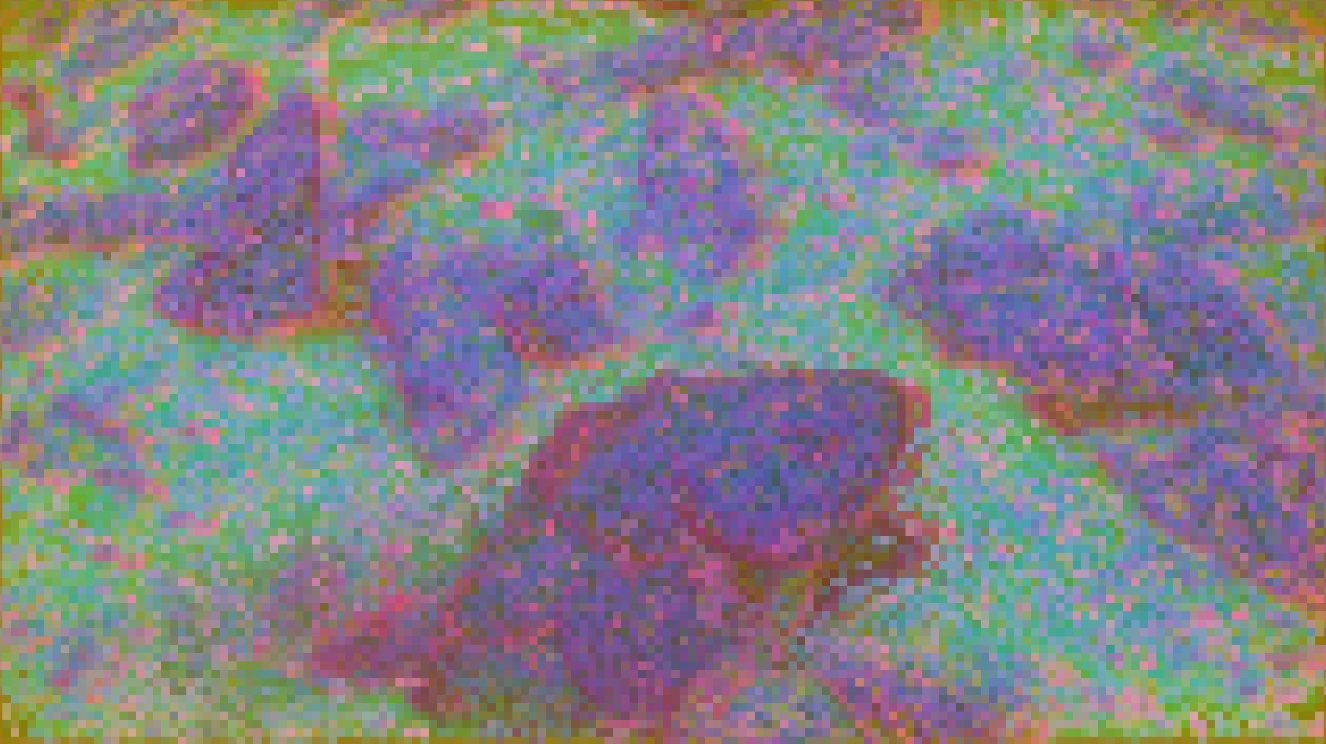}
        \includegraphics[width=0.19\linewidth]{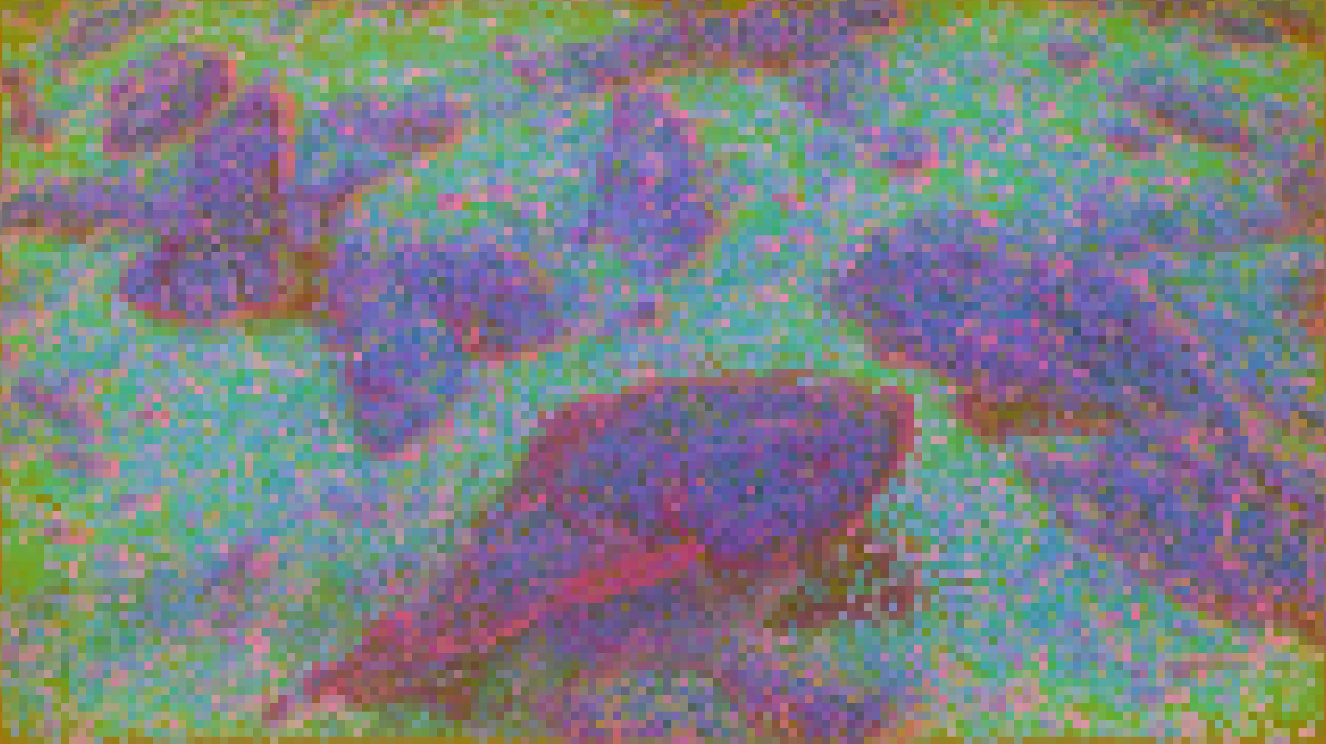}
        \includegraphics[width=0.19\linewidth]{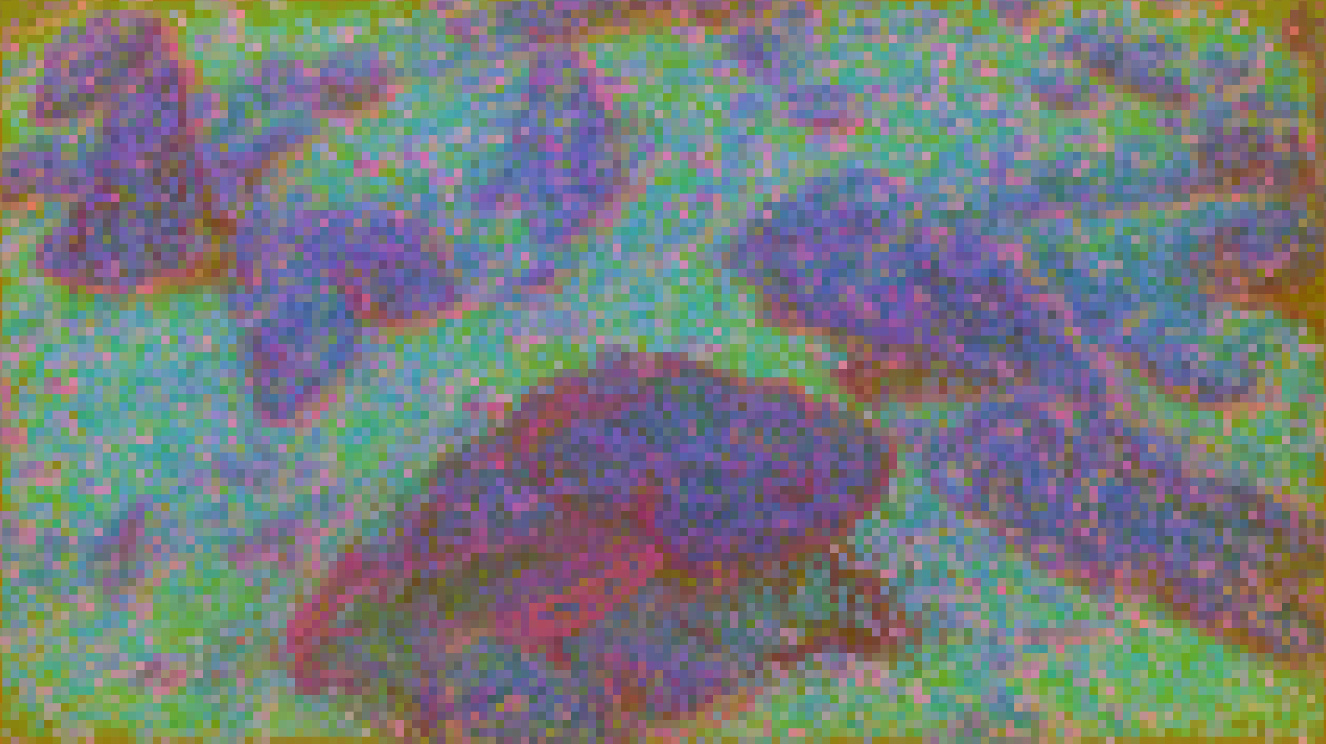}

    \vspace{0.1cm}

    \includegraphics[width=0.19\linewidth]{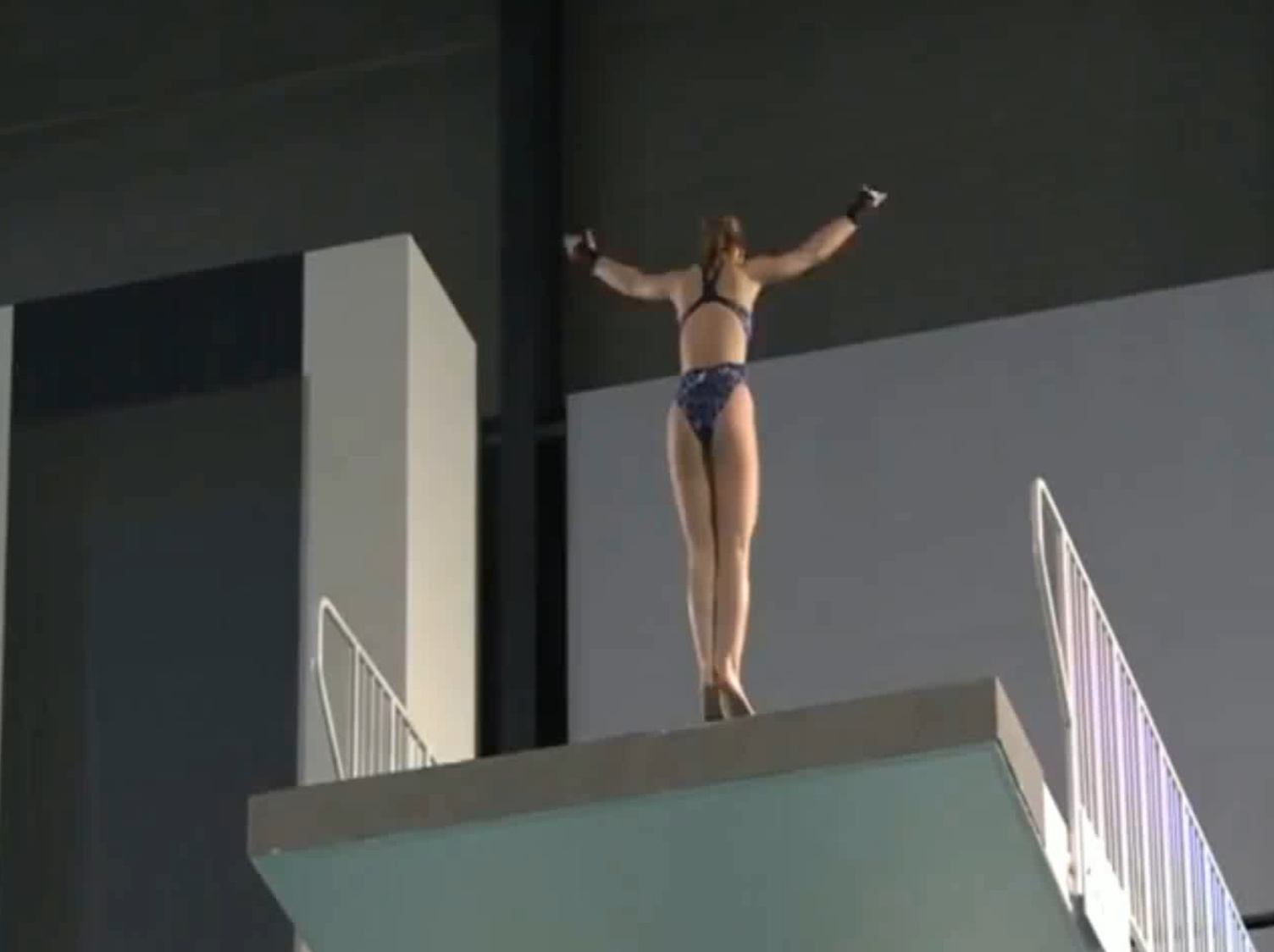}
    \includegraphics[width=0.19\linewidth]{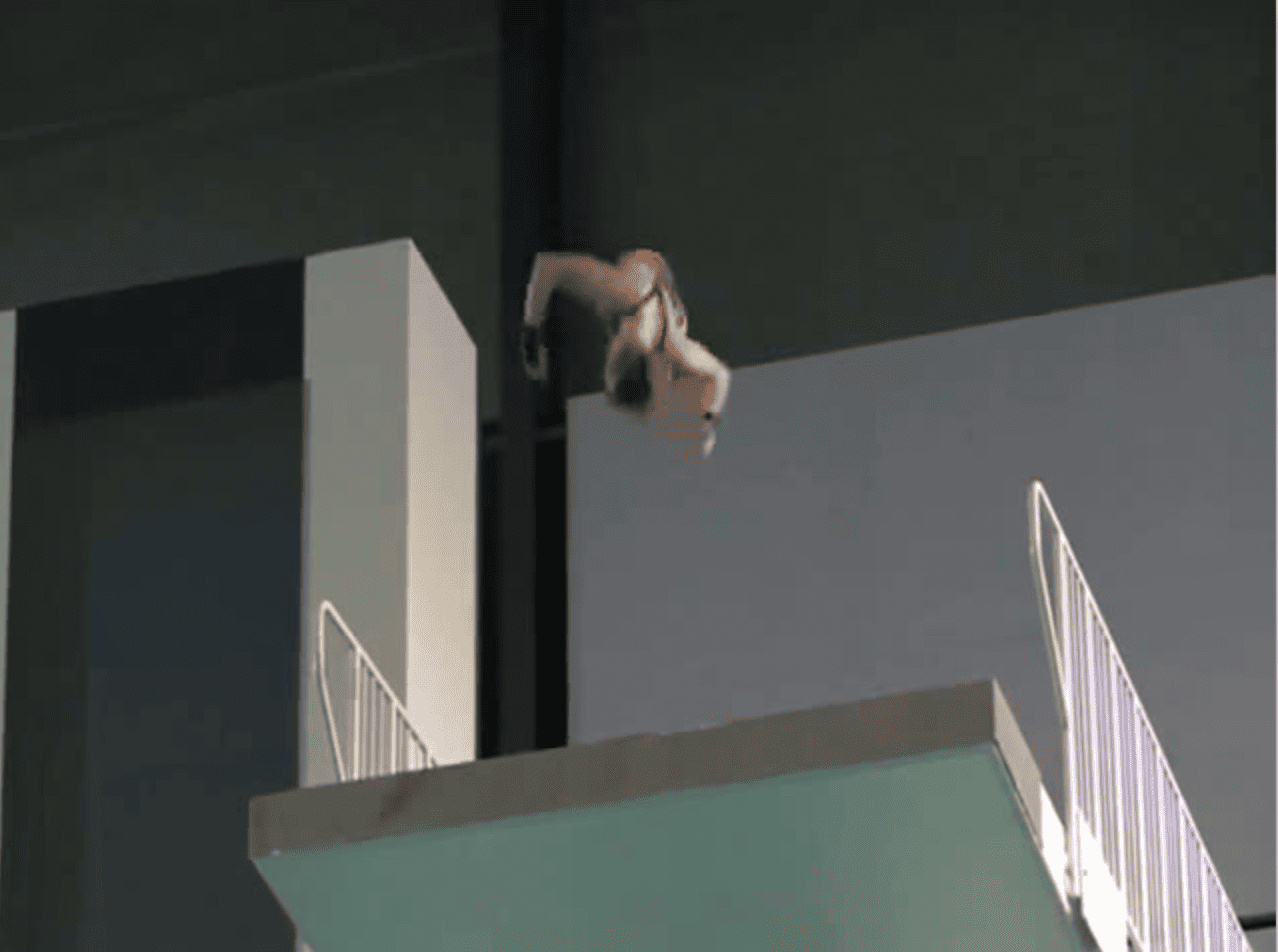}
    \includegraphics[width=0.19\linewidth]{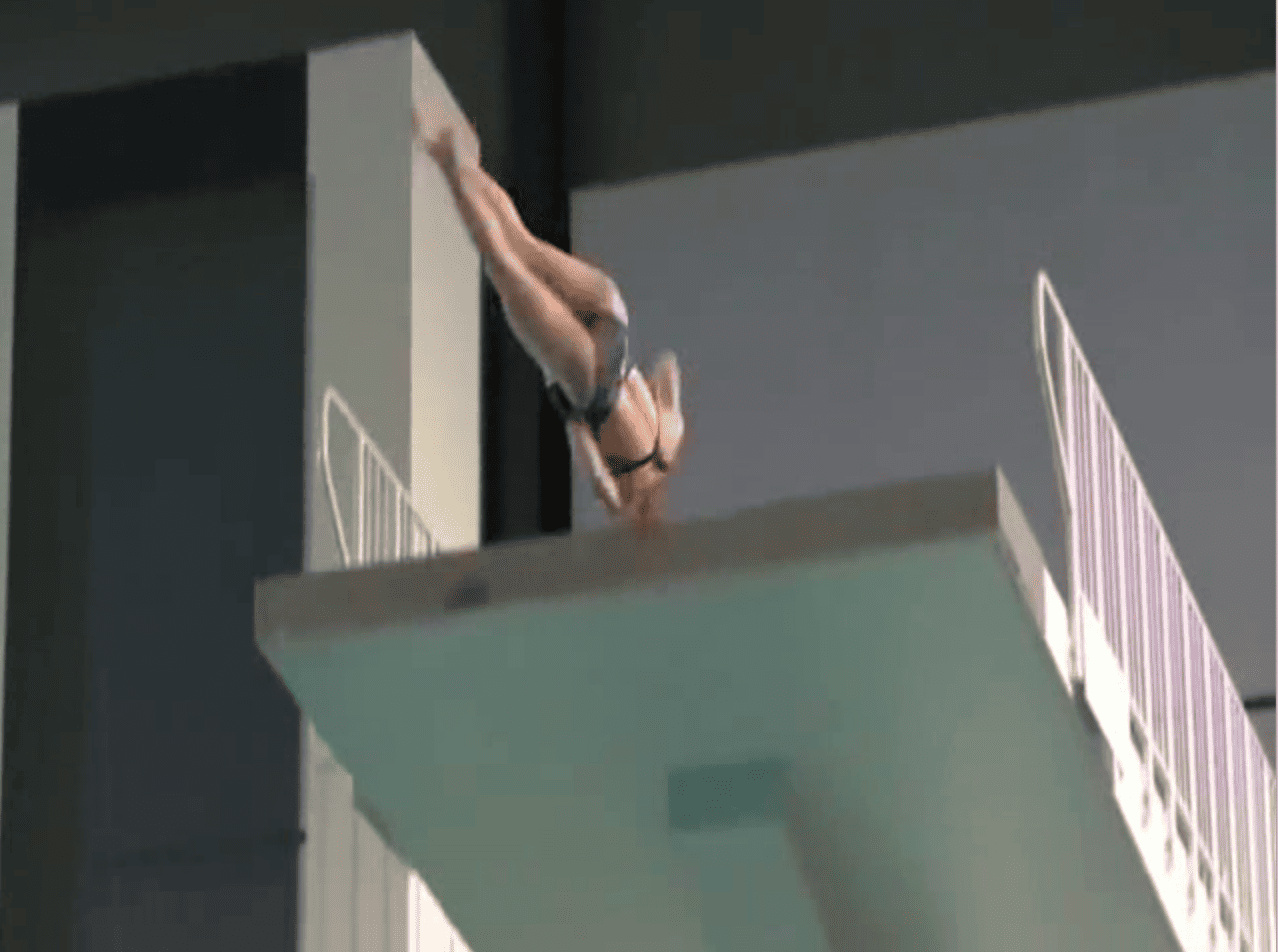}
    \includegraphics[width=0.19\linewidth]{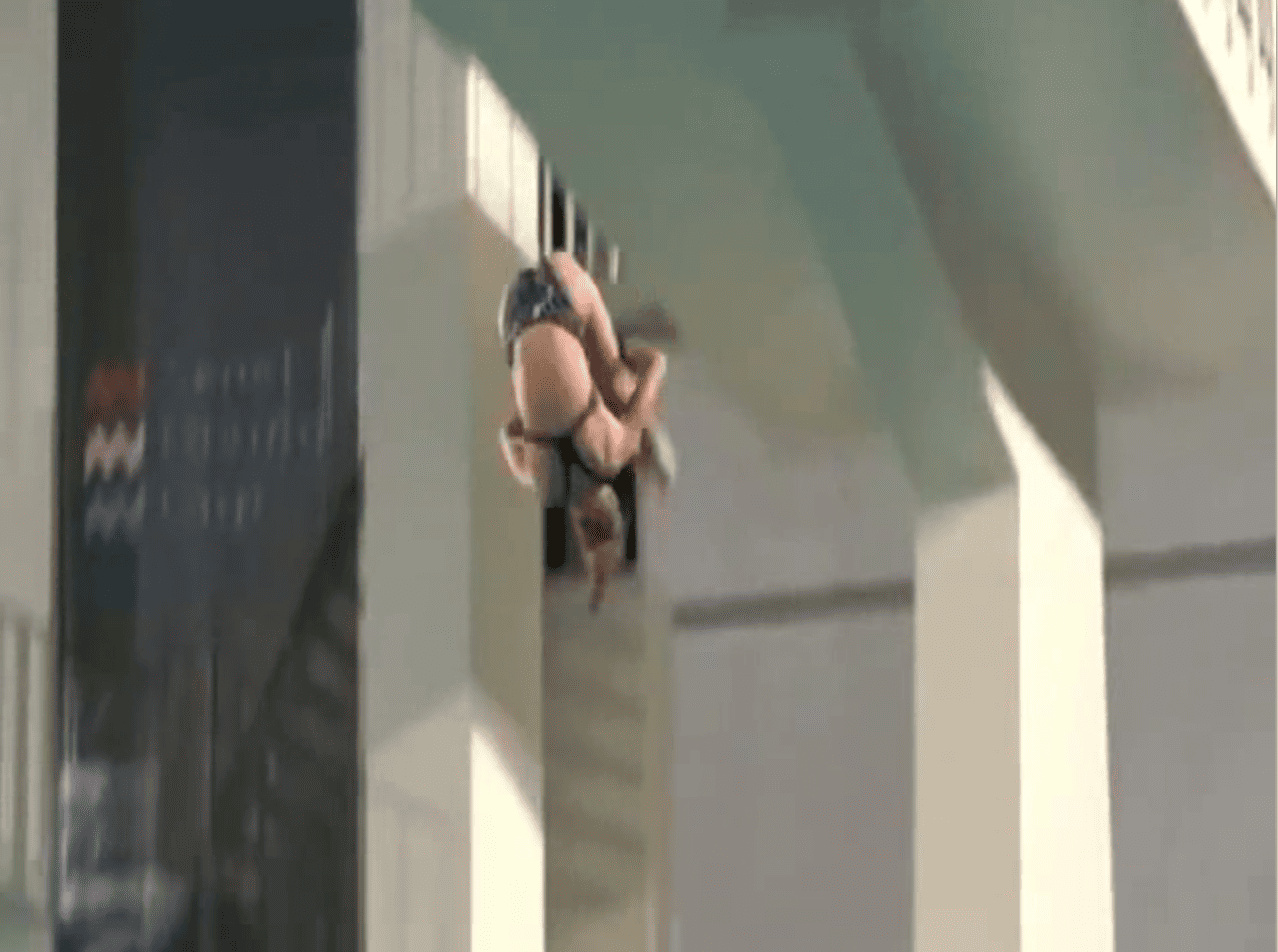}
    \includegraphics[width=0.19\linewidth]{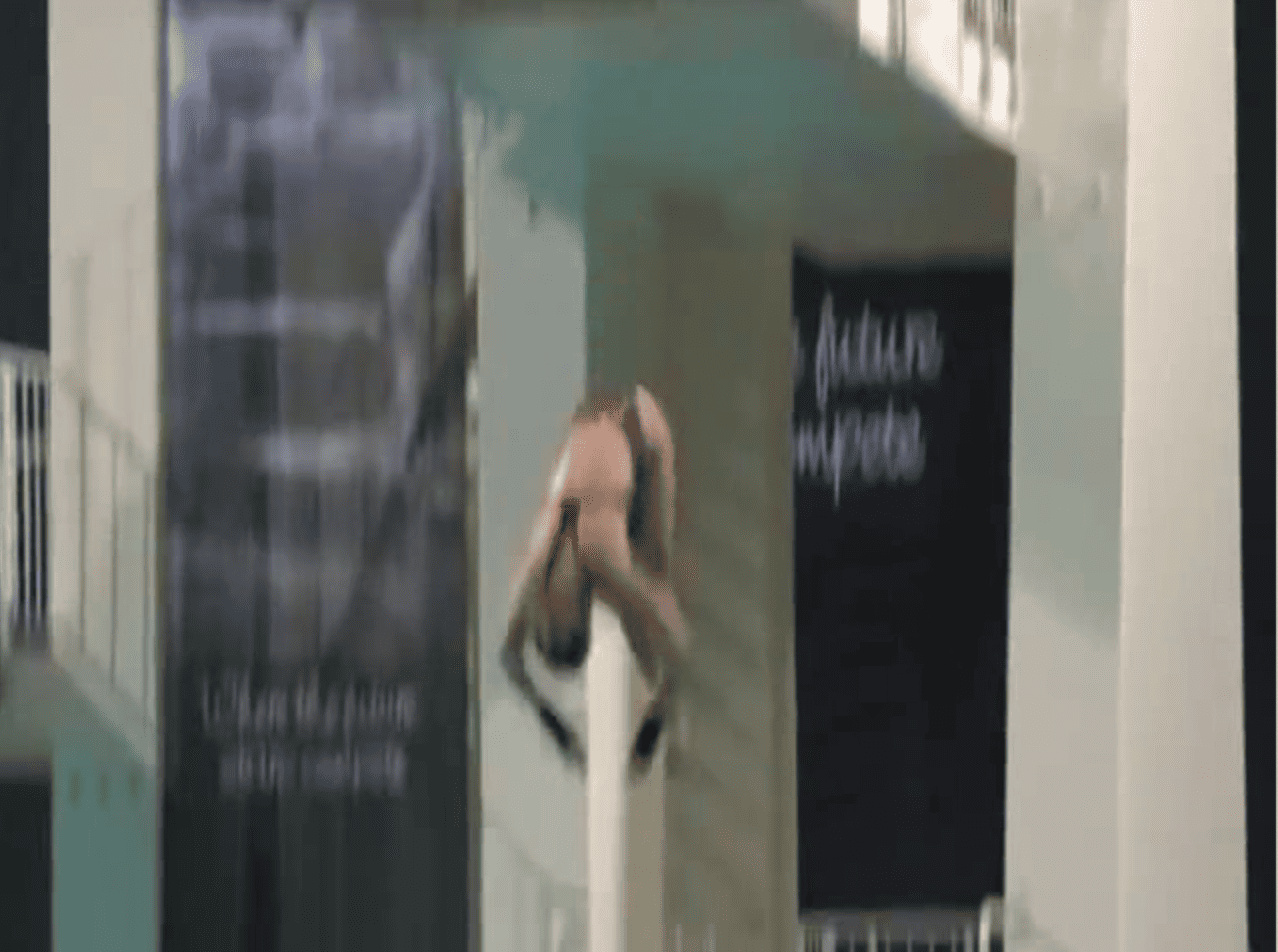}

        \includegraphics[width=0.19\linewidth]{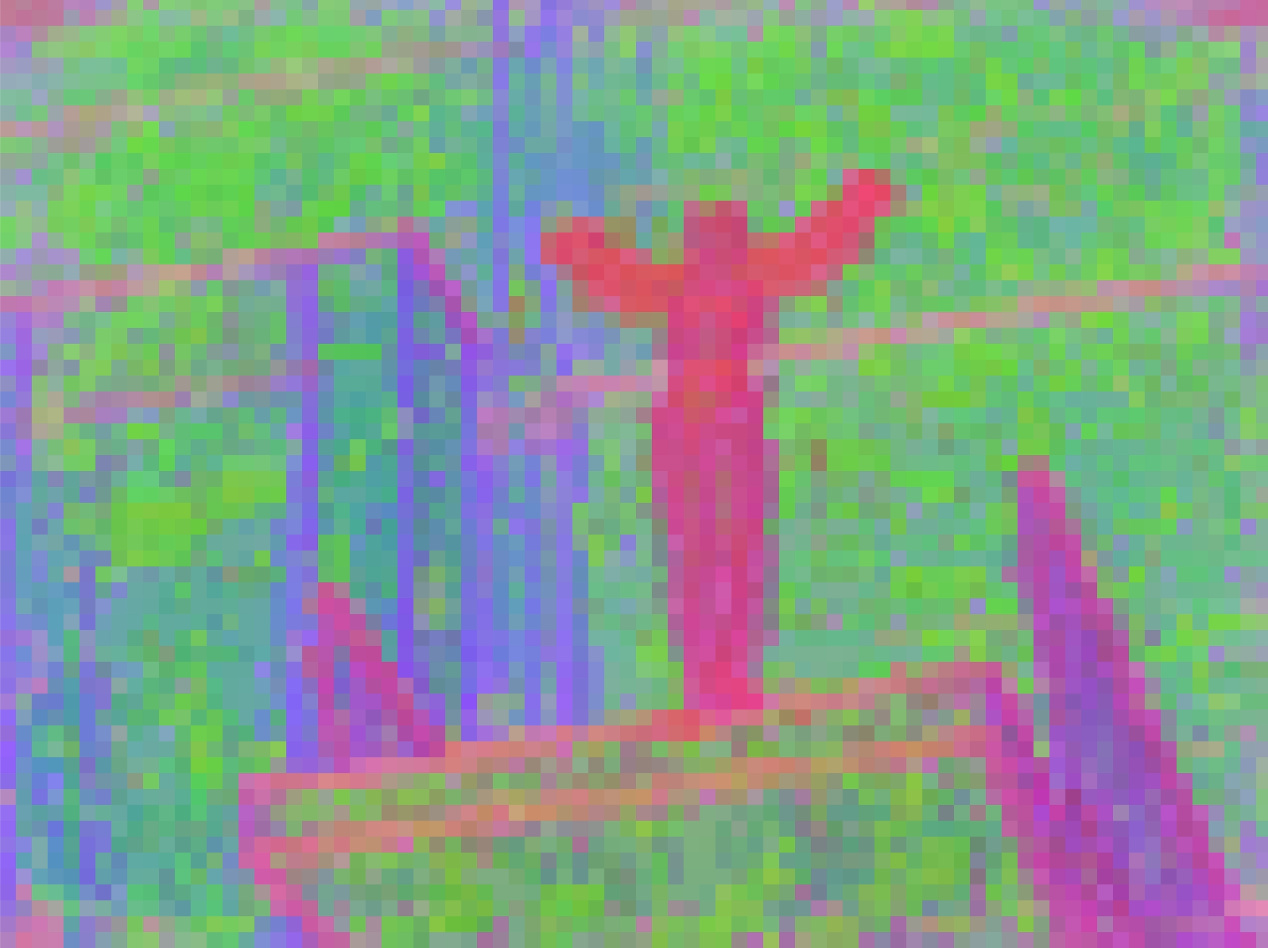}
    \includegraphics[width=0.19\linewidth]{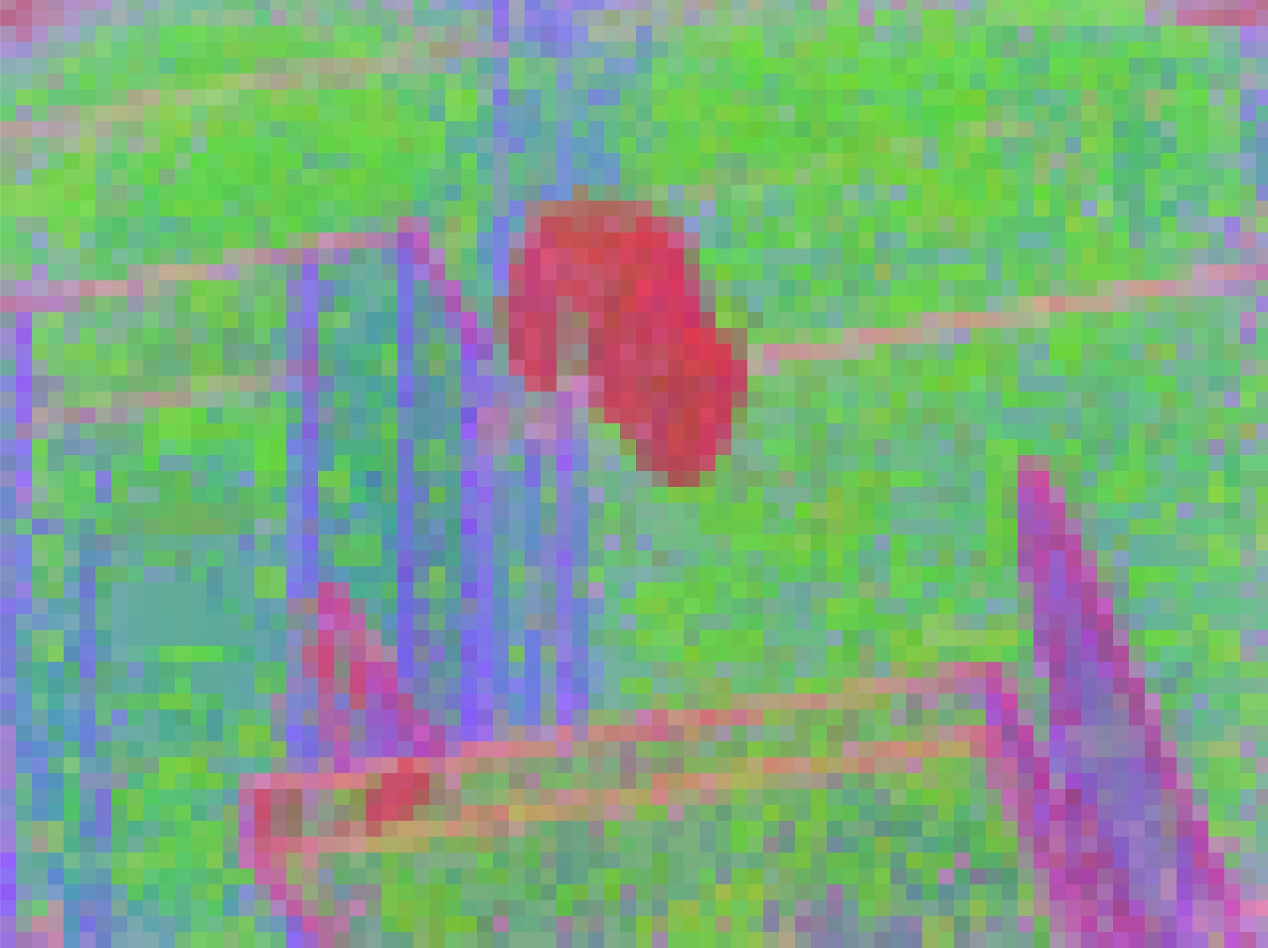}
    \includegraphics[width=0.19\linewidth]{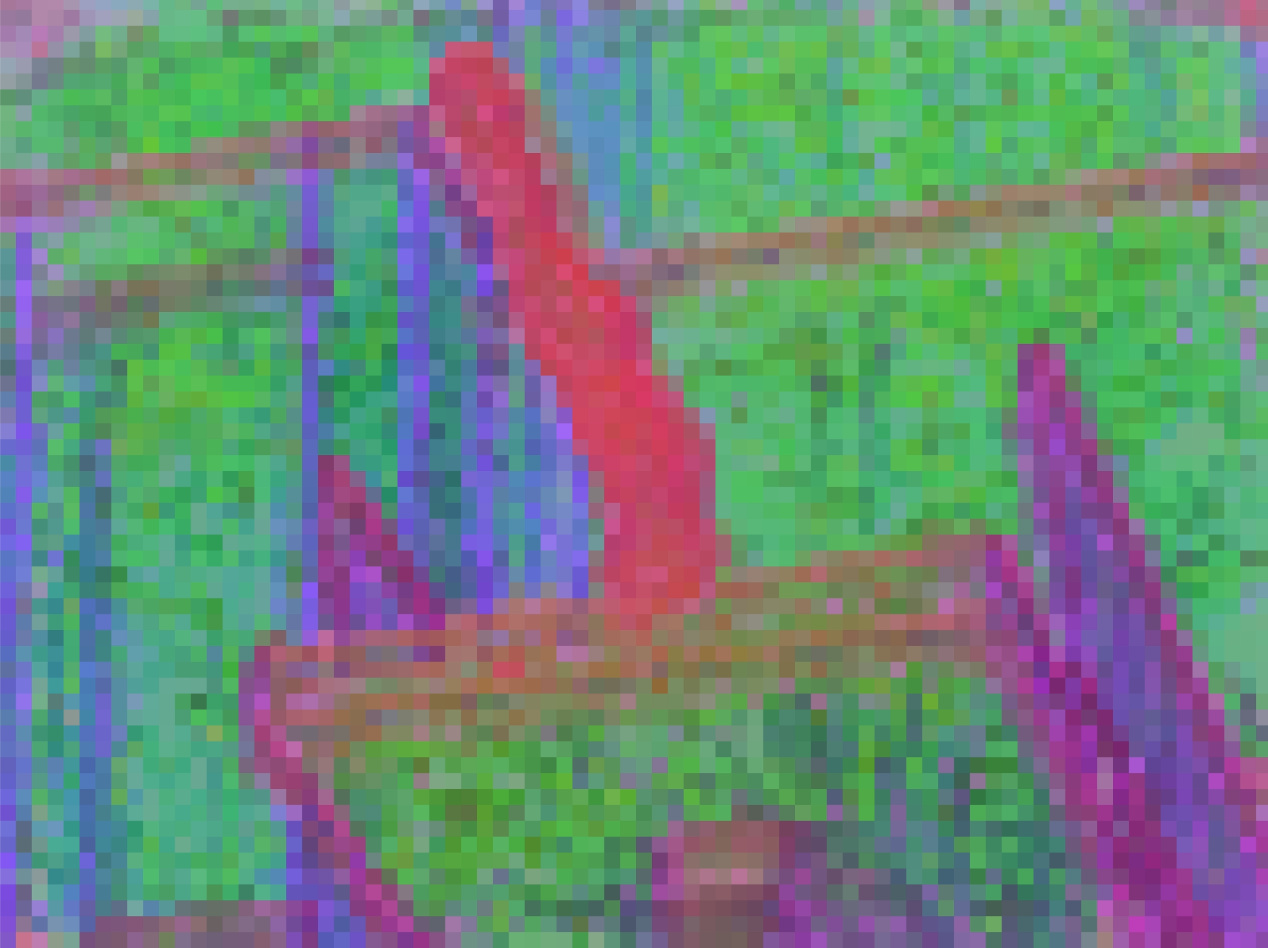}
    \includegraphics[width=0.19\linewidth]{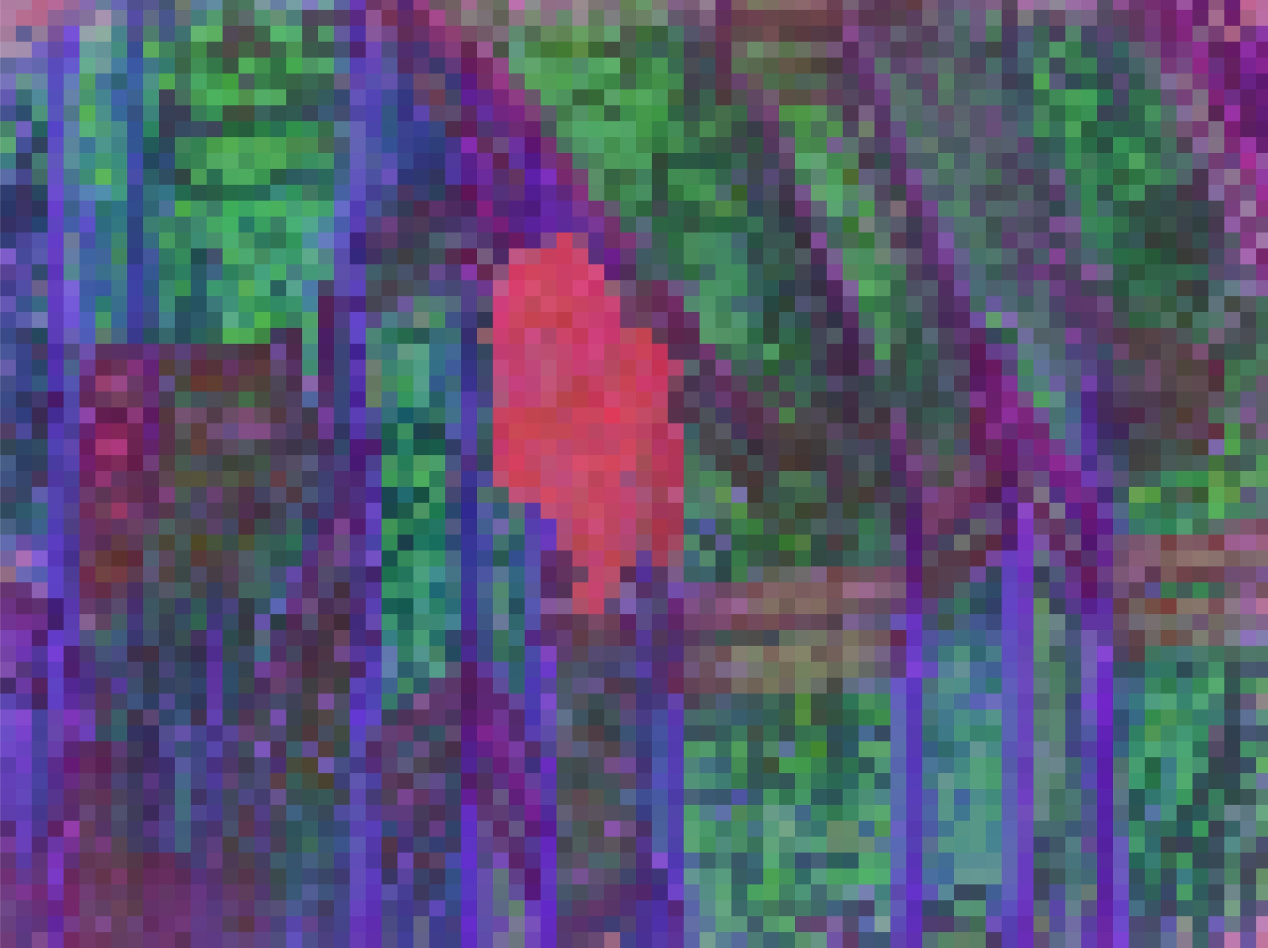}
    \includegraphics[width=0.19\linewidth]{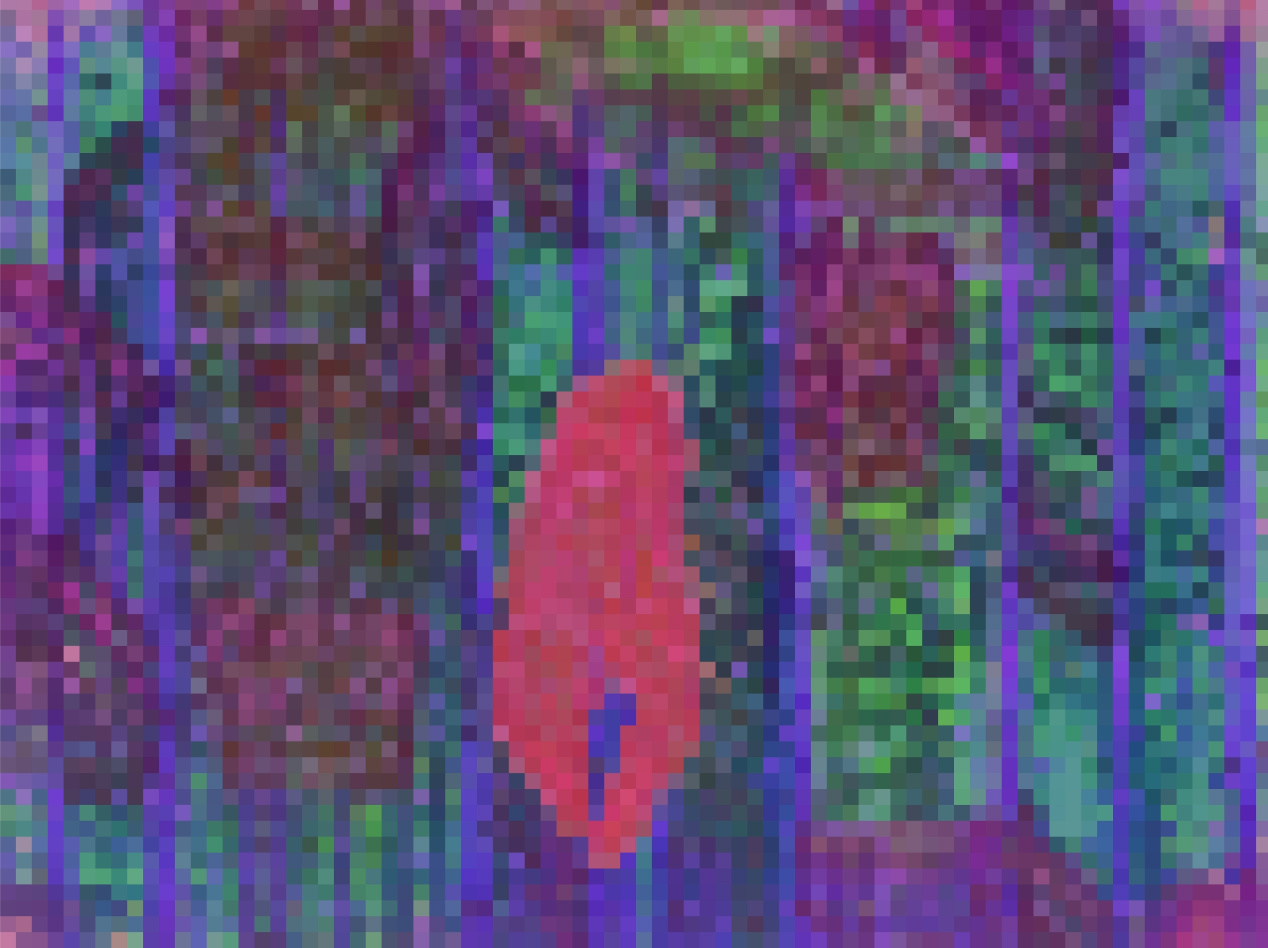}

    \caption{\textbf{V-JEPA 2.1 unlocks dense features from video.} \normalfont Qualitative results on Ego4D (1080px), Cityscapes (1080px), Diving48 (768px), and MOCA (768px) illustrate the strong temporal consistency of the dense features produced by our ViT-G model, particularly on dynamic objects. Since we process entire video sequences, we employ the 3D convolutional video tokenizer.}
    \label{fig:videos_qualit}
\end{figure}

\subsection{Qualitative results of V-JEPA 2.1 Dense Features}

Figure~\ref{fig: qualit_comparatie} presents a comparison of dense feature maps obtained from a single image using V-JEPA 2.1 ViT-G, DINOv2 ViT-g~\citep{oquab2023dinov2}, DINOv3 ViT-H+~\citep{simeoni2025dinov3} and V-JEPA 2 ViT-g~\citep{assran2025v}.
We project each encoder’s feature space into three dimensions using PCA.
Following \cite{simeoni2025dinov3}, we permute the three components across the six possible RGB channels permutations, and we display the most visually appealing.
The qualitative results in Figure~\ref{fig: qualit_comparatie} illustrate the high-quality of V-JEPA 2.1 dense feature maps obtained from a single image, which show fine-grained and precise identification of objects, such as the cat’s fur details or pedestrians in Cityscapes scenes.
Figure~\ref{fig:videos_qualit} further extends this comparison to full video sequences of 64 frames, showing temporal consistent dense maps across diverse scenarios, including egocentric videos from Ego4D \citep{grauman2022ego4d}, Cityscapes sequences~\citep{cordts2016cityscapes}, Diving48 videos~\citep{li2018resound} and animal camouflage footage from MOC~\citep{lamdouar2020betrayed}.

\subsection{Family of distilled models} \label{sec:results_distillation}

We follow the same protocol to evaluate our distilled ViT-L and ViT-B models. Table~\ref{tab:distill_vs_scratch} presents our results, where we observe a significant improvement in all our benchmarks between the ViT-L trained from scratch and the ViT-L distilled from ViT-G, almost closing the gap with ViT-G performance. On action recognition on SSv2, performance improves from 74.2\% to 76.5\%, almost reaching 77.7\% from ViT-G. In image segmentation on ADE20k, performance improves from 42.0 mIoU to 46.7 mIoU. In depth estimation and video segmentation, the ViT-L distilled is even closer to ViT-G, with 2.490 RMSE on KITTI, close to 2.461 RMSE, and 68.7 $\mathcal{J}\&\mathcal{F}$-Mean on DAVIS, close to 67.0 $\mathcal{J}\&\mathcal{F}$-Mean. Finally, our ViT-B offers competitive performance with our ViT-L trained from scratch.
\begin{table}[h]
    \centering

    \caption{\textbf{Benefits of Distillation.} Distilling the V-JEPA 2.1 ViT-G into smaller models (ViT-L, 300M; ViT-B, 80M) yields consistent improvements over training from scratch and approaches the performance of the ViT-G teacher.}
    \vspace{-1mm}
    \resizebox{\columnwidth}{!}{%
    \begin{tabular}{lcccccccccccc}
         Model & $\#$Params & Distilled & SSv2 & K400& IN1K& ADE20K& Citys. & VOC12& NYU ($\downarrow$) & KITTI ($\downarrow$) & DAVIS \\
         \hline
         ViT-G & 2B &  & 77.7& 87.7& 85.5& 47.9& 73.5 & 85.0& 0.307& 2.461&69.0\\
         ViT-L & 300M &  & 74.2& 84.3& 81.8& 42.0& 66.7 &79.6& 0.381& 2.914&64.5\\
         ViT-L & 300M & \checkmark & 76.5 & 86.9 & 84.9& 46.7& 71.0& 83.9 &0.336& 2.490&68.7\\
         ViT-B & 80M & \checkmark & 74.5 & 85.1 & 84.0& 42.0 & 64.5 &78.0& 0.408& 2.850&67.0\\
         \bottomrule
    \end{tabular}
    }
    \label{tab:distill_vs_scratch}
\end{table}

%\begin{table}[t]
%    \centering
%    \caption{\bf{Familly of distilled models.} .}
%    \begin{tabular}{lccccccccc}
%           Model&  SSv2&  K400&  IN1K&  ADE20K &Cityscapes&  VOC12& NYU &KITTI &DAVIS\\
%         \hline
%          Dinov2 ViT-L/14& 49.8& 82.4& 86.3& 48.8 &70.3& 82.1& 0.394& 2.78&64.4\\
%          DINOv3 ViT-L/16& 68.9& 84.9& 87.1& 54.9 &79.4& 86.2& 0.352& 2.64&72.2\\
%          \hdashline
%          V-JEPA 2.1 ViT-L/16& & & 84.9&  46.7&71.0& 83.9& 0.335& 2.49&68.7\\
%         \hline
%          Dinov2 ViT-B/14&  47.8&  79.6&  84.5&  48.4 &69.4&  82.5&  0.416& 2.90&64.9\\
%          Dinov3 ViT-B/16& 65.2& 82.2& 85.1& 51.8 &76.6& 85.8& 0.373& 2.750&70.3\\
%          \hdashline
%          V-JEPA 2.1 ViT-B/16& & & 84.0&  42.0&64.5& 78.0& 0.408& 2.85&67.0\\
%         \bottomrule
%    \end{tabular}
%    
%    \label{tab:distilled_vs_dino}
%\end{table}